\documentclass[10pt,twocolumn,letterpaper]{article}
\usepackage{cvpr}              

\usepackage[dvipsnames]{xcolor}

\usepackage{academicons}
\definecolor{orcidlogocol}{HTML}{A6CE39}

\usepackage{tikz}
\def\checkmark{\tikz\fill[scale=0.4](0,.35) -- (.25,0) -- (1,.7) -- (.25,.15) -- cycle;} 
\newcommand{\tikzcircle}[2][red,fill=red]{\tikz[baseline=-0.5ex]\draw[#1,radius=#2] (0,0) circle ;}%

\usepackage[table]{colortbl}
\definecolor{Gray}{gray}{0.9}
\definecolor{TUMBlue}{RGB}{0,101,189}

\usepackage{longtable}
\usepackage{xcolor}
\usepackage{threeparttablex}
\usepackage{multicol}
\usepackage[export]{adjustbox}

\usepackage{tocloft}
\usepackage[toc,page,header]{appendix}
\usepackage{minitoc}

\usepackage{titletoc}

\usepackage{amsmath,amssymb}
\usepackage{tabularray}
\usepackage{tabularx}
\usepackage{multirow}
\newcolumntype{x}[1]{!{\centering\arraybackslash\vrule width #1}}
\newcolumntype{L}[1]{>{\raggedright\arraybackslash}p{#1}} 

\definecolor{cvprblue}{rgb}{0.21,0.49,0.74}
\usepackage[pagebackref,breaklinks,colorlinks,citecolor=cvprblue]{hyperref}

\usepackage[capitalize]{cleveref}
\crefname{section}{Sec.}{Secs.}
\Crefname{section}{Section}{Sections}
\Crefname{table}{Table}{Tables}
\crefname{table}{Tab.}{Tabs.}

\def\paperID{16827} 
\def\confName{CVPR}
\def\confYear{2024}

\newcolumntype{N}{@{}m{0pt}@{}}
\newcolumntype{Y}{>{\centering\arraybackslash}X}

\title{{\color{TUMBlue}{TUM}}Traf V2X Cooperative Perception Dataset}

\author{
   Walter Zimmer\textsuperscript{\rm 1}\quad\quad\quad
   Gerhard Arya Wardana\textsuperscript{\rm 1}\quad\quad\quad
   Suren Sritharan\textsuperscript{\rm 1}\\
   Xingcheng Zhou\textsuperscript{\rm 1}\quad\quad
   Rui Song\textsuperscript{\rm 1,\rm 2}\quad\quad
   Alois C. Knoll\textsuperscript{\rm 1}\\
   \vspace{1ex}
    $^{1}$Technical University of Munich \quad $^{2}$Fraunhofer IVI\\
    \url{https://tum-traffic-dataset.github.io/tumtraf-v2x}
}

\begin{document}


\makeatletter
\let\@oldmaketitle\@maketitle
\renewcommand{\@maketitle}{\@oldmaketitle
  \includegraphics[width=1.0\linewidth,trim=1cm 21.05cm 1cm 1cm]{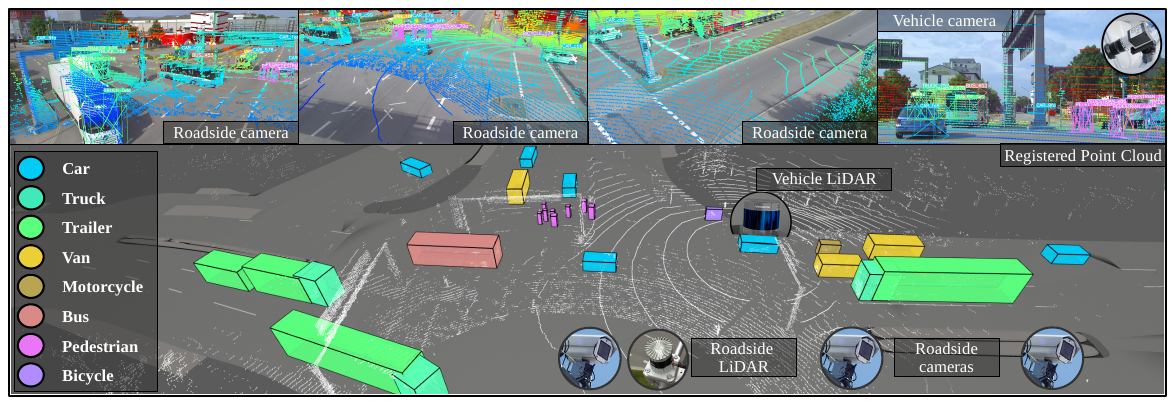}\bigskip
  \captionof{figure}{\textbf{Visualization} of 3D box labels and tracks in our \textbf{TUMTraf V2X Cooperative Perception Dataset}. The top part shows the labels projected into the four camera images. The part below shows a point cloud from two LiDARs with 3D box labels of the same scene.\\}
  }
\makeatother
\maketitle


\begin{abstract}
Cooperative perception offers several benefits for enhancing the capabilities of autonomous vehicles and improving road safety. Using roadside sensors in addition to onboard sensors increases reliability and extends the sensor range. External sensors offer higher situational awareness for automated vehicles and prevent occlusions. We propose CoopDet3D, a cooperative multi-modal fusion model, and TUMTraf-V2X, a perception dataset, for the cooperative 3D object detection and tracking task. Our dataset contains 2,000 labeled point clouds and 5,000 labeled images from five roadside and four onboard sensors. It includes 30k 3D boxes with track IDs and precise GPS and IMU data. We labeled eight categories and covered occlusion scenarios with challenging driving maneuvers, like traffic violations, near-miss events, overtaking, and U-turns. Through multiple experiments, we show that our CoopDet3D camera-LiDAR fusion model achieves an increase of +14.36 3D mAP compared to a vehicle camera-LiDAR fusion model. Finally, we make our dataset, model, labeling tool, and dev-kit publicly available on our website.
\end{abstract}    
\begin{table*}[htbp]
\caption{Comparison of 3D cooperative V2X perception datasets with our proposed TUMTraf-V2X Cooperative Perception dataset (I=Infrastructure, V=Vehicle). 
}
\label{tab:dataset2}
\centering
\begin{ThreePartTable}
\begin{tabularx}{\linewidth}{l|YYYYYYY}

\hline
Dataset & \small{OPV2V} \cite{xu2022opv2v} & \small{V2XSet} \cite{xu2022v2x} & \small{V2X-Sim} \cite{li2022v2x} & \small{V2V4Real} \cite{xu2023v2v4real} & \small{DAIR-V2X-C} \cite{yu2022dair} & \small{V2X-Seq (SPD)} \cite{yu2023v2x} & \small{\textbf{TUMTraf-V2X (Ours)}}\\
\hline
Year & \cellcolor{yellow!10}2022 & \cellcolor{yellow!10}2022 & \cellcolor{yellow!10}2022 & \cellcolor{yellow!10}2022 & \cellcolor{yellow!10}2022 & \cellcolor{green!10}2023 & \cellcolor{green!10}2024 \\
V2X & V2V & V2V\&I & V2V\&I & V2V & V2I & V2I & V2I\\
Real data & \cellcolor{red!10}- & \cellcolor{red!10}- &        \cellcolor{red!10}- & \cellcolor{green!10}\checkmark        & \cellcolor{green!10}\checkmark & \cellcolor{green!10}\checkmark & \cellcolor{green!10}\checkmark \\
Annotation range & \cellcolor{yellow!10}120 m & \cellcolor{yellow!10} 120 m&\cellcolor{red!10} 70 m& \cellcolor{green!10}200 m& \cellcolor{green!10}280 m& \cellcolor{green!10}280 m& \cellcolor{green!10} 200 m \\
Day \& night scenes& \cellcolor{red!10}-& \cellcolor{red!10}-& \cellcolor{red!10}-& \cellcolor{red!10}-& \cellcolor{green!10}\checkmark& \cellcolor{green!10}\checkmark & \cellcolor{green!10}\checkmark                                                 \\
\# object classes & \cellcolor{red!10}1 & \cellcolor{red!10}1 & \cellcolor{red!10}1 & \cellcolor{yellow!10}5 & \cellcolor{green!10}10 & \cellcolor{green!10}9 & \cellcolor{green!10}8 \\
Track IDs & \cellcolor{red!10}- & \cellcolor{red!10}- & \cellcolor{green!10}\checkmark & \cellcolor{green!10}\checkmark  & 
 \cellcolor{red!10}- & \cellcolor{green!10}\checkmark
 & \cellcolor{green!10}\checkmark                                                        \\
HD Maps & \cellcolor{green!10}\checkmark & \cellcolor{green!10}\checkmark & \cellcolor{green!10}\checkmark & \cellcolor{green!10}\checkmark        &  \cellcolor{red!10}-  & \cellcolor{green!10}\checkmark & \cellcolor{green!10}\checkmark \\
\begin{tabular}[c]{@{}l@{}}\# of sensors (I $\vert$ V)\end{tabular} & \cellcolor{yellow!10}- $\vert$ 6\tnote{*} &\cellcolor{yellow!10} - $\vert$ 6\tnote{*} & \cellcolor{green!10} 5 $\vert$ 7   & \cellcolor{green!10}- $\vert$ 8\tnote{$\ddagger$} & \cellcolor{red!10}2 $\vert$ 3 & \cellcolor{red!10}2 $\vert$ 3 & \cellcolor{green!10} 5 $\vert$ 4 \\
Available worldwide & \cellcolor{green!10}\checkmark & \cellcolor{green!10}\checkmark & \cellcolor{green!10}\checkmark       & \cellcolor{green!10}\checkmark        & \cellcolor{red!10}-  & \cellcolor{red!10}- & \cellcolor{green!10}\checkmark \\
Traffic violations & \cellcolor{red!10}-& \cellcolor{red!10}-& \cellcolor{red!10}-& \cellcolor{red!10}-& \cellcolor{red!10}-& \cellcolor{red!10}- & \cellcolor{green!10}\checkmark                                                 \\
Labeled attributes$^\#$ & \cellcolor{red!10}-& \cellcolor{red!10}-& \cellcolor{red!10}-& \cellcolor{red!10}-& \cellcolor{red!10}-& \cellcolor{red!10}- & \cellcolor{green!10}\checkmark                                                 \\
OpenLABEL format & \cellcolor{red!10}-& \cellcolor{red!10}-& \cellcolor{red!10}-& \cellcolor{red!10}-& \cellcolor{red!10}-& \cellcolor{red!10}- & \cellcolor{green!10}\checkmark                                                 \\
\# Point Clouds & 11k & 11k & 10k & 20k & 39k & 15k & 2.0k\\
\# Images & 44k & 44k & 60k & 40k\tnote{$\dagger$} & 39k & 15k & 5.0k \\
\# 3D Boxes & 233k & 233k & 26k & 240k & 464k & 10.45k & 29.38k\\
Location & CARLA & CARLA & CARLA & USA & China & China & Germany \\
\hline
\end{tabularx}
\begin{tablenotes}
\setlength{\columnsep}{0.4cm}
\setlength{\multicolsep}{0cm}
  \begin{multicols}{2}
    \footnotesize
    \item [$\dagger$] Image dataset has not been released yet.
    \item [*] Value per vehicle. Multiple Conn. and Autom. Vehicles (CAVs) are used.
    \item [$\ddagger$] Total sensors from 2 CAVs.
    \item [$^\#$] Weather, time of day, orientation, number of LiDAR points
  \end{multicols}
\end{tablenotes}
\end{ThreePartTable}
\end{table*}


\section{Introduction}

\label{sec:introduction}
Cooperative perception involves the fusion of onboard sensor data and roadside sensor data, and it offers several advantages for enhancing the capabilities of autonomous vehicles and improving road safety. Using data from multiple sources makes the perception more robust to sensor failures or adverse environmental conditions. Roadside sensors provide an elevated view that helps to detect obstacles early. Moreover, they are also beneficial for precise vehicle localization and reduce the computational load of automated vehicles by offloading some perception tasks to the roadside sensors. Roadside sensors provide a global perspective of the traffic and offer a comprehensive situational awareness when fused with onboard sensor data. There are also fewer false positives or negatives because cooperative perception cross-validates the information from different sensors.

Infrastructure sensors can share perception-related information with vehicles through V2X. Due to minimal delay, and real-time capabilities, the infrastructure-based perception systems can further enhance the situational awareness and decision-making processes of vehicles.

Intelligent Transportation Systems (ITS) like the Testbed for Autonomous Driving \cite{krammer2022providentia} aim to improve safety by providing real-time traffic information. According to \cite{Cre.10122021}, testbeds extensively start using LiDAR sensors in their setups to create an accurate live digital twin of the traffic. Connected vehicles get a far-reaching view which enables them to react to breakdowns or accidents early. ITS systems also provide lane and speed recommendations to improve the traffic flow.\\
The key challenge with ego-centric vehicle datasets is that there are many occlusions from a vehicle perspective, e.g., if a large truck in front of the ego vehicle obscures the view. Roadside sensors located at a smart intersection provide a broad overview of the intersection and a full-surround view. Given the immense potential of ITS, there is a specific need for V2X datasets. Despite the high costs associated with collecting and labeling such datasets, this work addresses this challenge as a crucial step toward realizing large-scale ITS implementations.

\noindent
\textbf{Our contributions are as follows:}
\begin{itemize}
    \item We provide a high-quality V2X dataset for the cooperative 3D object detection and tracking task with 2,000 labeled point clouds and 5,000 labeled images. In total, 30k 3D bounding boxes with track IDs were labeled in challenging traffic scenarios like near-miss events, overtaking scenarios, U-turn maneuvers, and traffic violation events.
    \item We open-source our 3D bounding box annotation tool (\textit{3D BAT} v24.3.2) to label multi-modal V2X datasets.
    \item We propose \textit{CoopDet3D}, a cooperative 3D object detection model, and show in extensive experiments and ablation studies that it outperforms single view models on our V2X dataset by  +14.3 3D mAP.
    \item Finally, we provide a development kit to load the annotations in the widely recognized and standard format OpenLABEL \cite{hagedorn2021open}, to facilitate a seamless integration and utilization of the dataset. Furthermore, it can preprocess, visualize, and convert labels to and from different dataset formats, and evaluate perception and tracking methods.
\end{itemize}

\begin{figure*}
    \centering
    \includegraphics[width=1.0\linewidth,trim=0 0 0 0]{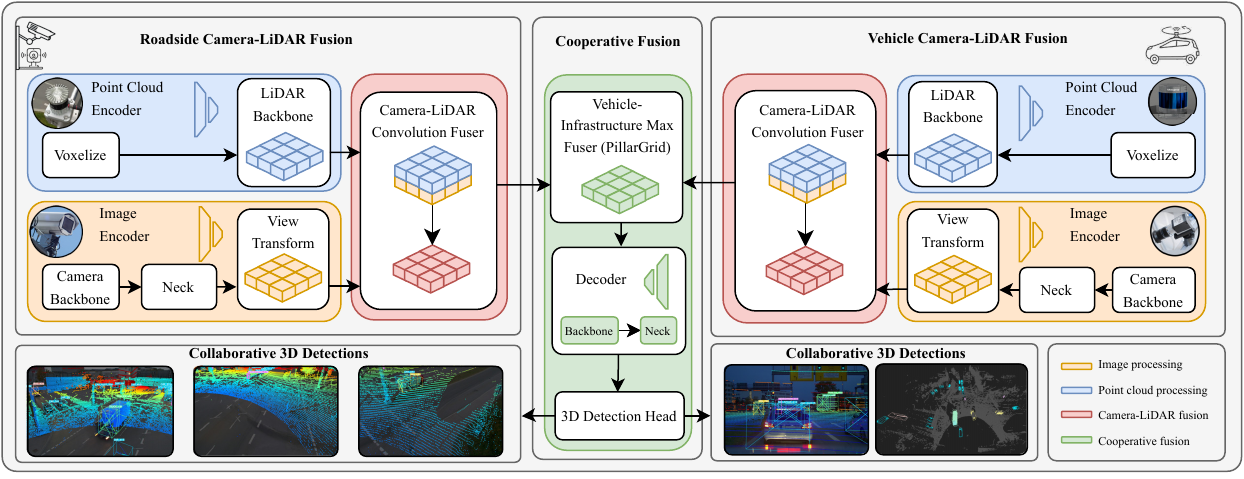}
    \caption{Our CoopDet3D framework is a multi-modal cooperative fusion system, comprising three distinct fusion pipelines. 1) The roadside camera-LiDAR fusion pipeline fuses three camera images and one LiDAR point cloud by extracting features and transforming them into a BEV representation. 2) The vehicle camera-LiDAR fusion pipeline fuses the vehicle camera feature map with the vehicle point cloud feature map using a convolutional fuser. 3) The vehicle and infrastructure feature maps are then fused by applying an element-wise max-pooling operation (Max Fuser). In the end, we use the TransFusion \cite{bai2022transfusion} 3D detection head to obtain 3D bounding box predictions.}
    \label{fig:coopdet3d}
\end{figure*}

\section{Related work}
\label{sec:related_work}

3D autonomous driving datasets are mainly categorized based on the viewpoint. Table \ref{tab:dataset2} highlights the main differences between our proposed dataset and other V2X datasets. 

\subsection{Single viewpoint datasets}
Single viewpoint datasets are obtained from a single point of reference, either an ego-vehicle or roadside infrastructure. Onboard sensor-based datasets like KITTI \cite{geiger2013vision}, nuScenes \cite{caesar2020nuscenes}, and Waymo \cite{sun2020scalability} contain a diverse set of sensor data collected from a moving vehicle equipped with multiple sensors, including high-resolution cameras, LiDARs, radars, and GPS/INS systems. These datasets are abundant and provide many annotated data, including bounding boxes, track IDs, segmentation masks, and depth maps under different urban driving scenarios.

On the other hand, roadside sensor-based datasets are in the infancy stage. High-quality multi-modal (camera and LiDAR) datasets are presented in \cite{cress2022a9,zimmer2023tumtraf,busch2022lumpi}, which are obtained from Infrastructure Perception Systems (IPS). Similarly, in \cite{ye2022rope3d}, the authors provide a dataset consisting of only images taken from different viewpoints and under varying traffic conditions. These datasets provide a top-down view of a crowded intersection under different conditions and, as such, can overcome issues such as occlusions created by other vehicles and thereby have a higher number of object labels than onboard sensor-based datasets.

\subsection{V2X datasets}
V2X datasets exploit the information from multiple viewpoints to gain additional knowledge regarding the environments. In this way, they overcome the limitations of single viewpoint datasets such as occlusion, limited field of view (FOV), and low point cloud density.

DAIR-V2X dataset family \cite{yu2022dair} is one of the foremost cooperative multi-modal datasets introduced. It contains three subsets: an intersection, a vehicle, and a cooperative dataset. 
The cooperative dataset contains 464k 3D box labels belonging to 10 classes, making it one of the largest cooperative datasets. The V2X-Seq dataset \cite{yu2023v2x} extends selected sequences of the DAIR-V2X dataset with track IDs and is partitioned into a sequential perception dataset (SPD) and a trajectory forecasting dataset. Despite these, the lack of specific information, such as the labeling methodology used, the exact models of the sensors deployed, the distribution of the classes, and the scenarios within the dataset, leads to uncertainty in the extendability and application of this dataset in varying conditions.


In V2V4Real \cite{xu2023v2v4real}, the authors propose a multi-modal cooperative dataset focusing only on V2V perception. Two vehicles equipped with cameras, LiDAR, and GPS/IMU integration systems are used to collect multi-modal sensor data for diverse scenarios. As opposed to all other V2X datasets, this focuses on V2V perception, and though it is of similar size to other cooperative datasets, it contains fewer classes and 3D bounding box information.

Simulated multi-agent perception datasets have been proposed in \cite{li2022v2x, xu2022opv2v, xu2022v2x}.
These datasets contain multi-modal sensor data (camera and LiDAR) obtained from 
roadside units (RSUs) and multiple ego vehicles, which enable collaborative perception. They use a combination of simulators such as SUMO \cite{SUMO2018}, CARLA \cite{Dosovitskiy17}, and OpenCDA \cite{xu2022opv2v} for flow simulation, data retrieval, and V2X communication.
However, the utility of the dataset is still limited due to the simulated nature of the data, and its extendability to real-life applications has not been studied in detail.


\subsection{V2X perception models for object detection}

The datasets presented above have been used to develop various models for a wide variety of tasks, with the majority focusing on 3D object detection. Different approaches have been taken depending on the availability and challenges, and these methods are grouped based on the number of nodes employed and the modalities used for detection.

Most 3D object detection models use multi-modal sensor data obtained from a single point of view, which is often an ego-vehicle. Due to the popularity and abundant availability of vehicle datasets  \cite{geiger2013vision,caesar2020nuscenes,sun2020scalability}, most models use images, point cloud data, or both modalities. Image-based models were the pioneers in 3D object detection due to their low cost and simplicity, and both vehicular camera-based models \cite{kumar2022deviant, wu2023monopgc} and infrastructure camera-based models \cite{yang2023bevheight} have been proposed. LiDAR-based 3D object detection models \cite{lang2019pointpillars, zimmer2023real} became popular since LiDAR point clouds provide 3D depth information and are robust, especially in adverse weather conditions and nighttime scenarios. Fusion models combine the information obtained from both images and point clouds and have been shown to outperform the prior methods \cite{zimmer2023infra}. Single viewpoint fusion models use either vehicular camera and LiDAR \cite{liu2023bevfusion, vora2020pointpainting, yan2023cross} or infrastructure camera and LiDAR \cite{zimmer2023infra} for 3D object detection.

Cooperative perception models, which use data from multiple viewpoints, have been shown to overcome issues related to occlusion, which were often present in vehicular sensor-based models. V2I cooperative perception models \cite{bai2023vinet,he2021vi,bai2022pillargrid,xu2022v2x,yu2023vehicle,wei2023robust,qiao2023cobevfusion} use the sensor data from both vehicles and infrastructure and V2V models \cite{hu2023collaboration,song2024collaborative} communicate the sensor data between multiple vehicles. In this work, our cooperative multi-modal dataset is one contribution among others. Thus, while most of the prior works focus on unimodal cooperative perception using either LiDAR point clouds \cite{chen2023transiff, bai2022pillargrid} or camera images \cite{hu2023collaboration}, we benchmark our dataset with CoopDet3D, a deep fusion based cooperative multi-modal 3D object detection model based on BEVFusion \cite{hu2023collaboration} and PillarGrid \cite{bai2022pillargrid}.

\section{TUMTraf-V2X Dataset}
\label{sec:dataset}
Our TUMTraf V2X Cooperative Perception Dataset focuses on challenging traffic scenarios and various day and nighttime scenes. The data is further annotated, emphasizing high-quality labels through careful labeling and high-quality review processes. It also contains dense traffic and fast-moving vehicles, which reveals the specific challenges in cooperative perception, such as pose estimation errors, latency, and synchronization. Furthermore, we provide sensor data from nine different sensors covering the same traffic scenes under diverse weather conditions and lighting variations. The infrastructure sensors are oriented in all four directions of the intersection to get a $360^{\circ}$ view, which leads to better perception results. Finally, it contains rare events like traffic violations where pedestrians cross the road at a busy four-way intersection while the crossing light is lit red.

\subsection{Sensor setup}
Our TUMTraf V2X Cooperative Perception Dataset was recorded on an ITS system with nine sensors.\\
The infrastructure sensor setup is the following:
\begin{itemize}
    \item 1x Ouster LiDAR OS1-64 (gen. 2), 64 vert. layers, 360° FOV, below horizon config., 10 cm acc. \emph{{@}}120 m range
    \item 4x Basler ace acA1920-50gc, 1920$\times$1200, Sony IMX174 with 8 mm lenses
\end{itemize}
On the vehicle, the following sensors were used:
\begin{itemize}
     \item 1x Robosense RS-LiDAR-32, 32 vert. layers, 360° FOV, 3 cm accuracy \emph{@}200 m range
     \item 1x Basler ace acA1920-50gc, 1920$\times$1200, Sony IMX174 with 16 mm lens
     \item 1x Emlid Reach RS2+ multi-band RTK GNSS receiver
     \item 1x XSENS MTi-30-2A8G4 IMU
\end{itemize}

\begin{figure}
    \centering
    \includegraphics[width=\linewidth]{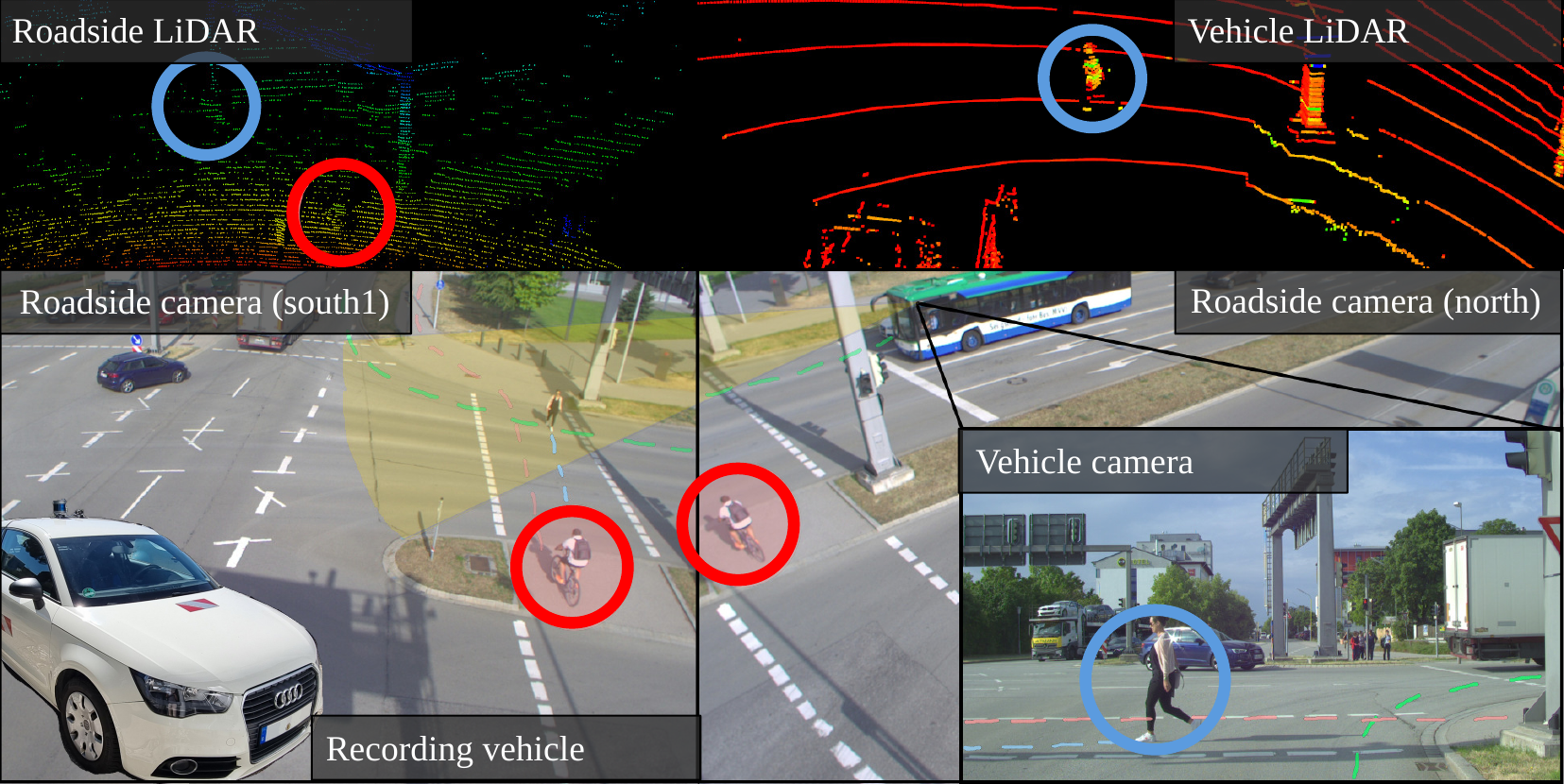}
    \caption{Demonstration of a possible V2X occlusion scenario. A pedestrian (blue) is crossing the road in front of the ego vehicle. An occluded bicycle is marked in red. The recording vehicle with the sensor setup is shown in the bottom left corner.}
    \label{fig:occlusion_scenario}
\end{figure}

\subsection{Sensor calibration and registration}
We synchronize the cameras and LiDARs in the spatial and temporal domain. First, we determine the intrinsic camera parameters and the radial and tangential image distortions by using a checkerboard target. We then calibrate the roadside LiDAR with the roadside cameras by picking 100-point pairs in the point cloud and camera image. Extrinsic parameters (rotation and translation) are calculated by minimizing the reprojection error of 2D-3D point correspondences \cite{madsen2004methods}. We follow the same procedure for onboard camera-LiDAR calibration. Finally, we calibrate the onboard LiDAR to the roadside LiDAR. This spatial registration is done by first estimating a coarse transformation. We pick ten 3D point pairs in each point cloud and minimize their distance using the least squares method. Then, we apply the point-to-point Iterative Closest Point (ICP) algorithm \cite{besl1992method} to get the fine transformation between the point clouds.

We label the vehicle and infrastructure point clouds after registering them. The coarse registration was done by measuring the GPS position of the onboard LiDAR and the roadside LiDAR. Then, we transform every 10th onboard point cloud to the coordinate system of the infrastructure point cloud. The fine registration was done by applying the point-to-point ICP to get an accurate V2I transformation matrix. All rotations of the point cloud frames in between are interpolated based on the spherical linear interpolation (SLERP) \cite{shoemake1985animating} method:
\begin{equation}
    SLERP(q_0,q_1,t) = q_0(q_0^{-1}q_1)^t,
\end{equation}
where $q_0$ and $q_1$ are the quaternions representing the rotations of the start and end frames and $t\in[0,1]$.
Translation vectors $\mathbf{T}_0$ and $\mathbf{T}_1$  were obtained using linear interpolation:
\begin{equation}
    \mathbf{T}(t) = \mathbf{T}_0 + t(\mathbf{T}_1 - \mathbf{T}_0).
\end{equation}
This dual interpolation strategy ensures that the estimated transformations between the frames are smooth and geometrically accurate, thus adhering closely to the actual movements of the vehicle over time.

\subsection{Data selection and labeling}
We selected the data based on challenging traffic scenarios, like U-turns, tailgate events, and traffic violation maneuvers. Besides the high traffic density of 31 objects per frame, we selected frames with high-class coverage. We selected 700 frames during sunny daytime and 100 frames during cloudy nighttime for labeling. The camera and LiDAR data were recorded into rosbag files at 15 Hz and 10 Hz, respectively. We extracted and synchronized the data based on ROS \cite{quigley2009ros} timestamps and labeled it with our \textit{3D BAT} (v24.3.2) annotation tool\footnote{\scriptsize{\url{https://github.com/walzimmer/3d-bat}}}. We improved the 3D BAT \cite{zimmer20193d} baseline labeling tool to label 3D objects faster and more precisely with a one-click annotation feature. The annotators were instructed to label traffic participants while examining the images. Objects are still labeled, even if they have no 3D points inside, but are visible in the images. Extremities (e.g., pedestrian limbs) are included in the bounding box, but side mirrors of vehicles aren't. If a pedestrian carries an object, that object is included in the bounding box. If two or more pedestrians are carrying an object, only the box of one will include the object. After labeling, each annotator checked the work of other annotators manually frame-by-frame. When errors were found, the original annotator was notified, and they fixed it. This helps ensure that the labels in our dataset are high quality.

\begin{figure*}[h!] 
\centering
\subfloat[Distribution of objects between day and night. \label{fig:distribution_of_object_classes}]{%
    \includegraphics[width=0.33\linewidth,trim={0 0 0 0.32cm},clip]{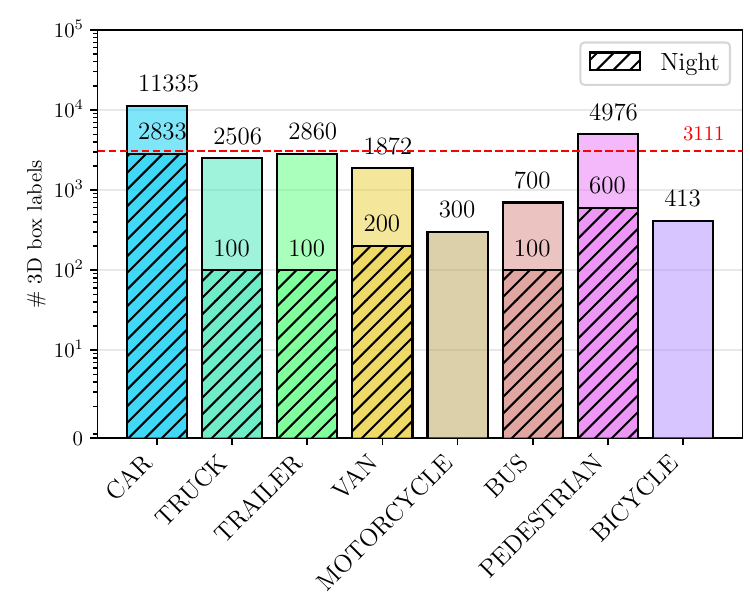}}
\hfill
\subfloat[Avg. and max. num. of 3D points for each class. \label{fig:tbd}]{%
    \includegraphics[width=0.33\linewidth,trim={0 0 0 0.32cm},clip]{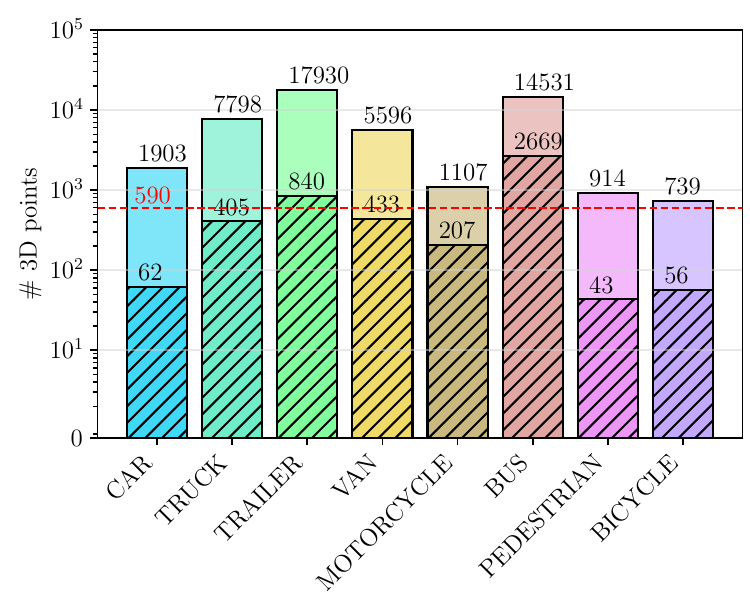}}
\hfill
\subfloat[Avg. and max. track length for all classes. \label{fig:tbd}]{%
    \includegraphics[width=0.33\linewidth,trim={0 0 0 0.32cm},clip]{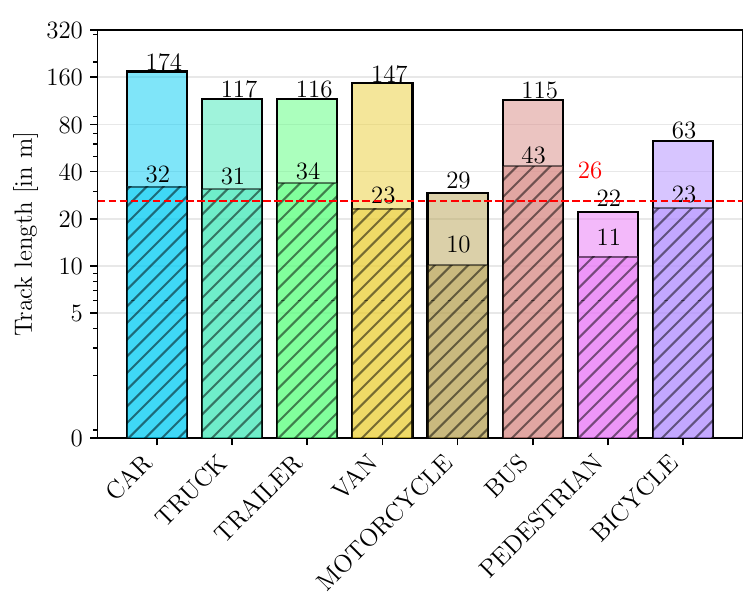}}
\caption{Our TUMTraf-V2X dataset (version 1.0) contains 25k 3D box labels in total and is balanced among eight different object classes. (a) Cars (11,203) and pedestrians (4,781) are highly represented in the dataset. (b) 3D box labels contain on average 590 points inside, which shows the density of the labeled objects. The BUS class has the highest point density. (c) All traffic participants are tracked for 26 m on average. Buses have the highest average track length of 43 m, whereas the CAR class contains the max. track length of 173.95 m.}
\label{fig:statistic_plots_01}
\end{figure*}
\begin{figure*}[h!] 
\centering
\subfloat[Visualization of rotations (yaw). \label{fig:rotations}]{%
    \includegraphics[width=0.33\linewidth]{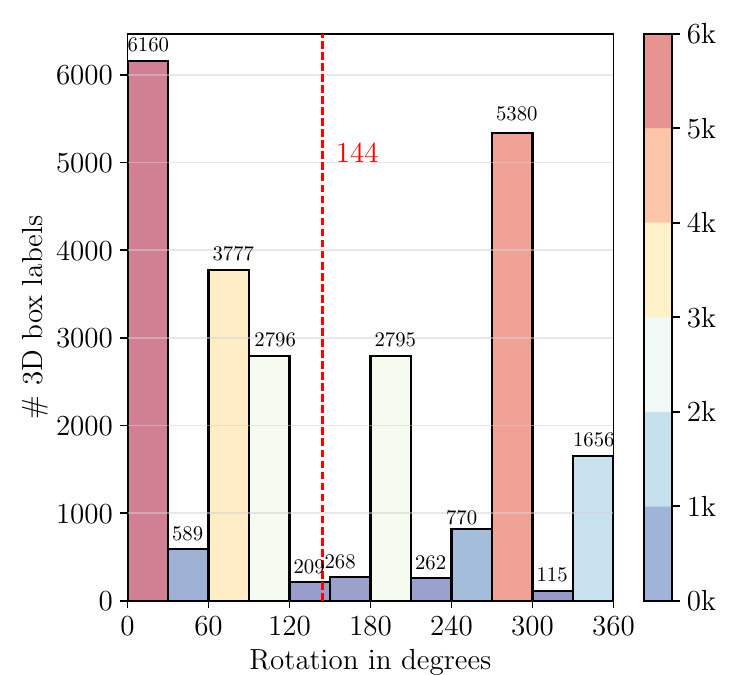}}
\hfill
\subfloat[3D points grouped by distance. \label{fig:num_points_with_distance}]{%
    \includegraphics[width=0.33\linewidth]{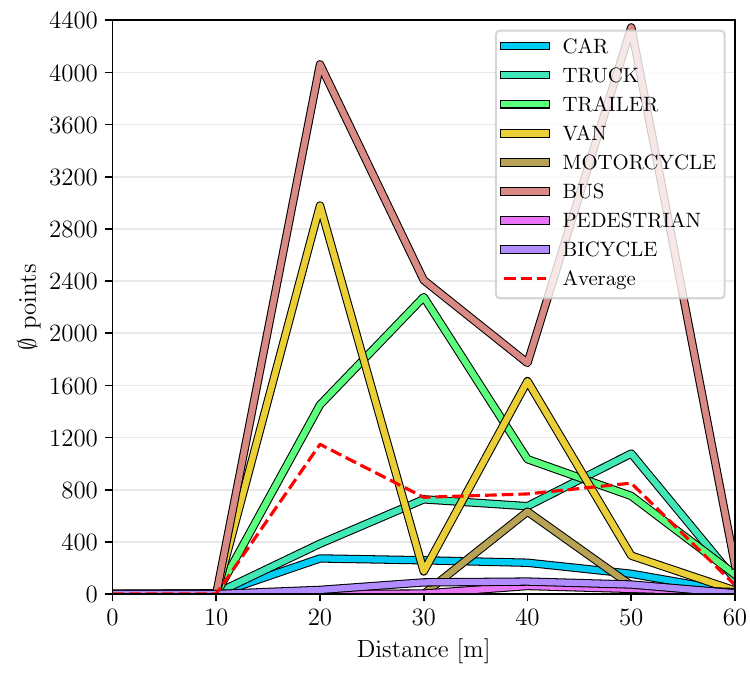}}
\hfill
\subfloat[BEV visualization of tracks. \label{fig:bev_plot}]{%
    \includegraphics[width=0.33\linewidth]{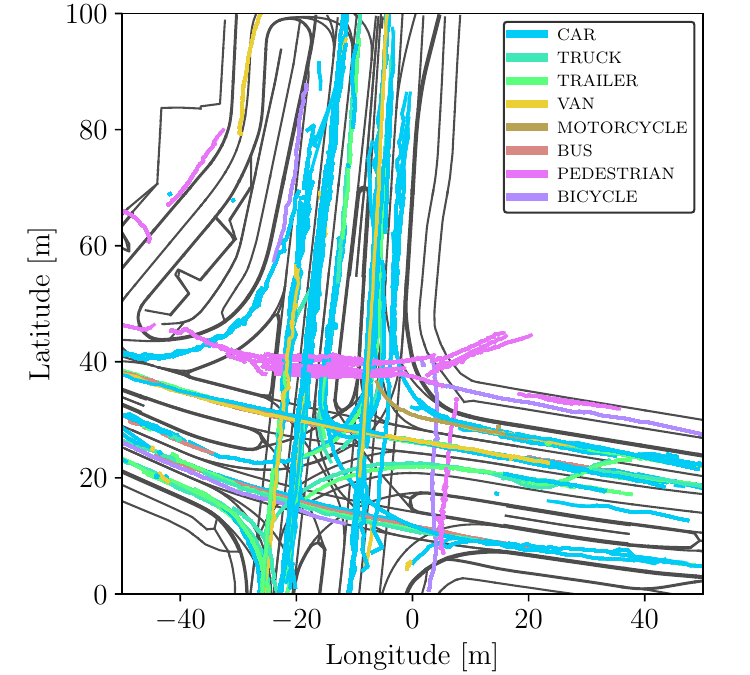}}
\caption{Our dataset was recorded at a crowded intersection with many left and right turns. (a) Most of the vehicles (6,160) are driving in the east direction (0 degree). (b) 3D boxes were labeled up to 200 m range and are very dense between 10 and 60 m. (c) The visualization of BEV tracks shows where pedestrians and bicycles are crossing the road.}
\label{fig:statistic_plots_02}
\end{figure*}
\begin{figure*}[h!] 
\centering
\subfloat[Histogram of 3D points. \label{fig:distribution_of_object_classes}]{%
    \includegraphics[width=0.33\linewidth]{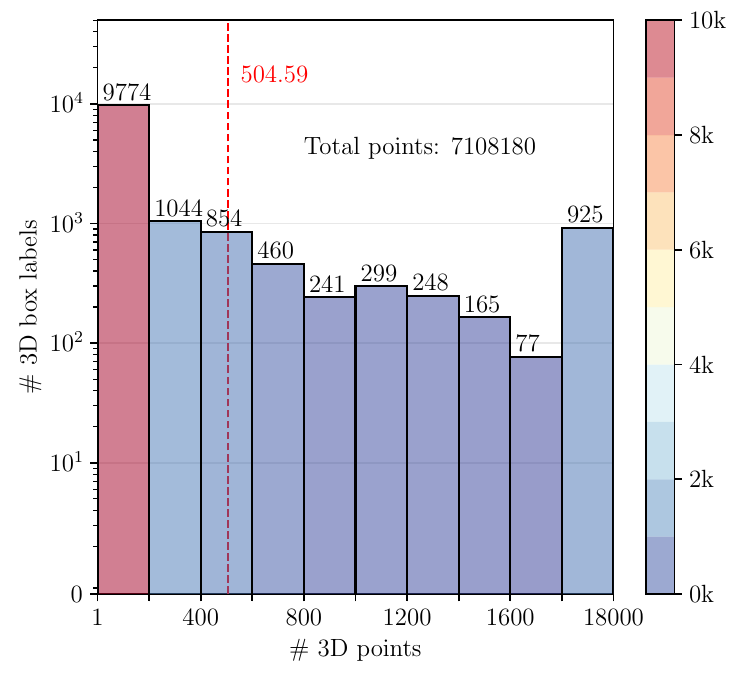}}
\hfill
\subfloat[Histogram of 3D box labels. \label{fig:histogram_objects_in_frame}]{%
    \includegraphics[width=0.33\linewidth]{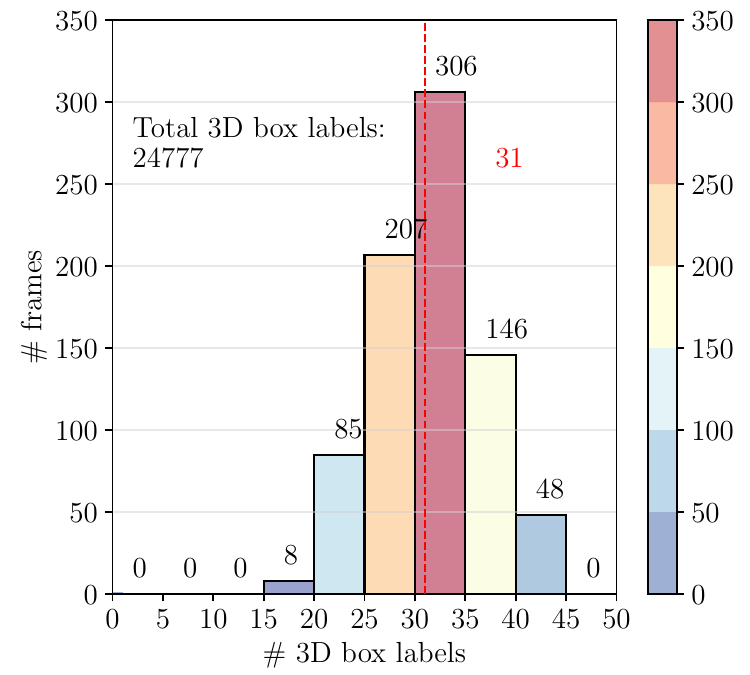}}
\hfill
\subfloat[Histogram of track lengths. \label{fig:histogram_track_lengths}]{%
    \includegraphics[width=0.33\linewidth]{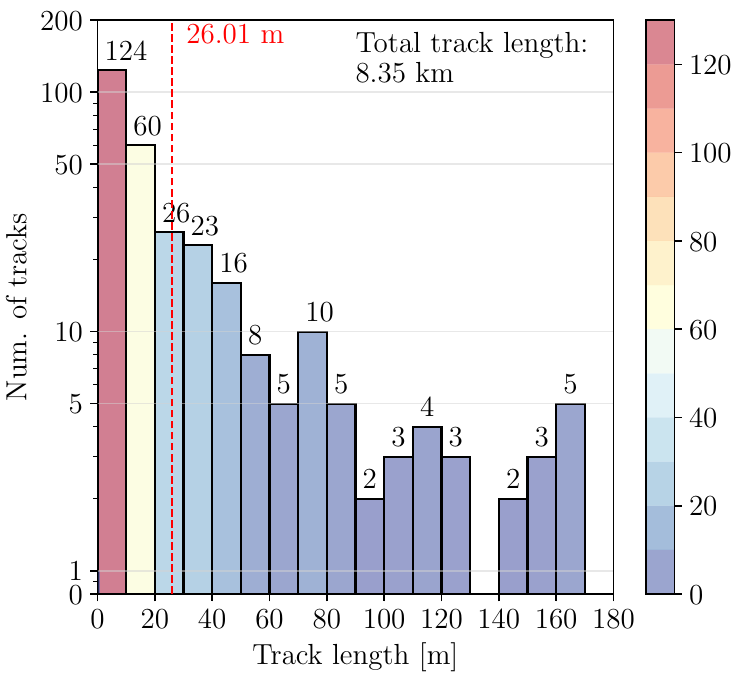}}
\caption{(a) Most 3D box labels contain between 1 and 200 3D points inside, with an average of 505 3D points, excluding empty boxes. Objects that were close to both LiDAR sensors even contained up to 18k 3D points. (b) Frames contain between 20 and 45 traffic participants, with an average of 31. (c) Objects were tracked up to 180 m and the average track length is 26 m.}
\label{fig:statistic_plots_03}
\end{figure*}

\subsection{Data structure and format}
We record eight different scenes, each 10 sec. long, from vehicle and infrastructure perspectives using nine sensors and split the data into a train (80\%), val. (10\%), and test (10\%) set. We use stratified sampling to distribute all sets' object classes equally (see Fig. \ref{fig:distribution_of_object_classes}). Labels are provided in the ASAM OpenLABEL \cite{hagedorn2021open} standard.

\subsection{Dataset development kit}
We provide a dev-kit to work with our dataset. In addition to generating the data statistics, it provides modules for multi-class stratified splitting (train/val/test), point cloud registration, loading annotations in OpenLABEL format, evaluation of detection and tracking results, pre-processing steps such as point cloud filtering, and post-processing such as bounding box filtering. 
The statistics from Figures \ref{fig:statistic_plots_01}, \ref{fig:statistic_plots_02}, and \ref{fig:statistic_plots_03} were created using our dataset dev kit.
It also contains modules to convert the labels from OpenLABEL to KITTI or our custom nuScenes format with timestamps instead of tokens and vice versa. This dev kit enables users of popular datasets to migrate their models and make them compatible with our dataset format. We release our dev kit\footnote{\scriptsize{\url{https://github.com/tum-traffic-dataset/tum-traffic-dataset-dev-kit}}} under the MIT license and the dataset under the Creative Commons (CC) BY-NC-ND 4.0 license.
\begin{table}[htbp]
  \caption{Evaluation results ($mAP_{BEV}$ and $mAP_{3D}$) of CoopDet3D on our TUMTraf-V2X test set in south2 FOV.}
  \label{tbl:quantitativeResultsS2}
  \centering
  \resizebox{\columnwidth}{!}{%
  \begin{tabular}{ll|cccccN}
    \hline
    \multicolumn{2}{c|}{\textbf{Config.}} & {$\mathbf{mAP_{BEV}\uparrow}$} & \multicolumn{4}{c}{$\mathbf{mAP_{3D}\uparrow}$} \\
    \textbf{Domain} & \textbf{Modality} & & \textbf{Easy$\uparrow$} & \textbf{Mod.$\uparrow$} & \textbf{Hard$\uparrow$} & \textbf{Avg.$\uparrow$}\\
    \hline
    Vehicle & Camera & 46.83 & 31.47 & 37.82 & 30.77 & 30.36 \\
    Vehicle & LiDAR & 85.33 & 85.22 & 76.86 & 69.04 & 80.11 \\
    Vehicle & Cam+LiDAR & 84.90 & 77.60 & 72.08 & 73.12 & 76.40 \\
    Infra. & Camera & 61.98 & 31.19 & 46.73 & 40.42 & 35.04 \\
    Infra. & LiDAR & 92.86 & 86.17 & 88.07 & 75.73 & 84.88 \\
    Infra. & Cam+LiDAR & 92.92 & 87.99 & \textbf{89.09} & \textbf{81.69} & \underline{87.01} \\
    Coop. & Camera & 68.94 & 45.41 & 42.76 & 57.83 & 45.74 \\
    Coop. & LiDAR & \underline{93.93} & \underline{92.63} & 78.06 & 73.95 & 85.86 \\
    \rowcolor{Gray}
    Coop. & Cam+LiDAR & \textbf{94.22} & \textbf{93.42} & \underline{88.17} & \underline{79.94} & \textbf{90.76} \\
    \hline\\[-8pt]
  \end{tabular}
  }
\end{table}

\begin{table}[htbp]
  \caption{Evaluation results of infrastructure-only CoopDet3D vs. InfraDet3D \cite{zimmer2023infra} on TUMTraf Intersection test set \cite{zimmer2023tumtraf}.}
  \label{tbl:earlyVsDeep}
  \centering
  \resizebox{\columnwidth}{!}{%
  \begin{tabular}{lll|llllN}
    \hline
     & & & \multicolumn{3}{c}{$\mathbf{mAP_{3D}}\uparrow$} \\
    \textbf{Config.} & \textbf{FOV} &\textbf{Mod.} & \textbf{Easy$\uparrow$} & \textbf{Mod.$\uparrow$} & \textbf{Hard$\uparrow$} & \textbf{Avg.$\uparrow$}\\
    \hline
    InfraDet3D & south 1 & LiDAR & 75.81 & 47.66 & \textbf{42.16} & 55.21 \\
    \rowcolor{Gray}
    CoopDet3D & south 1 & LiDAR & \textbf{76.24} & \textbf{48.23} & 35.19 & \textbf{69.47} \\
    InfraDet3D & south 2 & LiDAR & 38.92 & 46.60 & \textbf{43.86} & 43.13 \\
    \rowcolor{Gray}
    CoopDet3D& south 2 & LiDAR & \textbf{74.97} & \textbf{55.55} & 39.96 & \textbf{69.94} \\
    InfraDet3D& south 1 & Cam+LiDAR & 67.08 & 31.38 & 35.17 & 44.55 \\
    \rowcolor{Gray}
    CoopDet3D& south 1 & Cam+LiDAR & \textbf{75.68} & \textbf{45.63} & \textbf{45.63} & \textbf{66.75} \\
    InfraDet3D& south 2 & Cam+LiDAR & 58.38 & 19.73 & 33.08 & 37.06 \\
    \rowcolor{Gray}
    CoopDet3D& south 2 & Cam+LiDAR & \textbf{74.73} & \textbf{53.46} & \textbf{41.96} & \textbf{66.89} \\
    \hline\\[-8pt]
  \end{tabular}
  }
\end{table}

\begin{table}[htbp]
  \caption{Ablation study on cooperative 3D object detection with 11 combinations of camera and LiDAR backbones. The best tradeoff between speed and accuracy is highlighted in gray.}
  \label{tbl:ablationStudies1}
  \centering
  \resizebox{\columnwidth}{!}{%
   \begin{tabular}{lccc}
    \hline
    \textbf{Backbone Configuration} & $\mathbf{mAP_{BEV}\uparrow}$ & \textbf{FPS$\uparrow$} & \textbf{VRAM$\downarrow$}\\
    \hline
    VoxelNet non-deterministic + SwinT & 93.47 & 6.30 & 6.69 GiB \\
    VoxelNet non-deterministic + YOLOv8 s & 92.94 & 7.24 & 6.39 GiB \\
    VoxelNet Torchsparse + SwinT & 93.51 & 8.84 & \underline{4.61 GiB} \\
    VoxelNet Torchsparse + YOLOv8 s & 92.94 & 10.66 & \textbf{4.28 GiB} \\
    VoxelNet Torchsparse + YOLOv8 s (retrained) & \underline{94.31} & 10.66 & \textbf{4.28 GiB} \\
    PointPillars 512 + Swin T & \textbf{94.43} & 9.00 & 4.94 GiB \\
    PointPillars 512 + YOLOv8 s & 94.27 & \underline{11.14} & 4.63 GiB \\
    PointPillars 512 + YOLOv8 s (retrained) & 94.25 & \underline{11.14} & 4.63 GiB \\
    PointPillars 512\_2x + Swin T & 92.79 & 9.06 & 4.94 GiB \\
    PointPillars 512\_2x + YOLOv8 s & 94.16 & \textbf{11.20} & 4.63 GiB \\
    \rowcolor{Gray}
    PointPillars 512\_2x + YOLOv8 s (retrained) & 94.22 & \textbf{11.20} & 4.63 GiB \\
    \hline
  \end{tabular}
  }
\end{table}


\setlength{\fboxsep}{0pt}%
\setlength{\fboxrule}{1pt}%
\begin{figure*}[h!]
\centering
\minipage{0.25\textwidth}
  \fbox{\includegraphics[width=\linewidth,trim={0 0 2.8cm 0},clip]{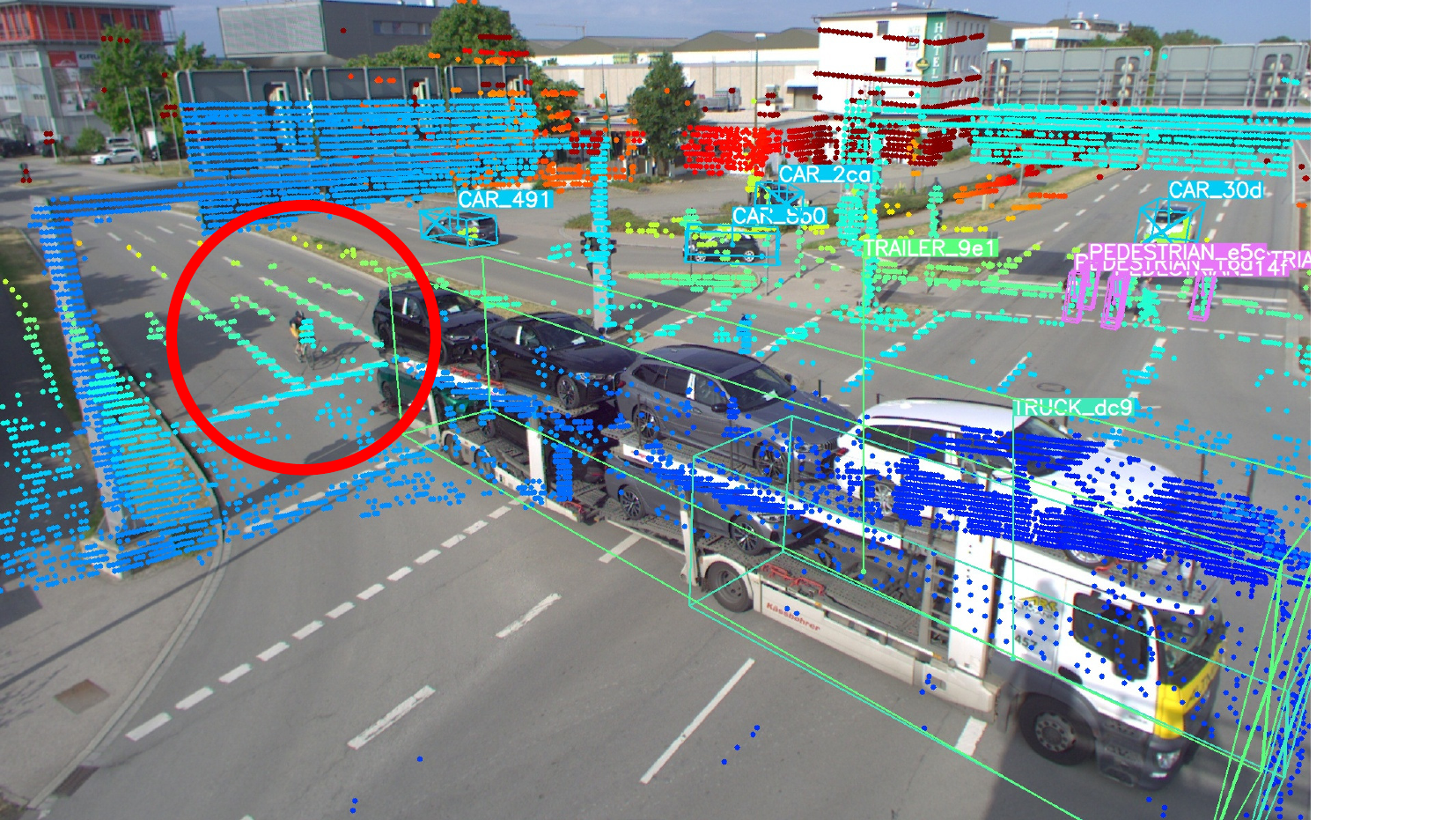}}
\endminipage
\minipage{0.25\textwidth}
  \fbox{\includegraphics[width=\linewidth]{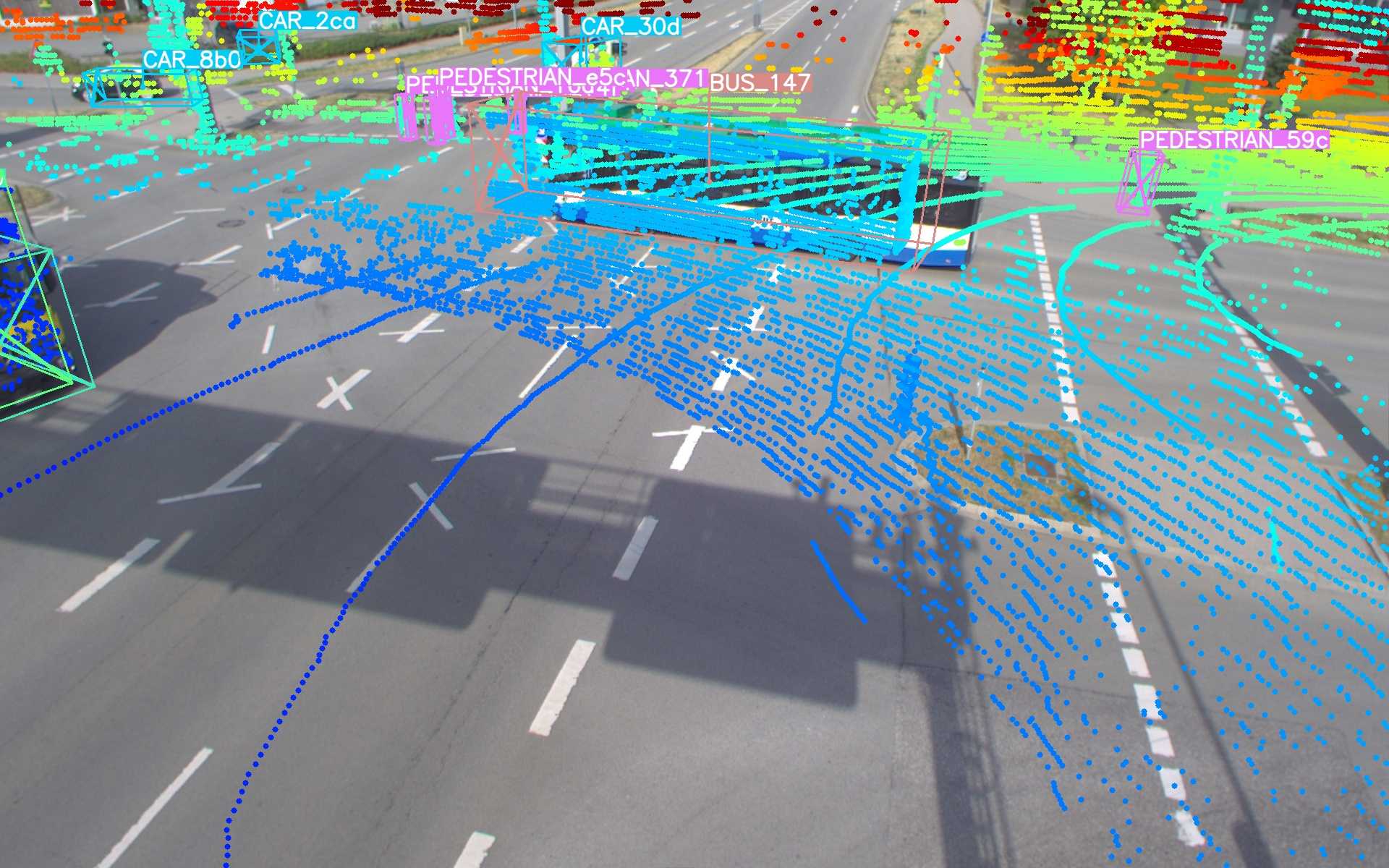}}
\endminipage
\minipage{0.25\textwidth}%
  \fbox{\includegraphics[width=\linewidth]{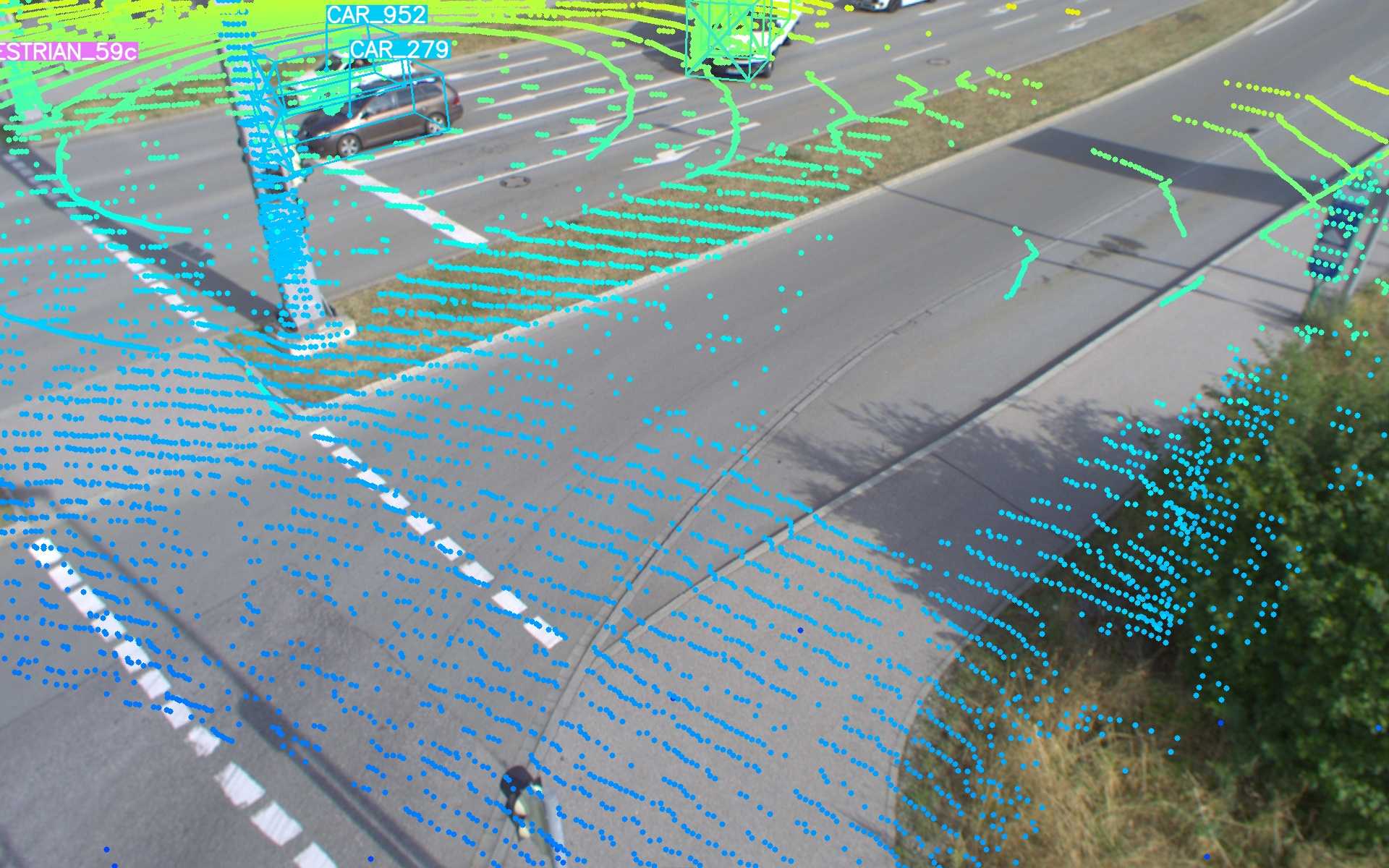}}
\endminipage
\minipage{0.25\textwidth}
  \fbox{\includegraphics[width=\linewidth,trim={0 0 2.8cm 0},clip]{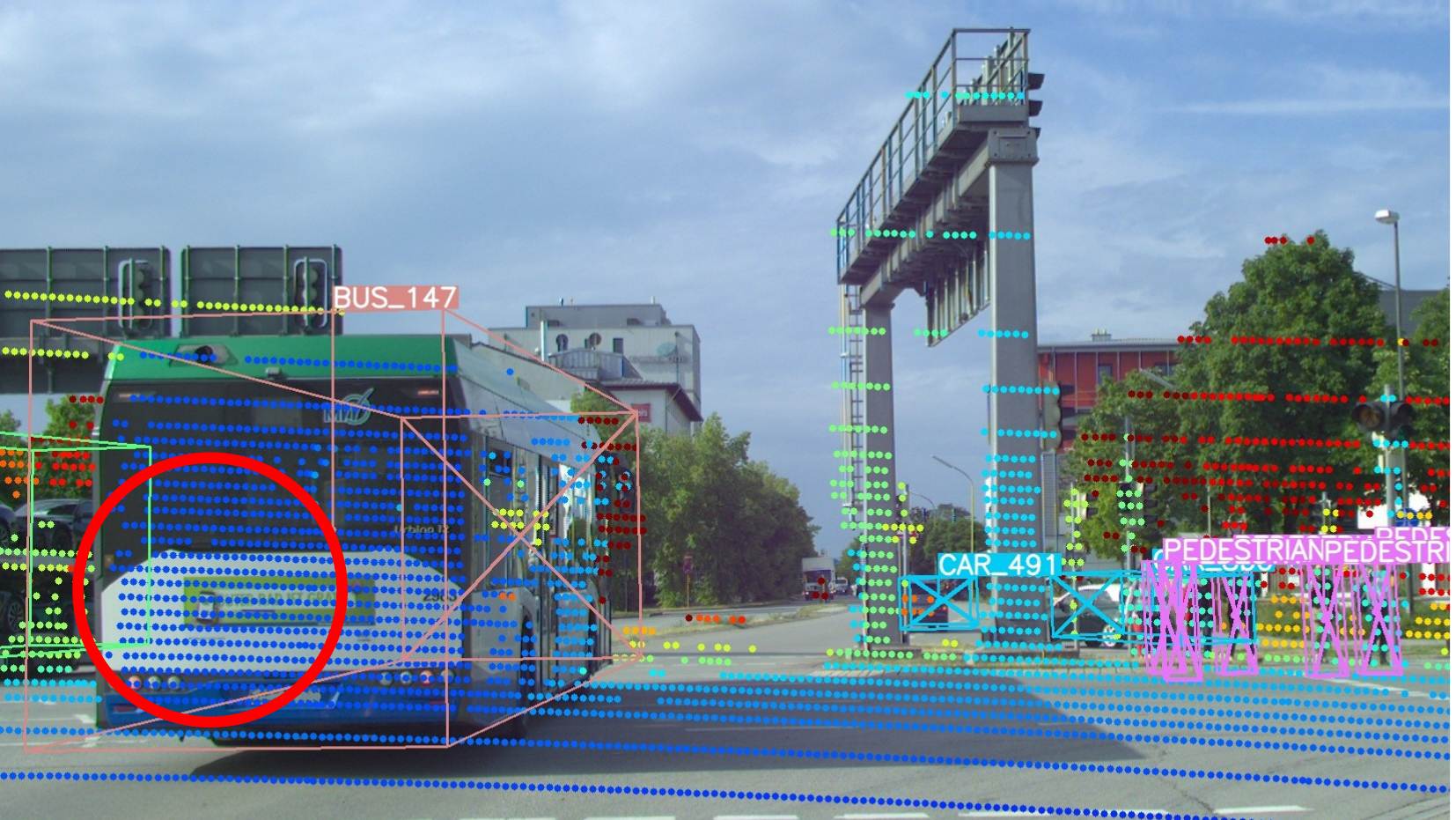}}
\endminipage\\
\vspace{-0.07cm}
\minipage{0.25\textwidth}
  \fbox{\includegraphics[width=\linewidth,trim={0 0 2.8cm 0},clip]{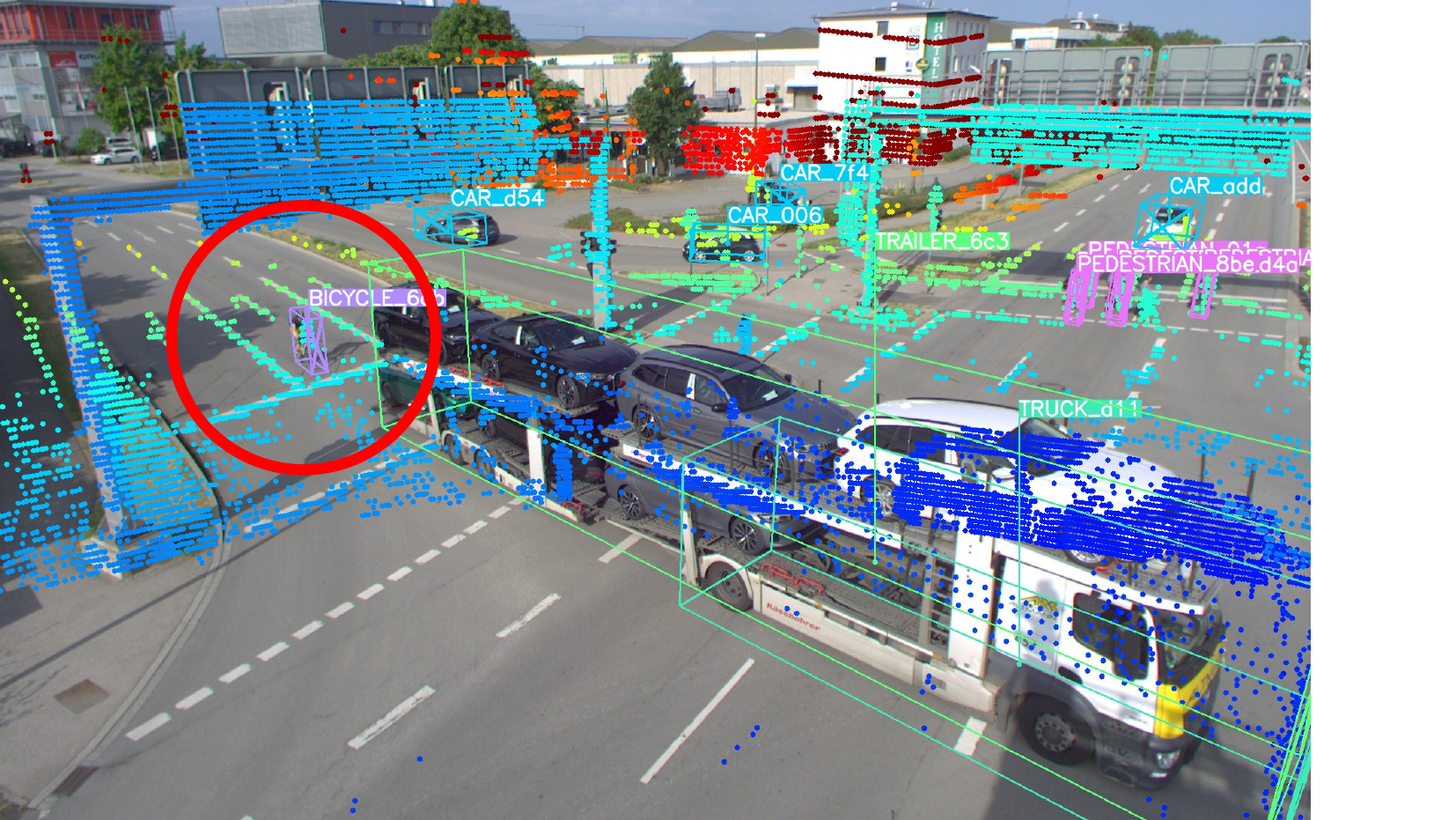}}
\endminipage
\minipage{0.25\textwidth}%
  \fbox{\includegraphics[width=\linewidth]{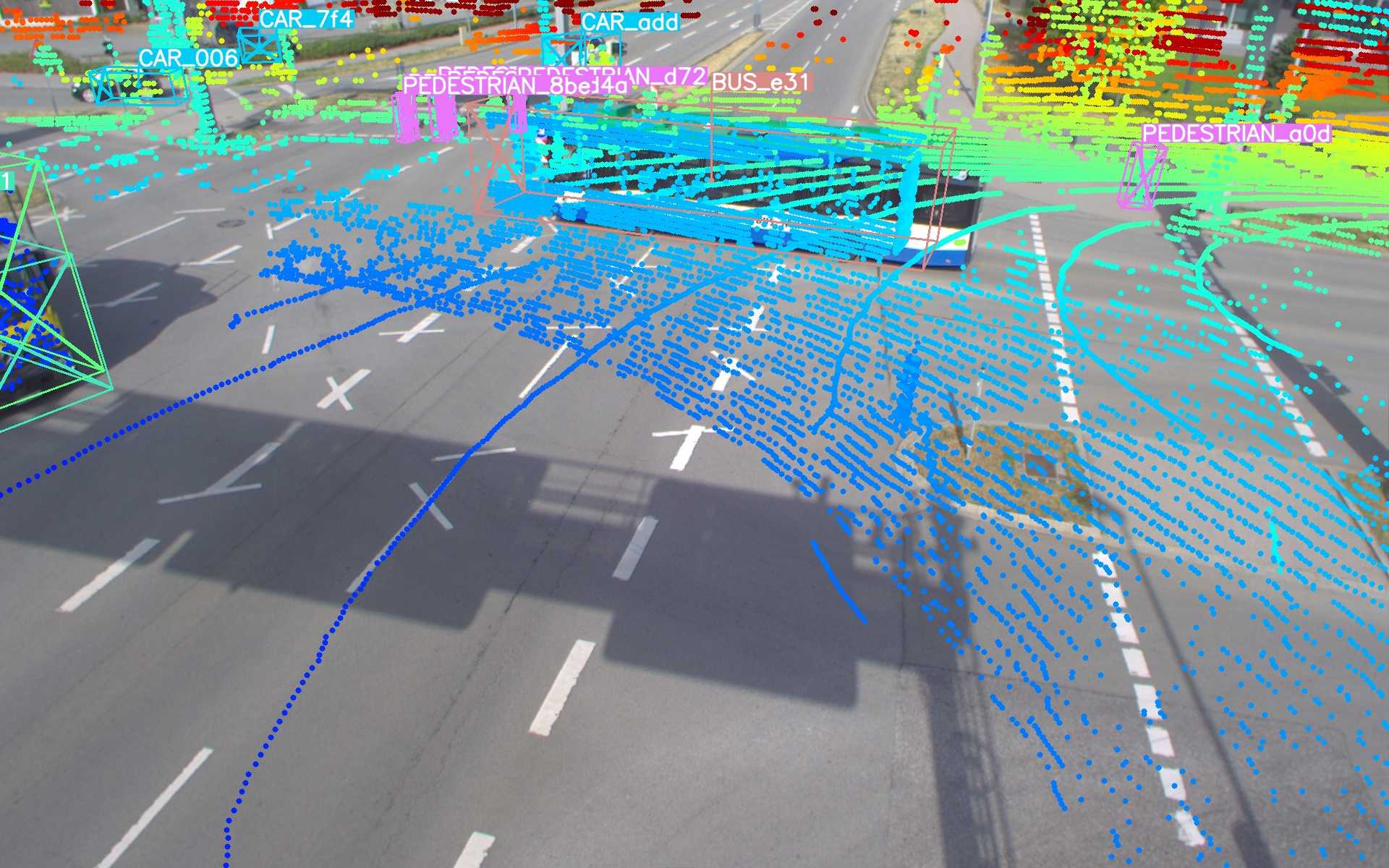}}
\endminipage
\minipage{0.25\textwidth}%
  \fbox{\includegraphics[width=\linewidth]{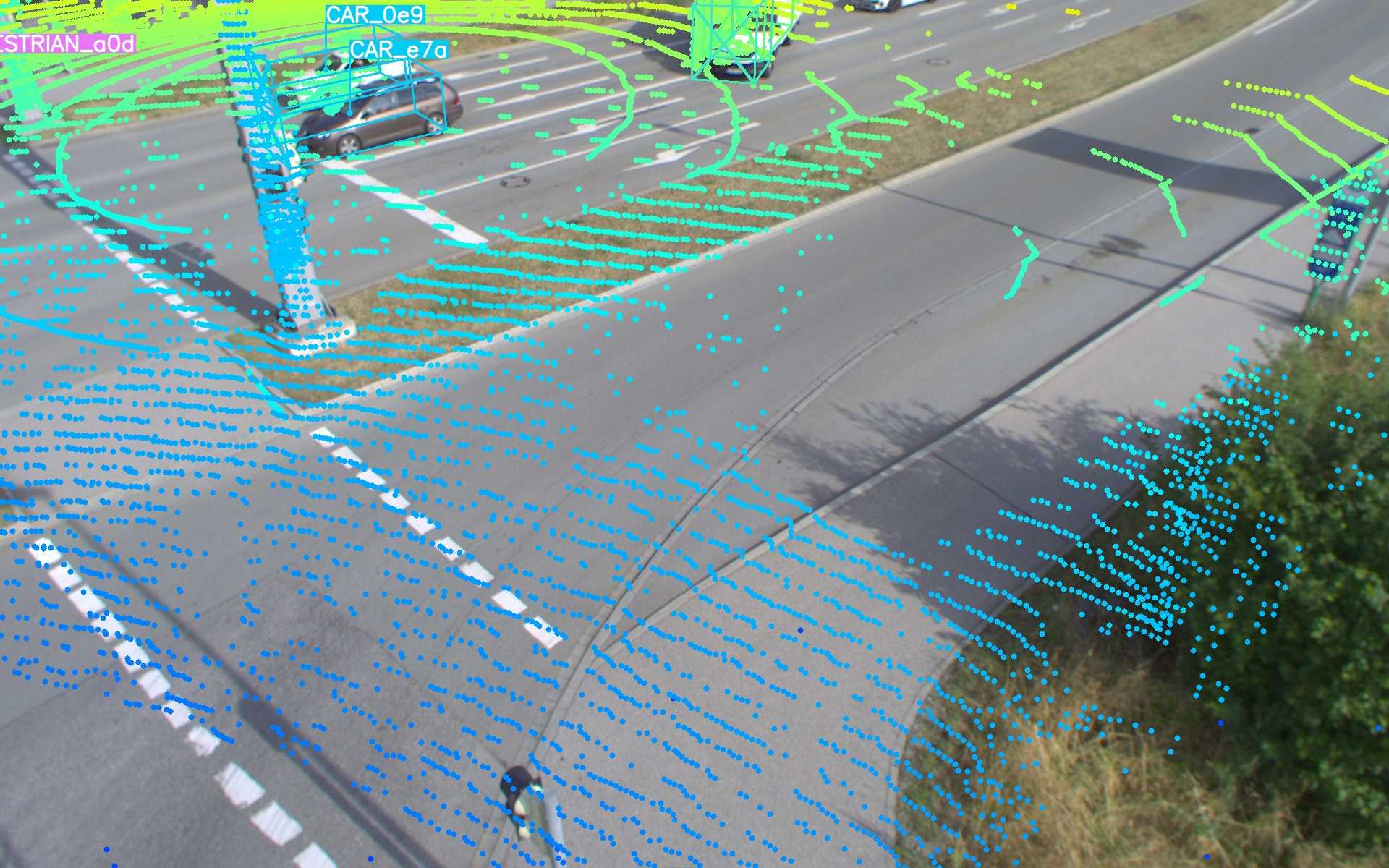}}
\endminipage
\minipage{0.25\textwidth}%
  \fbox{\includegraphics[width=\linewidth,trim={0 0 2.8cm 0},clip]{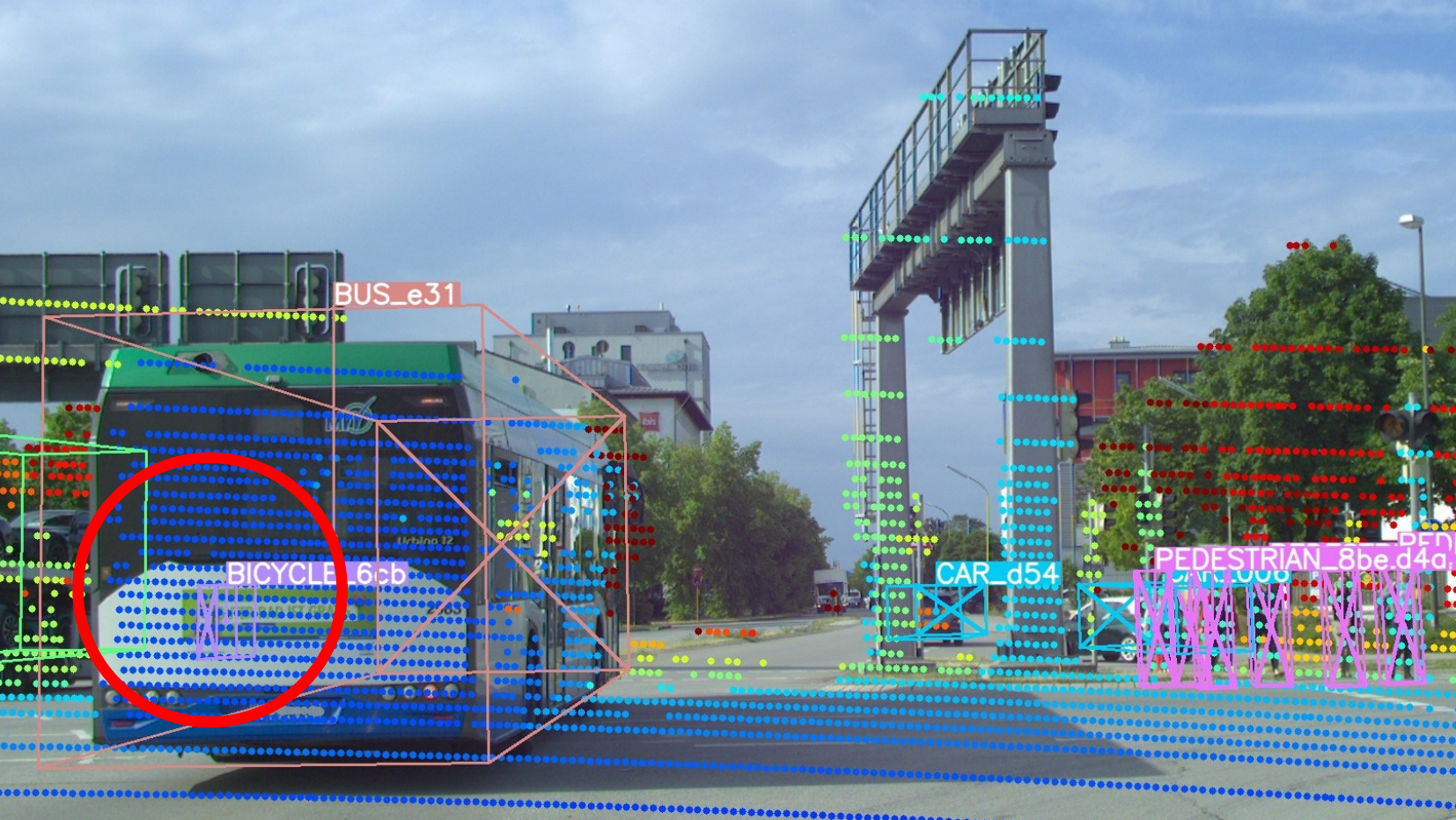}}
\endminipage
\caption{Qualitative results on the TUMTraf-V2X Cooperative Perception test set. The first row shows the inference results of the onboard (vehicle-only) camera-LiDAR fusion with 23 detected objects. In the second row, the results of the cooperative vehicle-infrastructure camera-LiDAR fusion are visualized. Here, the 25 traffic participants could be detected with the support of roadside sensors. }
\label{fig:qualitative_results1} 
\end{figure*}

\section{Benchmark}
\label{sec:benchmark}
We propose CoopDet3D, an extension of BEVFusion \cite{liu2023bevfusion} and PillarGrid \cite{bai2022pillargrid} for deep cooperative multi-modal 3D object detection and benchmark it on our dataset.

\subsection{Evaluation metrics}
The accuracy is measured in terms of the mean average precision (mAP). Two types of mAP measures are used: BEV mAP considers the BEV center distance, and the results are obtained using the same evaluation methodology used in BEVFusion, which in turn uses the evaluation protocol of nuScenes \cite{caesar2020nuscenes}. Similarly, the 3D mAP measure considers the intersection in 3D, and the results are obtained using the evaluation script of our TUM Traffic dataset Devkit. The runtime is evaluated using frames per second (FPS) as the metric and the results were obtained by measuring the time needed by the model to run one full inference, including data preprocessing and voxelization. The first five iterations are skipped as a warmup since they are usually considerably slower than the average. Finally, the complexity of the model is measured in terms of the maximal VRAM usage across all GPUs during training and testing. 


\subsection{CoopDet3D model}
Our \textit{CoopDet3D} uses a BEVFusion-based backbone for camera-LiDAR fusion on the vehicle and the infrastructure sides separately to obtain the vehicle and infrastructure features. The best backbone for image and point cloud feature extraction was chosen through multiple ablation studies. Then, inspired by the method proposed by PillarGrid \cite{bai2022pillargrid}, an element-wise max-pooling operation is proposed to fuse the resulting fused camera-LiDAR features of vehicle and infrastructure together. Finally, the detection head from BEVFusion is used for 3D detection from the fused feature. The architecture of \textit{CoopDet3D} is shown in Fig. \ref{fig:coopdet3d}.

First, we disable the camera feature extraction nodes and train the LiDAR-only model for 20 epochs. Then, we use pre-trained weights for the cooperative model and fine-tune the entire model for eight further epochs. Hyperparameter tuning revealed that the default hyperparameters of BEVFusion \cite{liu2023bevfusion} gave the best results, and such were not modified. The preprocessing steps are also the same as the BEVFusion, but we change the point cloud range to $[-75, 75]$ in the x- and y-scale and $[-8, 0]$ in the z-scale since the dataset used in this case is different. Furthermore, we use 3x NVIDIA RTX 3090 GPUs with 24 GB VRAM for training and a single GPU for evaluation. We open-source our model and provide pre-trained weights\footnote{\scriptsize{\url{https://github.com/tum-traffic-dataset/coopdet3d}}}.

\subsection{Experiments and ablation studies}

The objective of these experiments is to highlight the importance of our V2X multi-viewpoint dataset as opposed to single-viewpoint datasets. As such, we conduct multiple experiments and ablation studies with data obtained from each viewpoint and compare the results on the proposed model.

\vspace{.5\baselineskip}
\noindent
\textbf{Cooperative perception compared to single-viewpoint}

We conduct multiple experiments with all possible combinations of a) viewpoints: vehicle-only, infrastructure-only, cooperative, and b) modalities: camera-only, LiDAR-only, and camera-LiDAR fusion. Table \ref{tbl:quantitativeResultsS2} shows the mAP achieved by CoopDet3D for each of these combinations.

We observe that the results follow a general pattern of cooperative performance being better than infrastructure-only, which is, in turn, better than vehicle-only. Furthermore, fusion models perform better than LiDAR-only models, which in turn are better than camera-only models. 
Figure \ref{fig:qualitative_results1} shows qualitative results between our vehicle-only camera-LiDAR fusion model and our cooperative vehicle-infrastructure camera-LiDAR fusion model. Again, we observe from these samples that the cooperative perception model is able to detect 25 traffic participants, whereas the vehicle-only model is only able to detect 23 objects due to occlusions and a limited field of view.

\vspace{.5\baselineskip}
\noindent
\textbf{Deep fusion compared to late fusion}

\begin{figure}[t!]
\centering
\minipage{0.23\textwidth}
  \includegraphics[width=\linewidth,frame]{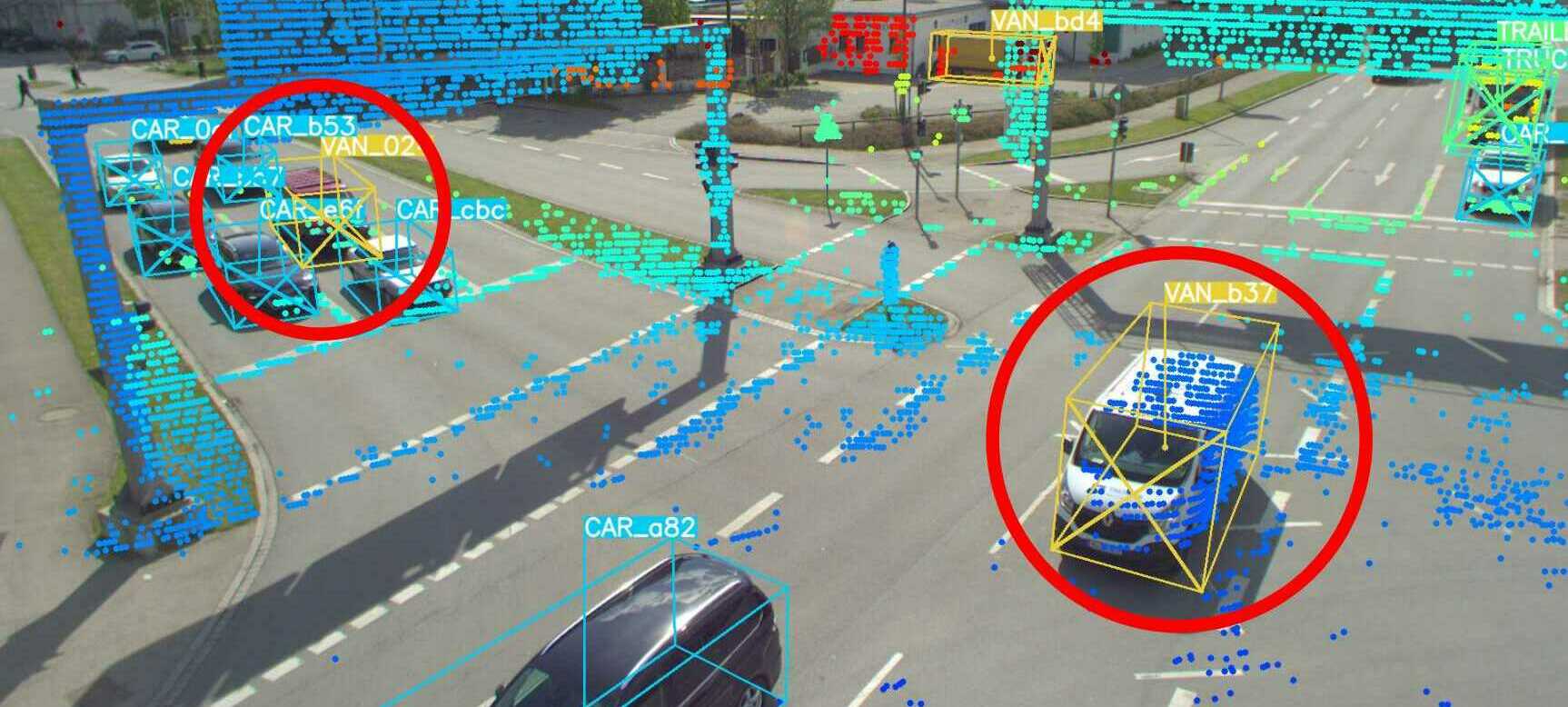}
\endminipage
\minipage{0.23\textwidth}
  \includegraphics[width=\linewidth,frame]{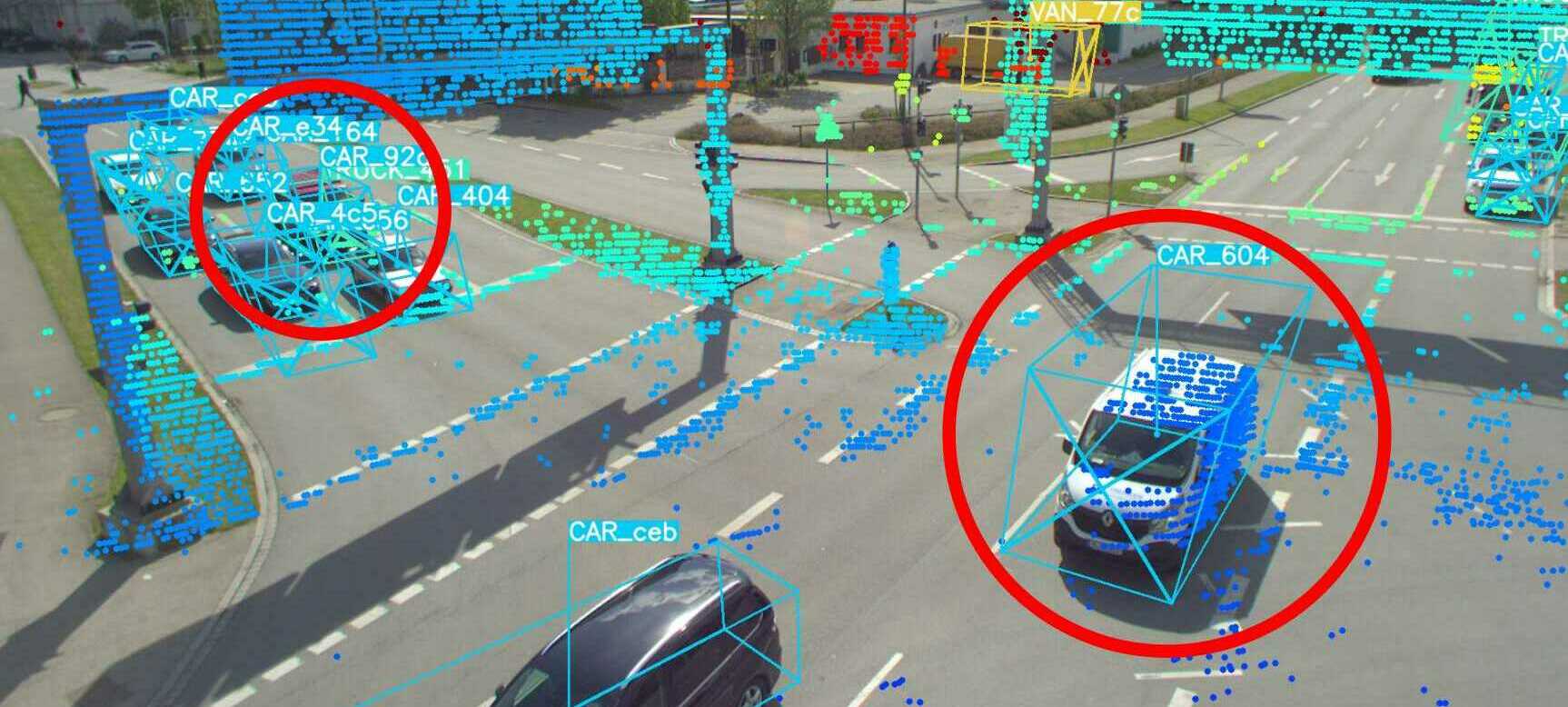}
\endminipage\\
\vspace{-0.04cm}
\minipage{0.23\textwidth}%
  \includegraphics[width=\linewidth,frame]{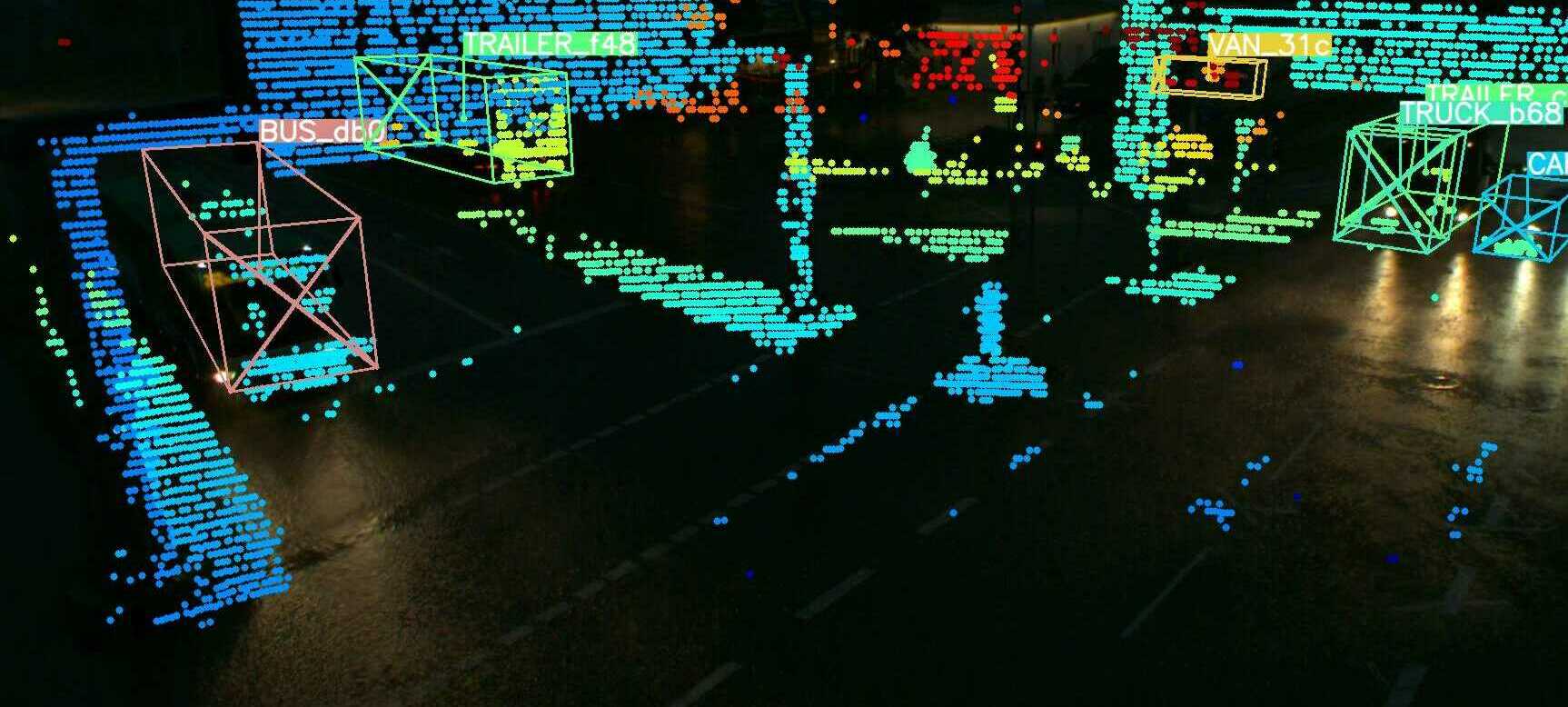}
\endminipage
\minipage{0.23\textwidth}
  \includegraphics[width=\linewidth,frame]{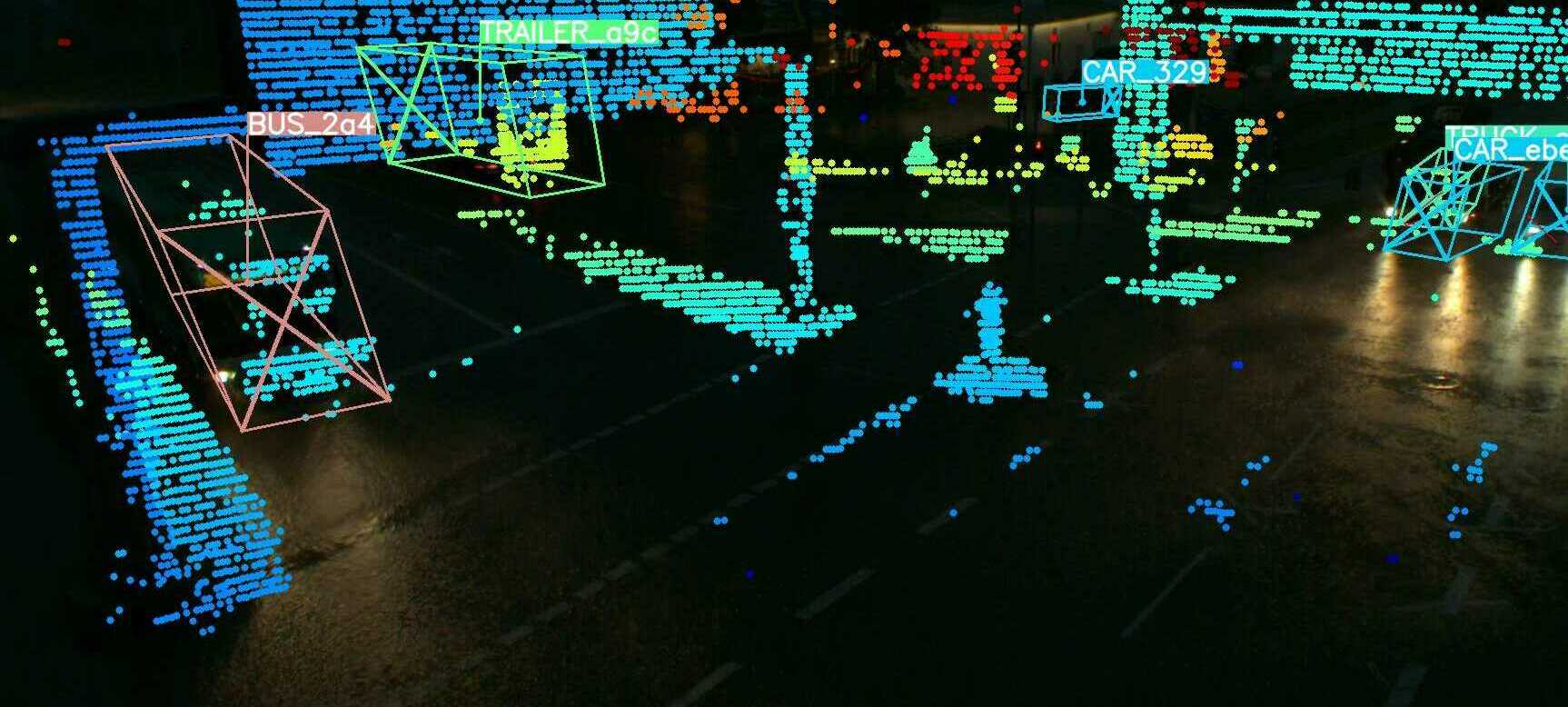}
\endminipage
\caption{Qualitative results of our CoopDet3D (left) and the InfraDet3D (right) model on the TUMTraf Intersection test set during day and nighttime. Detected objects marked with a red circle were classified correctly by CoopDet3D.}
\label{fig:qualitative_results2} 
\end{figure}

Next, we compare our proposed CoopDet3D model to the current SOTA camera-LiDAR fusion method on the TUMTraf Intersection test set \cite{zimmer2023tumtraf}, InfraDet3D \cite{zimmer2023infra}. The proposed method uses deep fusion, whereas the InfraDet3D method is a late fusion method. Table \ref{tbl:earlyVsDeep} shows the performance of our model against InfraDet3D, and the results show that the proposed deep fusion method outperforms the SOTA late fusion model in all metrics, except in the hard difficulty in LiDAR-only mode. Furthermore, Figure \ref{fig:qualitative_results2} shows two sample images taken during day and nighttime, wherein deep fusion again outperforms late fusion. South 1 and South 2 refer to sensors covering different FOVs.

We note that these experiments were conducted in an offline setting, disregarding other considerations for simplicity. However, when deploying it in real life, factors such as the transmission bandwidth should also be considered. Since we observed that deep feature fusion generally leads to higher efficacy, the V2I transmissions should contain these features instead of infrastructure bounding boxes.

\vspace{.5\baselineskip}

\noindent
\textbf{Model performance with different backbones}

As an ablation study, we present the results of the experiments to find the best backbone and model configuration for the cooperative camera-LiDAR fusion model. For the camera backbone, SwinT \cite{liu2021swin} and MMYOLO's \cite{mmyolo2022} implementation of YOLOv8 \cite{glenn2023ultralytics} were considered. For the LiDAR backbone, VoxelNet \cite{zhou2018voxelnet} and PointPillars \cite{lang2019pointpillars} were considered. In addition, VoxelNet was implemented with two different backends, namely SPConv v2 and Torchsparse \cite{tangandyang2023torchsparse++}. For PointPillars, two grid sizes are considered $512 \times 512$ for both train and test grids (PointPillars 512) and $512 \times 512$ train grid with $1024 \times 1024$ test grid (PointPillars 512\_2x). The results of these experiments are shown in Table \ref{tbl:ablationStudies1}.

The results show that only models that use any combination of VoxelNet Torchsparse, both PointPillars variants, and YOLOv8 are able to run above 10 FPS. From these configurations, we choose PointPillars 512\_2x with YOLOv8 as the best configuration for all the above experiments as it achieves the best results across all the ablation studies. This is a promising result since we also know that this backbone configuration is able to run in real-time (11.2 FPS) on an RTX 3090 without using TensorRT acceleration.
 


\setlength{\fboxsep}{0pt}%
\setlength{\fboxrule}{1pt}%

An interesting observation is that utilizing pre-trained weights for transfer learning of YOLOv8 is not always beneficial, as the results from PointPillars 512 + YOLOv8 s show. This is likely because the pre-trained weights were from MS COCO \cite{lin2014microsoft}, and they have a very different data domain compared to our dataset. Since MS COCO is also much larger than our dataset in terms of camera images, retraining harms the performance of the model slightly.

In terms of efficiency, the goal of these experiments was to verify that the proposed CoopDet3D model with the best configuration provides the highest accuracy while also being able to run in real-time (minimum of 10 Hz). Furthermore, it should also be feasible to train the model on a high-performance GPU and perform inference on a mid-range consumer GPU deployable on an edge device. The results concerning the VRAM usage during inference show that the complexity of the model makes this feasible.

\section{Conclusion and future work}
\label{sec:conclusion}
This work proposes the TUMTraf-V2X dataset, a multi-modal multi-view V2X dataset for cooperative 3D object detection and tracking. Our dataset focuses on challenging traffic scenarios at an intersection and provides views from the infrastructure and the ego vehicle. To benchmark the dataset, we propose CoopDet3D -- a baseline model for cooperative perception. Experiments show that cooperative fusion leads to higher efficacy than its unimodal and single-view camera-LiDAR fusion counterparts. Furthermore, cooperative fusion leads to an improvement of +14.3 3D mAP compared to vehicle-only perception, highlighting the need for V2X datasets. Finally, we provide our \textit{3D BAT} v24.3.2 labeling tool and dev kit to load, parse, and visualize the dataset. It also includes modules for pre- and postprocessing and evaluation. Future efforts will integrate this platform into online environments, enabling a broader range of infrastructure-based, real-time perception applications.

\section*{Acknowledgment}
This research was supported by the Federal Ministry of Education and Research in Germany within the $\text{\textit{AUTOtech.agil}}$ project, Grant Number: 01IS22088U.

\clearpage
\setcounter{page}{1}

\setlength{\fboxsep}{-1pt}%
\setlength{\fboxrule}{1pt}%

\twocolumn[{%
\renewcommand\twocolumn[1][]{#1}%
\maketitlesupplementary
\vspace{0.5cm}
\small{\url{https://tum-traffic-dataset.github.io/tumtraf-v2x}}
\vspace{0.5cm}
\begin{center}
    \centering
    \captionsetup{type=figure}
    \fbox{\includegraphics[width=0.99\textwidth]{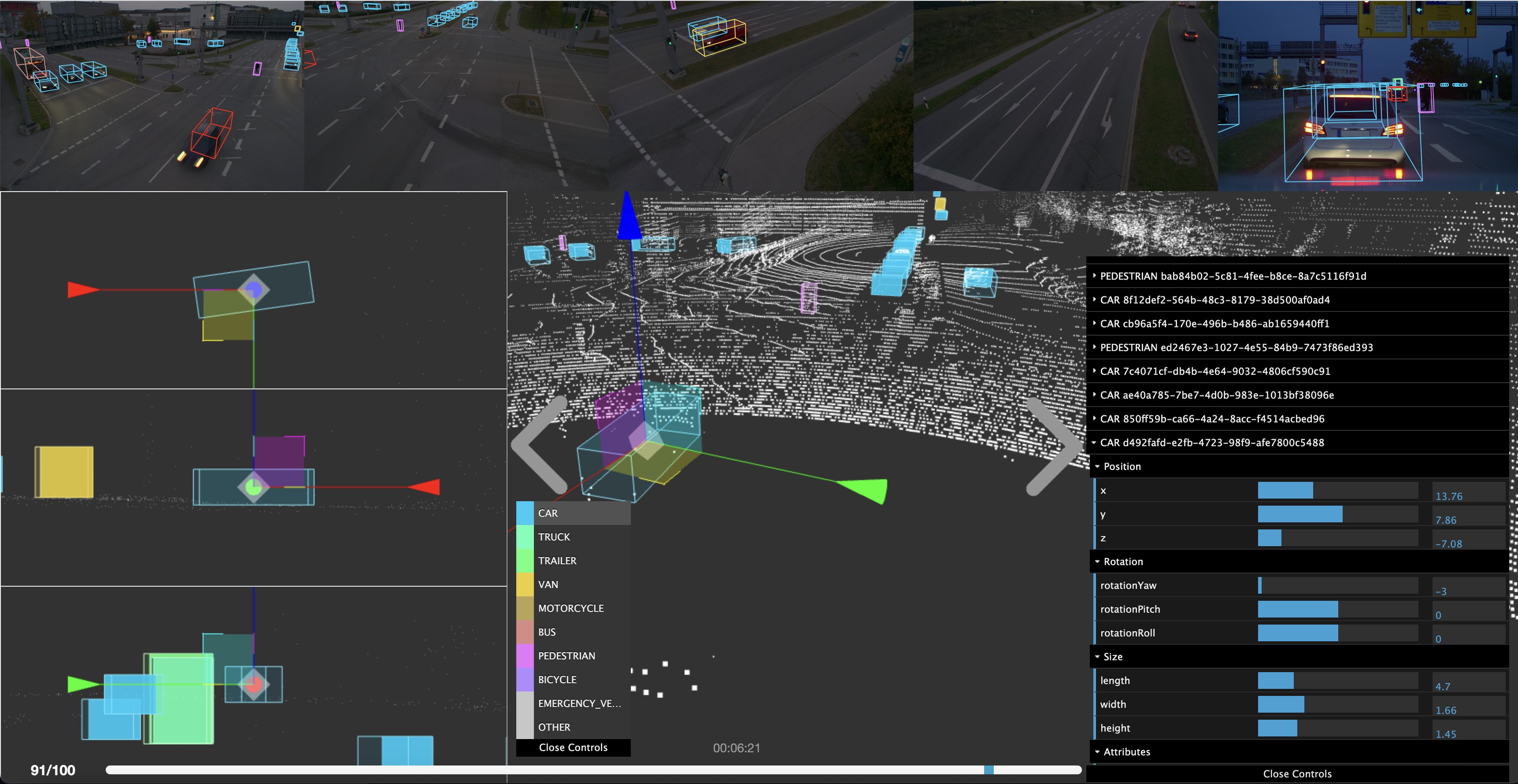}}
    \captionof{figure}{Visualization of our web-based \textit{3D BAT} (v24.3.2) labeling tool. It shows the registered point cloud and five camera images on the top. On the left side, there are three helper views: top-down view, side view, and front view. The control pane on the right side contains a download button, an undo button, a drop-down menu to switch between a perspective (3D) and orthographic (BEV) view, a slider to change the point size, a drop-down menu to choose the dataset and sequence, some checkboxes for filtering the scene and hiding other annotations, a button to copy labels to the next frame, an auto-label button, a button for active learning, an interpolation button, and a reset button. In the bottom right corner, all labeled objects are displayed. Each object can be translated, scaled, and rotated using sliders or keyboard shortcuts. The scaling of an object will change the dimensions in all frames.}
    \label{fig:proanno}
\end{center}%
\vspace{0.1cm}
}]

\section*{Contents}
\vspace{-0.3cm}
{
  \hypersetup{linkcolor=black}
  \appendix 
  \startcontents[sections]
  \printcontents[sections]{l}{1}{\setcounter{tocdepth}{2}}
}

\section{Task Definition}
\subsection{Detection and tracking}
Detection and tracking are two crucial perception tasks for autonomous driving. In 3D object detection, the surrounding objects are located with their 3D position, dimensions (length, width, height), and rotation at each timestamp. In multi-object tracking (MOT), the correspondences between different objects are found across timestamps. Objects are associated temporally and given a unique track ID. The final detection and tracking output is a series of associated 3D boxes in each frame. \\
\subsection{Cooperative fusion}
The cooperative fusion approach combines data from several sensors from different perspectives to optimize the detection and tracking performance. Data from roadside cameras and LiDARs is fused with onboard camera and LiDAR sensor data to prevent occlusions.

\newcolumntype{Y}{>{\centering\arraybackslash}X}
\begin{table*}[t!]
\caption{Comparison of 3D annotation tools. \tikzcircle[fill=green!20]{3pt} Feature provided \tikzcircle[fill=yellow!20]{3pt} Feature unknown \tikzcircle[fill=red!20]{3pt} Feature not provided}
\footnotesize
\label{tab:annotation_tools}
\centering
\resizebox{\linewidth}{!}{%
\begin{tabularx}{\linewidth}{l|YYYYYYYYY}
\hline
Tool & 3D BAT \cite{zimmer20193d} & LATTE \cite{wang2019latte}& SAnE \cite{arief2020sane} &SUSTech POINTS\cite{sustech} & Label Cloud\cite{sager2021labelcloud} & ReBound \cite{chen2023rebound} &  PointCloud Lab\cite{PointCloudLab} & Xtreme1 \cite{Xtreme1} & \textbf{3D BAT (Ours)$^*$}\\
\hline
Year & \cellcolor{yellow!10}2019 & \cellcolor{yellow!10}2020 & \cellcolor{yellow!10}2020 & \cellcolor{yellow!10}2020 & \cellcolor{yellow!10}2021 & \cellcolor{green!10}2023 &  \cellcolor{green!10}2023 & \cellcolor{green!10}2023 & \cellcolor{green!10}2023 \\
Support V2X &\cellcolor{red!10}- & \cellcolor{red!10}- & \cellcolor{red!10}- & \cellcolor{red!10}- & \cellcolor{red!10}- & \cellcolor{red!10}- &  \cellcolor{red!10}- & \cellcolor{green!10}\checkmark & \cellcolor{green!10}\checkmark \\
2D/3D cam.+LiDAR fusion &\cellcolor{green!10}\checkmark & \cellcolor{green!10}\checkmark & \cellcolor{red!10}- & \cellcolor{green!10}\checkmark & \cellcolor{green!10}\checkmark & \cellcolor{green!10}\checkmark&  \cellcolor{red!10}-& \cellcolor{green!10}\checkmark & \cellcolor{green!10}\checkmark\\
AI assisted labeling & \cellcolor{green!10}\checkmark& \cellcolor{green!10}\checkmark& \cellcolor{green!10}\checkmark & \cellcolor{green!10}\checkmark & \cellcolor{red!10}- & \cellcolor{green!10}\checkmark & \cellcolor{green!10}\checkmark & \cellcolor{green!10}\checkmark & \cellcolor{green!10}\checkmark\\
Batch-mode editing &\cellcolor{red!10}- & \cellcolor{red!10}-  & \cellcolor{red!10}- & \cellcolor{green!10}\checkmark& \cellcolor{red!10}- & \cellcolor{red!10}-  & \cellcolor{red!10}- & \cellcolor{green!10}\checkmark& \cellcolor{green!10}\checkmark \\
Interpolation mode &\cellcolor{red!10}- & \cellcolor{red!10}-  & \cellcolor{red!10}- & \cellcolor{red!10}\checkmark& \cellcolor{red!10}- & \cellcolor{red!10}-  & \cellcolor{red!10}- & \cellcolor{red!10}-& \cellcolor{green!10}\checkmark \\
Active learning support &\cellcolor{red!10}- & \cellcolor{red!10}-  & \cellcolor{red!10}- & \cellcolor{red!10}\checkmark& \cellcolor{red!10}- & \cellcolor{red!10}-  & \cellcolor{red!10}- & \cellcolor{red!10}-& \cellcolor{green!10}\checkmark \\
Label custom attributes &\cellcolor{red!10}- & \cellcolor{red!10}- & \cellcolor{red!10}-& \cellcolor{green!10}\checkmark & \cellcolor{red!10}- & \cellcolor{green!10}\checkmark & \cellcolor{yellow!10}(?) & \cellcolor{green!10}\checkmark& \cellcolor{green!10}\checkmark \\
3D tracking & \cellcolor{green!10}\checkmark& \cellcolor{green!10}\checkmark & \cellcolor{green!10}\checkmark & \cellcolor{green!10}\checkmark & \cellcolor{red!10}- & \cellcolor{red!10}- &  \cellcolor{green!10}\checkmark & \cellcolor{red!10}- & \cellcolor{green!10}\checkmark \\
Support multiple cameras & \cellcolor{green!10}\checkmark& \cellcolor{red!10}-& \cellcolor{red!10}- & \cellcolor{green!10}\checkmark & \cellcolor{red!10}- & \cellcolor{green!10}\checkmark & \cellcolor{red!10}- & \cellcolor{green!10}\checkmark & \cellcolor{green!10}\checkmark\\
HD Maps & \cellcolor{red!10}-& \cellcolor{red!10}-& \cellcolor{red!10}- & \cellcolor{green!10}\checkmark  & \cellcolor{red!10}- & \cellcolor{red!10}- &   \cellcolor{red!10}-  & \cellcolor{green!10}\checkmark & \cellcolor{green!10}\checkmark \\
Web-based & \cellcolor{green!10}\checkmark& \cellcolor{red!10}-& \cellcolor{red!10}-&  \cellcolor{green!10}\checkmark& \cellcolor{red!10}-& \cellcolor{red!10}-& \cellcolor{red!10}-& \cellcolor{green!10}\checkmark & \cellcolor{green!10}\checkmark\\
3D navigation & \cellcolor{green!10}\checkmark & \cellcolor{green!10}\checkmark & \cellcolor{green!10}\checkmark& \cellcolor{green!10}\checkmark & \cellcolor{red!10}- & \cellcolor{green!10}\checkmark & \cellcolor{green!10}\checkmark & \cellcolor{green!10}\checkmark & \cellcolor{green!10}\checkmark \\
3D transform controls & \cellcolor{green!10}\checkmark& \cellcolor{green!10}\checkmark & \cellcolor{green!10}\checkmark& \cellcolor{green!10}\checkmark & \cellcolor{red!10}-& \cellcolor{green!10}\checkmark& \cellcolor{green!10}\checkmark& \cellcolor{green!10}\checkmark& \cellcolor{green!10}\checkmark \\
Side views (top/front/side)& \cellcolor{green!10}\checkmark& \cellcolor{red!10}-& \cellcolor{green!10}\checkmark& \cellcolor{green!10}\checkmark  & \cellcolor{red!10}-& \cellcolor{green!10}\checkmark  & \cellcolor{red!10}- & \cellcolor{green!10}\checkmark & \cellcolor{green!10}\checkmark\\
Perspective view editing & \cellcolor{green!10}\checkmark& \cellcolor{green!10}\checkmark & \cellcolor{green!10}\checkmark & \cellcolor{green!10}\checkmark & \cellcolor{green!10}\checkmark& \cellcolor{green!10}\checkmark & \cellcolor{green!10}\checkmark & \cellcolor{green!10}\checkmark & \cellcolor{green!10}\checkmark\\
Orthographic view editing & \cellcolor{green!10}\checkmark& \cellcolor{green!10}\checkmark & \cellcolor{green!10}\checkmark & \cellcolor{red!10}- & \cellcolor{red!10}-& \cellcolor{green!10}\checkmark &  \cellcolor{green!10}\checkmark& \cellcolor{green!10}\checkmark & \cellcolor{green!10}\checkmark\\
Object coloring & \cellcolor{green!10}\checkmark& \cellcolor{red!10}-& \cellcolor{green!10}\checkmark& \cellcolor{green!10}\checkmark &\cellcolor{red!10}- &\cellcolor{green!10}\checkmark & \cellcolor{green!10}\checkmark & \cellcolor{green!10}\checkmark & \cellcolor{green!10}\checkmark\\
Focus mode& \cellcolor{red!10}-& \cellcolor{red!10}- & \cellcolor{red!10}- & \cellcolor{green!10}\checkmark & \cellcolor{red!10}- & \cellcolor{red!10}- & \cellcolor{green!10}\checkmark& \cellcolor{green!10}\checkmark& \cellcolor{green!10}\checkmark \\
Support JPG/PNG files &\cellcolor{red!10}- & \cellcolor{yellow!10}(?)& \cellcolor{yellow!10}(?) & \cellcolor{green!10}\checkmark& \cellcolor{red!10}- & \cellcolor{red!10}-& \cellcolor{red!10}- & \cellcolor{yellow!10}(?) & \cellcolor{green!10}\checkmark\\
Keyboard-only support& \cellcolor{red!10}-& \cellcolor{red!10}- & \cellcolor{red!10}- & \cellcolor{red!10}- & \cellcolor{red!10}- & \cellcolor{red!10}- & \cellcolor{red!10}-& \cellcolor{red!10}-& \cellcolor{green!10}\checkmark \\
Offline annotation support& \cellcolor{red!10}-& \cellcolor{red!10}- & \cellcolor{red!10}- & \cellcolor{red!10}- & \cellcolor{red!10}- & \cellcolor{red!10}- & \cellcolor{red!10}-& \cellcolor{red!10}-& \cellcolor{green!10}\checkmark \\
OpenLABEL support & \cellcolor{red!10}-& \cellcolor{red!10}- & \cellcolor{red!10}- & \cellcolor{red!10}-&  \cellcolor{red!10}-& \cellcolor{red!10}-& \cellcolor{red!10}-& \cellcolor{red!10}- & \cellcolor{green!10}\checkmark                                                 \\
Open-source & \cellcolor{green!10}\checkmark& \cellcolor{green!10}\checkmark & \cellcolor{green!10}\checkmark & \cellcolor{green!10}\checkmark        & \cellcolor{green!10}\checkmark & \cellcolor{green!10}\checkmark       &  \cellcolor{red!10}-  & \cellcolor{green!10}\checkmark & \cellcolor{green!10}\checkmark \\
Github stars & \cellcolor{green!10}529& \cellcolor{yellow!10}374& \cellcolor{red!10}62 & \cellcolor{green!10}670        & \cellcolor{yellow!10}461 & \cellcolor{red!10}20       &  \cellcolor{red!10}-  & \cellcolor{green!10}542 & \cellcolor{green!10}543 \\
Citations & \cellcolor{green!10}46& \cellcolor{yellow!10}36& \cellcolor{red!10}16 & \cellcolor{yellow!10}33        &  \cellcolor{red!10}16 & \cellcolor{red!10}0       &  \cellcolor{red!10}2  & \cellcolor{red!10}0 & \cellcolor{green!10}54 \\
License & \cellcolor{green!10}MIT& \cellcolor{green!10}Apach. 2.0&  \cellcolor{green!10} Apach. 2.0&\cellcolor{yellow!10}GPL 3.0        &  \cellcolor{yellow!10}GPL 3.0 & \cellcolor{green!10}Apach. 2.0       & \cellcolor{red!10}-  & \cellcolor{green!10}Apach. 2.0 & \cellcolor{green!10}MIT\\
\hline
\end{tabularx}
}
\begin{tablenotes}
    \setlength{\columnsep}{0.4cm}
    \setlength{\multicolsep}{0cm}
        \footnotesize
        \item $^*$We use the latest release of 3D BAT version v24.3.2.
    \end{tablenotes}
\end{table*}

\section{Problem statement}
We consider a cooperative perception system with roadside and vehicle sensors symbolized by $r_s \; s \in [C, L]$ and $v_s \; s \in [C, L]$ notations, respectively. The cooperative system introduced in this work uses three infrastructure cameras $r_{Ci} \; i \in {1, 2, 3}$ where $i$ denotes the camera IDs, an infrastructure LiDAR $r_{L}$, one onboard vehicle camera $v_{C}$ and one onboard LiDAR $v_{L}$. Consequently, the vehicle sensors produce a set of images $v_{I}(\hat{t})$ and point clouds $v_{P}(\hat{t})$, and the infrastructure sensors produce a set of images $r_{Ii}(t')$, and point clouds $r_{P}(t')$. 
Here, $\hat{t}$ and $t'$ denote the vehicle and infrastructure data timestamps respectively. Note that a small synchronization error is still present though the infrastructure and roadside sensors are all synchronized to the same NTP time server. The average difference in timestamps between these two systems $\mathbf{E}[\hat{t}-t']$ is 24.91 ms and the two data sources are matched in our proposed dataset using the nearest neighbor matching algorithm.

The objective of cooperative 3D detection is to predict 3D bounding boxes of objects given a set of multi-modal multi-viewpoint data. Our proposed cooperative detection model takes the set of images and point clouds as the input $X(t) = [v_I(t), v_P(t), r_{I1}(t), r_{I2}(t), r_{I3}(t), r_{P1}(t)]$ at a given time $t$ and predicts the 3D bounding boxes as the output $\tilde{Y}(t)$. Here, $t$ denotes the shared timestamp after the matching algorithm. In addition to identifying the boxes' position, dimensions, and orientation, the proposed model also predicts the class of the corresponding object. Thus, we can represent the task of 3D object detection as:
\begin{equation}
    \min 
    \mathop{\mathbb{E}}_{y_j\,\in\,Y(t)} \left[
        \min_{\tilde{y}_k\,\in\,\tilde{Y}(t)}
        d_\theta \left( y_j, \tilde{y}_k \right) 
    \right]
\end{equation}
where $Y(t) = [y_1(t), y_2(t), ...]$ is the set of ground truth 3D box labels at time $t$, and $\tilde{Y}(t) = [\tilde{y}_1(t), \tilde{y}_2(t), ...]$ are the corresponding predicted 3D boxes. $d_\theta (y_j, y_k)$ is a parameterized discriminator function which measures the error between ground truth 3D label $y_j$ and the predicted 3D box $y_k$. Thus, our objective is to reduce the total error.

\section{Data anonymization}
We anonymize all our camera raw images $I = [v_I, r_{I1}, r_{I2}, r_{I3}, r_{I4}]$ in the roadside and vehicle domain by obfuscating all license plate numbers and faces. We use a medium \textit{YOLOv5} model \cite{jocher2020yolov5} for this purpose, which was pre-trained on 1080p images with labeled license plates and faces. During training, mosaic augmentation was applied to teach the model to recognize objects in different locations without relying too much on one specific context. At inference, we downscale the input images $I$ from a $1920\times1200$ resolution to $640\times400$ and pad the extra space to $640\times640$. A score threshold of 0.1 worked best to detect all private information. We set the granularity of the blurring filter to a blur size of 6 for the detected regions and set the ROI multiplier to 1.1. 

\section{Further related work}
This section compares our proposed \textit{3D BAT} v24.3.2 annotation tool and development kit to similar open-source tools.


\subsection{Annotation tools}
This work proposes our annotation tool \textit{3D BAT} v24.3.2, which supports combining LiDAR point clouds and simultaneously labels both the point clouds and images from multiple views.


\textit{3D BAT} \cite{zimmer20193d} is an open-source, web-based annotation framework designed for efficient and accurate 3D annotation of objects in LiDAR point clouds and camera images. With this tool, 2D and 3D box labels can be obtained, as well as track IDs. Its key features include semi-automatic labeling using interpolation of objects between frames. Labeled 3D boxes are automatically projected into all camera images, which requires extrinsic camera-LiDAR calibration data. Selected objects are displayed in a bird's eye view, side view, and front view, in addition to a perspective and orthographic view.

\textit{SUSTechPoints} \cite{sustech} is a multi-modal 3D object annotation tool. It first allows the addition of 3D bounding boxes in point clouds and then updates them in six degrees of freedom. It furthermore allows updating bounding boxes' type, attributes, and ID to create labeled datasets for detection and tracking tasks. It also allows users to visualize these boxes projected onto multiple camera images and lets the user enable or disable the point clouds and images for clear visualization. One major advantage of \textit{SUSTechPoints} is that it enables auto box fitting based on the point cloud shape, but the accuracy of the fitted box is highly dependent on the point cloud density.


\textit{labelCloud} \cite{sager2021labelcloud} is a domain-agnostic, lightweight tool designed specifically to label 3D objects. It offers two labeling modes namely picking and spanning. In the picking mode, objects with known sizes can be quickly adjusted. The spanning mode simplifies labeling by reducing the process to four clicks. Box dimensions and orientations of objects on flat surfaces can be efficiently defined. 

\textit{ReBound} \cite{chen2023rebound} is an open source 3D bounding box annotation tool designed to utilize active learning. It supports loading, visualizing, and extending existing datasets like nuScenes \cite{caesar2020nuscenes}, Waymo \cite{sun2020scalability} or Argoverse 2.0 \cite{wilson2023argoverse}. Model predictions can be analyzed and corrected in a 3D view and exported to specific formats.

\textit{PointCloudLab} \cite{PointCloudLab} leverages virtual and augmented reality (VR/AR) devices for 3D point cloud annotation. The annotator utilizes the controller of a HTC Vive to perform object-level annotations in the 3D point cloud. The immersive visual aid accelerates the labeling speed, improves the labeling quality, and enhances the labeling experience.

The \textit{Xtreme1} \cite{Xtreme1} labeling tool provides most of the functionalities of \textit{SUSTechPoints}. In addition to providing automated 3D labeling, it also provides support for automated 2D detection and segmentation tasks. Furthermore, it also supports multi-view point cloud data as the input. The tool also provides an interface for identifying specific errors in the labeling process, and a mechanism to evaluate different models on the labeled dataset. Moreover, it uses modern cloud-based standards, databases, Kubernetes for managing containers, and GitLab CI automation.

\begin{figure}[t]
    \centering
    \includegraphics[width=1.0\linewidth]{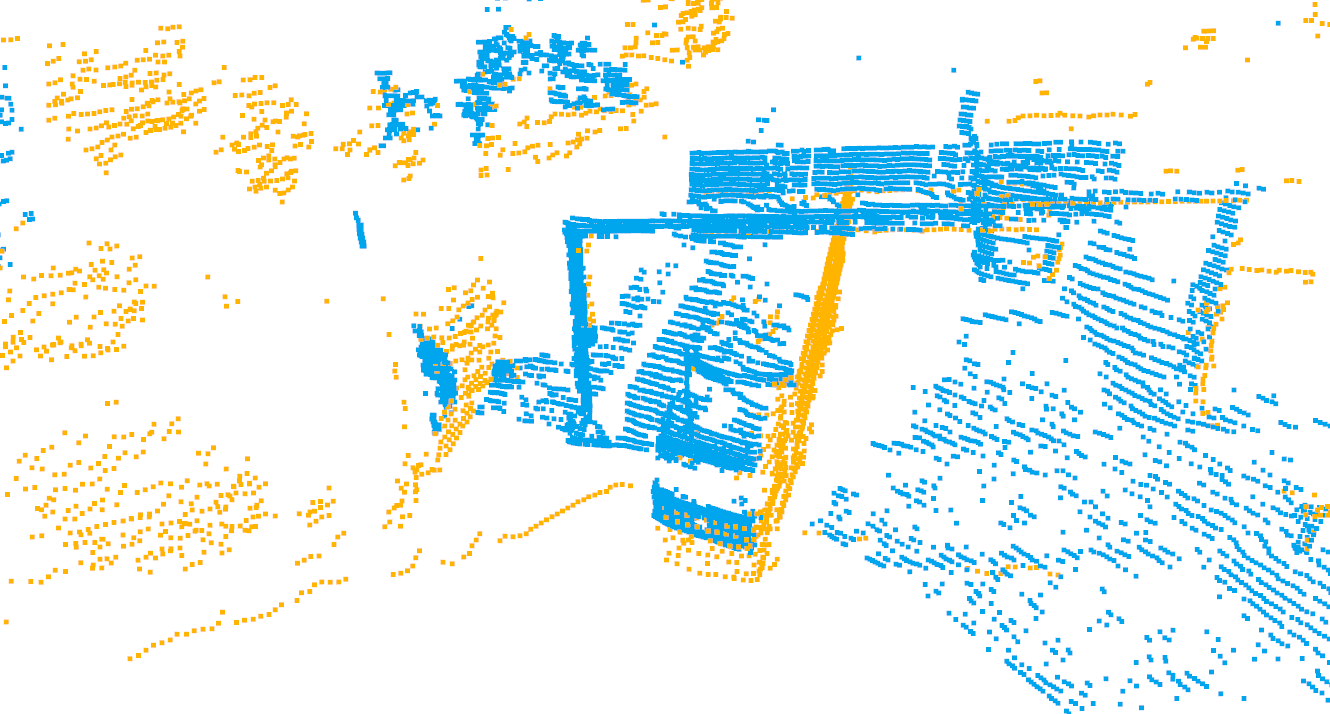}
    \caption{Point cloud registration results of an onboard LiDAR point cloud (orange) and an infrastructure LiDAR point cloud (blue).}
    \label{fig:point_cloud_registration_results}
\end{figure}

\begin{table*}[t!]
    \centering
    \caption{Tracking results of SORT and PolyMOT on drive\_41. P = Precision, R = Recall, MT = Mostly Tracked, PT = Partially Tracked, ML = Mostly Lost, FM = Track Fragmentations\label{tbl:tracking_results}}
    \begin{ThreePartTable}
    \resizebox{\linewidth}{!}{
    \begin{tabular}{l|rrrrrrrrrrrrrrr}
         \hline
         Tracker                     & IDP$\uparrow$  & IDR$\uparrow$ & IDF1$\uparrow$ & Recall$\uparrow$ & Precision$\uparrow$ & GT  & MT$\uparrow$ & PT$\uparrow$ & ML$\downarrow$ & FP$\downarrow$ & FN$\downarrow$  & IDS$\downarrow$ & FM$\downarrow$ &MOTA$\uparrow$  & MOTP$\downarrow$ \\
         \hline
         SORT\tnote{$*$}~~\cite{bewley2016simple}&36.313&21.029& 26.634 & 43.235  &74.657     & 3400&5   & \textbf{18} & 11 & 499& 1920& 439 & 110&15.647& \textbf{100.185} \\
         PolyMOT \cite{li2023poly}   &\textbf{68.416}&\textbf{42.559}& \textbf{52.475} &\textbf{46.735}  &\textbf{75.130}     & 3400&\textbf{8}   & 15 & \textbf{11} & \textbf{526}& \textbf{1811}& \textbf{13}  & \textbf{30} &\textbf{30.882}& 102.288 \\
         \hline
    \end{tabular}
    }
    \begin{tablenotes}
    \setlength{\columnsep}{0.4cm}
    \setlength{\multicolsep}{0cm}
      \begin{multicols}{2}
        \footnotesize
        \item [$*$] We modify the SORT tracker to track objects in 3D.
      \end{multicols}
    \end{tablenotes}
    \end{ThreePartTable}
\end{table*}


\setlength{\fboxsep}{0pt}%
\setlength{\fboxrule}{1pt}%
\begin{figure*}[h]
\centering
\minipage{0.14\textwidth}
  \caption*{a)}
  \vspace{-0.3cm}
  \fbox{\includegraphics[width=.98\linewidth]{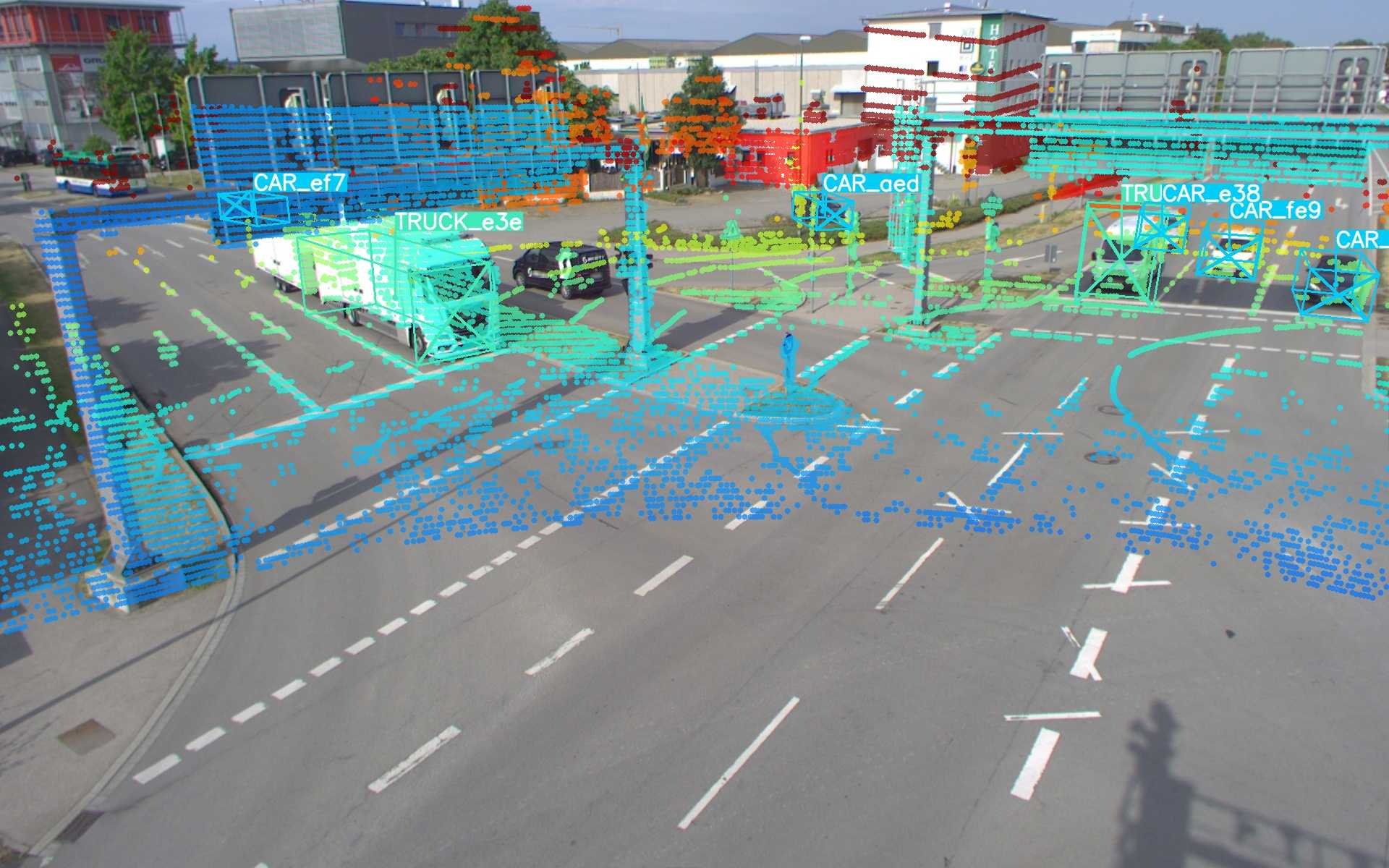}}
\endminipage
\minipage{0.14\textwidth}
\caption*{b)}
  \vspace{-0.3cm}
  \fbox{\includegraphics[width=.98\linewidth]{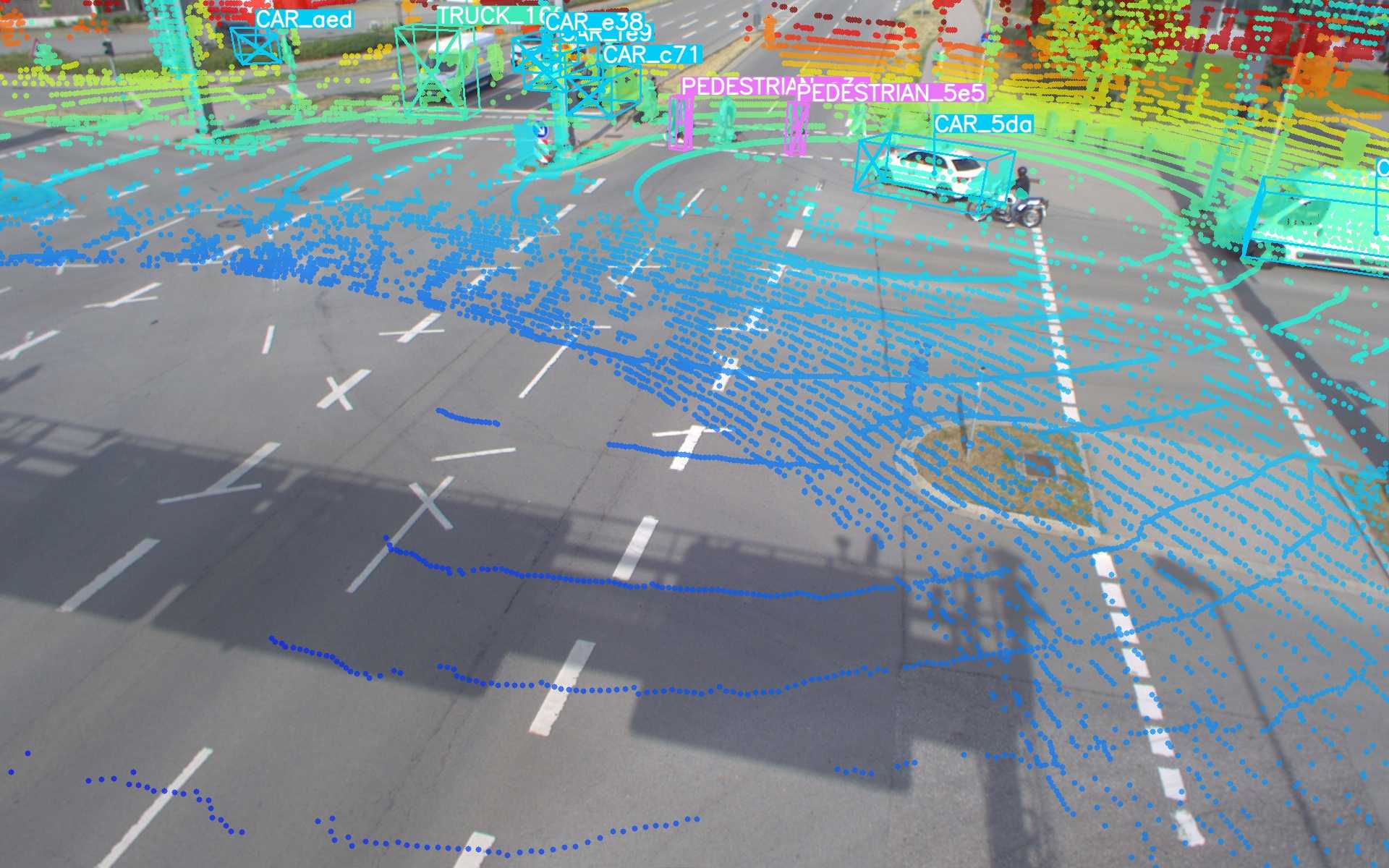}}
\endminipage
\minipage{0.14\textwidth}%
\caption*{c)}
  \vspace{-0.3cm}
  \fbox{\includegraphics[width=.98\linewidth]{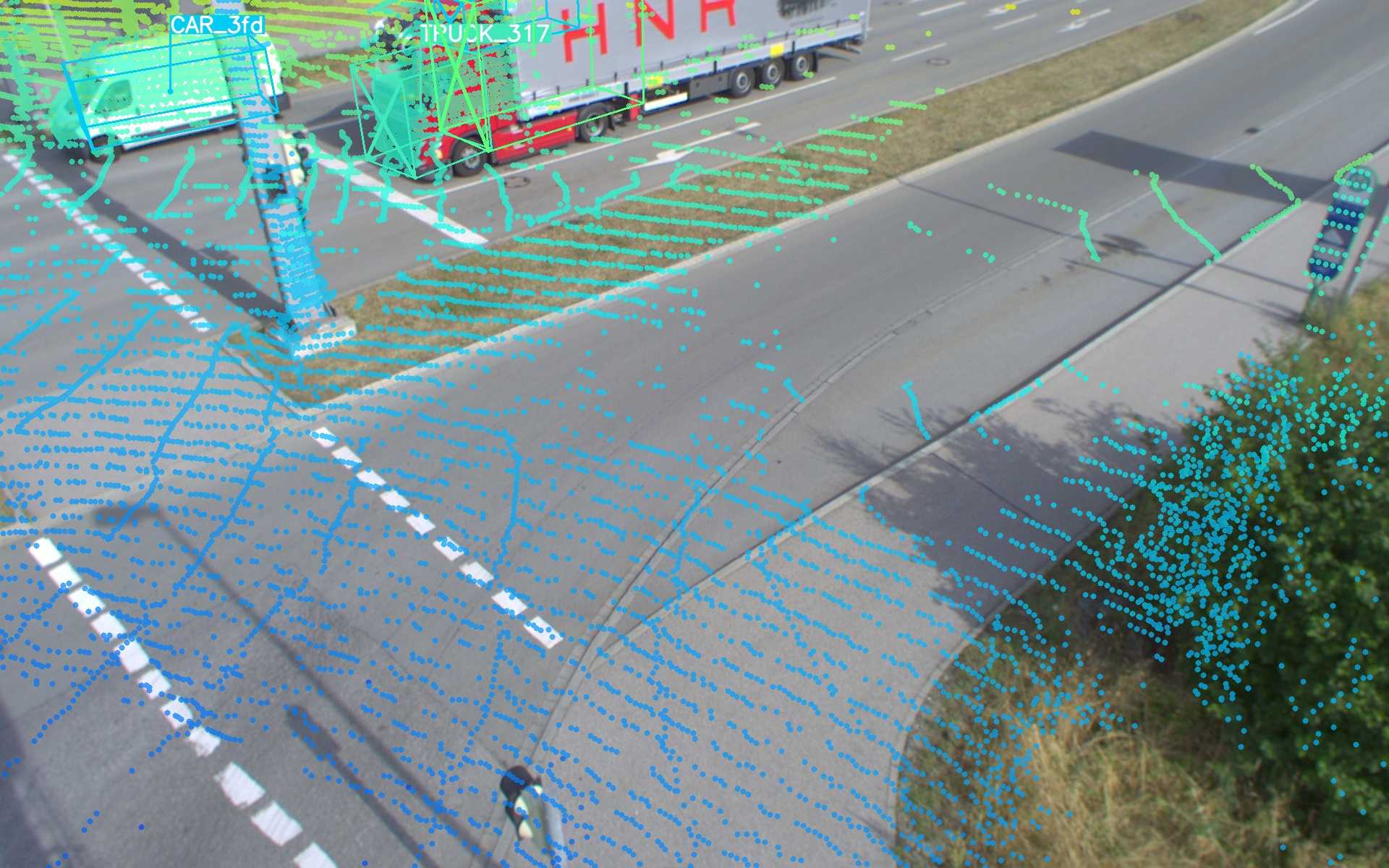}}
\endminipage
\minipage{0.14\textwidth}%
\caption*{d)}
  \vspace{-0.3cm}
  \fbox{\includegraphics[width=.98\linewidth]{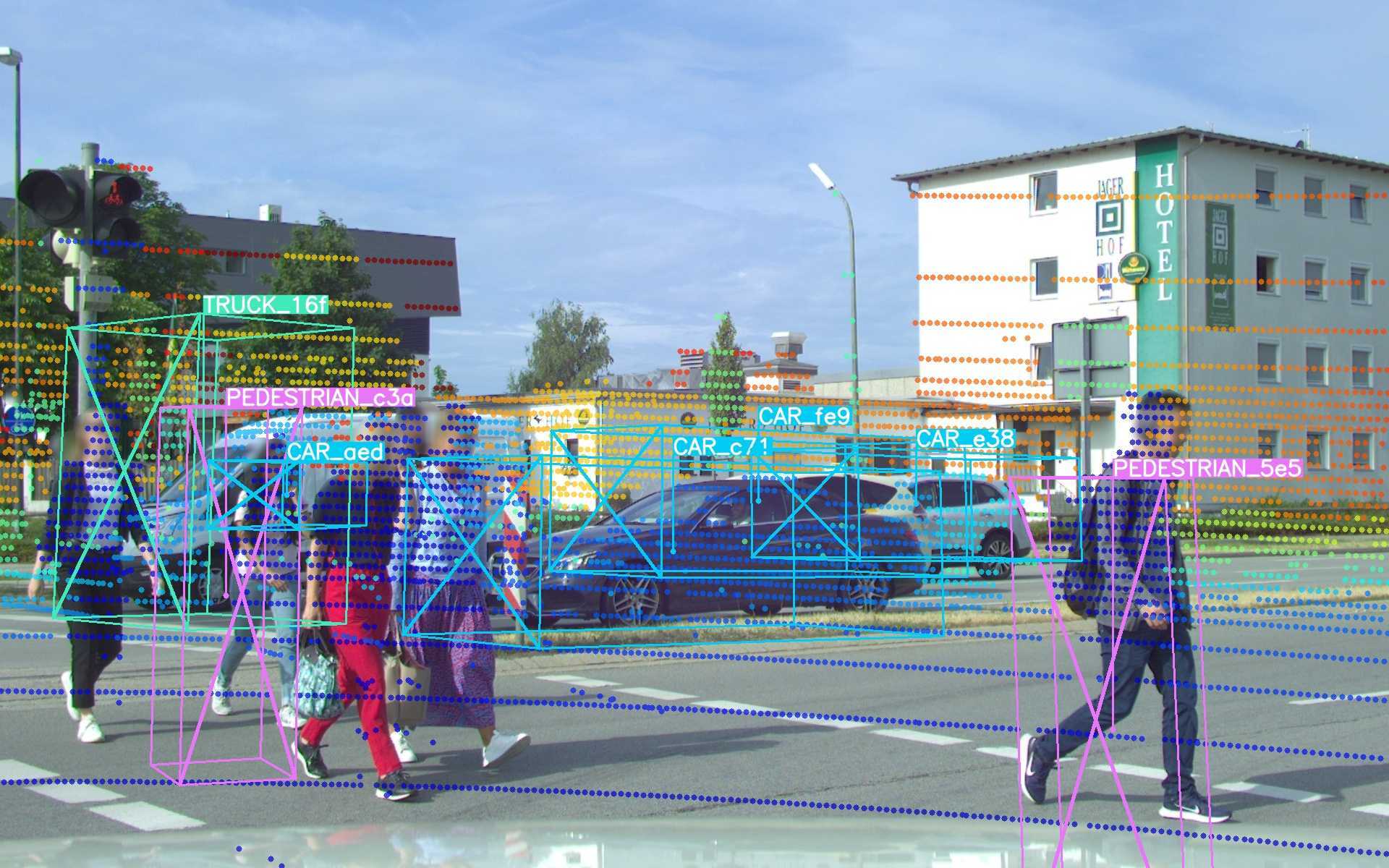}}
\endminipage
\minipage{0.14\textwidth}%
  \caption*{e)}
  \vspace{-0.3cm}
  \fbox{\includegraphics[width=.98\linewidth]{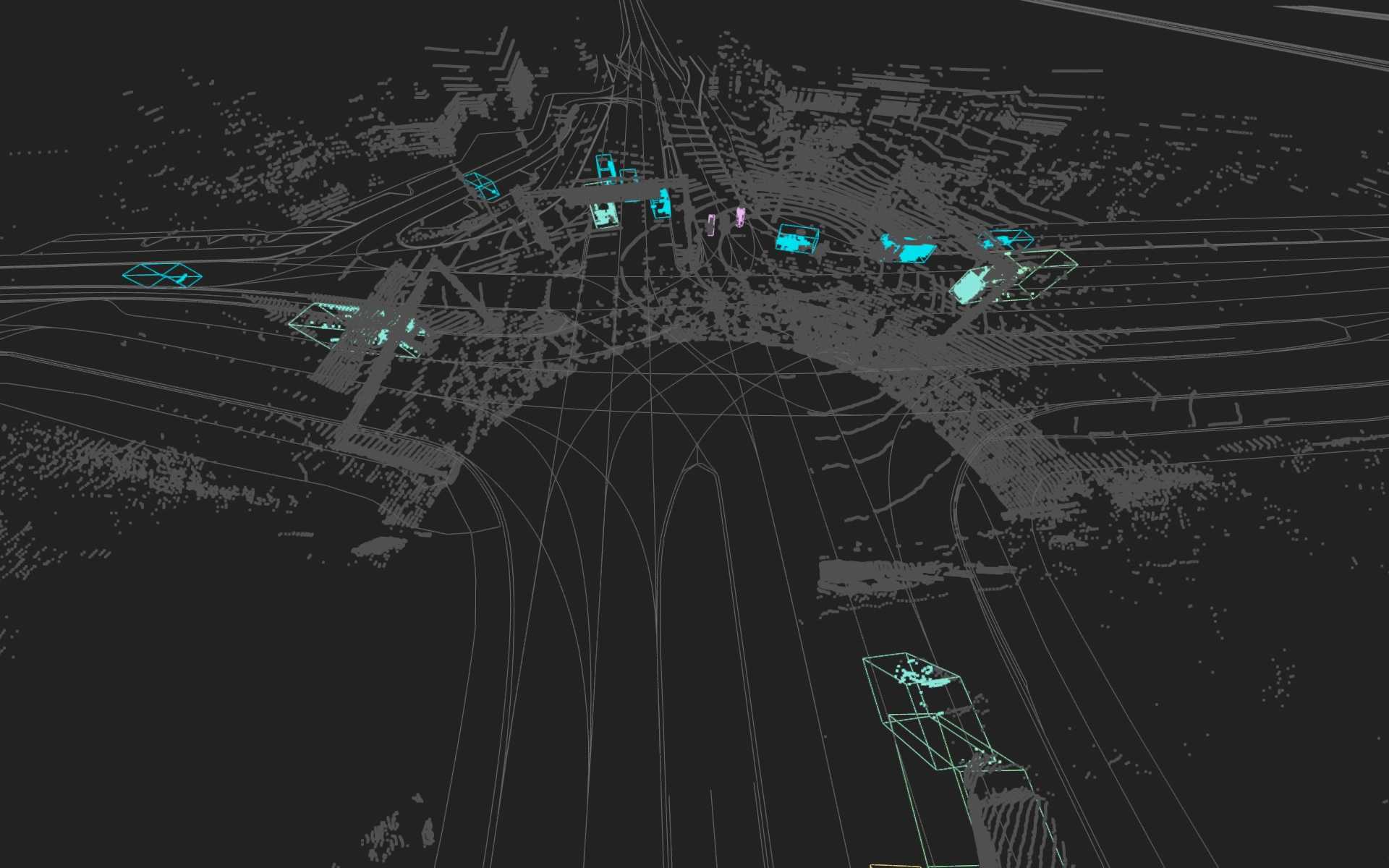}}
\endminipage
\minipage{0.14\textwidth}%
  \caption*{f)}
  \vspace{-0.3cm}
  \fbox{\includegraphics[width=.98\linewidth]{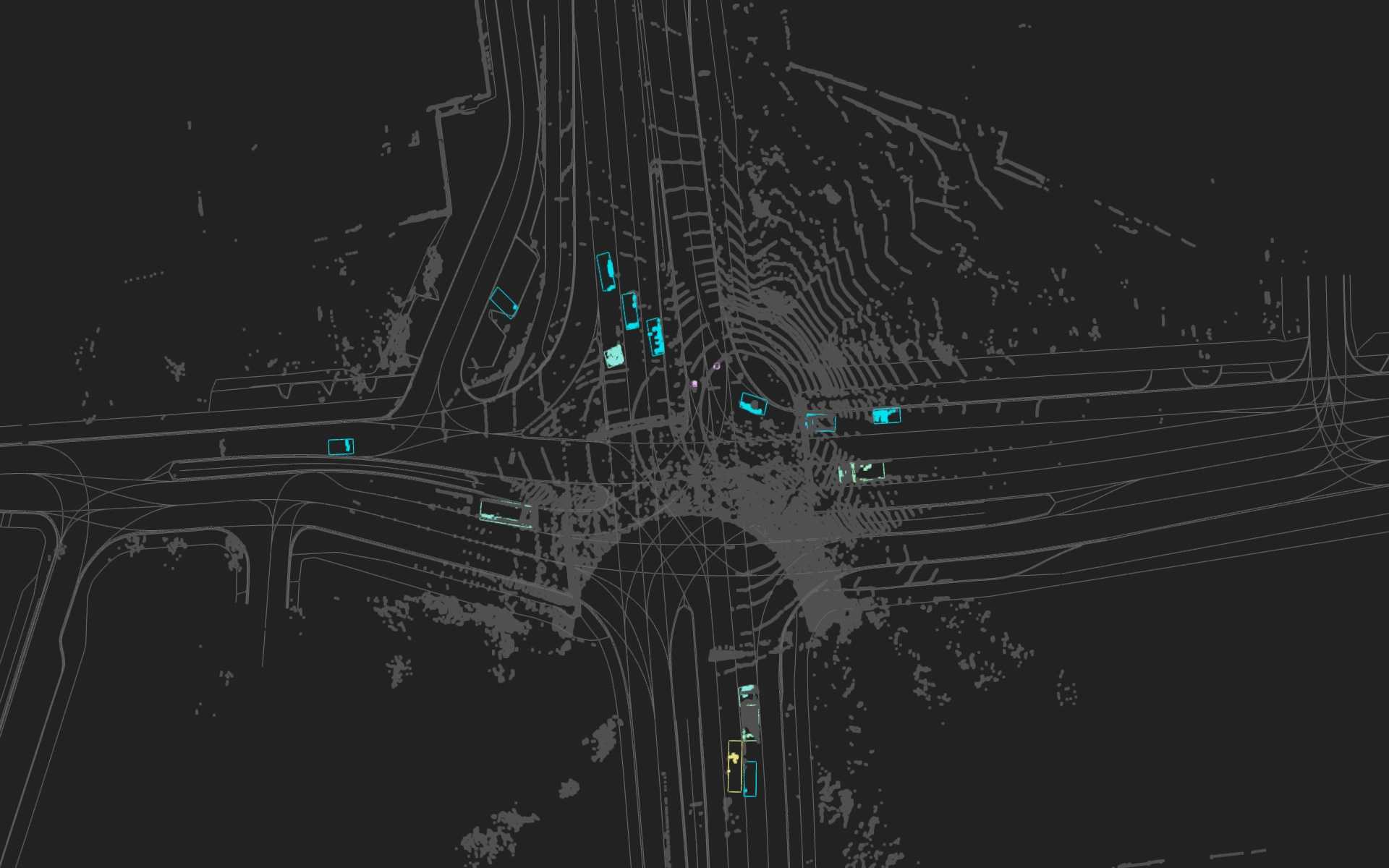}}
\endminipage
\minipage{0.14\textwidth}
  \caption*{g)}
  \vspace{-0.3cm}
  \fbox{\includegraphics[width=.98\linewidth]{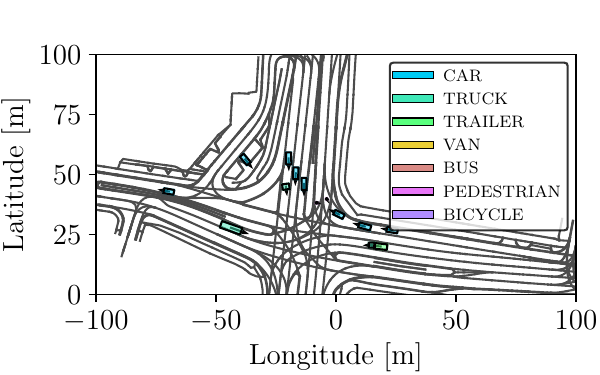}}
\endminipage\\
\minipage{0.14\textwidth}
  \vspace{-0.1cm}
  \fbox{\includegraphics[width=.98\linewidth]{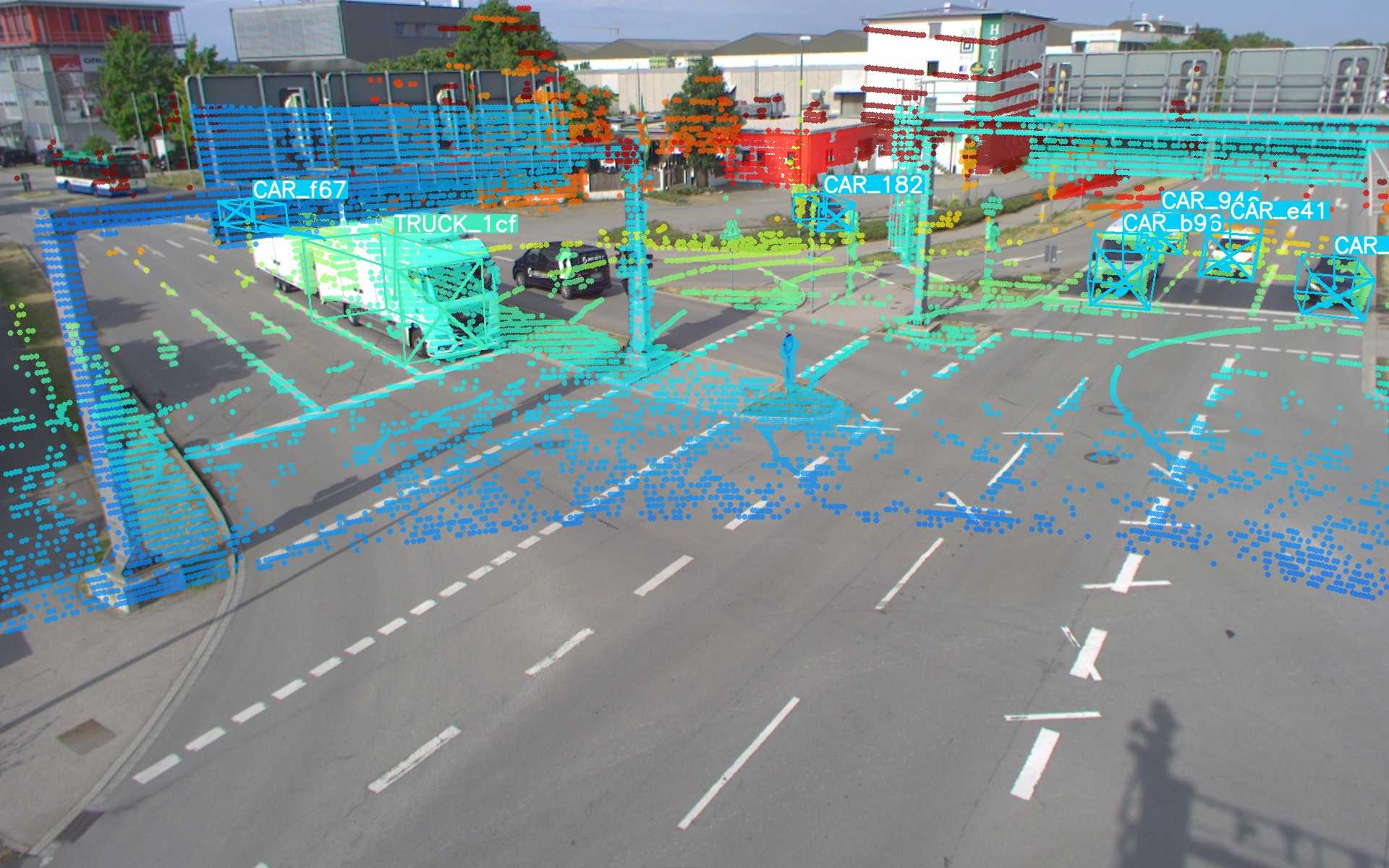}}
\endminipage
\minipage{0.14\textwidth}
\vspace{-0.1cm}
  \fbox{\includegraphics[width=.98\linewidth]{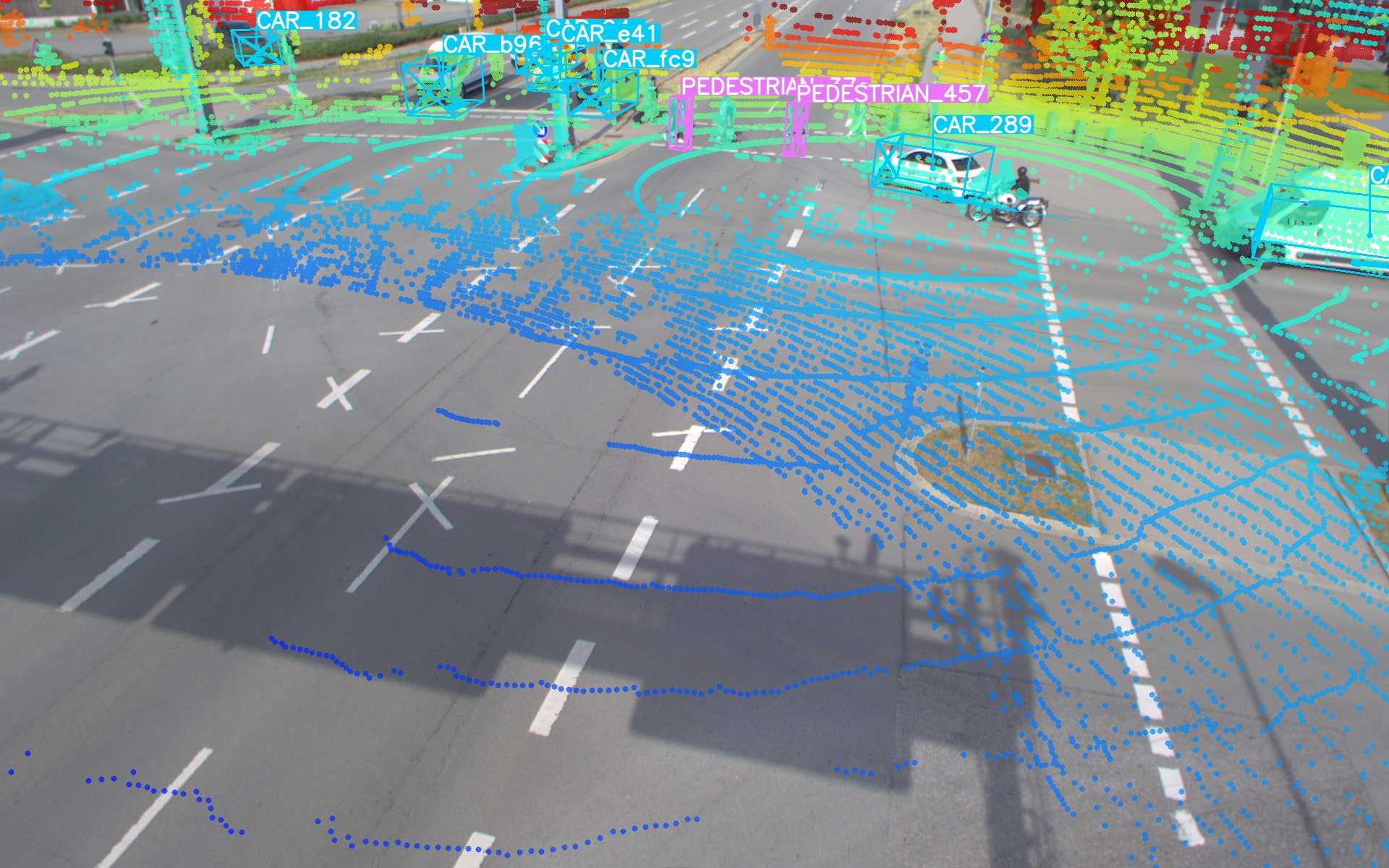}}
\endminipage
\minipage{0.14\textwidth}%
\vspace{-0.1cm}
  \fbox{\includegraphics[width=.98\linewidth]{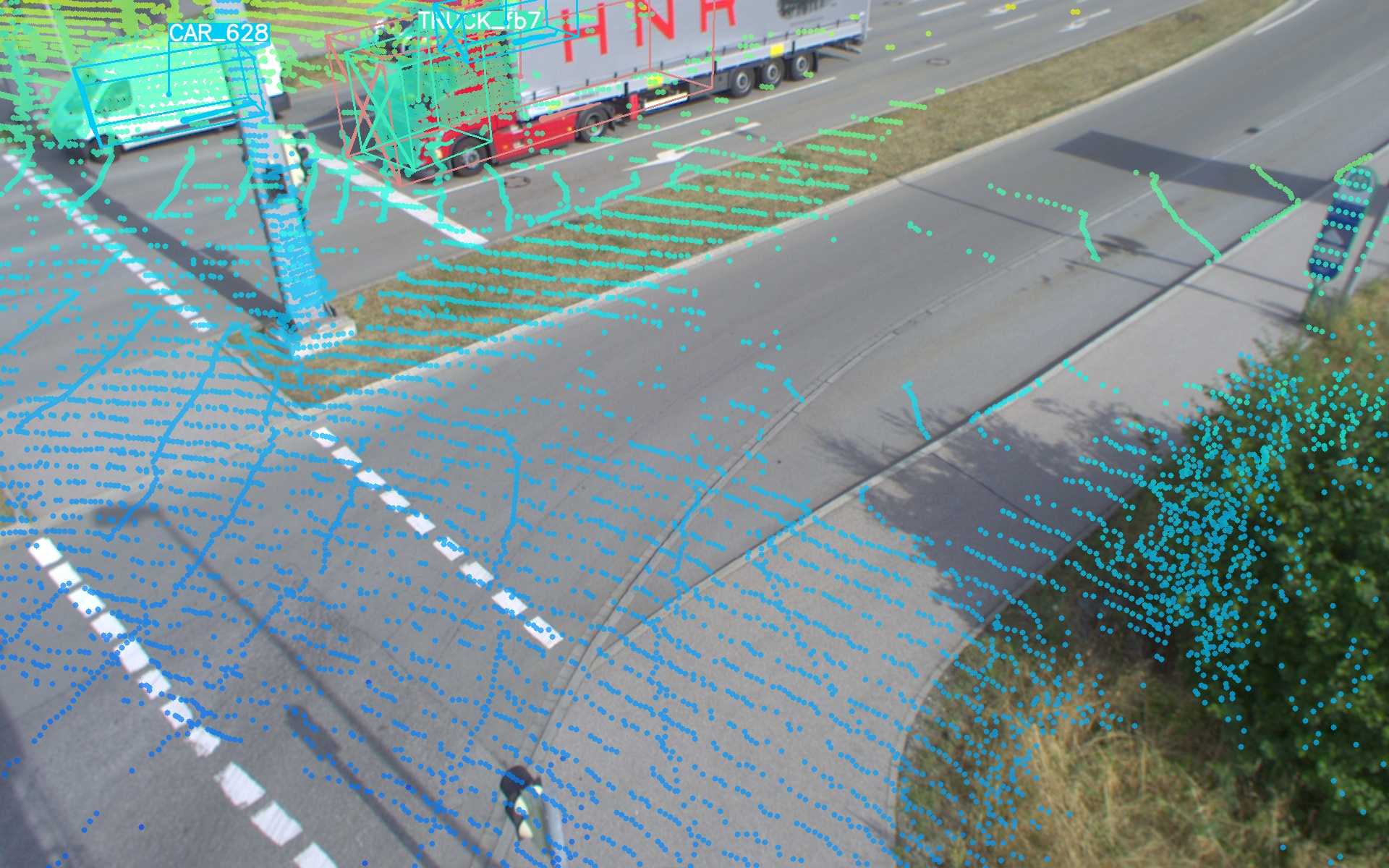}}
\endminipage
\minipage{0.14\textwidth}%
\vspace{-0.1cm}
  \fbox{\includegraphics[width=.98\linewidth]{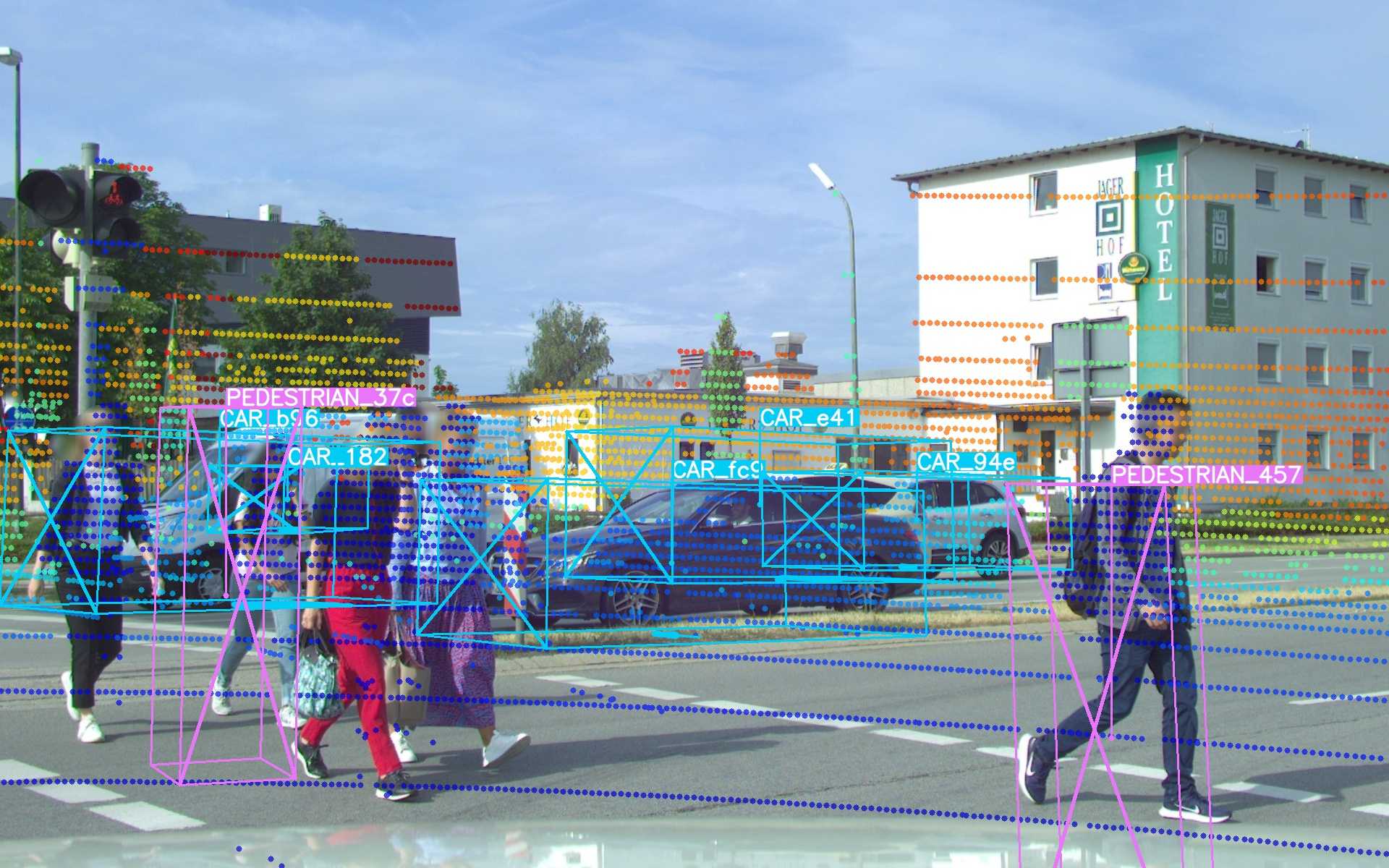}}
\endminipage
\minipage{0.14\textwidth}%
\vspace{-0.1cm}
  \fbox{\includegraphics[width=.98\linewidth]{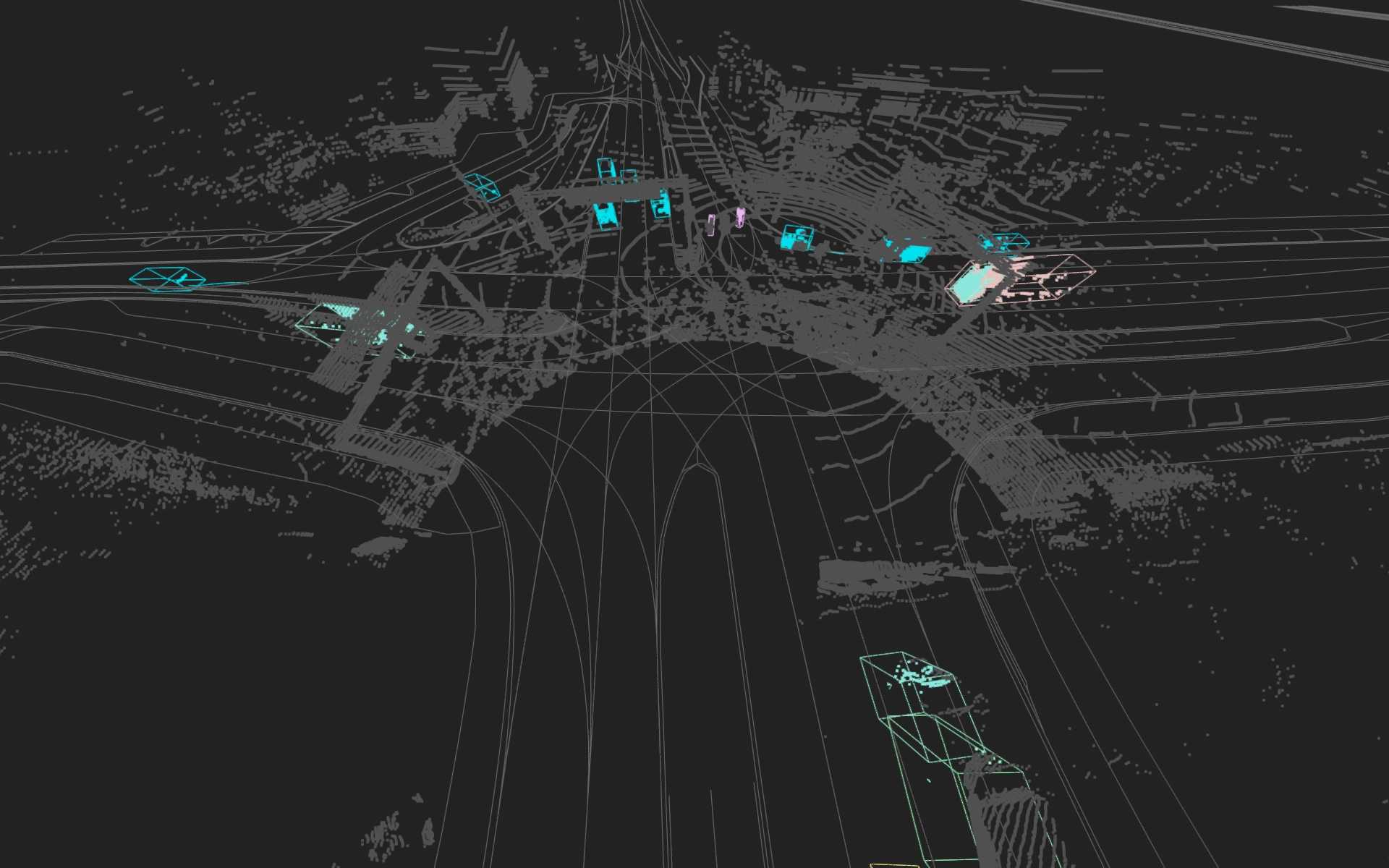}}
\endminipage
\minipage{0.14\textwidth}%
\vspace{-0.1cm}
  \fbox{\includegraphics[width=.98\linewidth]{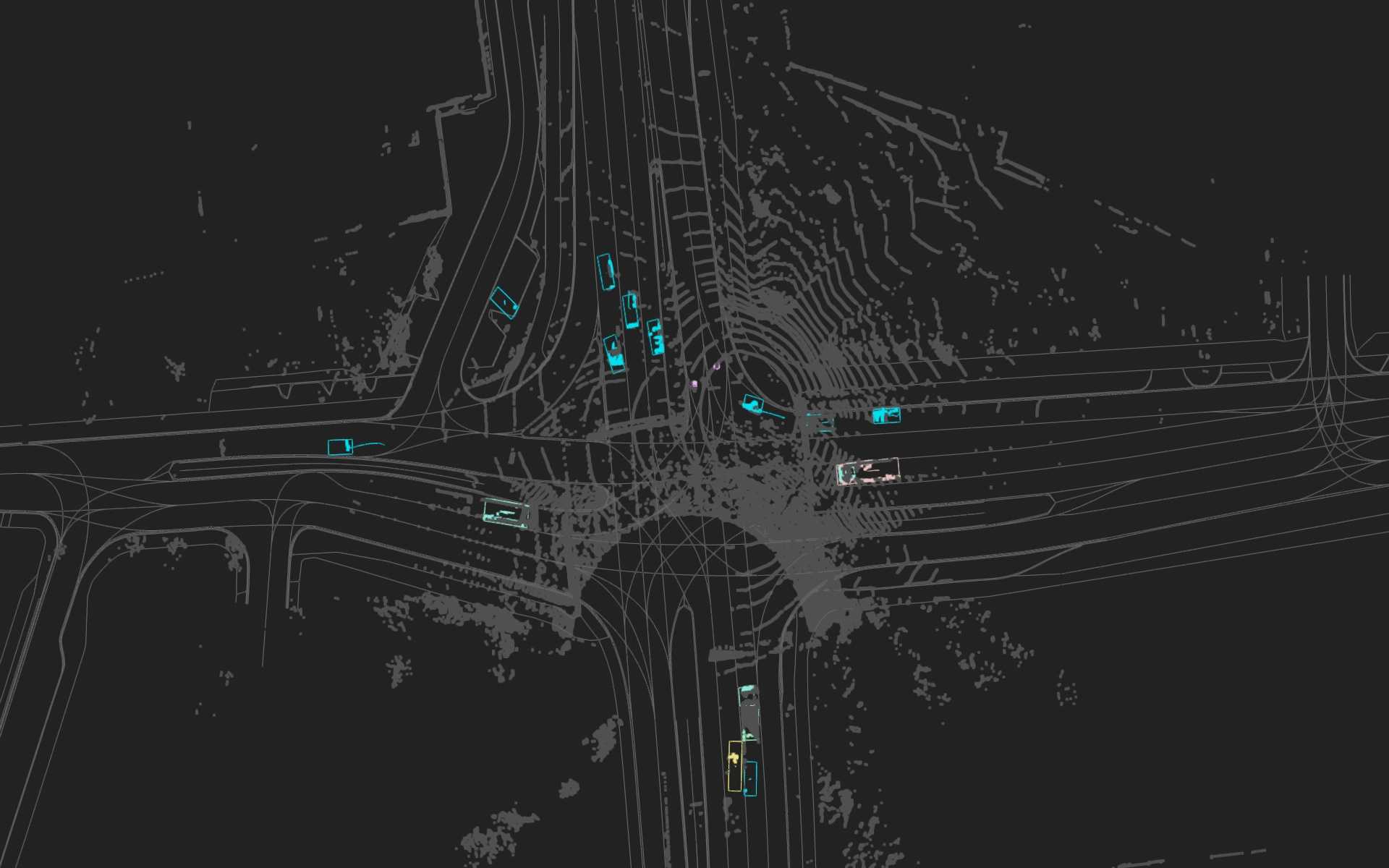}}
\endminipage
\minipage{0.14\textwidth}
\vspace{-0.1cm}
  \fbox{\includegraphics[width=.98\linewidth]{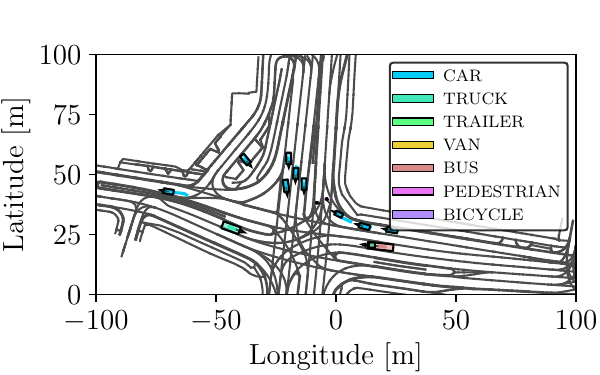}}
\endminipage\\
\minipage{0.14\textwidth}
\vspace{-0.1cm}
  \fbox{\includegraphics[width=.98\linewidth]{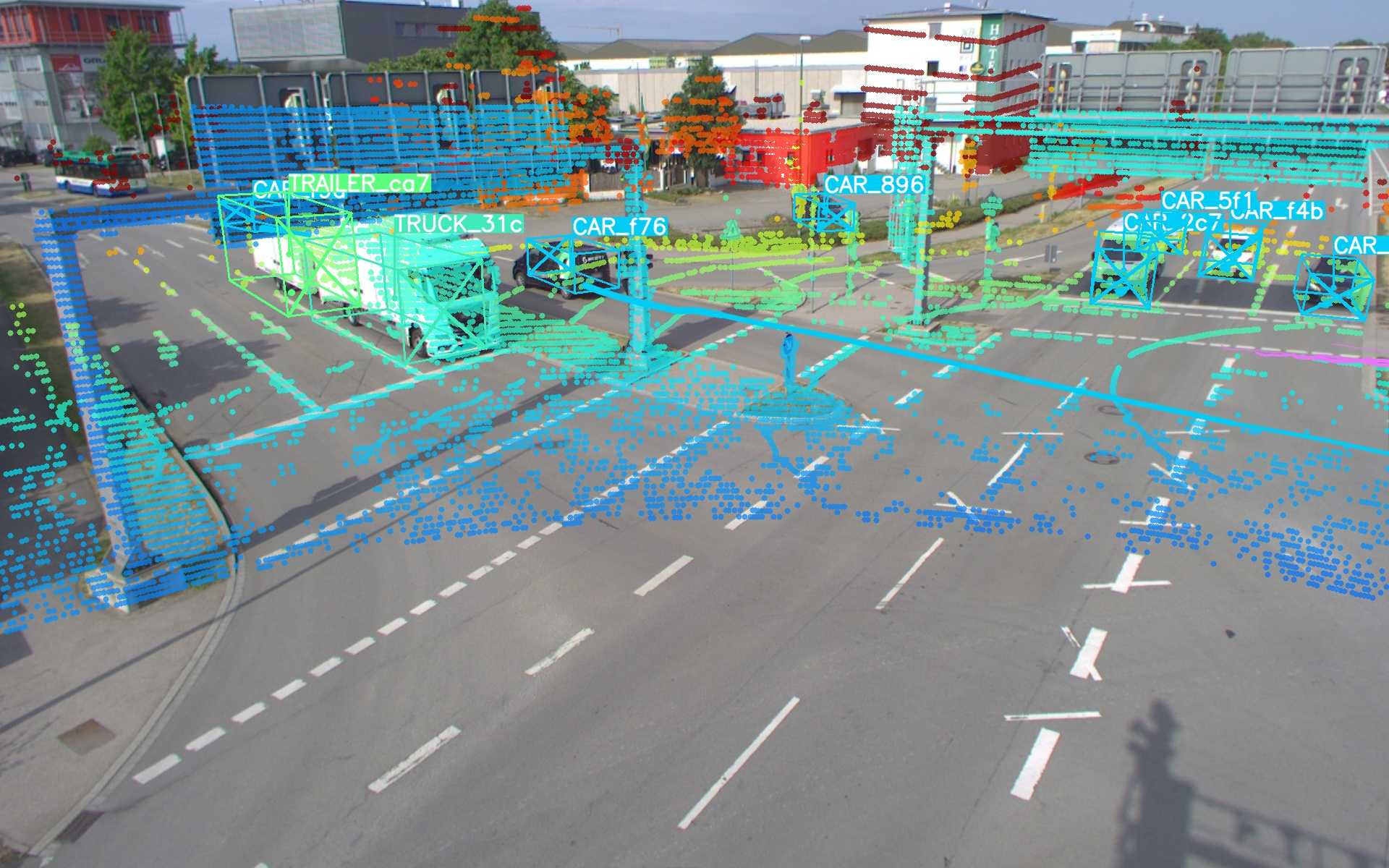}}
\endminipage
\minipage{0.14\textwidth}
\vspace{-0.1cm}
  \fbox{\includegraphics[width=.98\linewidth]{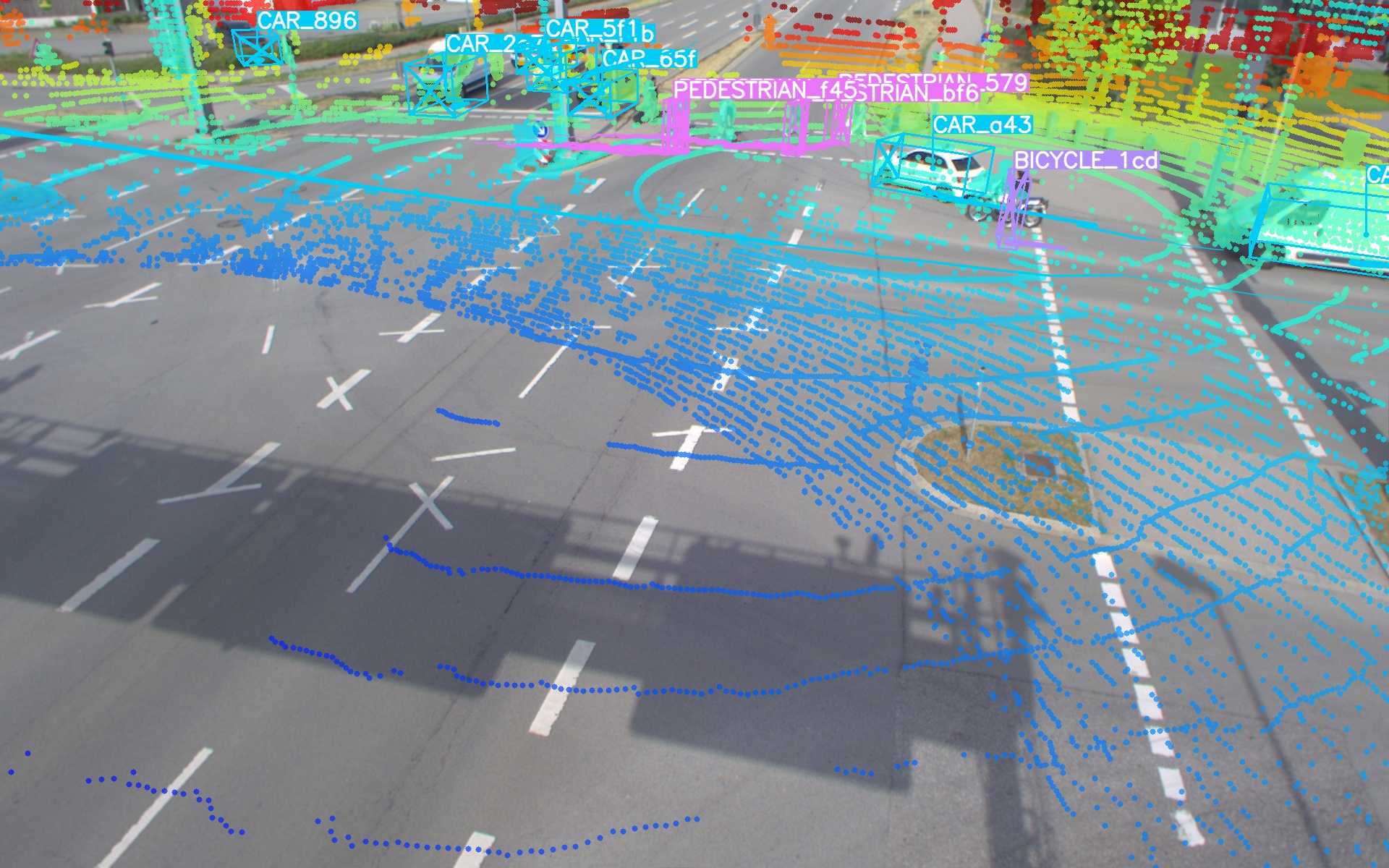}}
\endminipage
\minipage{0.14\textwidth}%
\vspace{-0.1cm}
  \fbox{\includegraphics[width=.98\linewidth]{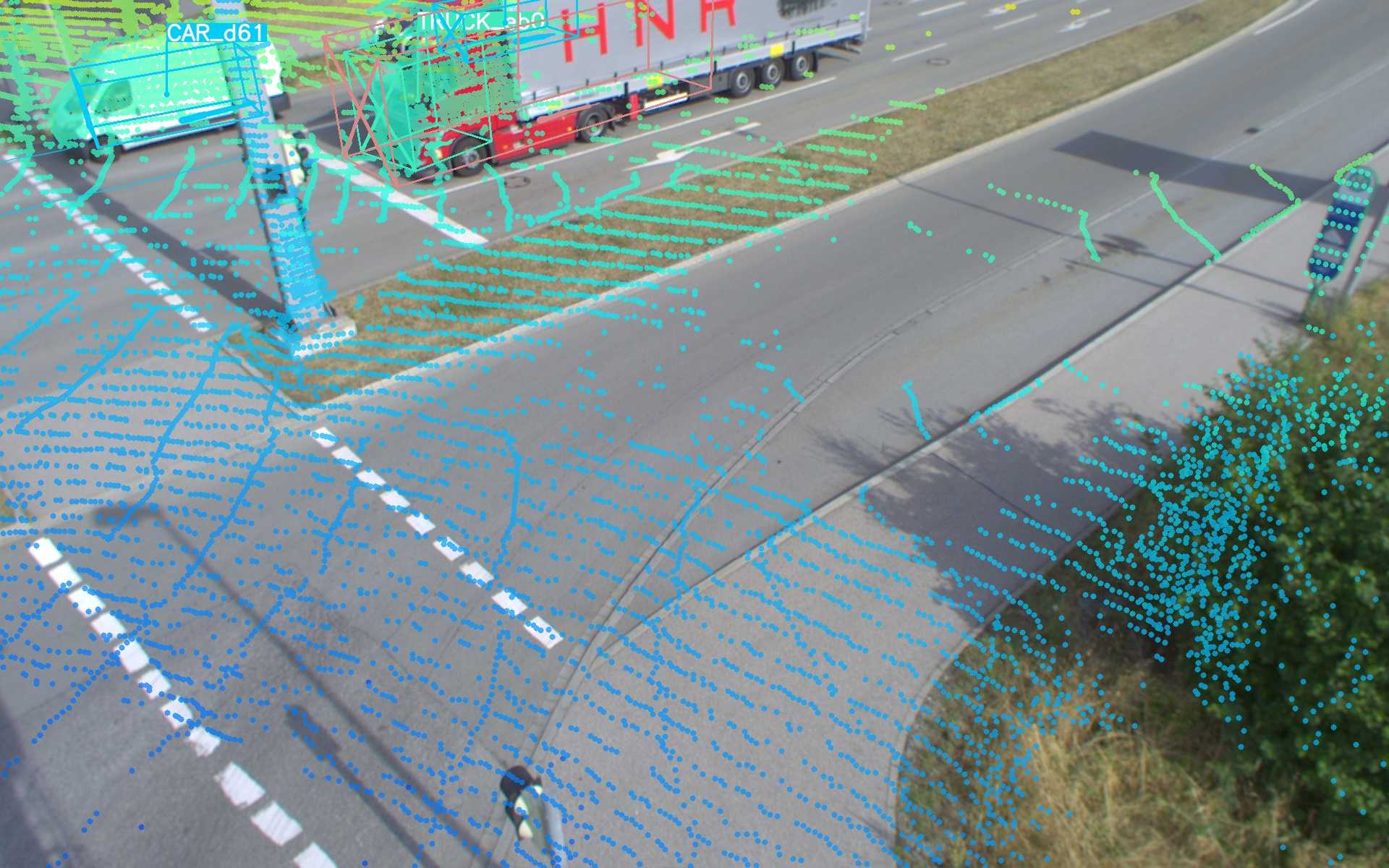}}
\endminipage
\minipage{0.14\textwidth}%
\vspace{-0.1cm}
  \fbox{\includegraphics[width=.98\linewidth]{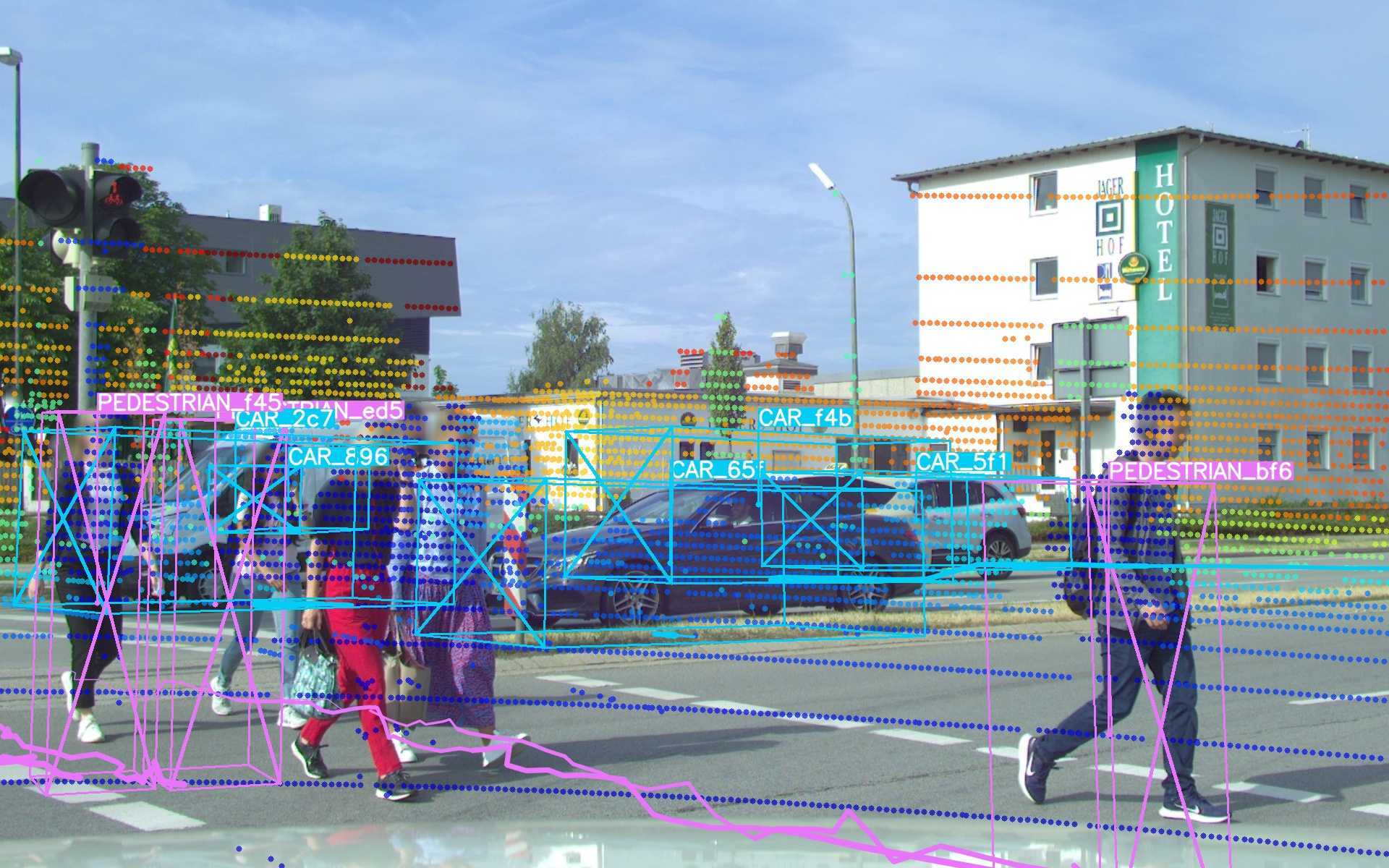}}
\endminipage
\minipage{0.14\textwidth}%
\vspace{-0.1cm}
  \fbox{\includegraphics[width=.98\linewidth]{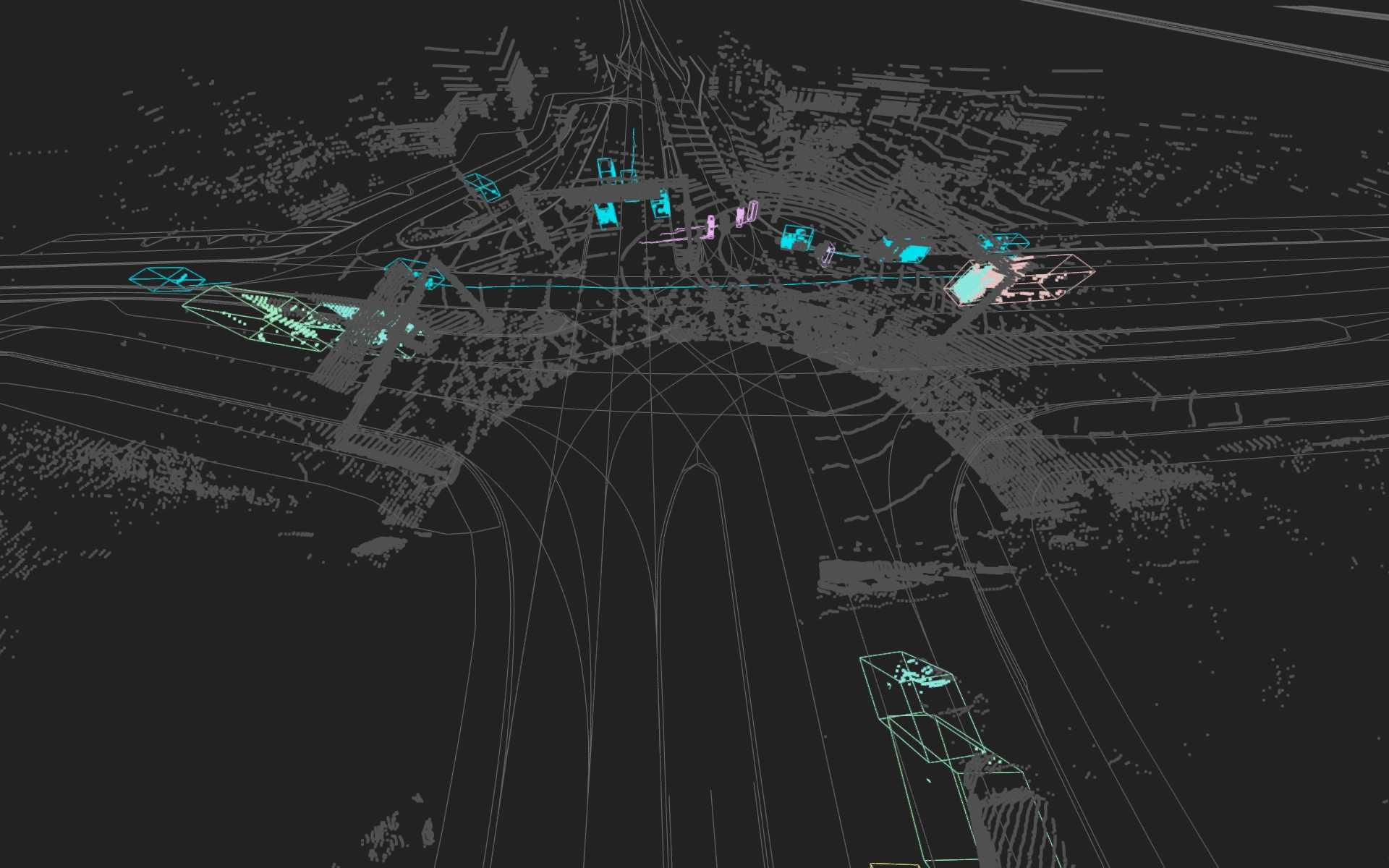}}
\endminipage
\minipage{0.14\textwidth}%
\vspace{-0.1cm}
  \fbox{\includegraphics[width=.98\linewidth]{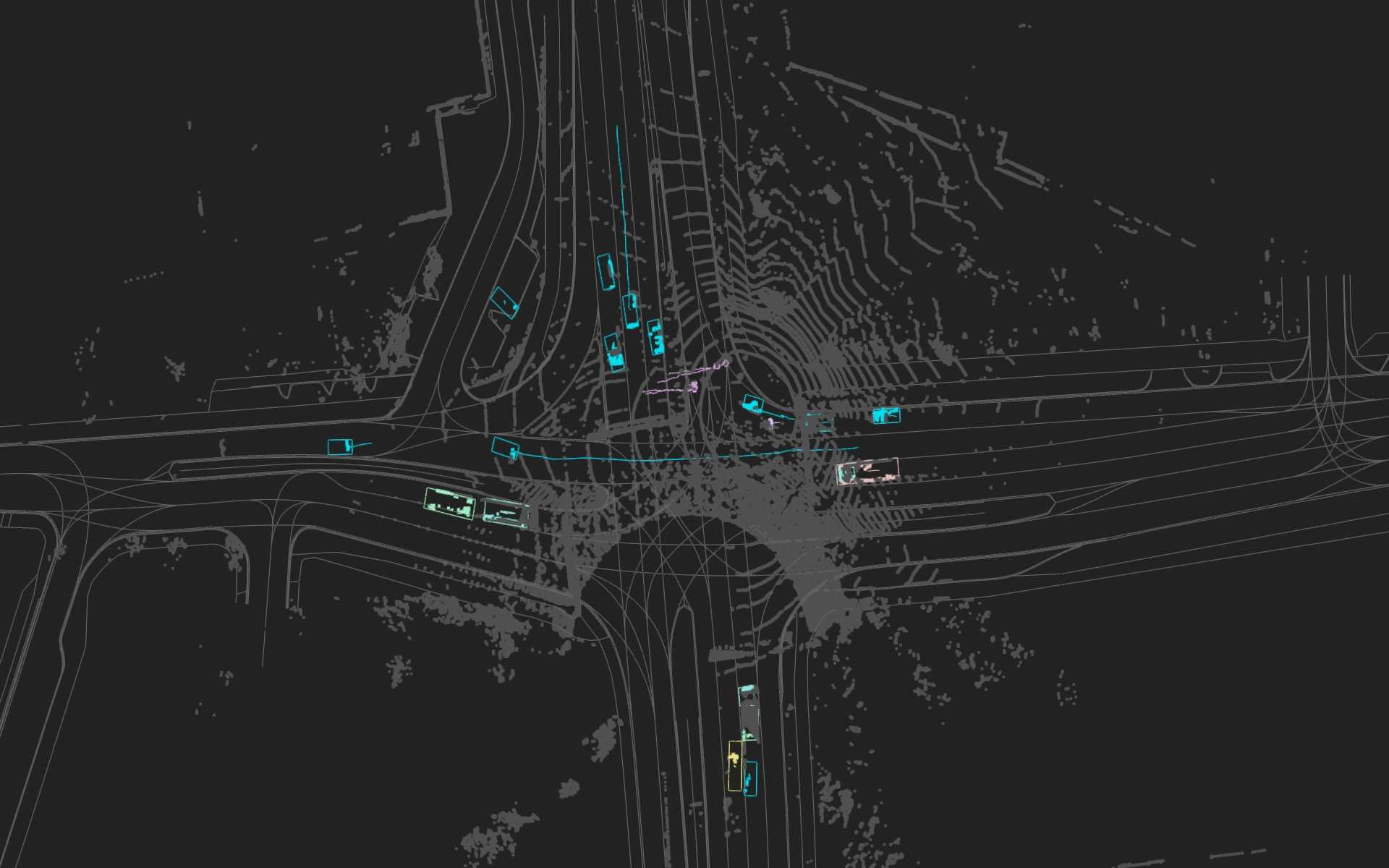}}
\endminipage
\minipage{0.14\textwidth}
\vspace{-0.1cm}
  \fbox{\includegraphics[width=.98\linewidth]{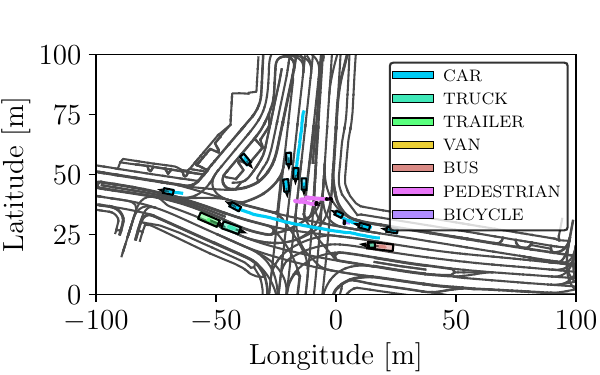}}
\endminipage\\
\minipage{0.14\textwidth}
\vspace{-0.1cm}
  \fbox{\includegraphics[width=.98\linewidth]{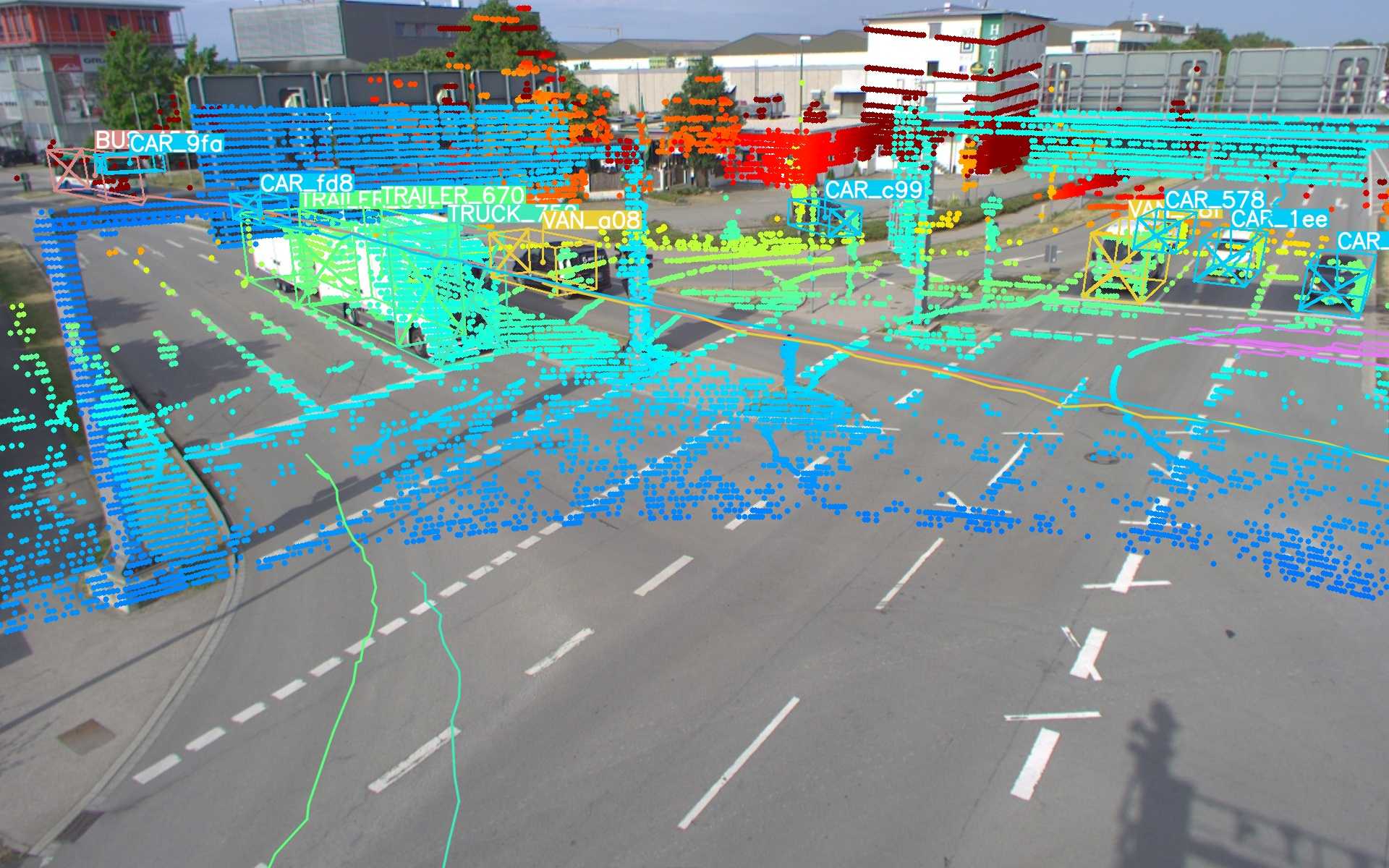}}
\endminipage
\minipage{0.14\textwidth}
\vspace{-0.1cm}
  \fbox{\includegraphics[width=.98\linewidth]{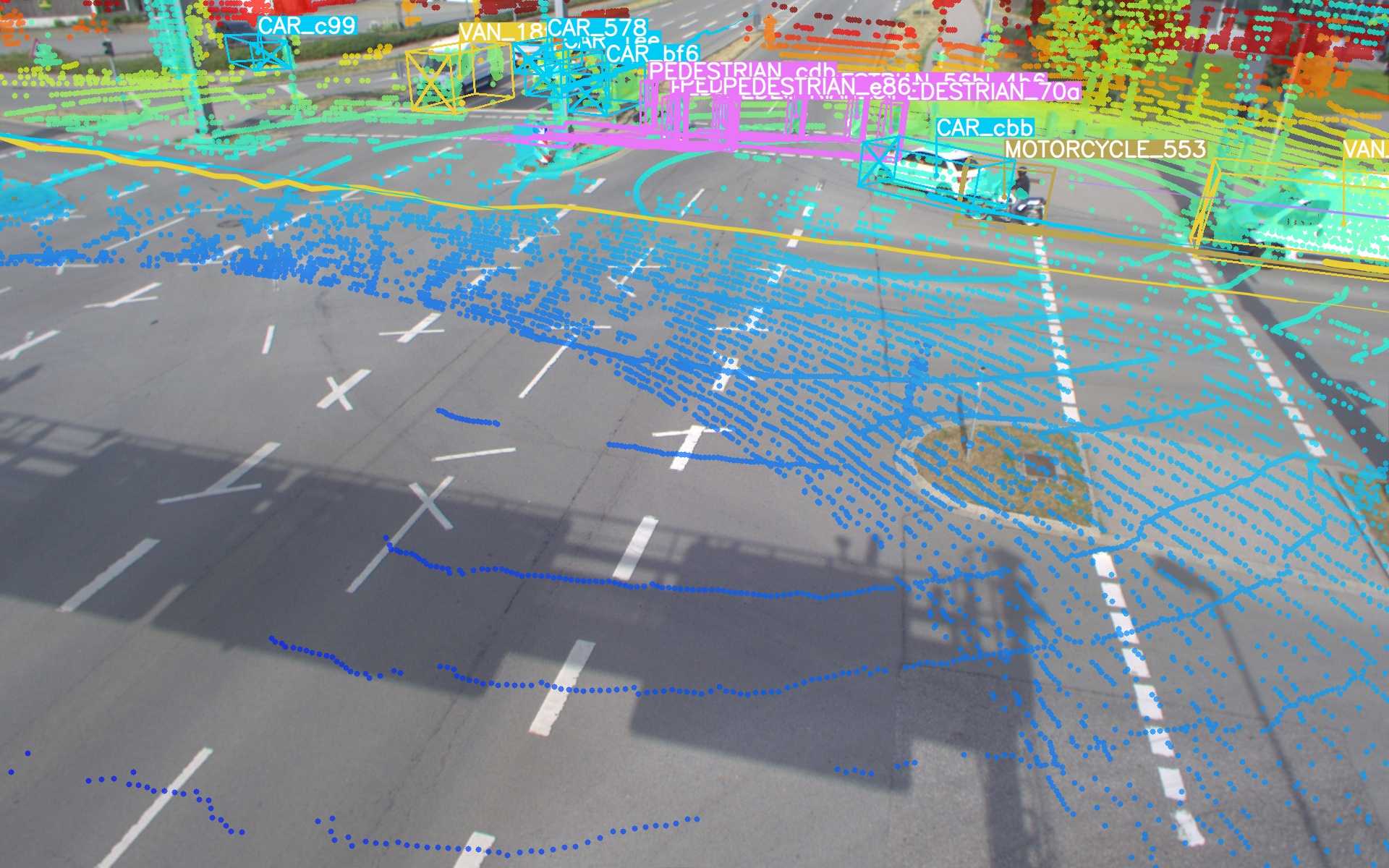}}
\endminipage
\minipage{0.14\textwidth}%
\vspace{-0.1cm}
  \fbox{\includegraphics[width=.98\linewidth]{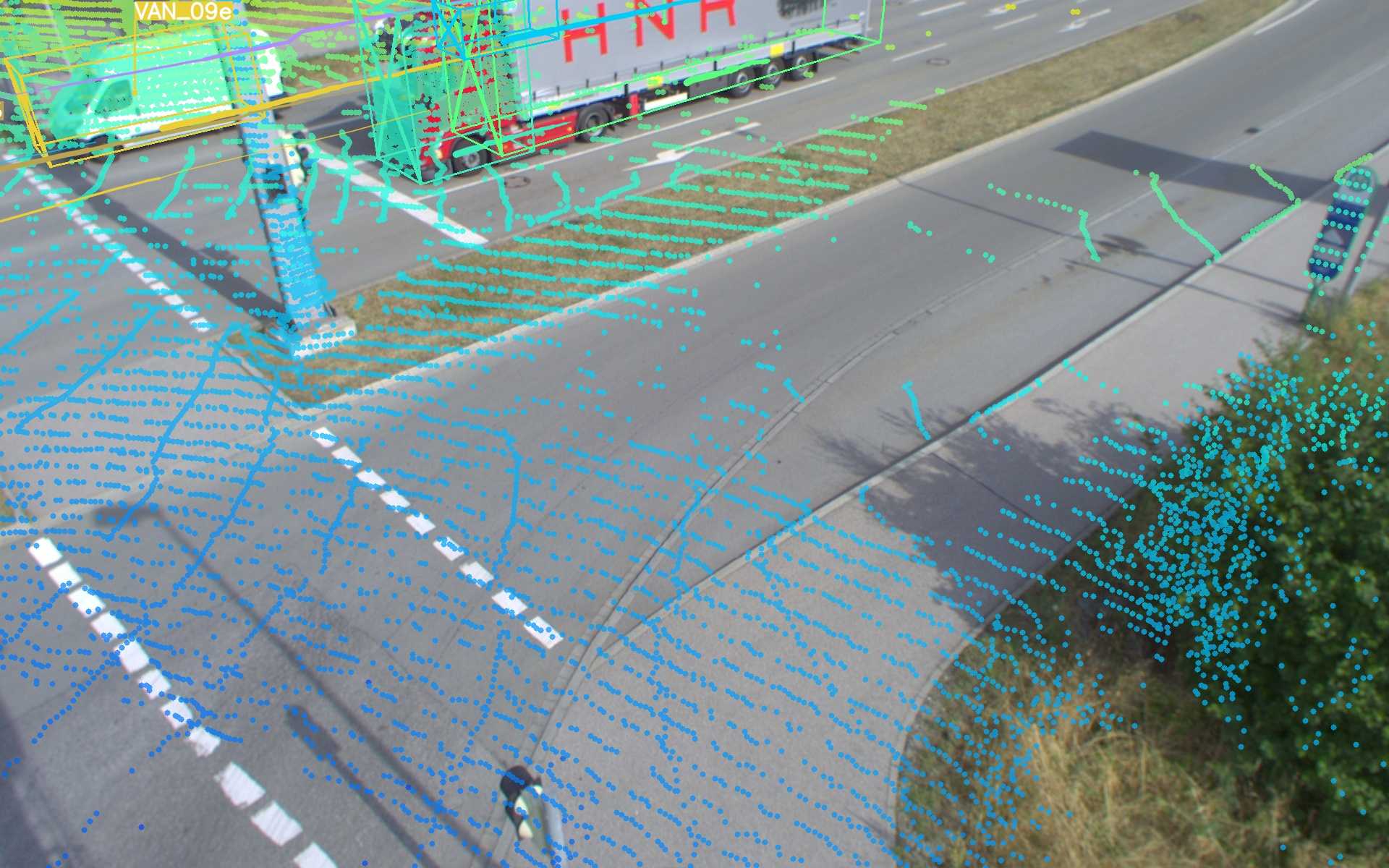}}
\endminipage
\minipage{0.14\textwidth}%
\vspace{-0.1cm}
  \fbox{\includegraphics[width=.98\linewidth]{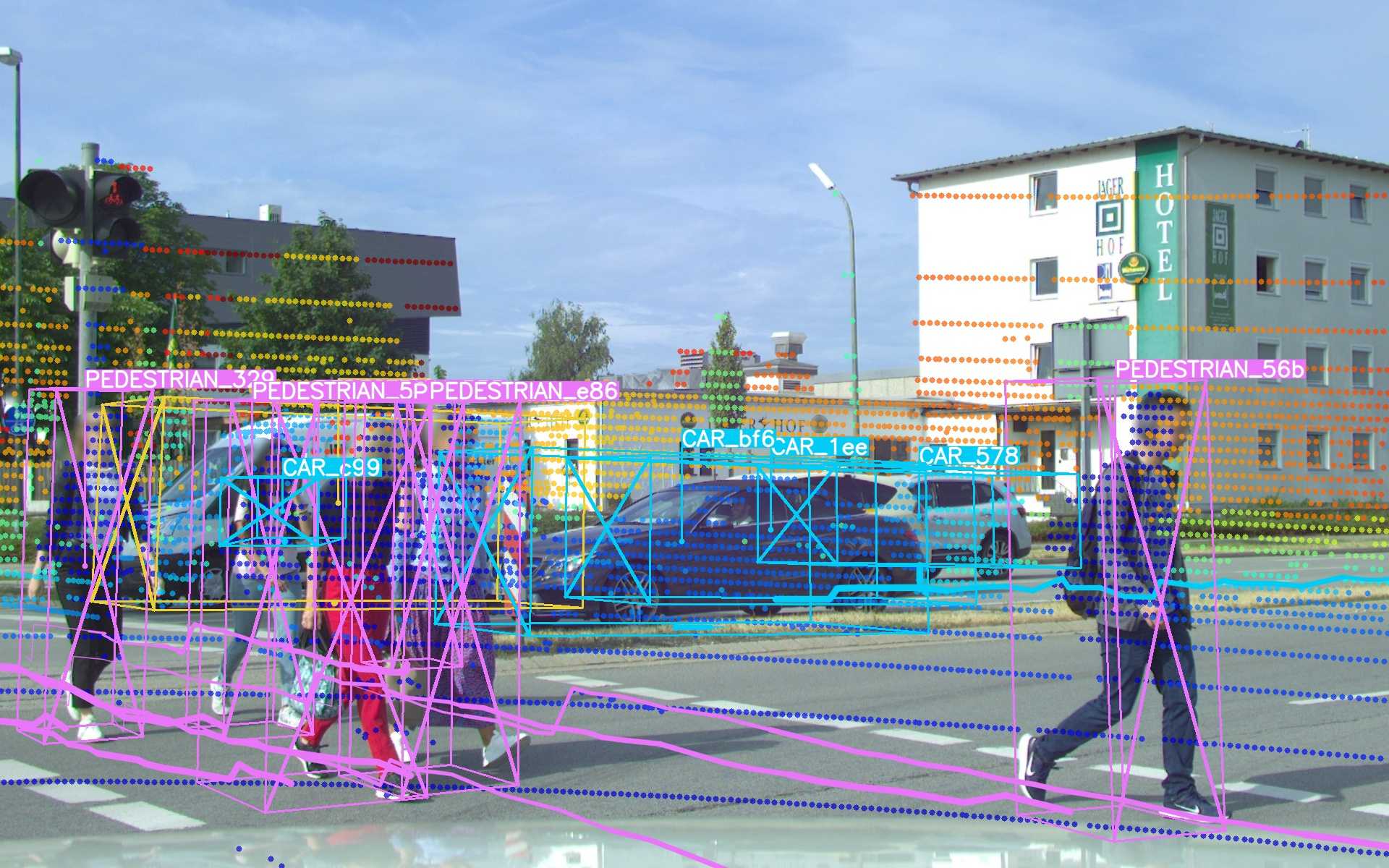}}
\endminipage
\minipage{0.14\textwidth}%
\vspace{-0.1cm}
  \fbox{\includegraphics[width=.98\linewidth]{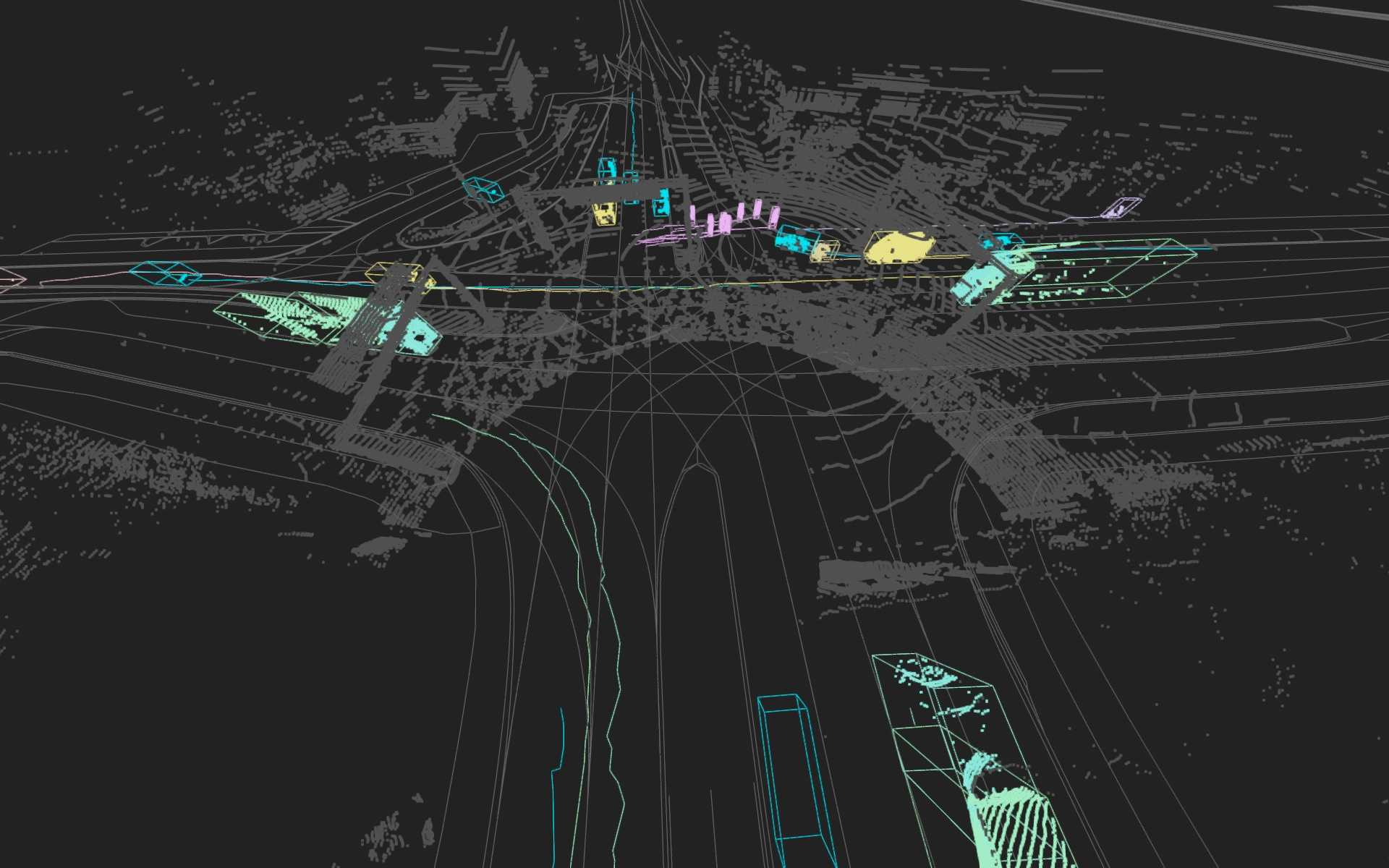}}
\endminipage
\minipage{0.14\textwidth}%
\vspace{-0.1cm}
  \fbox{\includegraphics[width=.98\linewidth]{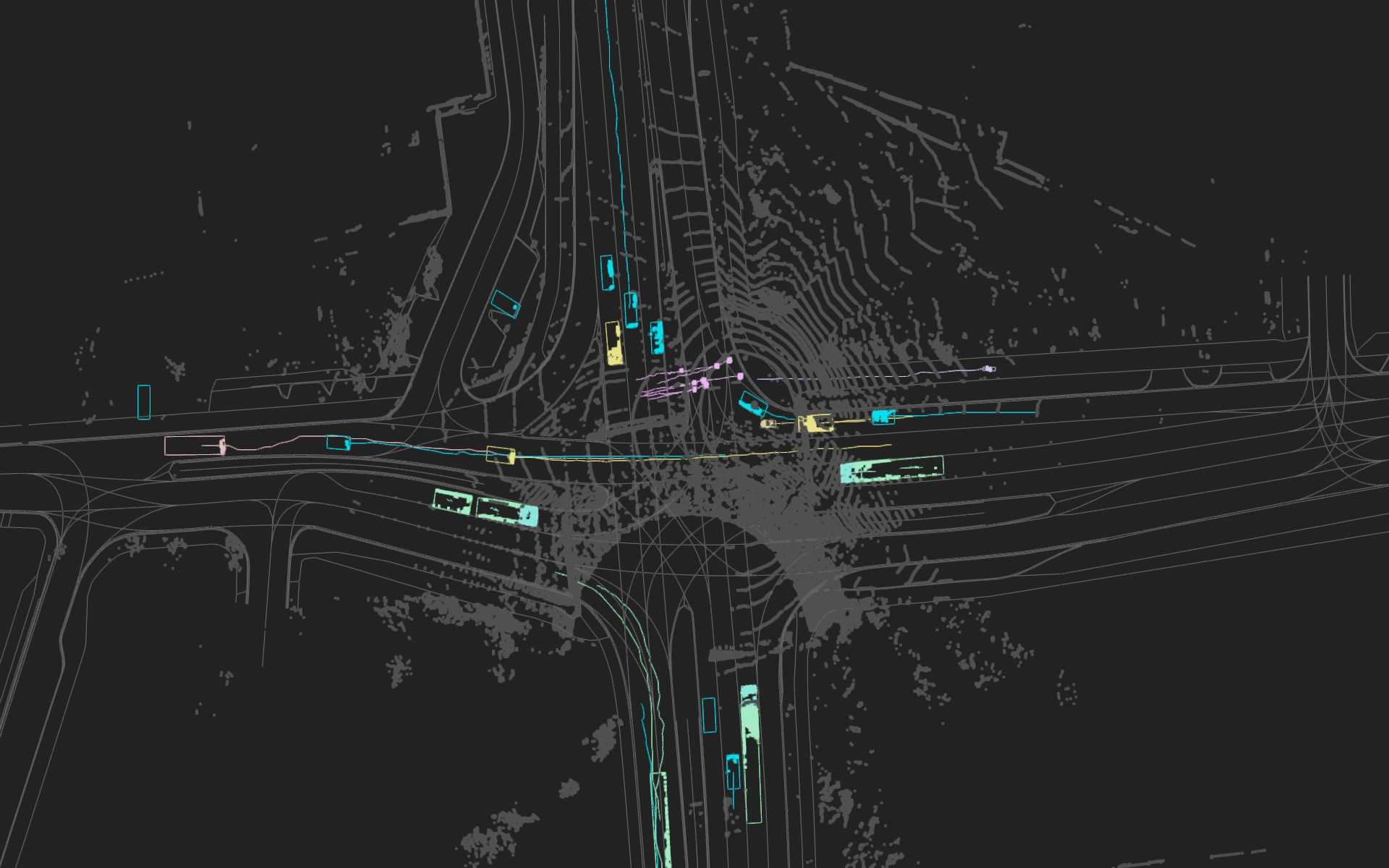}}
\endminipage
\minipage{0.14\textwidth}
\vspace{-0.1cm}
  \fbox{\includegraphics[width=.98\linewidth]{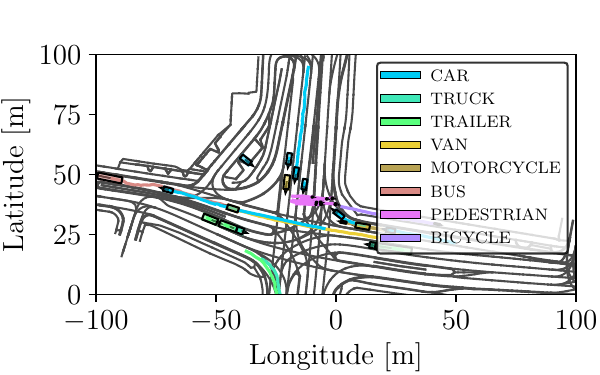}}
\endminipage
\caption{Tracking results on \textbf{drive\_42} test sequence of the \textit{TUMTraf-V2X} dataset. From top to bottom: \textit{CoopDet3D} detections, \textit{CoopDet3D} detections tracked by \textit{SORT}, \textit{CoopDet3D} detections tracked by \textit{PolyMOT}, ground truth. a-c) Tracking results projected into roadside camera images. d) Tracking results visualized in vehicle camera. e) Visualization of tracks in a point cloud and the HD map. f) Bird's eye view projection of tracks in a point cloud and the HD map. g) Visualization of all detected classes and their tracks on an HD map.}
\label{fig:tracking_qualitative_results} 
\end{figure*}

\subsection{Development kits}
\textit{OpenCOOD} \cite{xu2022opencood} is an open cooperative detection framework for autonomous driving which supports popular simulated datasets such as OPV2V \cite{xu2022opv2v} and V2XSet \cite{xu2022v2x}. Like the development kit proposed in this work, \textit{OpenCOOD} allows data preparation, pre/post-processing, and visualization. Furthermore, it also supports training and testing different benchmark models on these simulated datasets. However, the \textit{OpenCOOD} development kit only currently supports simulated datasets. Its full functionality is also limited to LiDAR-only cooperative perception, and images are only used for visualization. V2V4Real \cite{xu2023v2v4real} extends the OpenCOOD development kit, to support real-world data and additional perception tasks. Furthermore, data augmentation is also an additional feature that can be enabled when training the model.

Furthermore, the DAIR V2X \cite{yu2022dair}, proposes their own development kit, which provides data visualization and training tools. However, the access to the dataset is limited geographically.
Other development kits, such as the Nuscenes devkit \cite{nuscenes2019} and Rope3D devkit \cite{ye2022rope3d}, only support unimodal or single-view point datasets.

In comparison, our proposed development kit allows all the aforementioned functionalities in both image and LiDAR modes. Furthermore, our development kit contains modules for multi-modal cooperative data augmentation, while the model training and testing depend on the \textit{mmdetection} framework \cite{mmdet3d2020}.




\section{Point cloud registration details}

We first measure the GPS position (latitude and longitude) of the onboard LiDAR and the roadside LiDAR and convert it to UTM coordinates. For the coarse registration, we transform every 10th onboard point cloud $P_V$ to the infrastructure point cloud $P_I$ coordinate system using the initial transformation matrix shown in Eq. \ref{eq:transformation_matrix}. 
\begin{equation}
\label{eq:transformation_matrix}
T_{VI}^{0} =
\begin{bmatrix}
r_{11} & r_{12} & r_{13} & t_x\\
r_{21} & r_{22} & r_{23} & t_y\\
r_{31} & r_{32} & r_{33} & t_z\\
0 & 0 & 0 & 1
\end{bmatrix}
\end{equation}
The transformation matrix $T_{0}^{VI}$ contains as $3x3$ rotation matrix $R$ obtained by the IMU sensor and a $3x1$ translation vector $\vec{t}$ obtained by the GPS device. We then apply the point-to-point ICP for the fine registration to get an accurate V2I transformation matrix $T_{VI}$. 
\begin{equation}
    P_{VI} = P_I \oplus (P_V \cdot T_{VI})
\end{equation}
Fig. \ref{fig:point_cloud_registration_results} shows the point cloud registration results in two colors. The vehicle point cloud is displayed in orange, and the infrastructure point cloud is displayed in blue. We get an RMSE value of 0.02 m, which shows how well the point clouds were registered.

\section{Dataset labeling}

We provide a web-based labeling platform \textit{3D BAT} v24.3.2 to facilitate the development of V2X perception. It provides a one-click annotation feature to fit an oriented bounding box to a 3D object. It contains an interpolation mode that reduces the labeling time significantly and lets the user visualize the HD Map, which is highly beneficial for positioning 3D box labels accurately within lanes. The user interface of \textit{3D BAT} v24.3.2 is split into two main views: the upper portion displays the camera images captured by both infrastructure and vehicle-mounted cameras, while the lower portion renders the registered point cloud data obtained from the roadside and onboard LiDARs. The annotator first navigates the point cloud to identify objects of interest. Upon selecting an object, boxes are enclosed around it. These boxes are color-coded according to the object category (e.g., car, truck, trailer, van, motorcycle, bus, pedestrian, bicycle, and others) to allow for easy differentiation. After placing the 3D bounding box, they cross-check the predicted 2D bounding boxes in the camera images to ensure their correctness. Additional attributes can be modified and specified for each object on the right-hand side.

\section{Implementation details}
Here, we provide detailed information about the training schedule and the hyperparameters. We train our \textit{CoopDet3D} model in two stages. In stage one, we pre-train the \textit{PointPillars} backbone on onboard and roadside point clouds for 20 epochs. Then, in stage two, we finetune the model for eight further epochs on cooperative camera and LiDAR data. For the detection head, we use \textit{TransFusion} \cite{bai2022transfusion} to obtain 3D bounding box predictions. To calculate the matching cost $C_{match}$, it uses a weighted binary cross entropy loss $L_{cls}$, a weighted $L_1$ loss for the 3D box regression $\mathcal{L}_{reg}$, and a weighted IoU loss $\mathcal{L}_{IoU}$ \cite{zhou2019objects} (see Eq. \ref{eq:matching_cost}). 
\begin{equation}
\label {eq:matching_cost}
C_{match} = \lambda _1 \mathcal{L}_{cls} + \lambda _2 \mathcal{L}_{reg} + \lambda _3 \mathcal{L}_{IoU},
\end{equation}
where: $\lambda_1$, $\lambda_2$, $\lambda_3$ are the coefficients for the individual cost terms. Given all matched pairs, a focal loss \cite{lin2017focal} is computed for the final classification. A penalty-reduced focal loss \cite{yin2021center} is used for the heatmap prediction.

We use the following hyperparameters for training: the \textit{AdamW} optimizer with a learning rate of $1 \times e^{-4}$ and a weight decay of 0.01, a batch size of 4, a dropout rate of 0.1, the \textit{ReLU} activation function, and cyclic momentum.
We use the BEV encoder to transform the image into a BEV representation of $512\times512$ size. The point clouds are cropped to the following range: $[-75,75]~m$ for the X and Y axis, and $[-8,0]~m$ for the Z axis. For training, we use 3 x NVIDIA RTX 3090 GPUs.

\begin{table}[t]
  \caption{Evaluation results ($mAP_{BEV}$ and $mAP_{3D}$) of \textit{CoopDet3D} on our \textit{TUMTraf-V2X} test set in south1 FOV.}
  \label{tbl:quantitativeResultsS1}
  \centering
  \resizebox{\columnwidth}{!}{%
  \begin{tabular}{ll|rrrrrN}
    \hline
    \multicolumn{2}{c|}{\textbf{Config.}} & {$\mathbf{mAP_{BEV}\uparrow}$} & \multicolumn{4}{c}{$\mathbf{mAP_{3D}\uparrow}$} \\
    \textbf{Domain} & \textbf{Modality} & & \textbf{Easy$\uparrow$} & \textbf{Moderate$\uparrow$} & \textbf{Hard$\uparrow$} & \textbf{Avg.$\uparrow$}\\
    \hline
    Vehicle & Camera & 46.83 & 39.31 & 12.42 & 4.29 & 35.02 \\
    Vehicle & LiDAR & 85.33 & 77.30 & 31.26 & 53.76 & 76.68 \\
    Vehicle & Cam+LiDAR & 84.90 & 77.29 & 34.29 & 39.71 & 76.19 \\
    Infra. & Camera & 61.98 & 41.13 & 15.64 & 1.35 & 37.09 \\
    Infra. & LiDAR & 92.86 & 82.16 & 45.14 & 46.56 & 81.07 \\
    Infra. & Cam+LiDAR & 92.92 & \textbf{85.43} & 49.10 & 49.56 & \underline{84.13} \\
    Coop. & Camera & 68.94 & 52.04 & 29.26 & 10.28 & 49.81 \\
    Coop. & LiDAR & \underline{93.93} & \underline{84.61} & \underline{50.00} & \underline{53.78} & \textbf{84.15} \\
    \rowcolor{Gray}
    Coop. & Cam+LiDAR & \textbf{94.22} & 84.50 & \textbf{51.67} & \textbf{55.14} & 84.05 \\
    \hline\\[-8pt]
  \end{tabular}
  }
\end{table}

\section{Metrics}
This section presents the evaluation metrics for two main tasks in V2X perception, i.e. the \textit{Cooperative 3D Object Detection} (C3DOD) task and the \textit{Cooperative Multiple Object Tracking} (CMOT) task. Notably, we  adopt the mainstream metrics for the cooperative perception evaluation to make fair comparisons with the vehicle-only and infrastructure-only algorithms. \\
\setlength{\fboxsep}{0pt}%
\setlength{\fboxrule}{1pt}%
\begin{figure*}[h!]
\centering
\minipage{0.25\textwidth}
  \fbox{\includegraphics[width=.98\linewidth]{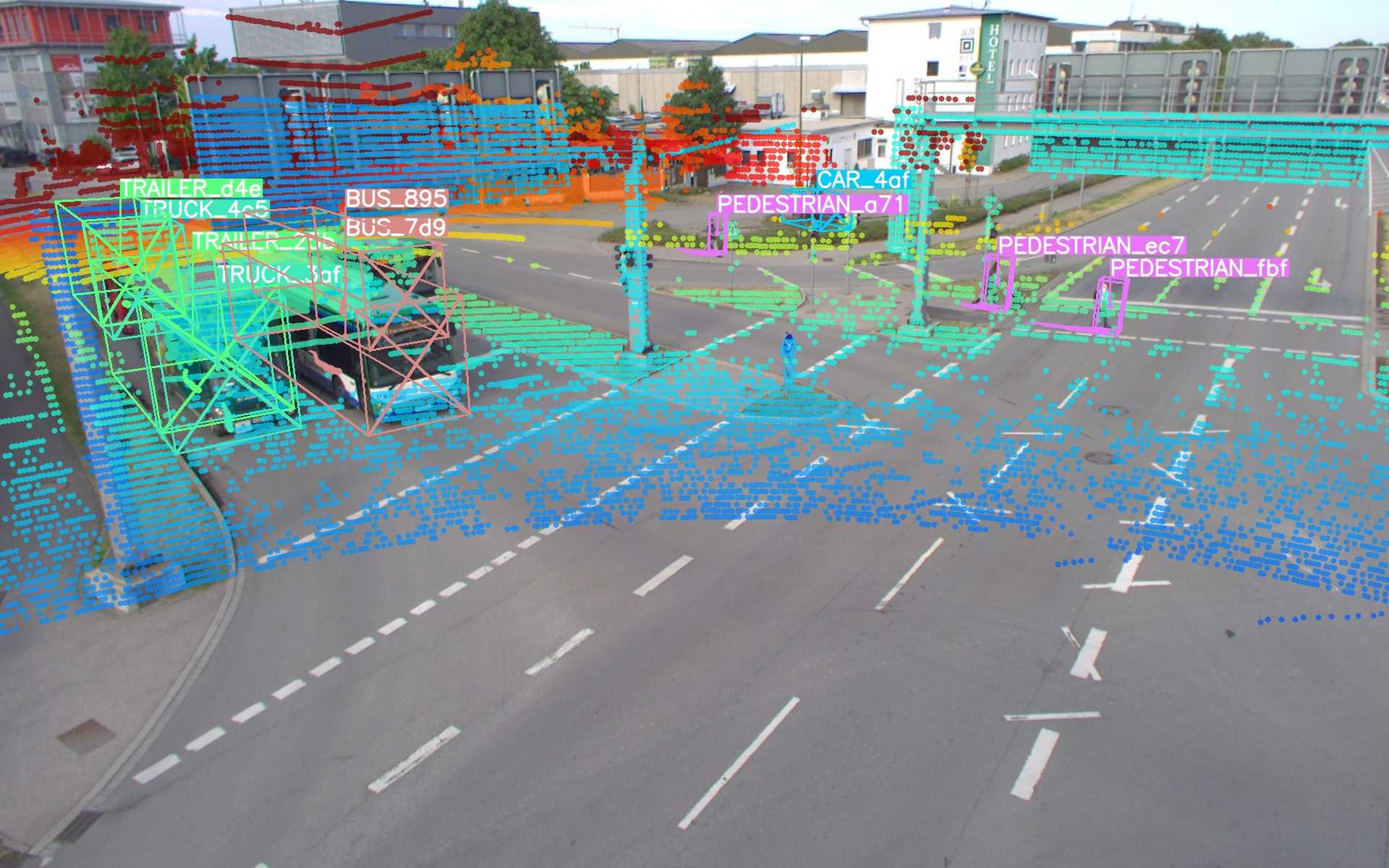}}
\endminipage
\minipage{0.25\textwidth}
  \fbox{\includegraphics[width=.98\linewidth]{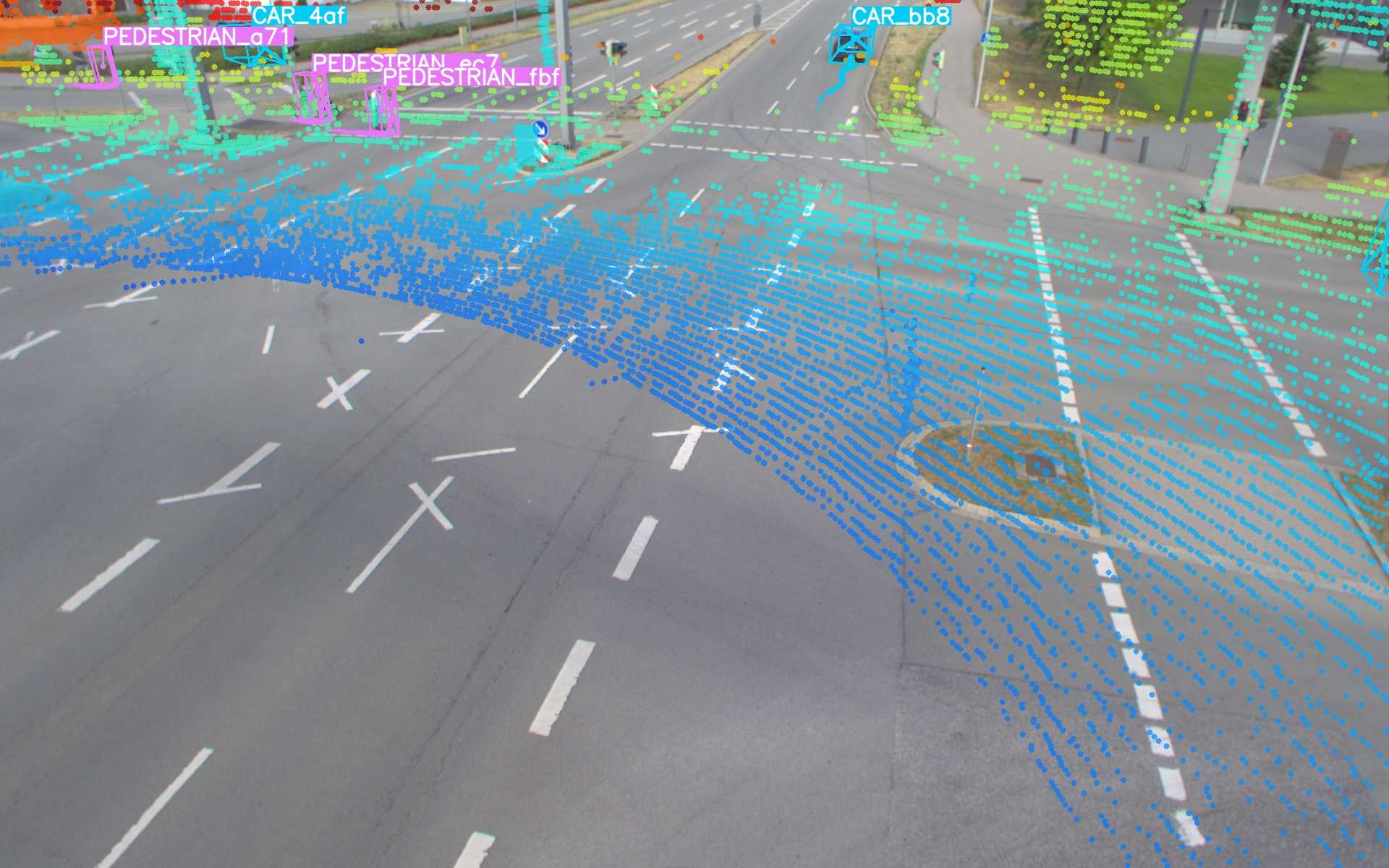}}
\endminipage
\minipage{0.25\textwidth}%
  \fbox{\includegraphics[width=.98\linewidth]{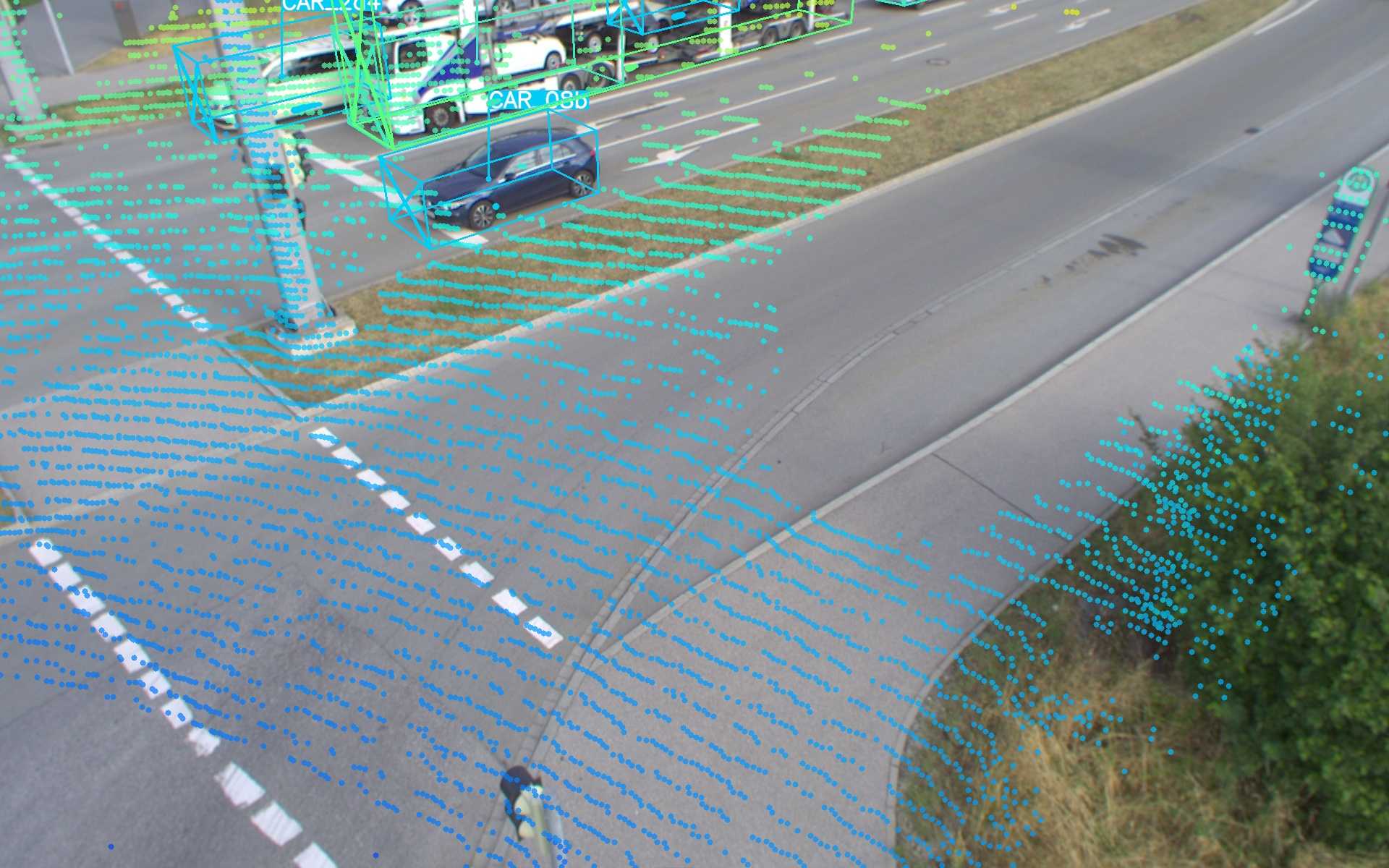}}
\endminipage
\minipage{0.25\textwidth}
  \fbox{\includegraphics[width=.98\linewidth]{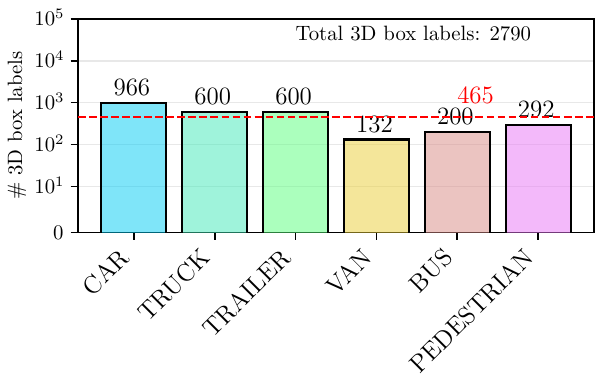}}
\endminipage\\
\vspace{-0.07cm}
\minipage{0.25\textwidth}
  \fbox{\includegraphics[width=.98\linewidth]{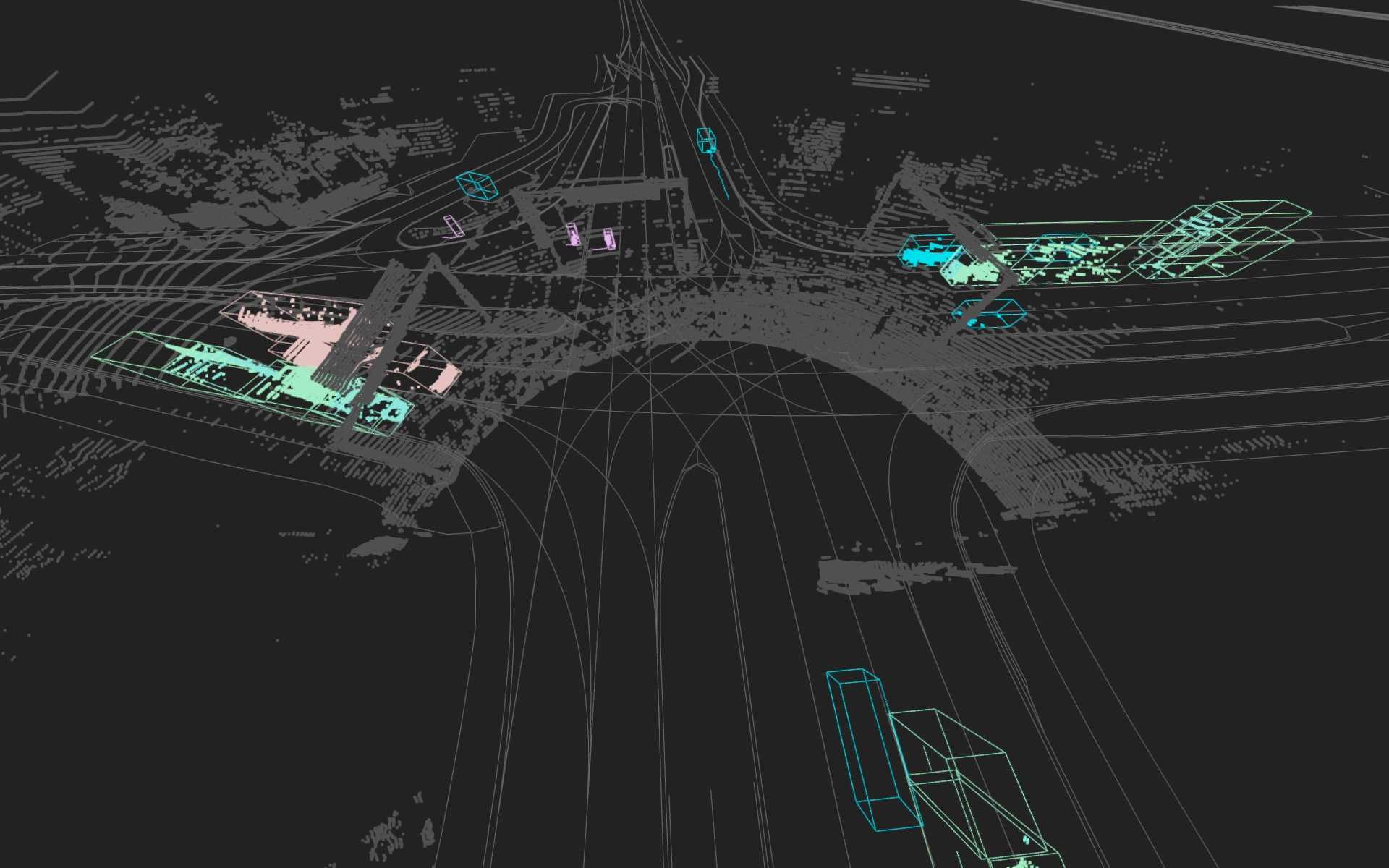}}
\endminipage
\minipage{0.25\textwidth}%
  \fbox{\includegraphics[width=.98\linewidth]{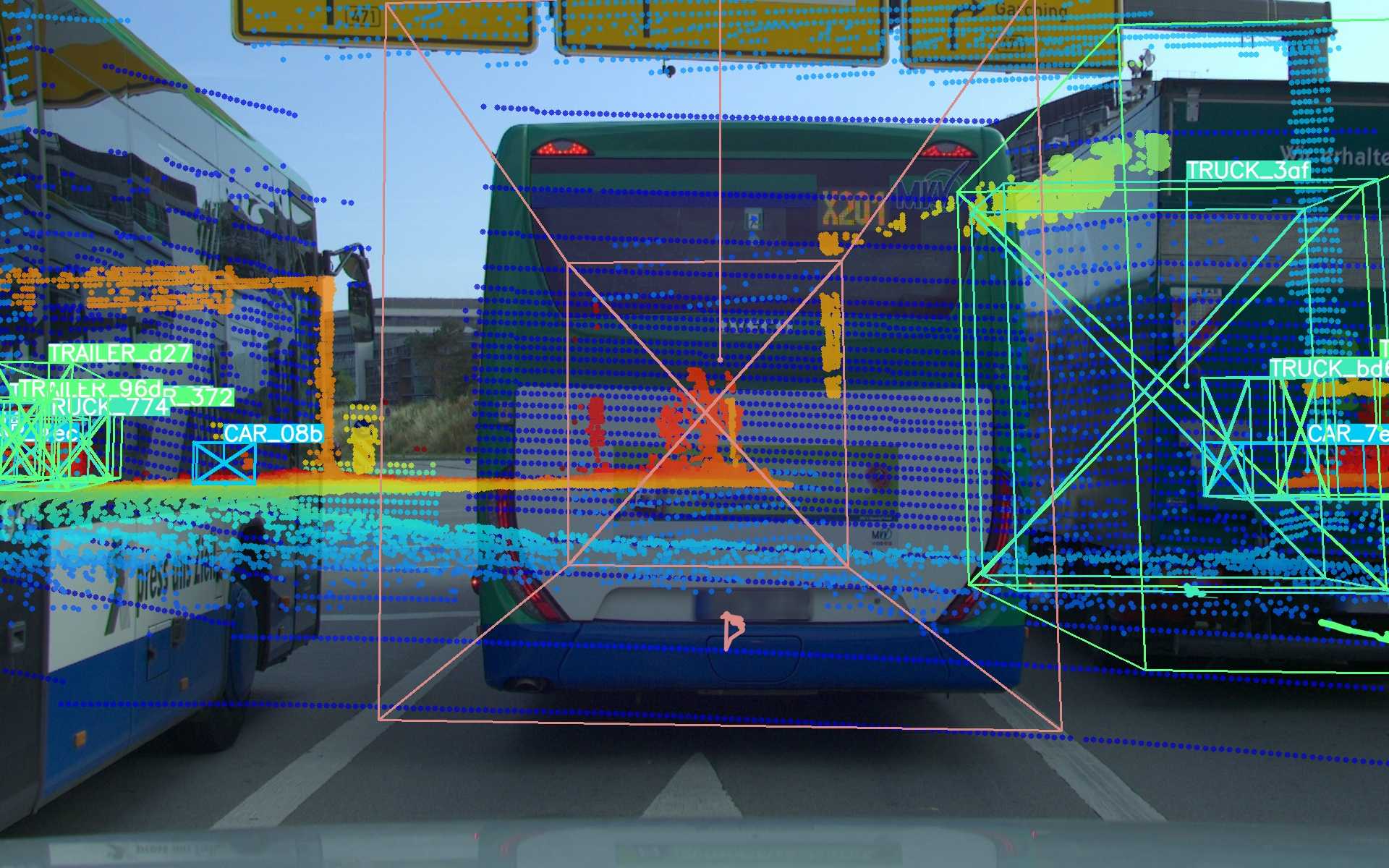}}
\endminipage
\minipage{0.25\textwidth}%
  \fbox{\includegraphics[width=.98\linewidth]{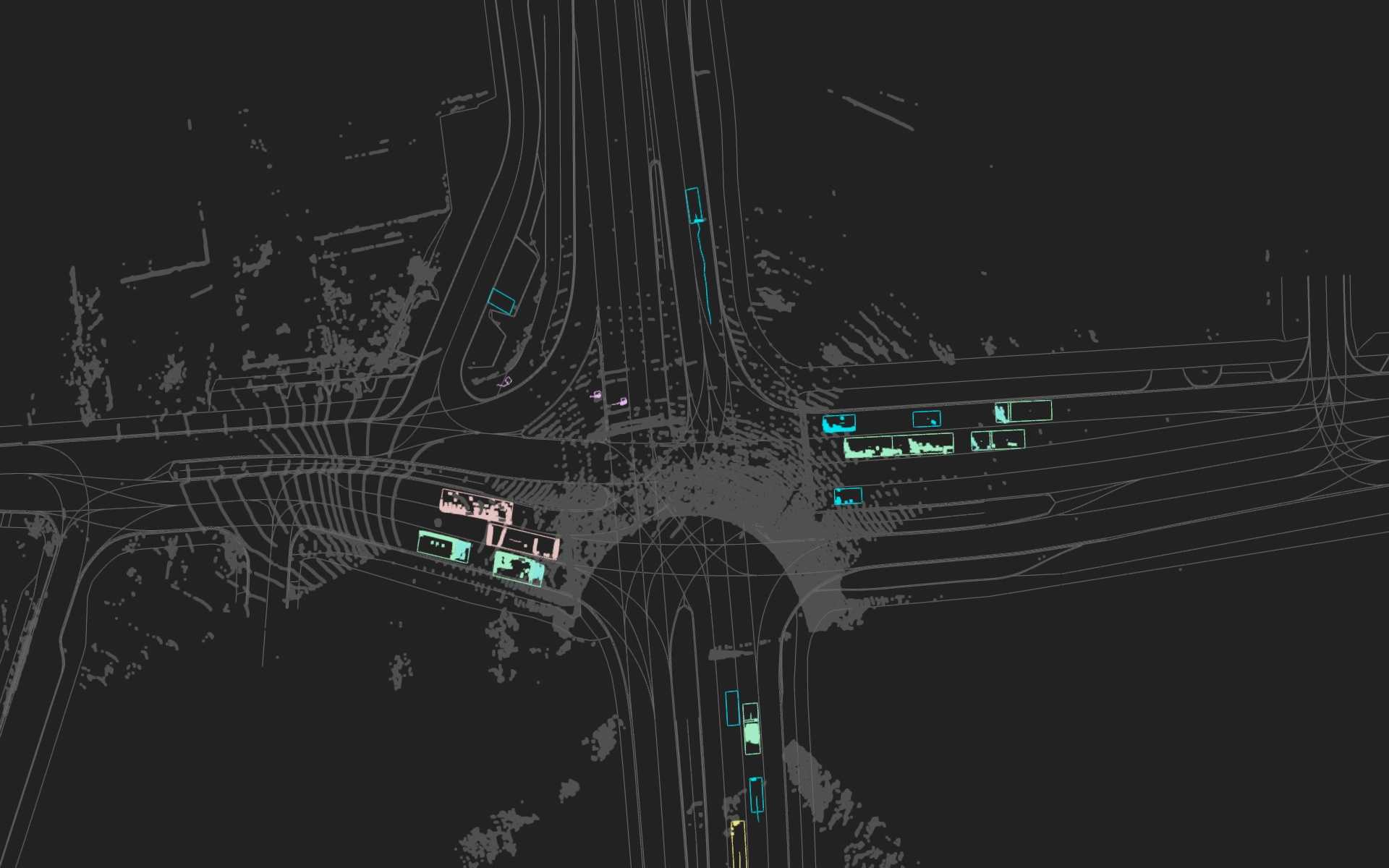}}
\endminipage
\minipage{0.25\textwidth}%
  \fbox{\includegraphics[width=.98\linewidth]{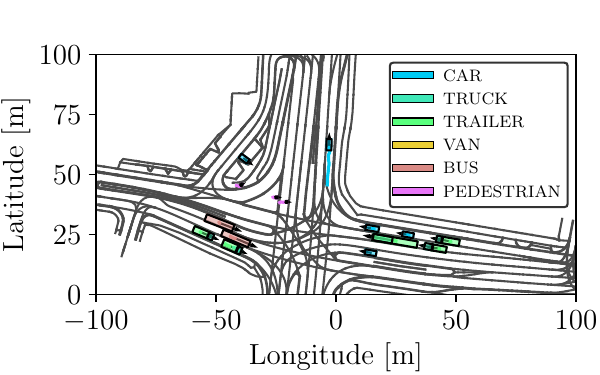}}
\endminipage
\caption{Visualization of \textbf{drive\_07} of the \textit{TUMTraf-V2X} dataset. In this example, the ego vehicle is occluded by two busses and two large trucks. The roadside sensors enhance the perception range, making traffic participants behind the buses visible. In total, this ten-second-long sequence contains 2,790 labeled 3D objects during the daytime.}
\label{fig:dataset_visualization_drive_07} 
\end{figure*}
\vspace{1cm}
\setlength{\fboxsep}{0pt}%
\setlength{\fboxrule}{1pt}%
\begin{figure*}[h!]
\centering
\minipage{0.25\textwidth}
  \fbox{\includegraphics[width=.98\linewidth]{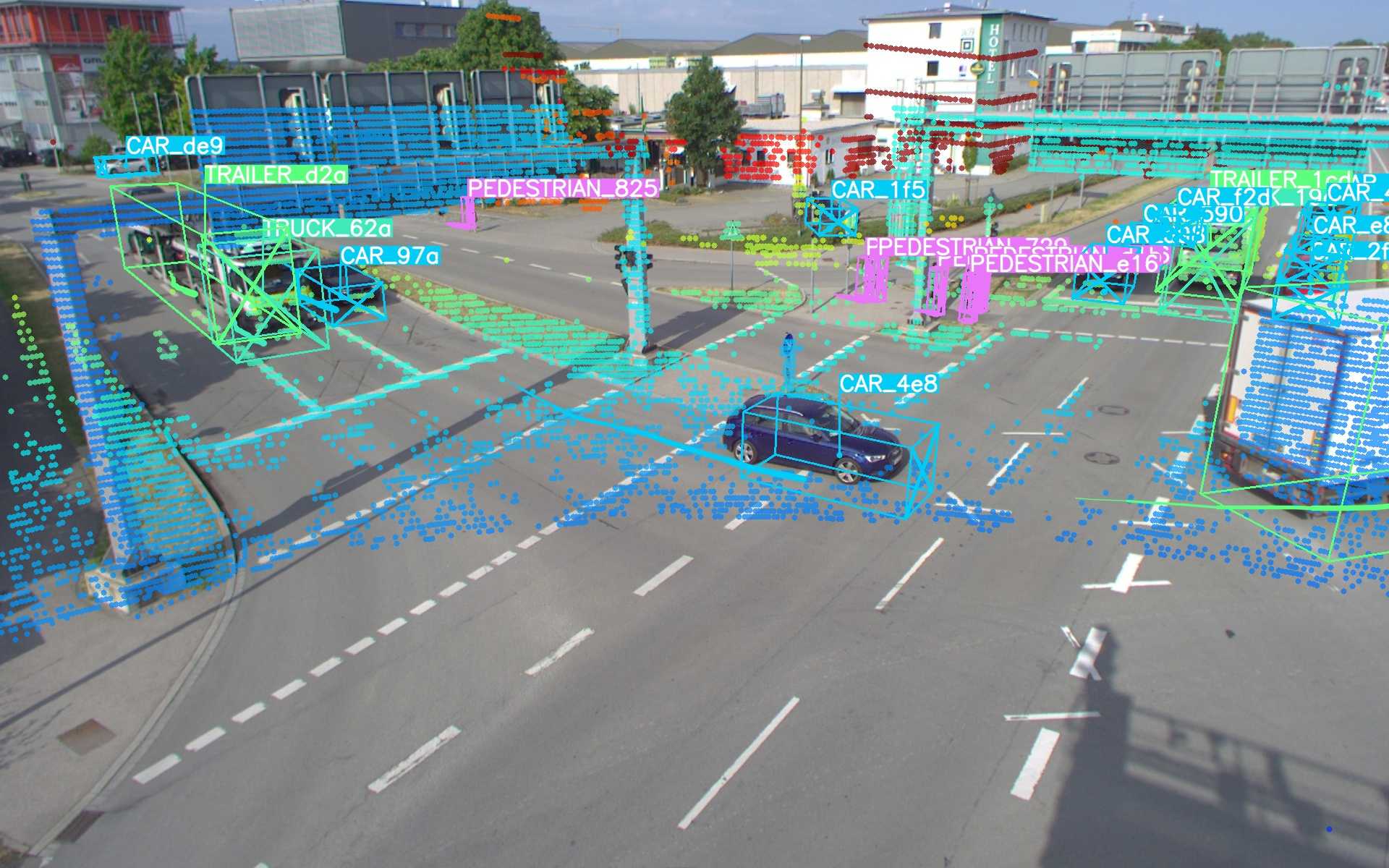}}
\endminipage
\minipage{0.25\textwidth}
  \fbox{\includegraphics[width=.98\linewidth]{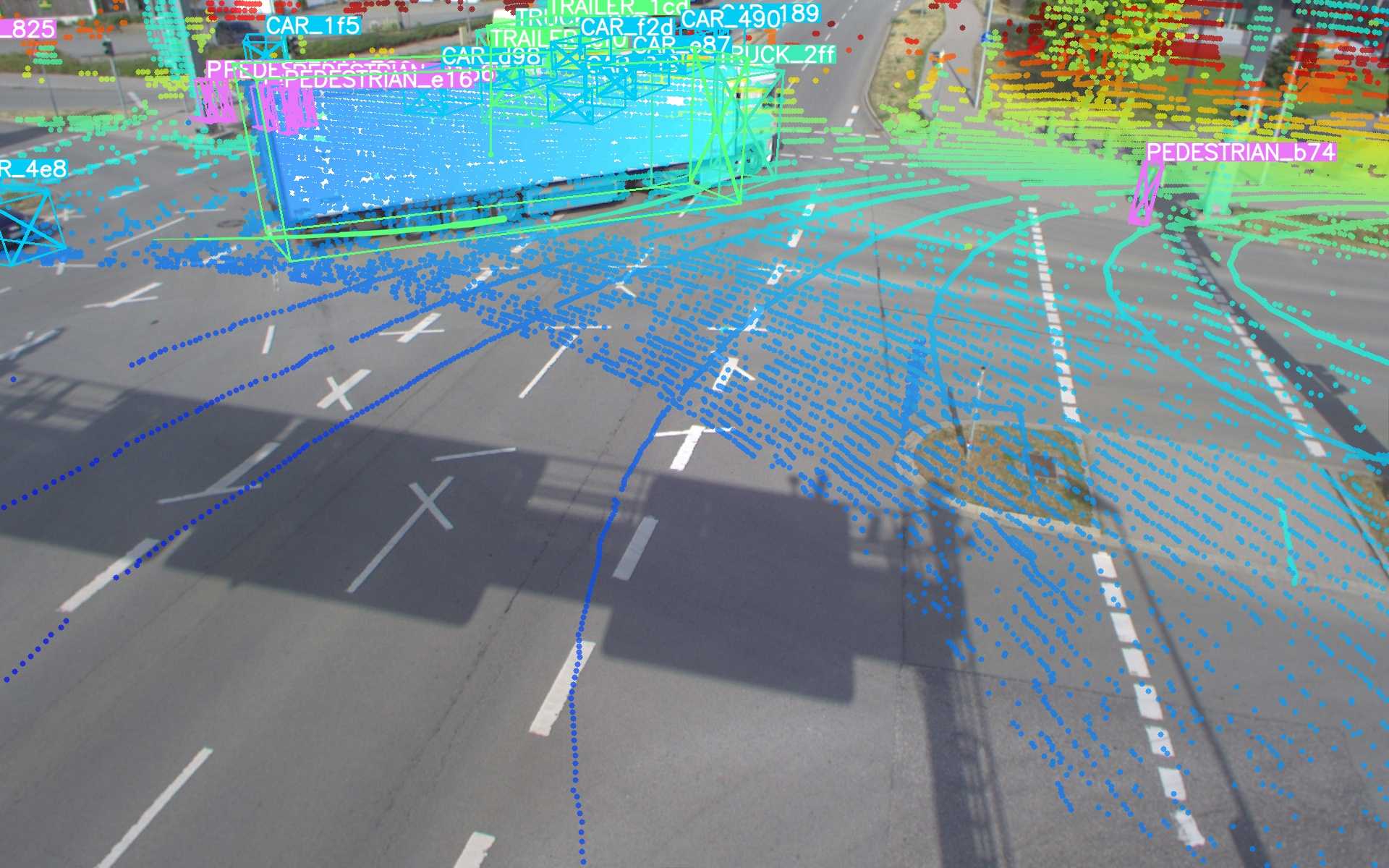}}
\endminipage
\minipage{0.25\textwidth}%
  \fbox{\includegraphics[width=.98\linewidth]{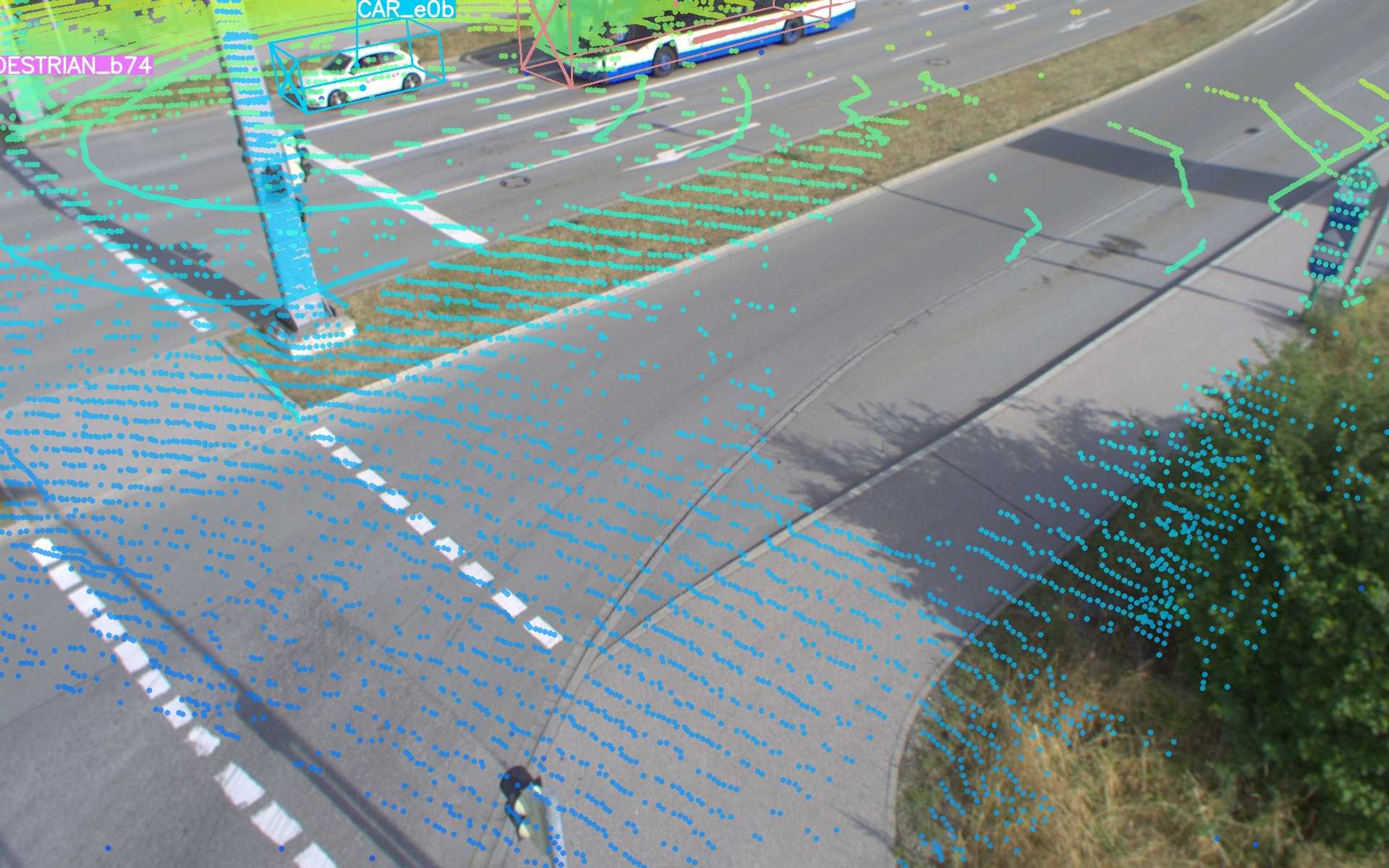}}
\endminipage
\minipage{0.25\textwidth}
  \fbox{\includegraphics[width=.98\linewidth]{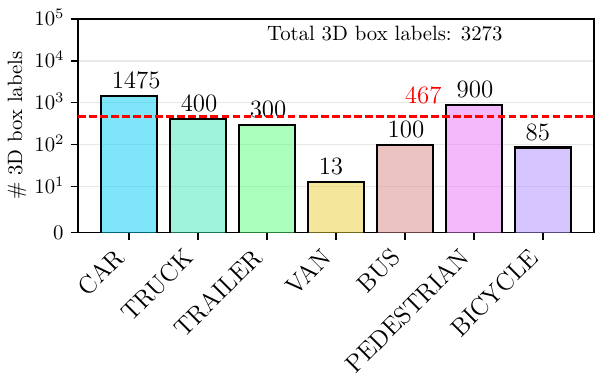}}
\endminipage\\
\vspace{-0.07cm}
\minipage{0.25\textwidth}
  \fbox{\includegraphics[width=.98\linewidth]{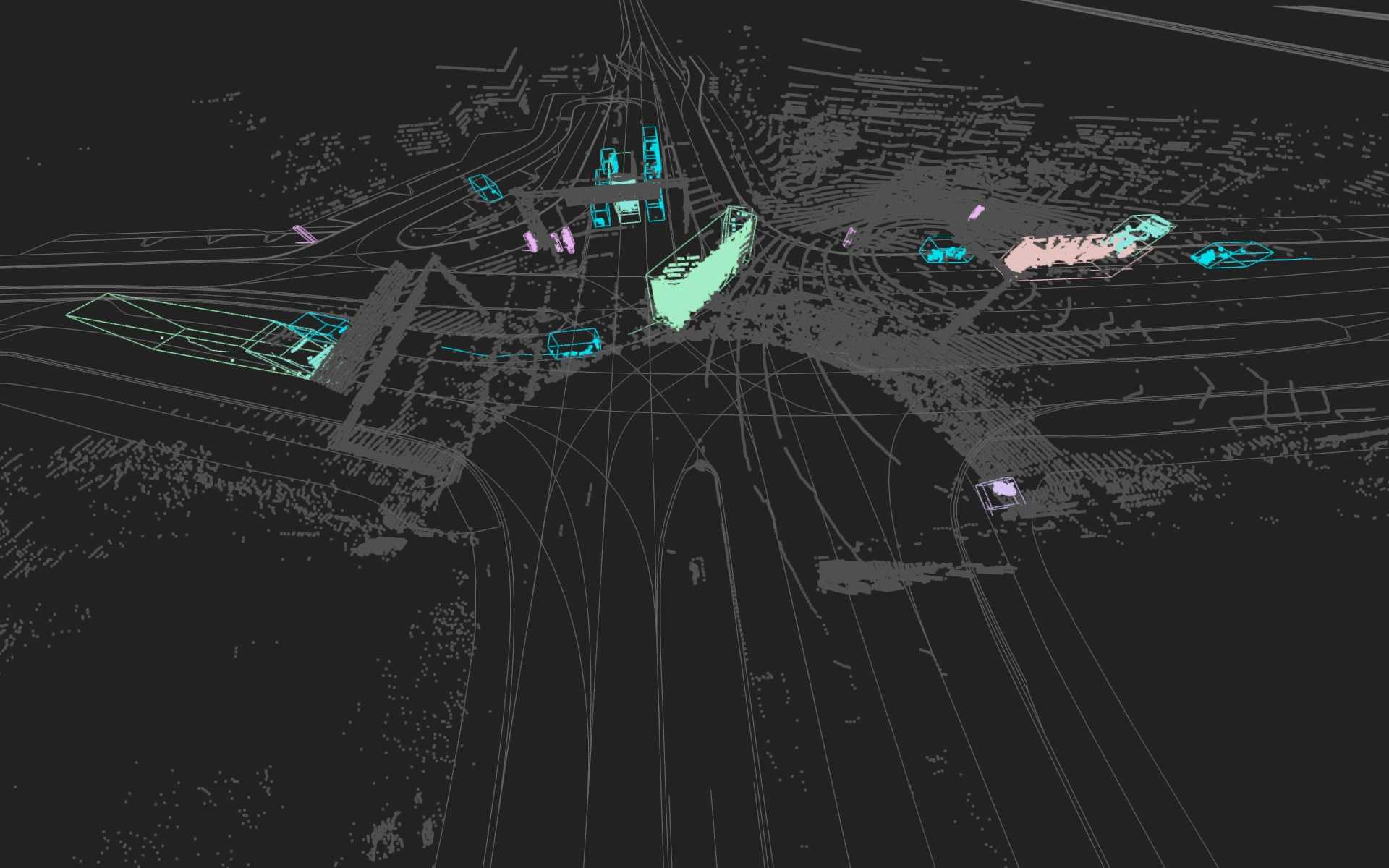}}
\endminipage
\minipage{0.25\textwidth}%
  \fbox{\includegraphics[width=.98\linewidth]{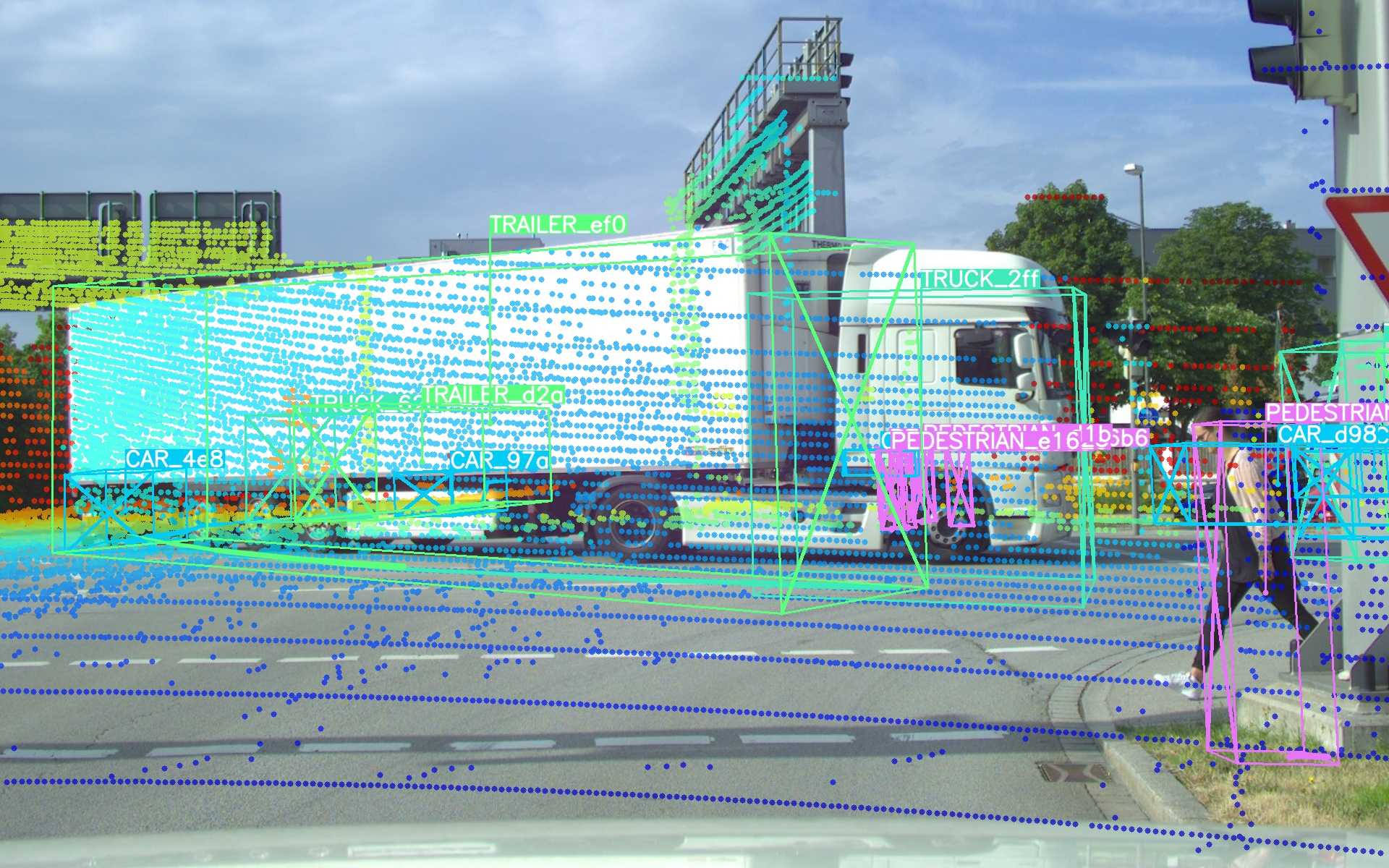}}
\endminipage
\minipage{0.25\textwidth}%
  \fbox{\includegraphics[width=.98\linewidth]{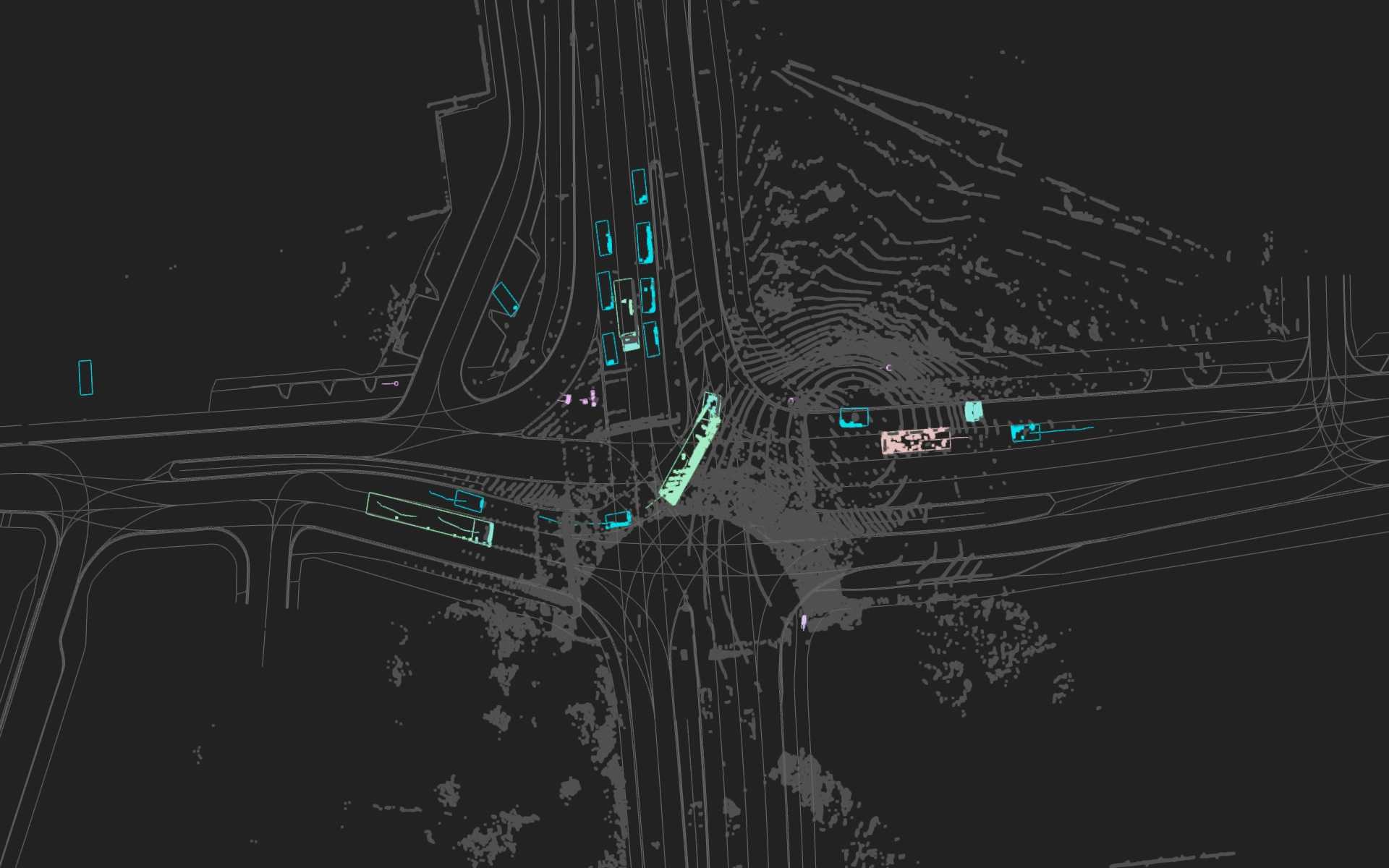}}
\endminipage
\minipage{0.25\textwidth}%
  \fbox{\includegraphics[width=.98\linewidth]{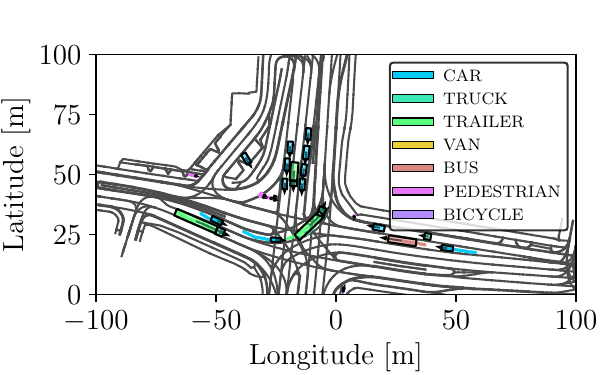}}
\endminipage
\caption{Visualization of \textbf{drive\_12} of the \textit{TUMTraf-V2X} dataset. This sequence with 3,273 3D boxes shows multiple occlusion scenarios. In one scenario a truck is occluding multiple pedestrians. The roadside sensors can perceive the objects behind the truck so that the ego vehicle becomes aware of them.}
\label{fig:dataset_visualization_drive_12} 
\end{figure*}
\setlength{\fboxsep}{0pt}%
\setlength{\fboxrule}{1pt}%
\begin{figure*}[h!]
\centering
\minipage{0.25\textwidth}
  \fbox{\includegraphics[width=.98\linewidth]{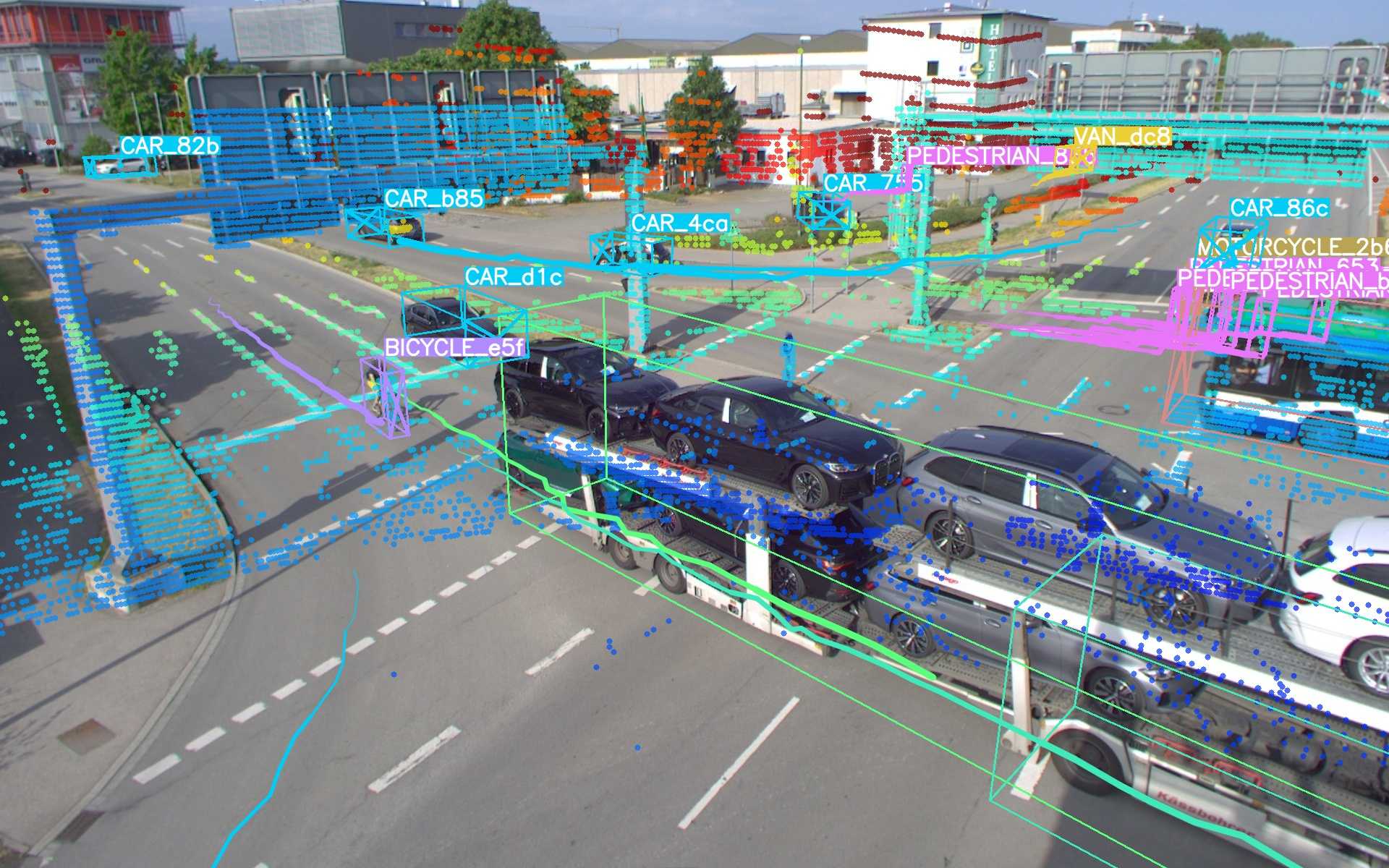}}
\endminipage
\minipage{0.25\textwidth}
  \fbox{\includegraphics[width=.98\linewidth]{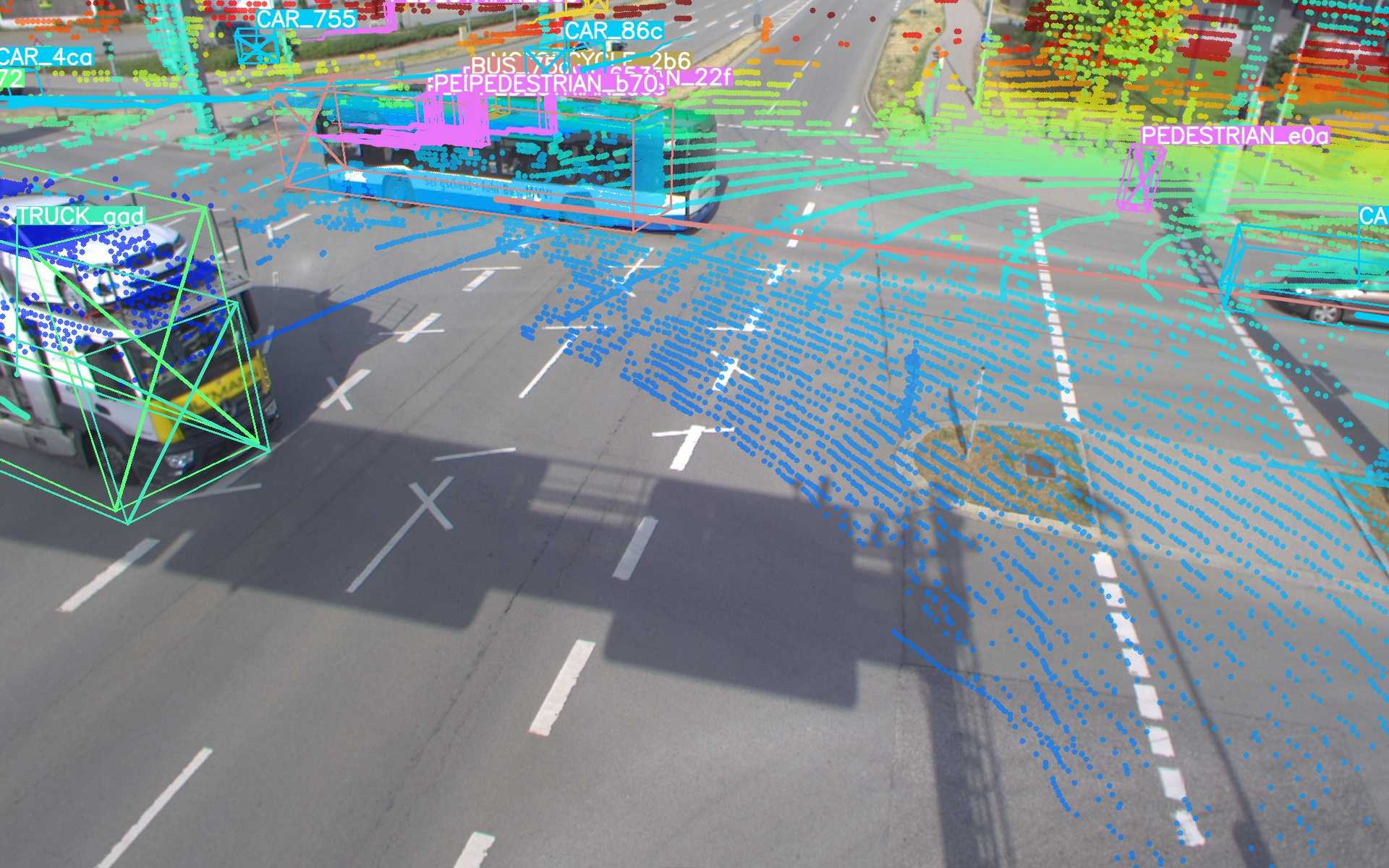}}
\endminipage
\minipage{0.25\textwidth}%
  \fbox{\includegraphics[width=.98\linewidth]{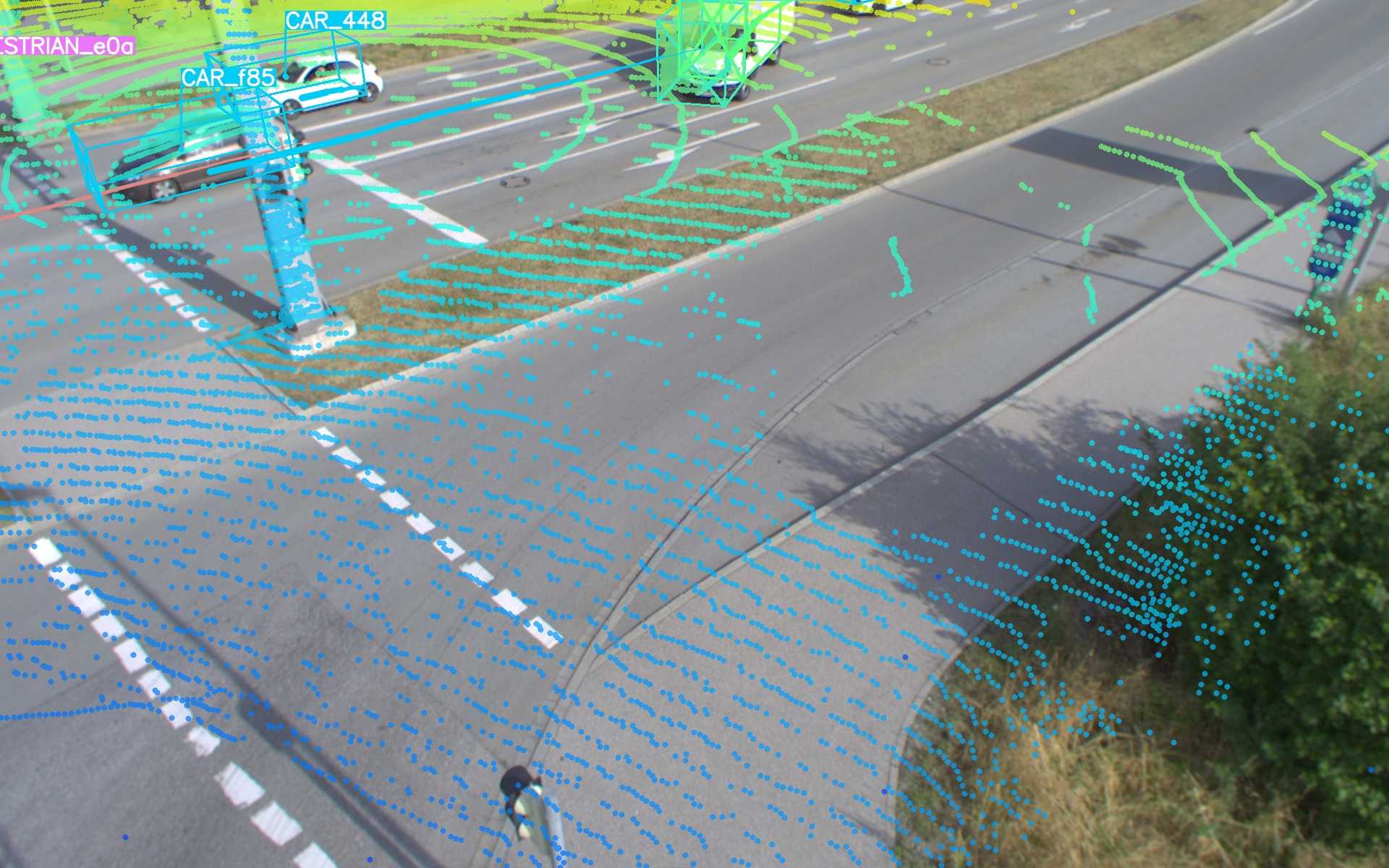}}
\endminipage
\minipage{0.25\textwidth}
  \fbox{\includegraphics[width=.98\linewidth]{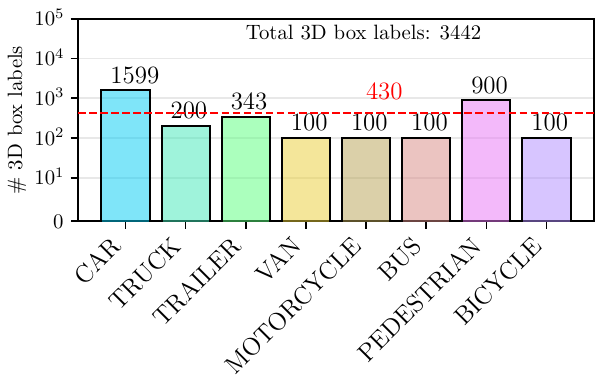}}
\endminipage\\
\vspace{-0.07cm}
\minipage{0.25\textwidth}
  \fbox{\includegraphics[width=.98\linewidth]{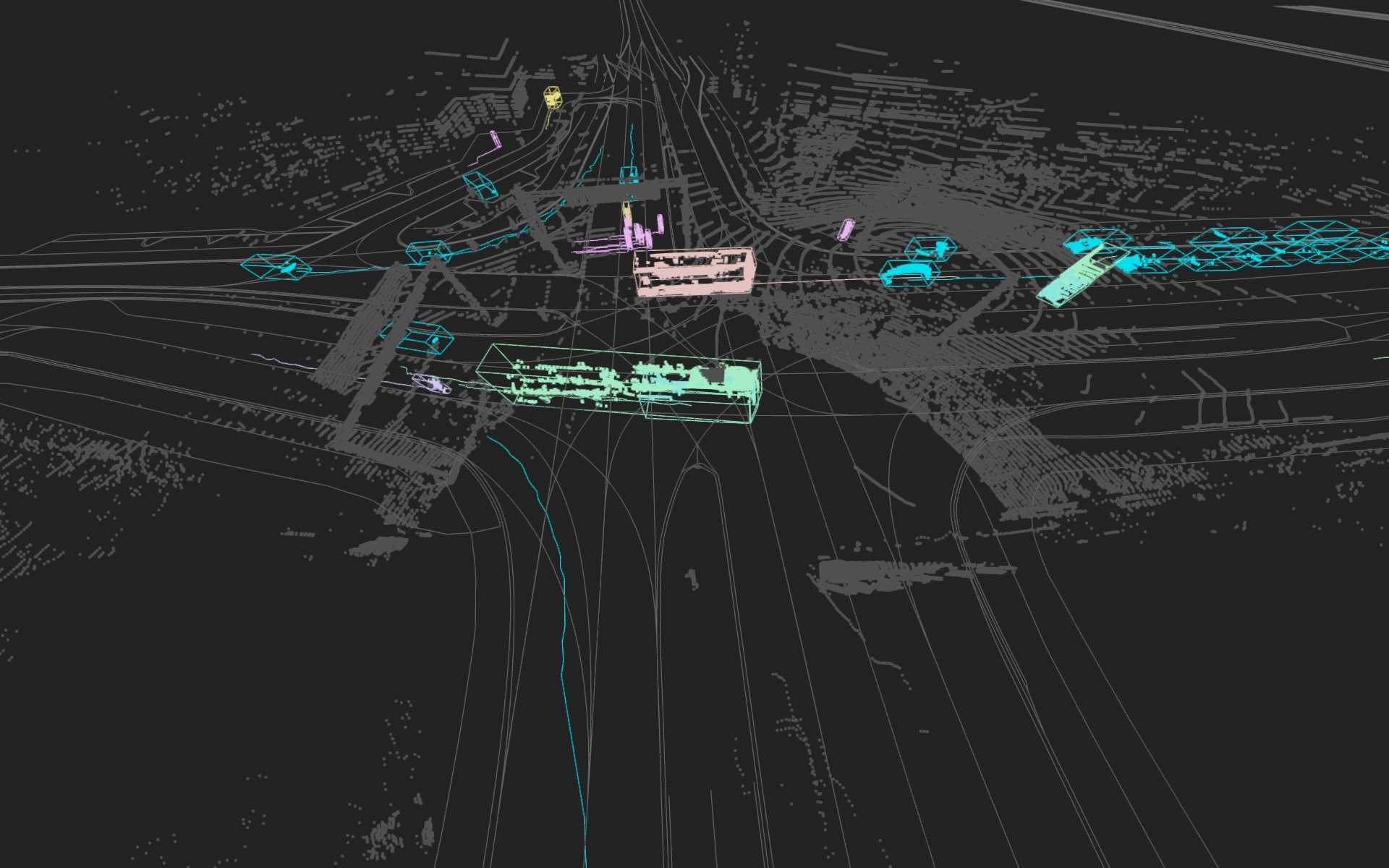}}
\endminipage
\minipage{0.25\textwidth}%
  \fbox{\includegraphics[width=.98\linewidth]{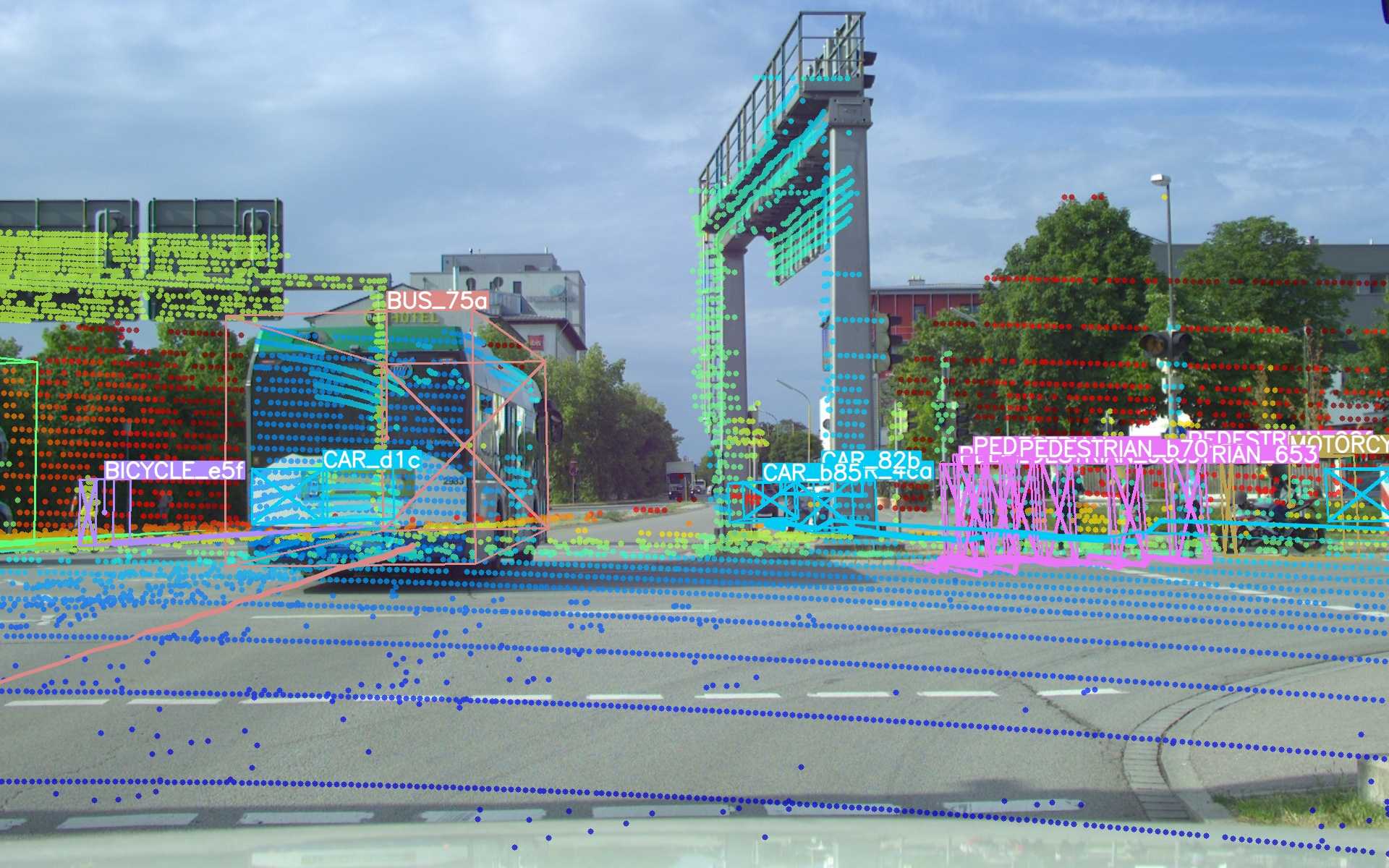}}
\endminipage
\minipage{0.25\textwidth}%
  \fbox{\includegraphics[width=.98\linewidth]{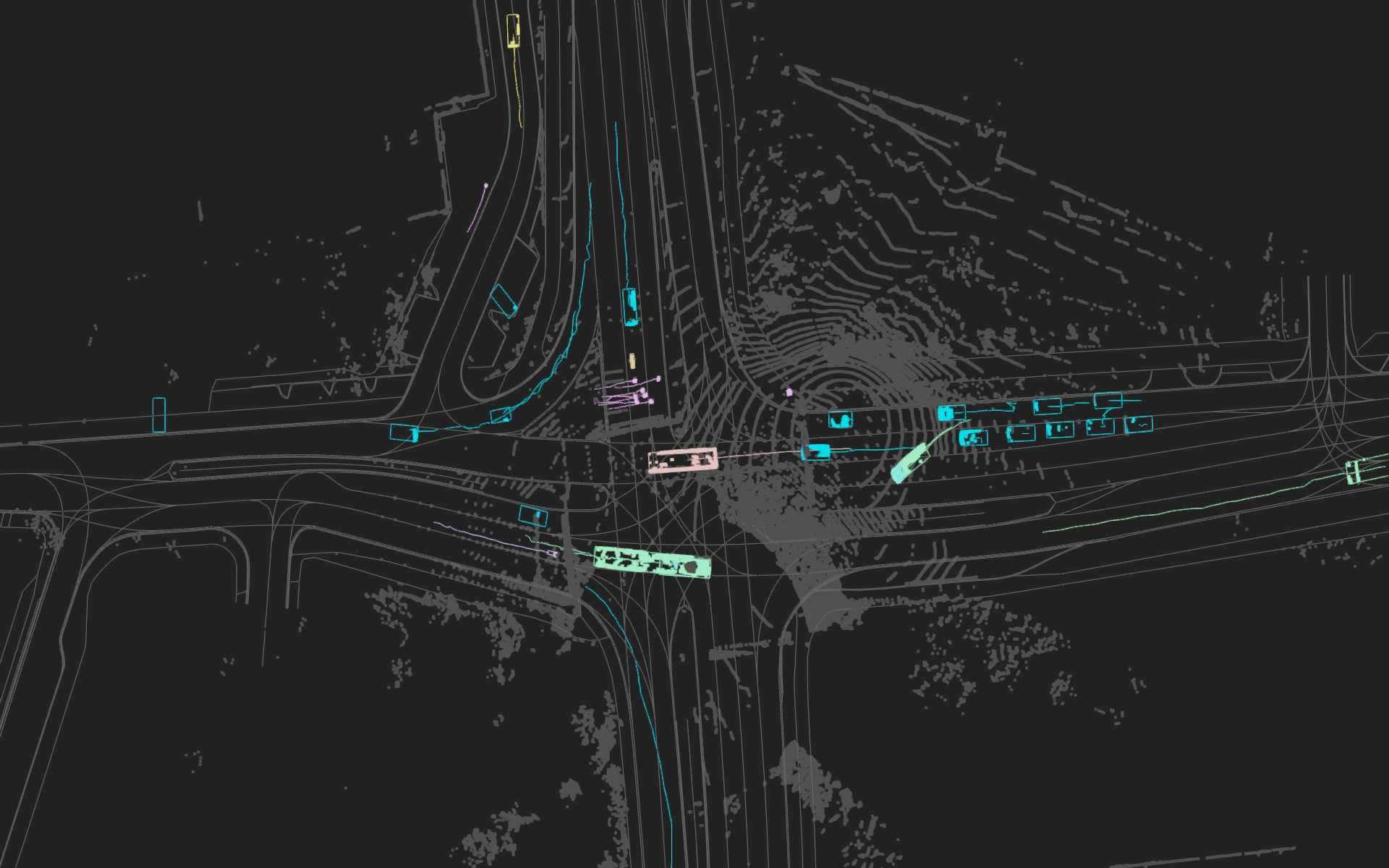}}
\endminipage
\minipage{0.25\textwidth}%
  \fbox{\includegraphics[width=.98\linewidth]{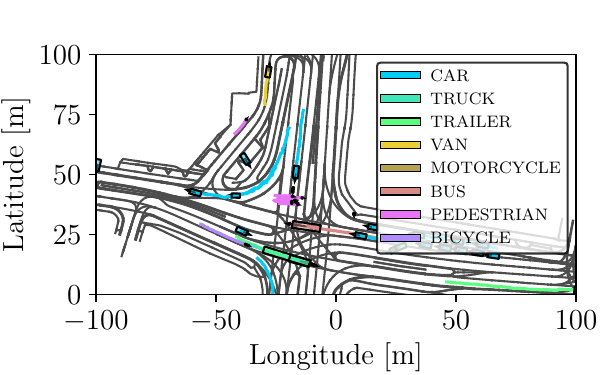}}
\endminipage
\caption{Visualization of \textbf{drive\_15} of the \textit{TUMTraf-V2X} dataset. In this drive, a bus is occluding a car which the roadside sensors can perceive. This is the largest sequence during daytime with 3,442 labeled 3D objects.}
\label{fig:dataset_visualization_drive_15} 
\end{figure*}
\setlength{\fboxsep}{0pt}%
\setlength{\fboxrule}{1pt}%
\begin{figure*}[h!]
\centering
\minipage{0.25\textwidth}
  \fbox{\includegraphics[width=.98\linewidth]{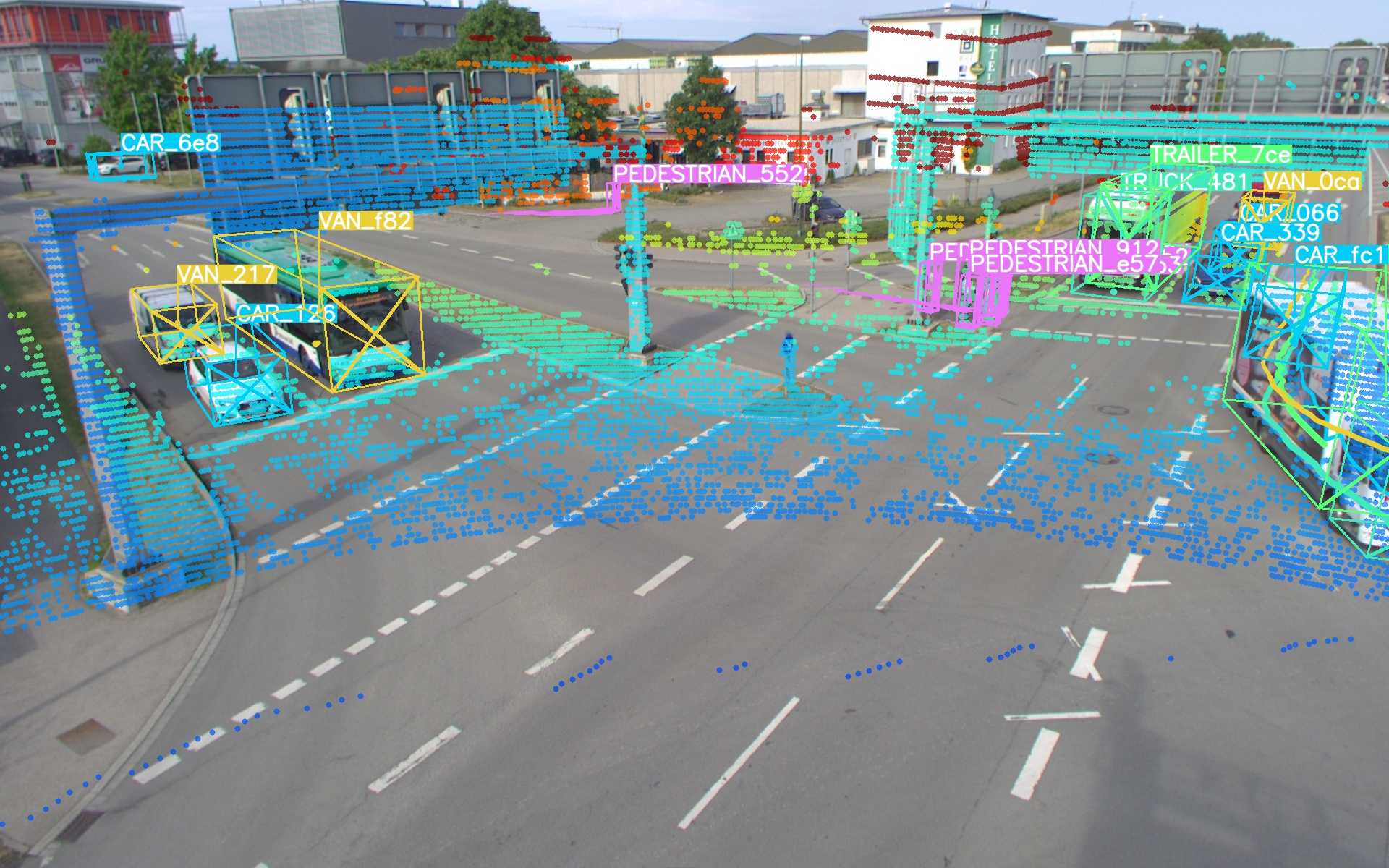}}
\endminipage
\minipage{0.25\textwidth}
  \fbox{\includegraphics[width=.98\linewidth]{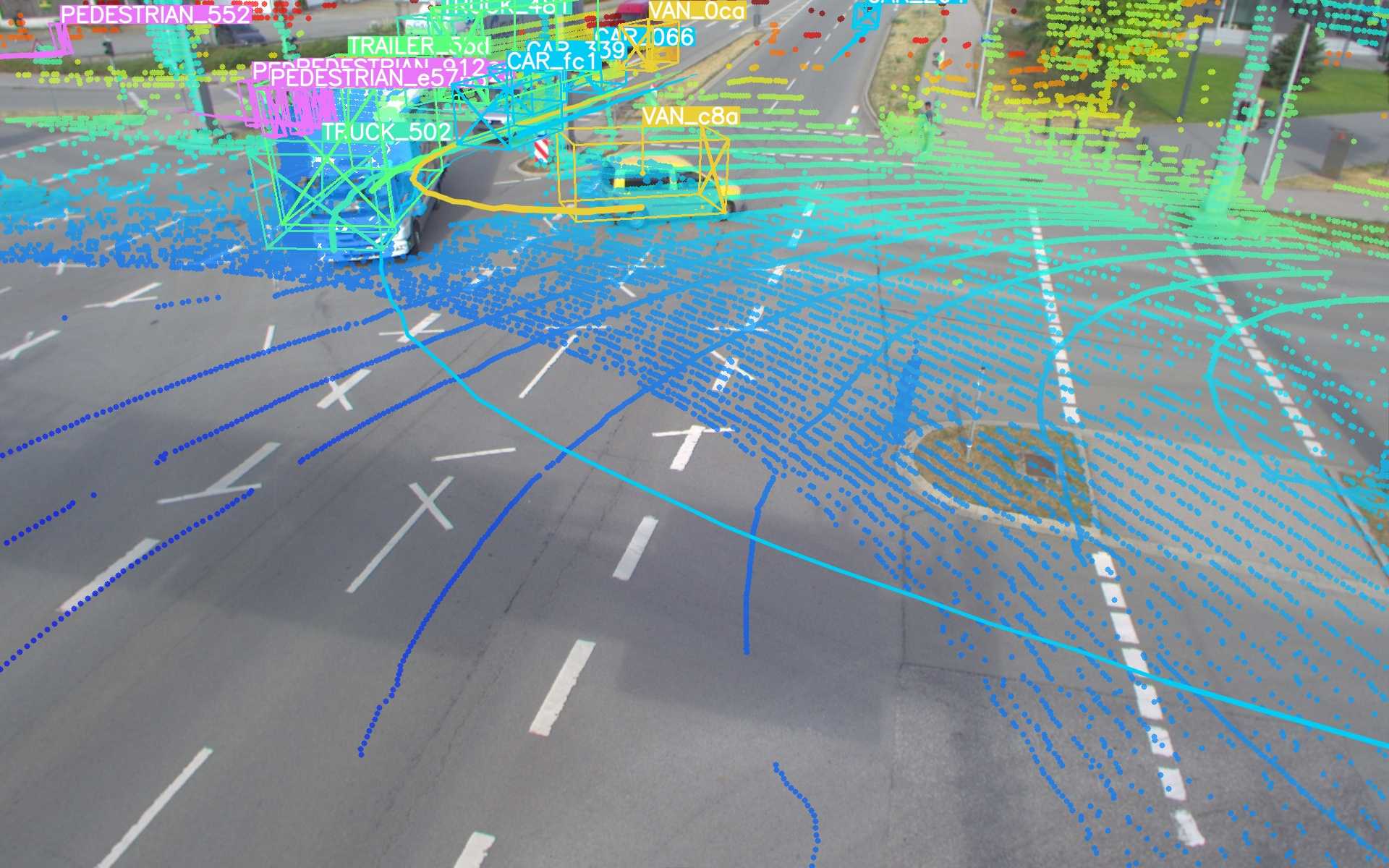}}
\endminipage
\minipage{0.25\textwidth}%
  \fbox{\includegraphics[width=.98\linewidth]{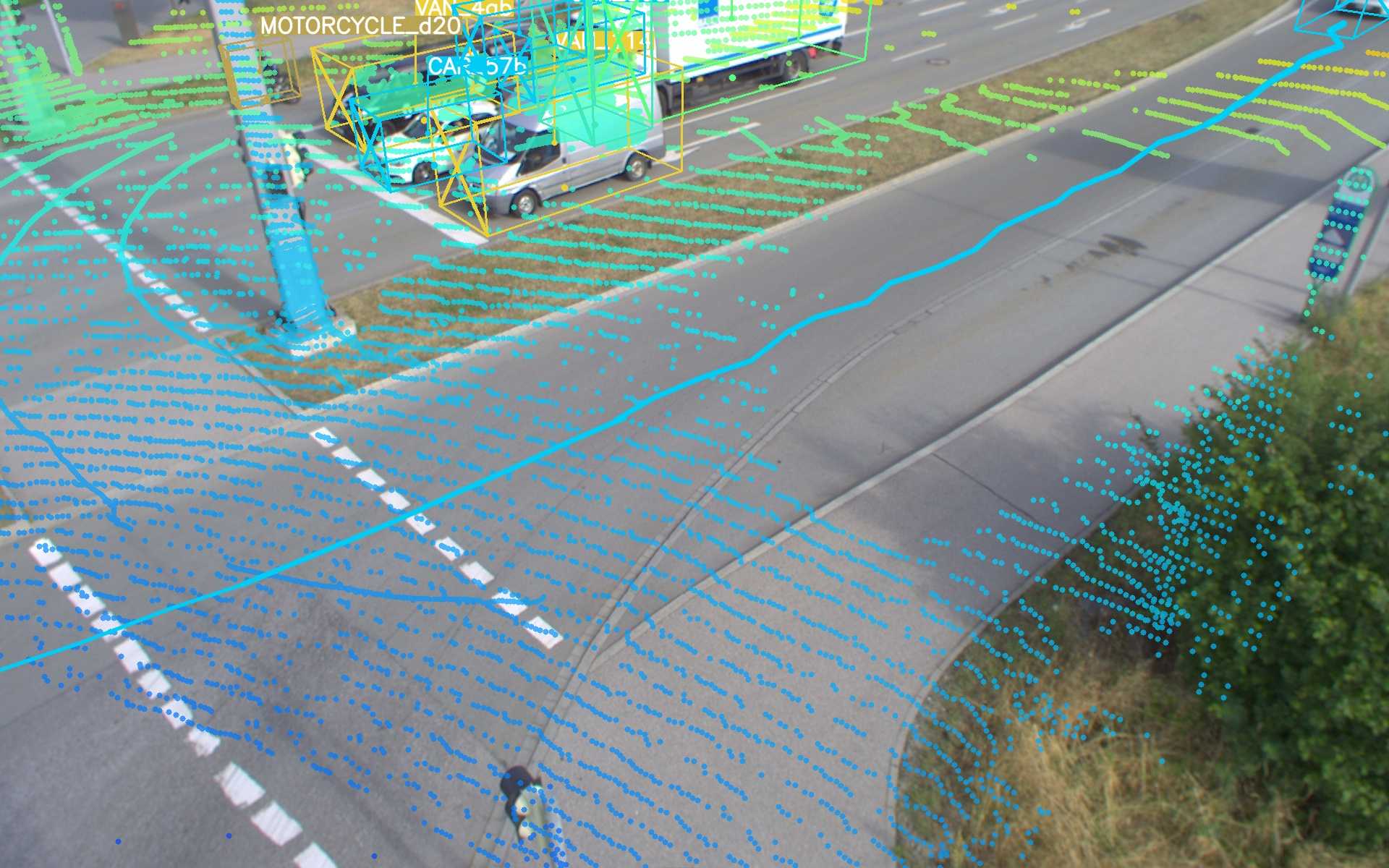}}
\endminipage
\minipage{0.25\textwidth}
  \fbox{\includegraphics[width=.98\linewidth]{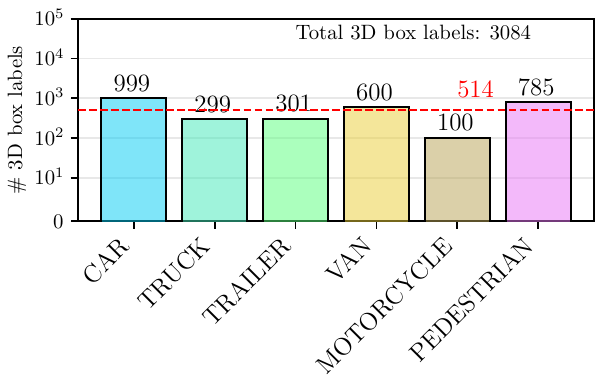}}
\endminipage\\
\vspace{-0.07cm}
\minipage{0.25\textwidth}
  \fbox{\includegraphics[width=.98\linewidth]{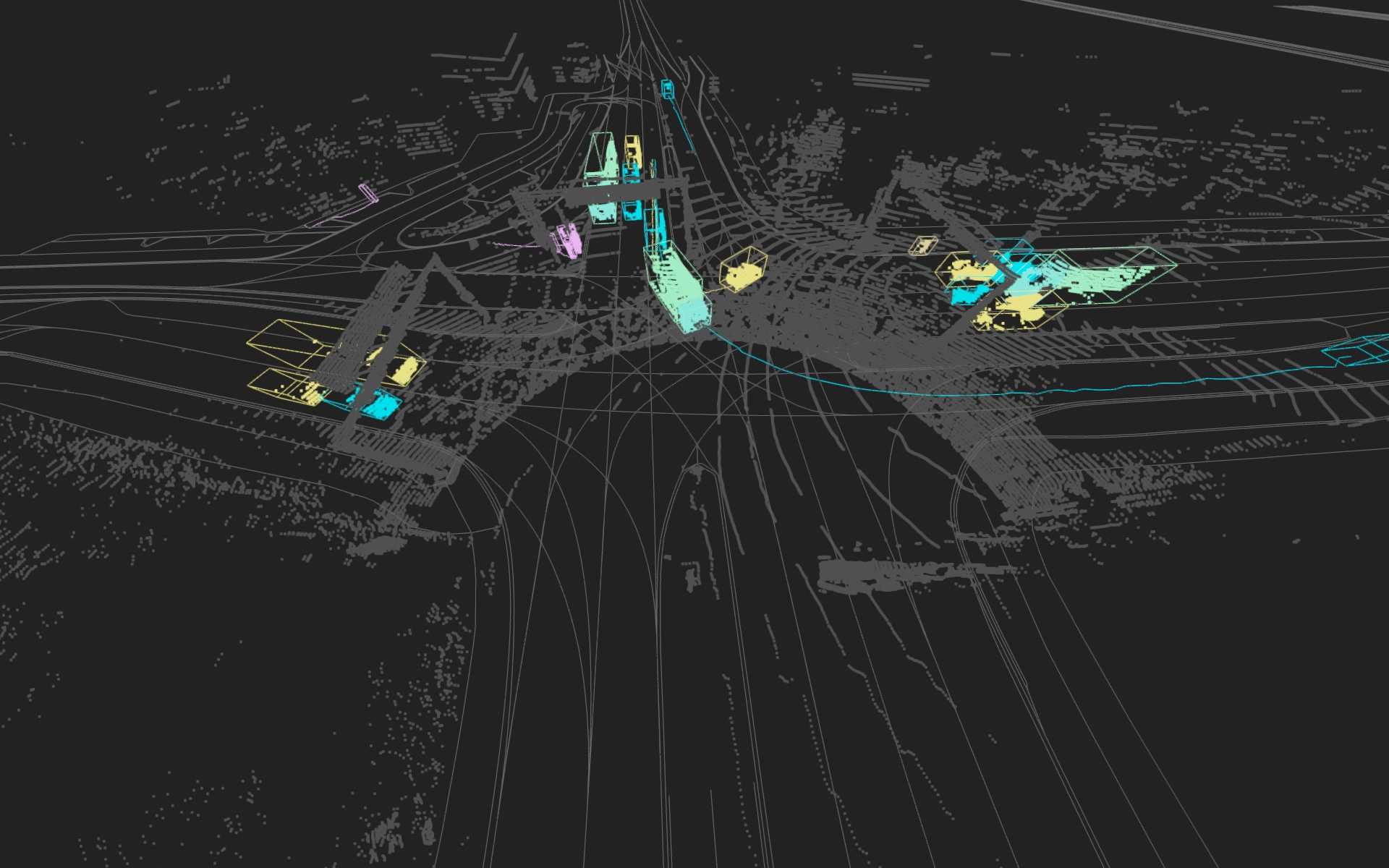}}
\endminipage
\minipage{0.25\textwidth}%
  \fbox{\includegraphics[width=.98\linewidth]{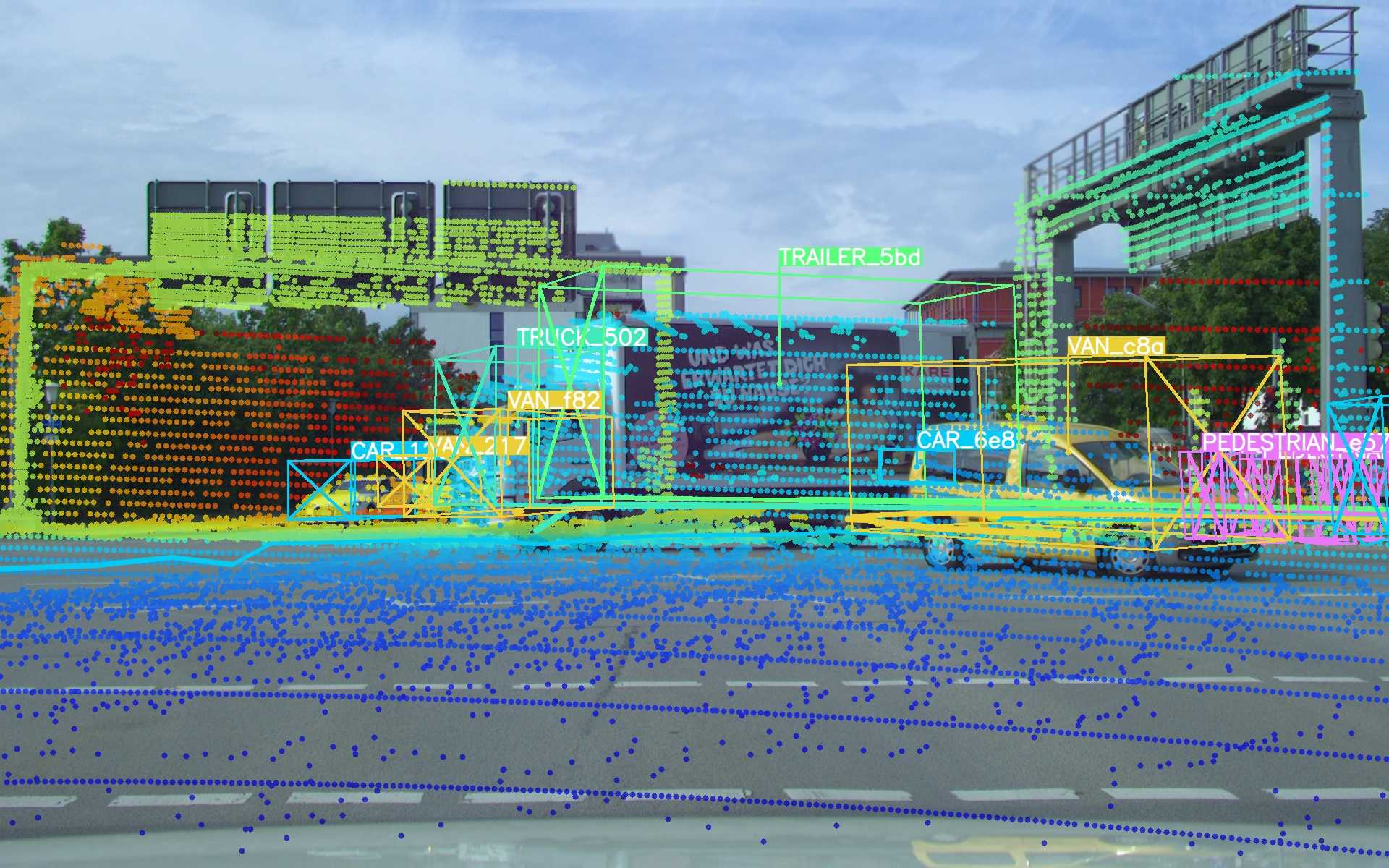}}
\endminipage
\minipage{0.25\textwidth}%
  \fbox{\includegraphics[width=.98\linewidth]{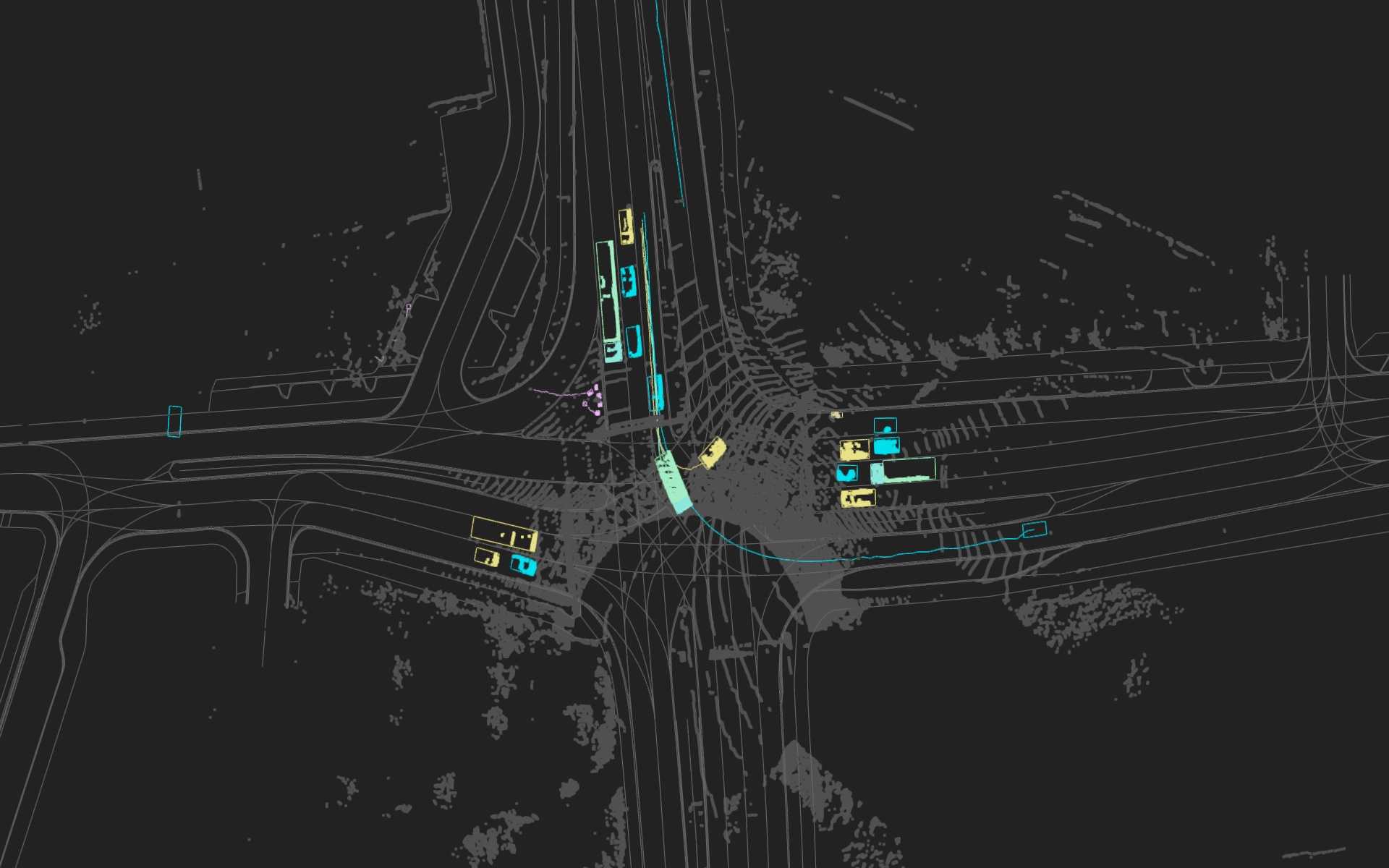}}
\endminipage
\minipage{0.25\textwidth}%
  \fbox{\includegraphics[width=.98\linewidth]{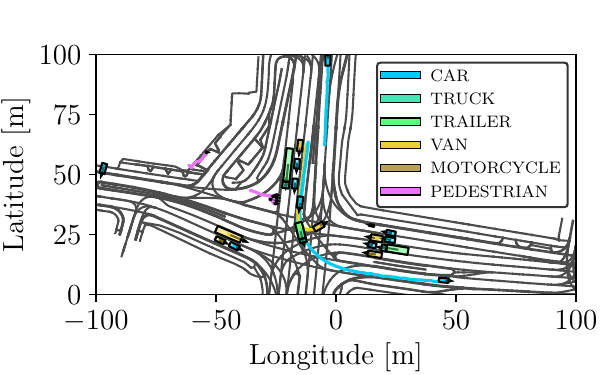}}
\endminipage
\caption{Visualization of \textbf{drive\_22} of the \textit{TUMTraf-V2X} dataset. In this drive, many vehicles are performing a U-turn maneuver and occlude some pedestrians waiting at a red traffic light. The pedestrians are within the field of view of the roadside sensors and can be perceived. This sequence contains 3,084 labeled 3D objects.}
\label{fig:dataset_visualization_drive_22} 
\end{figure*}

\setlength{\fboxsep}{0pt}%
\setlength{\fboxrule}{1pt}%
\begin{figure*}[h!]
\centering
\minipage{0.25\textwidth}
  \fbox{\includegraphics[width=.98\linewidth]{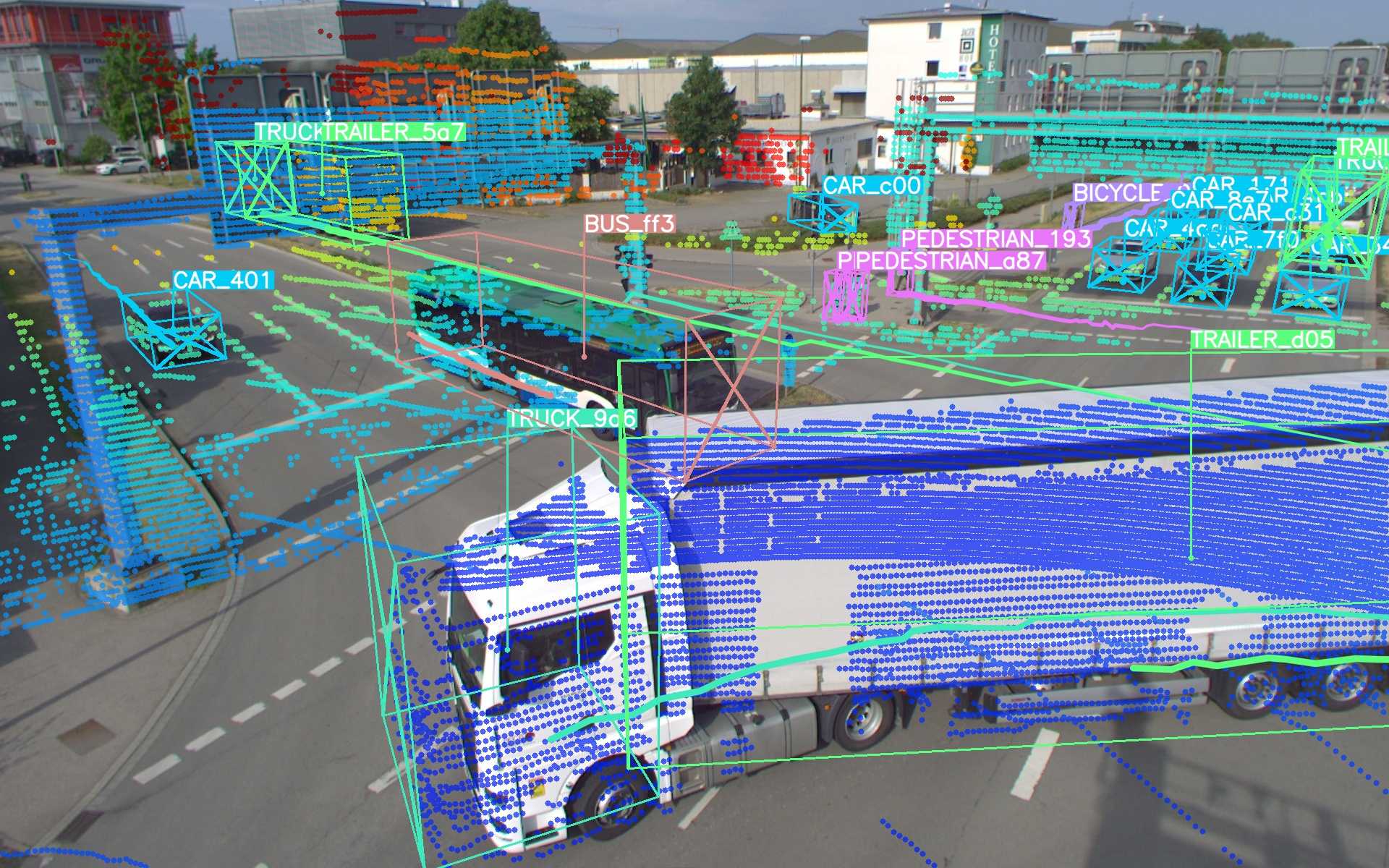}}
\endminipage
\minipage{0.25\textwidth}
  \fbox{\includegraphics[width=.98\linewidth]{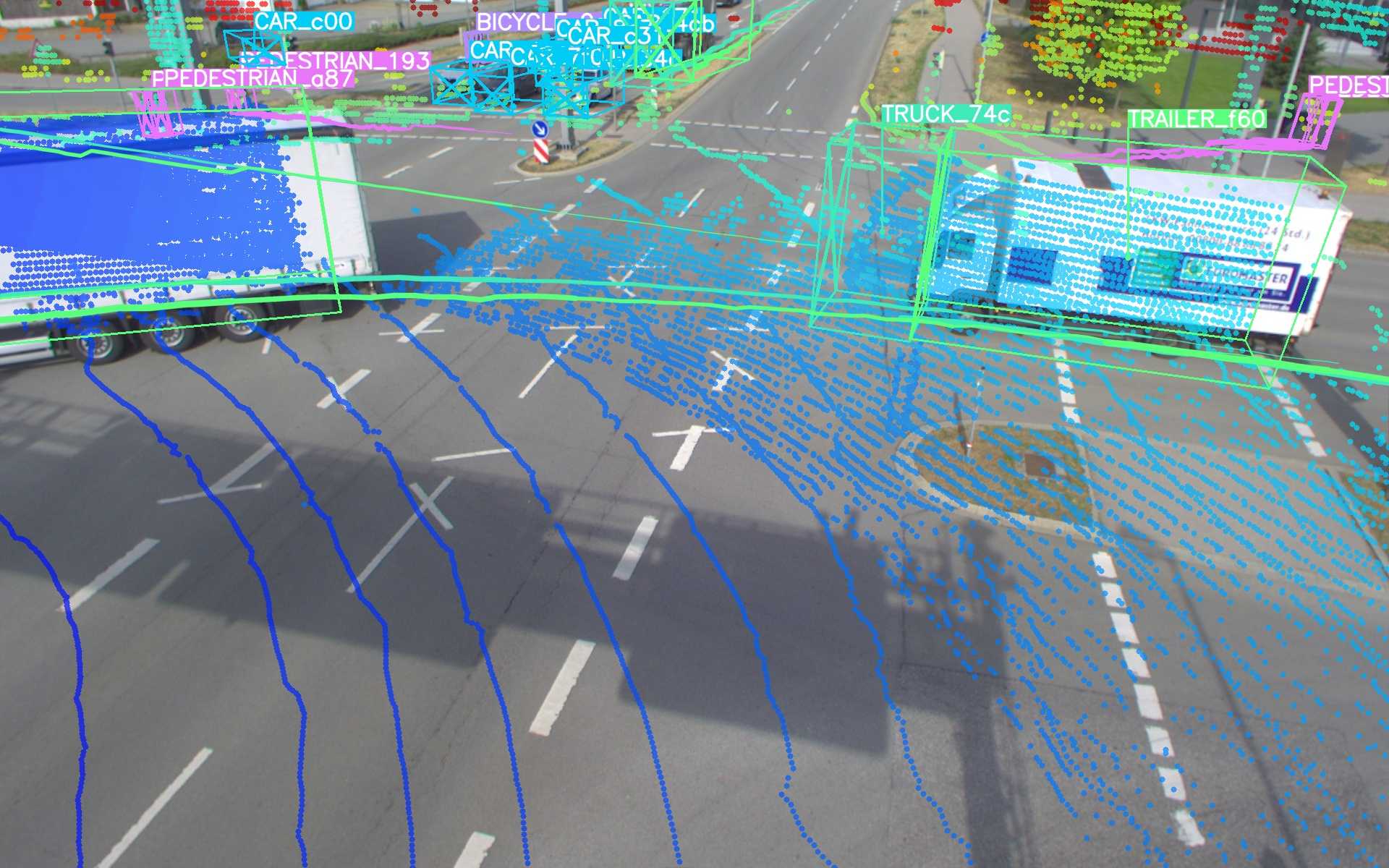}}
\endminipage
\minipage{0.25\textwidth}%
  \fbox{\includegraphics[width=.98\linewidth]{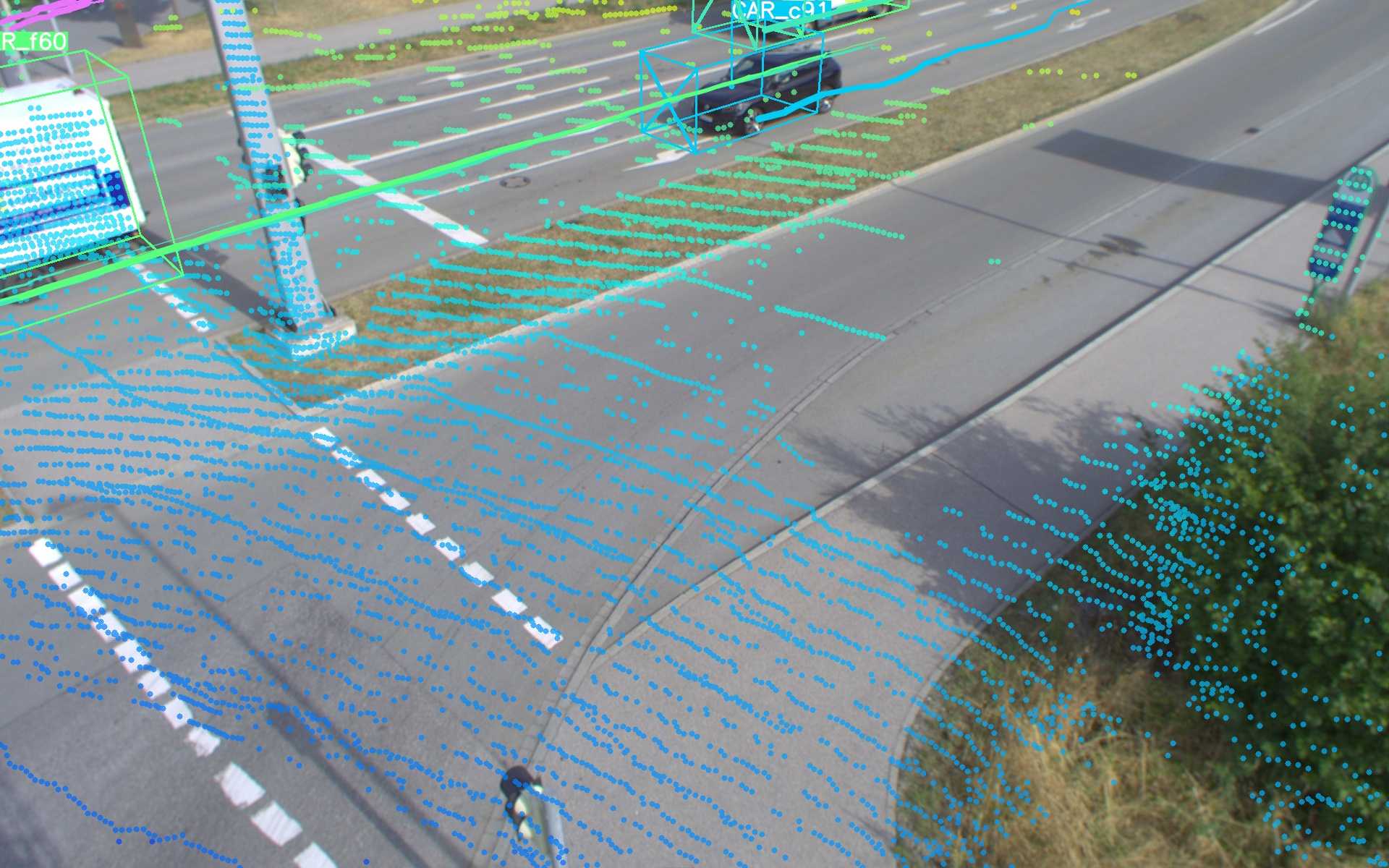}}
\endminipage
\minipage{0.25\textwidth}
  \fbox{\includegraphics[width=.98\linewidth]{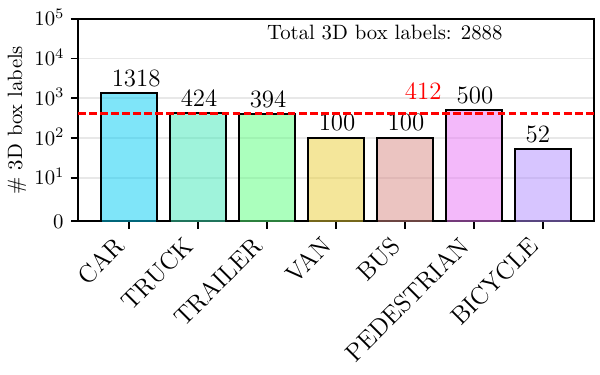}}
\endminipage\\
\vspace{-0.07cm}
\minipage{0.25\textwidth}
  \fbox{\includegraphics[width=.98\linewidth]{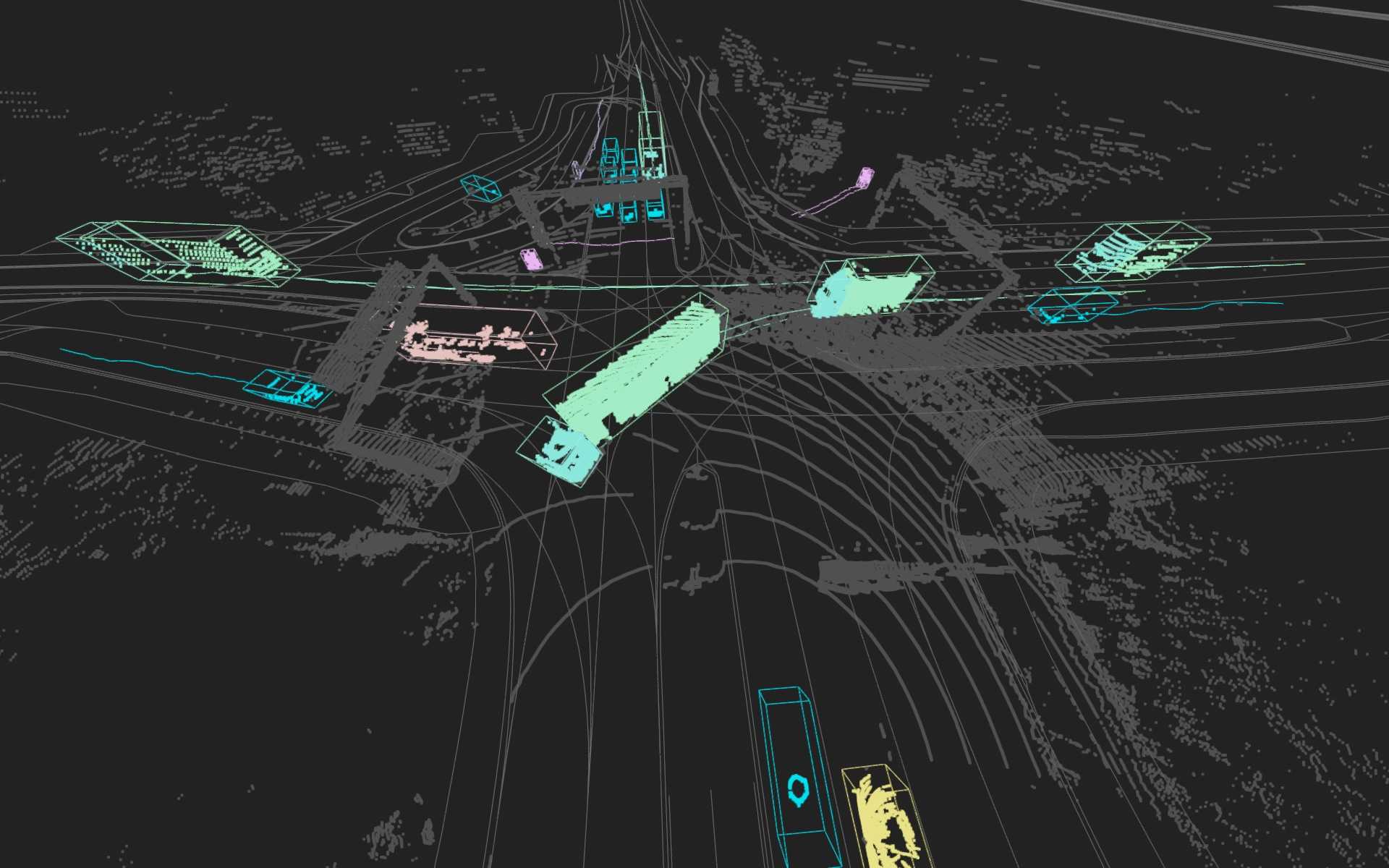}}
\endminipage
\minipage{0.25\textwidth}%
  \fbox{\includegraphics[width=.98\linewidth]{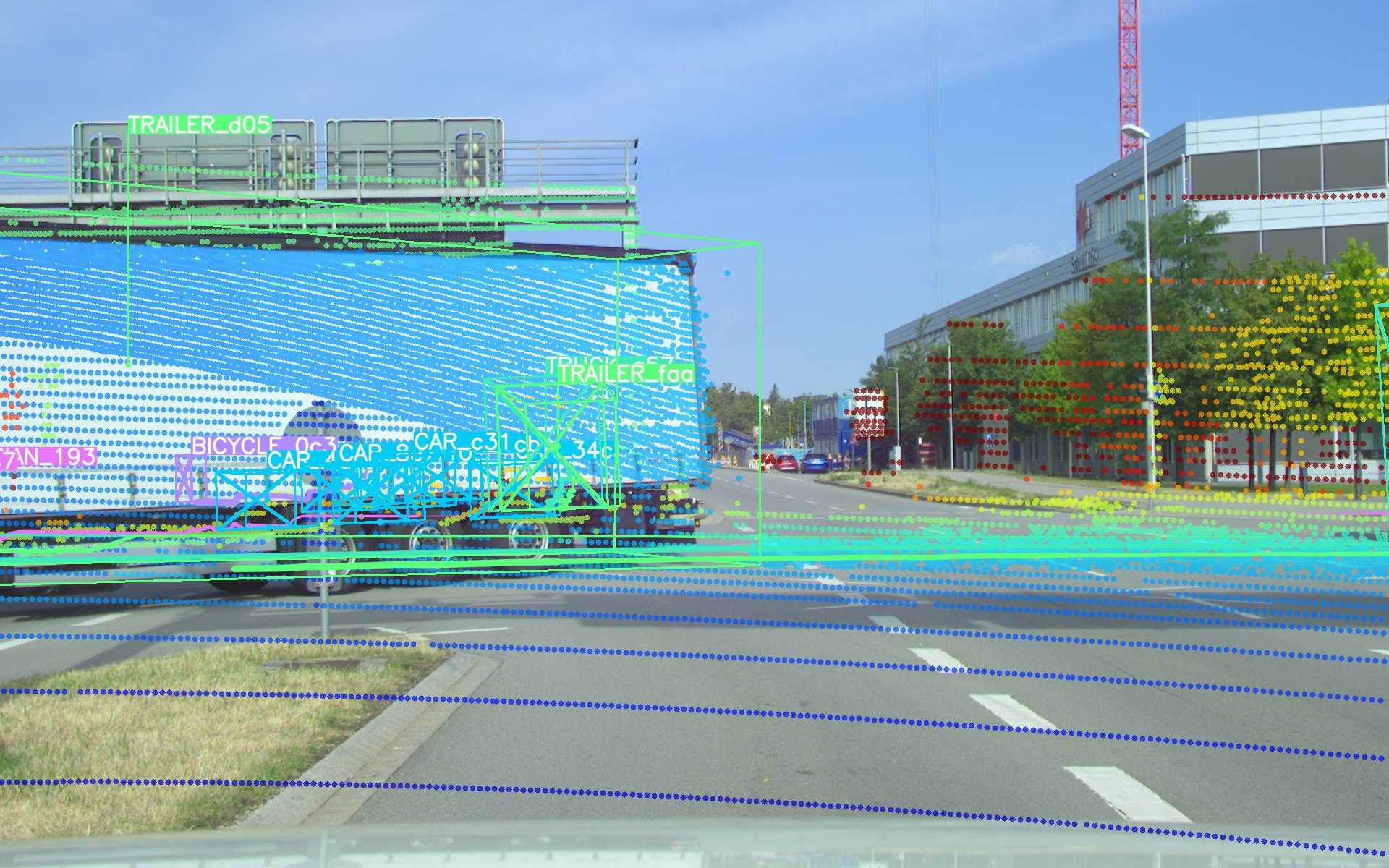}}
\endminipage
\minipage{0.25\textwidth}%
  \fbox{\includegraphics[width=.98\linewidth]{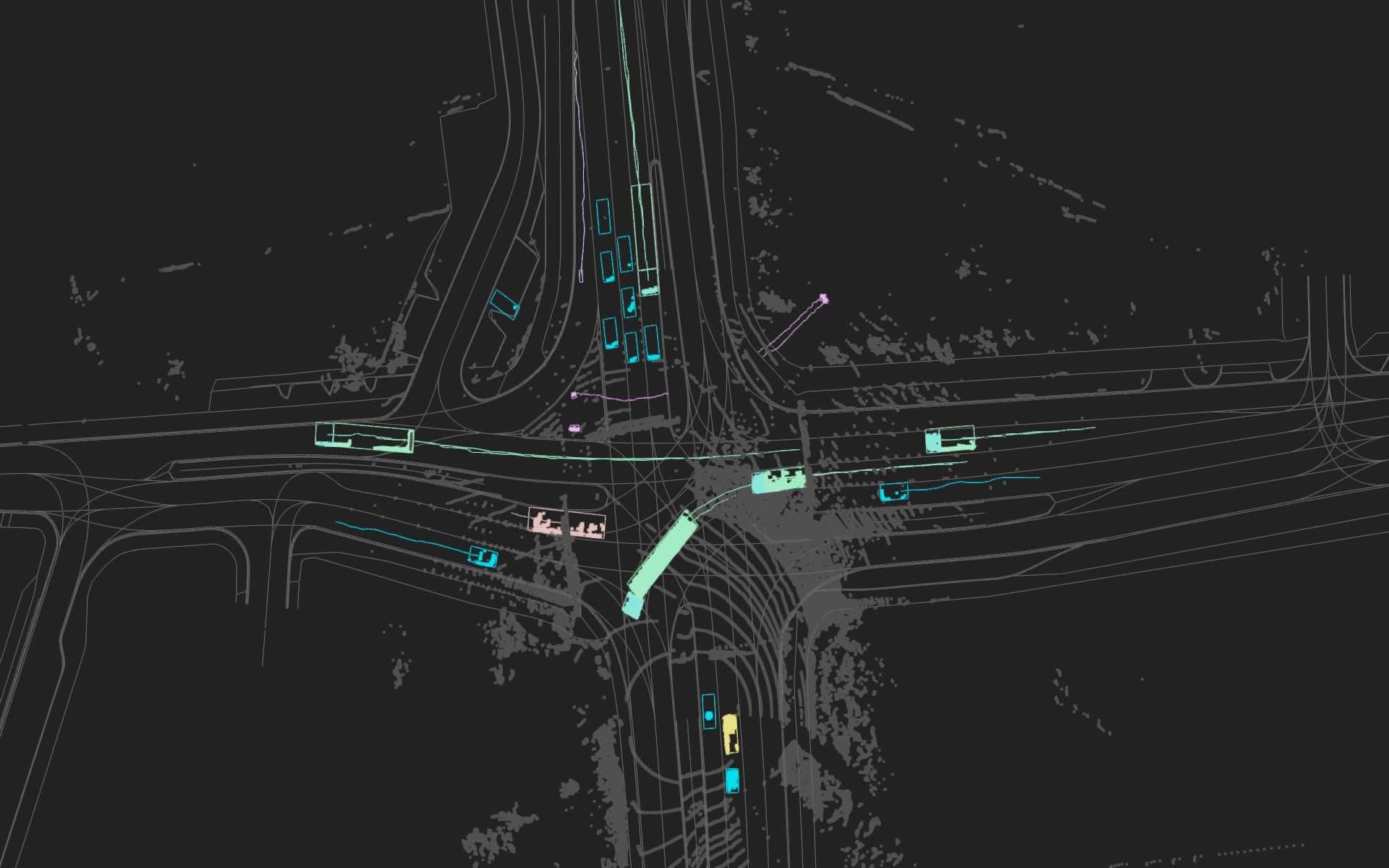}}
\endminipage
\minipage{0.25\textwidth}%
  \fbox{\includegraphics[width=.98\linewidth]{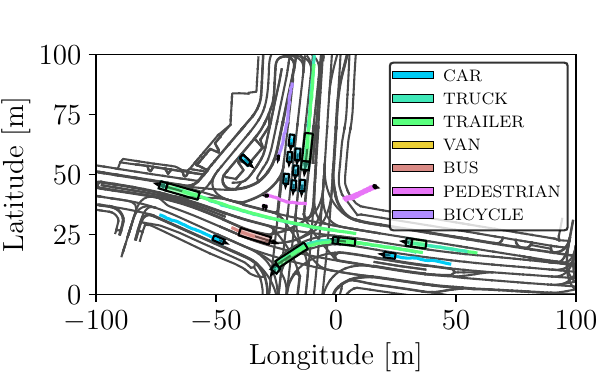}}
\endminipage
\caption{Visualization of \textbf{drive\_26} of the \textit{TUMTraf-V2X} dataset. In this drive, multiple trucks and trailers occlude traffic participants. These traffic participants are visible from the elevated roadside cameras and LiDAR mounted on the infrastructure. This sequence contains 2,888 labeled 3D objects.}
\label{fig:dataset_visualization_drive_26} 
\end{figure*}

\setlength{\fboxsep}{0pt}%
\setlength{\fboxrule}{1pt}%
\begin{figure*}[h!]
\centering
\minipage{0.25\textwidth}
  \fbox{\includegraphics[width=.98\linewidth]{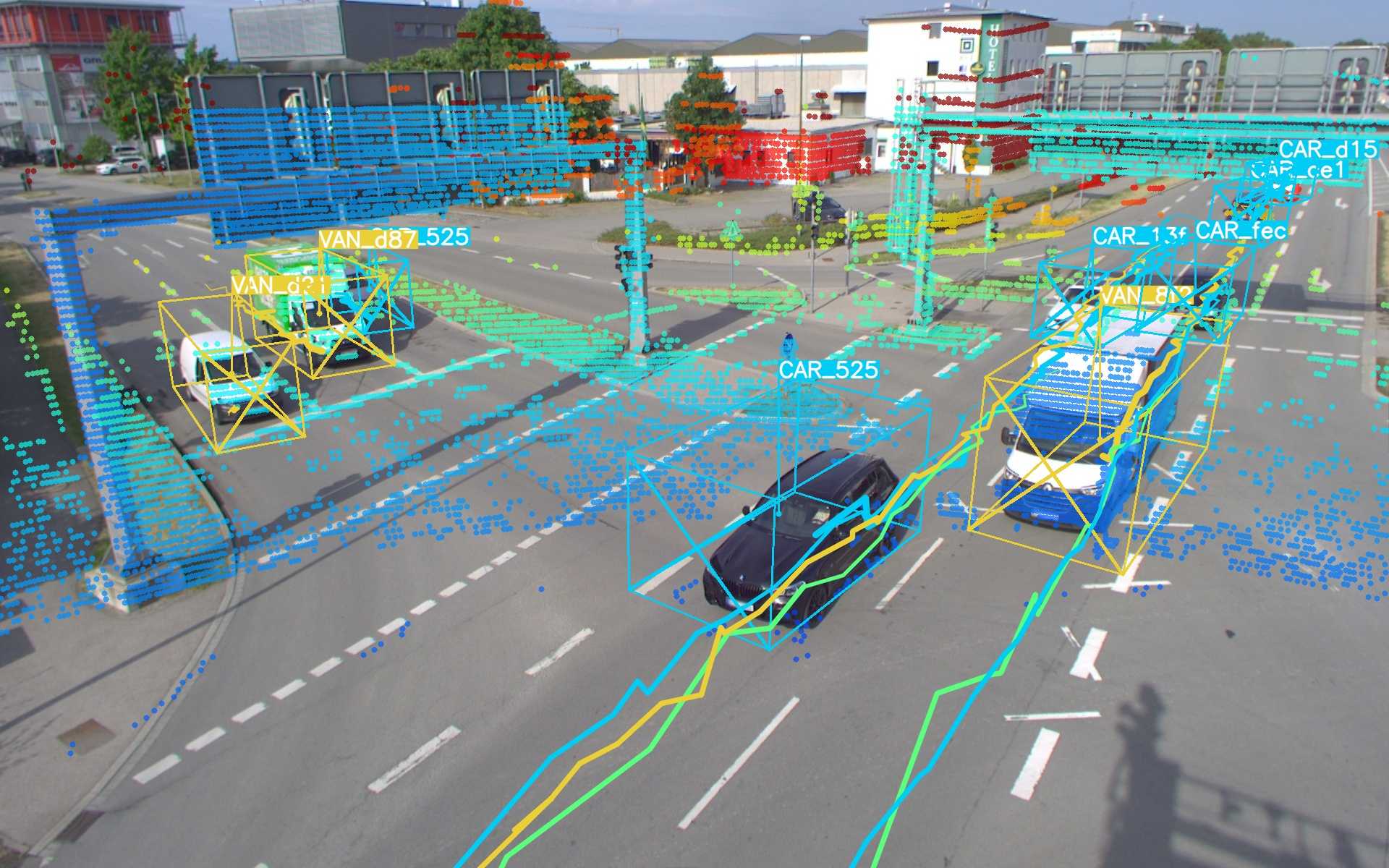}}
\endminipage
\minipage{0.25\textwidth}
  \fbox{\includegraphics[width=.98\linewidth]{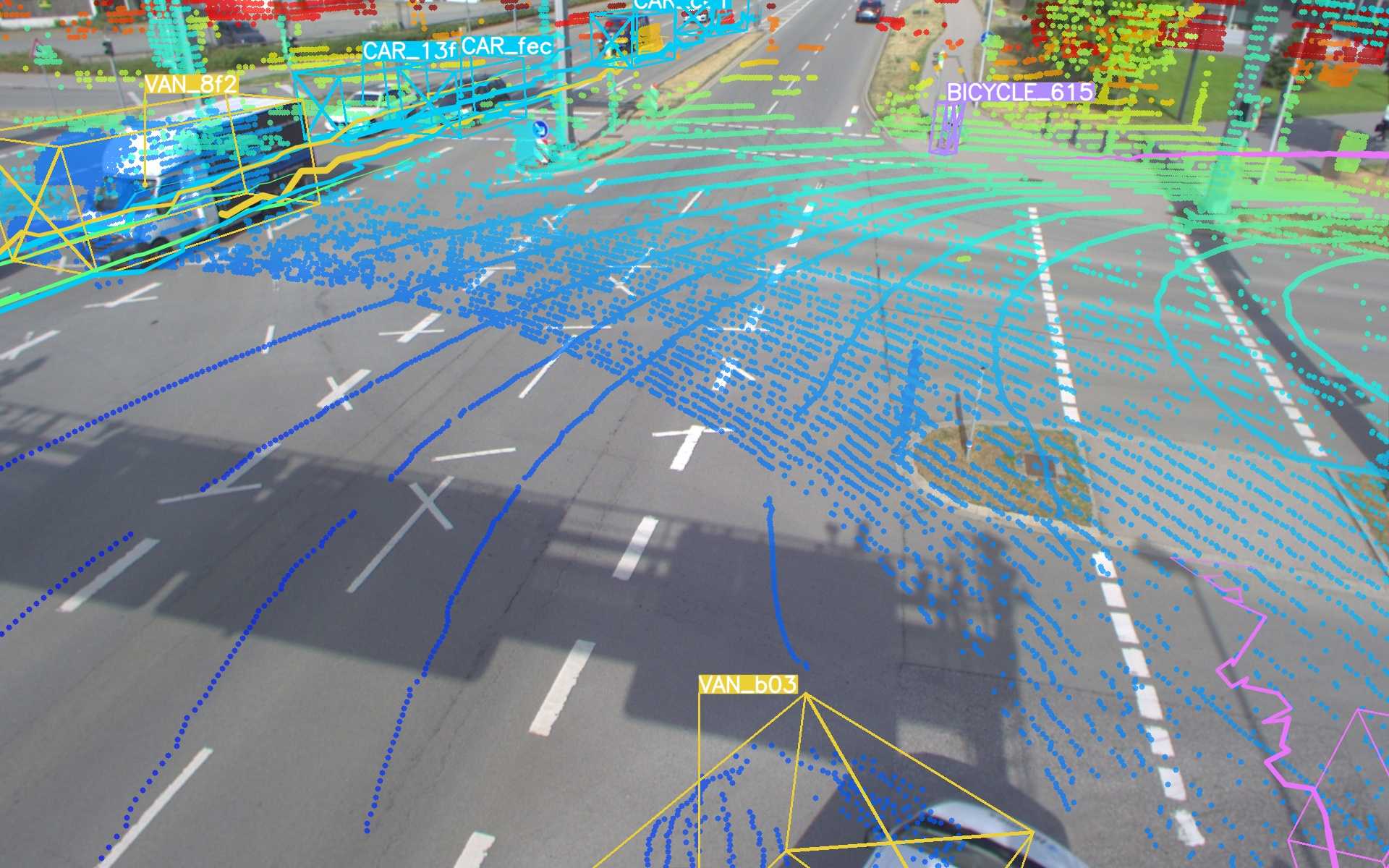}}
\endminipage
\minipage{0.25\textwidth}%
  \fbox{\includegraphics[width=.98\linewidth]{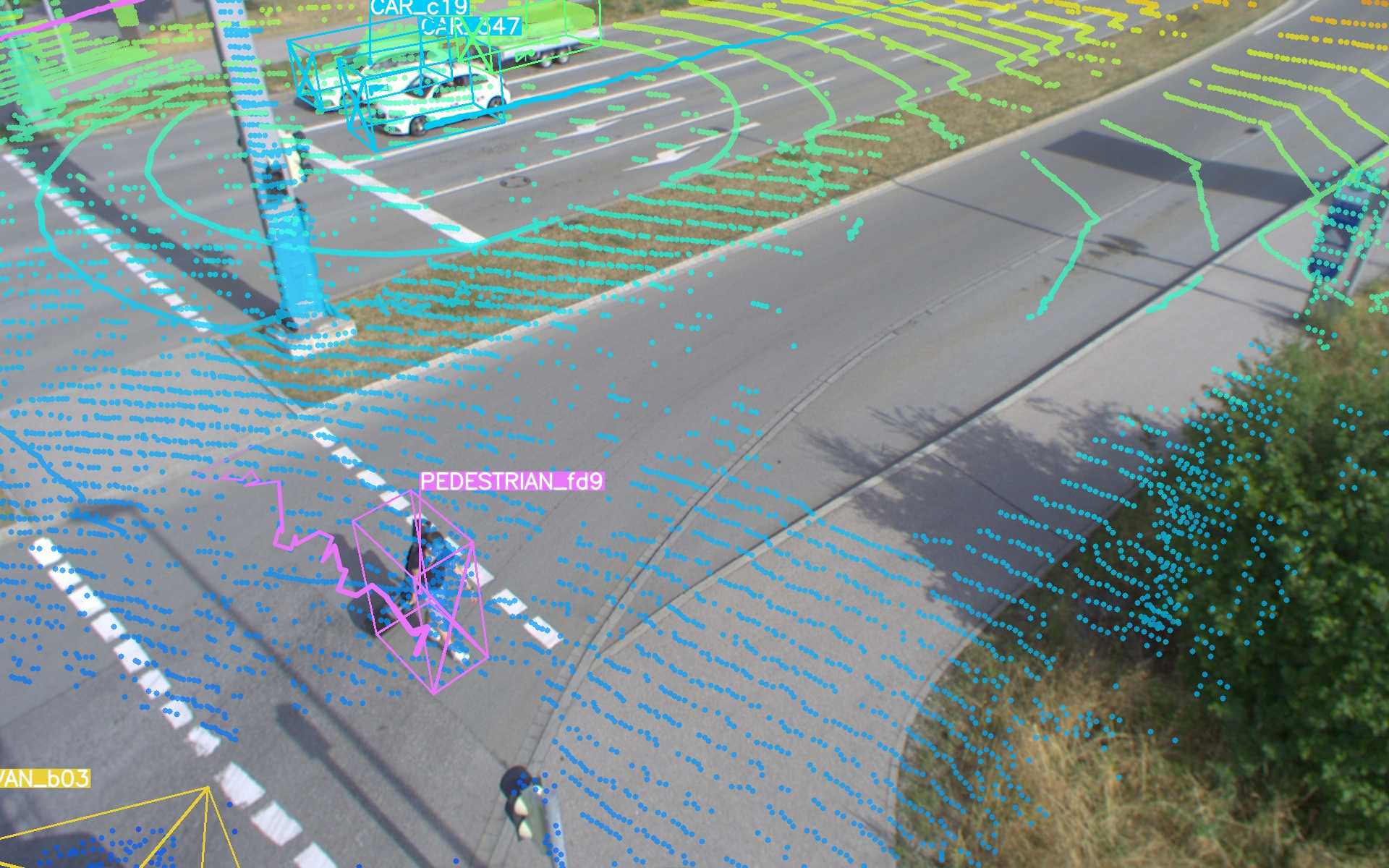}}
\endminipage
\minipage{0.25\textwidth}
  \fbox{\includegraphics[width=.98\linewidth]{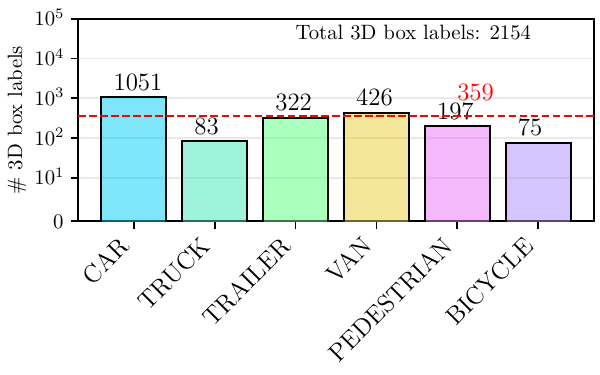}}
\endminipage\\
\vspace{-0.07cm}
\minipage{0.25\textwidth}
  \fbox{\includegraphics[width=.98\linewidth]{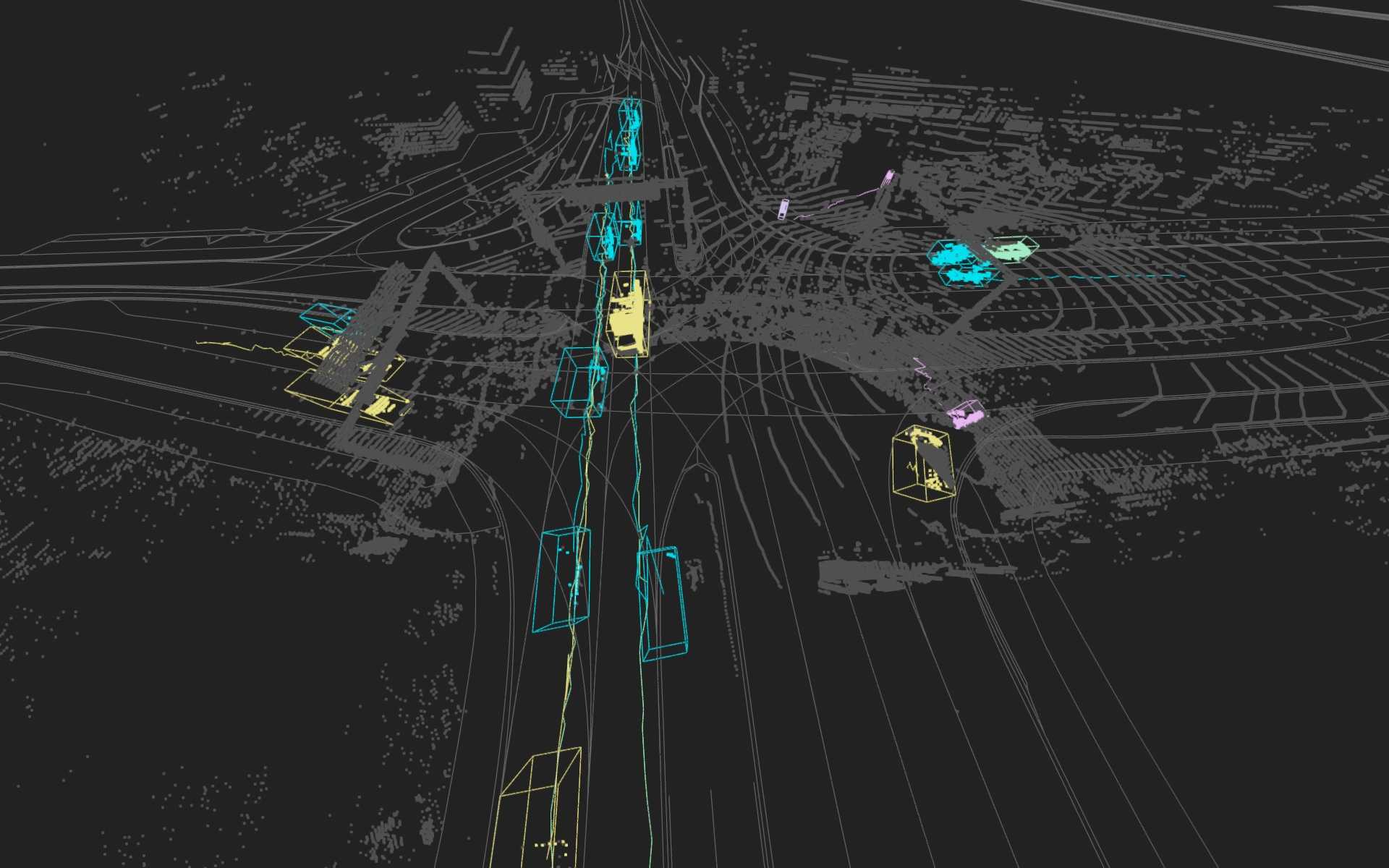}}
\endminipage
\minipage{0.25\textwidth}%
  \fbox{\includegraphics[width=.98\linewidth]{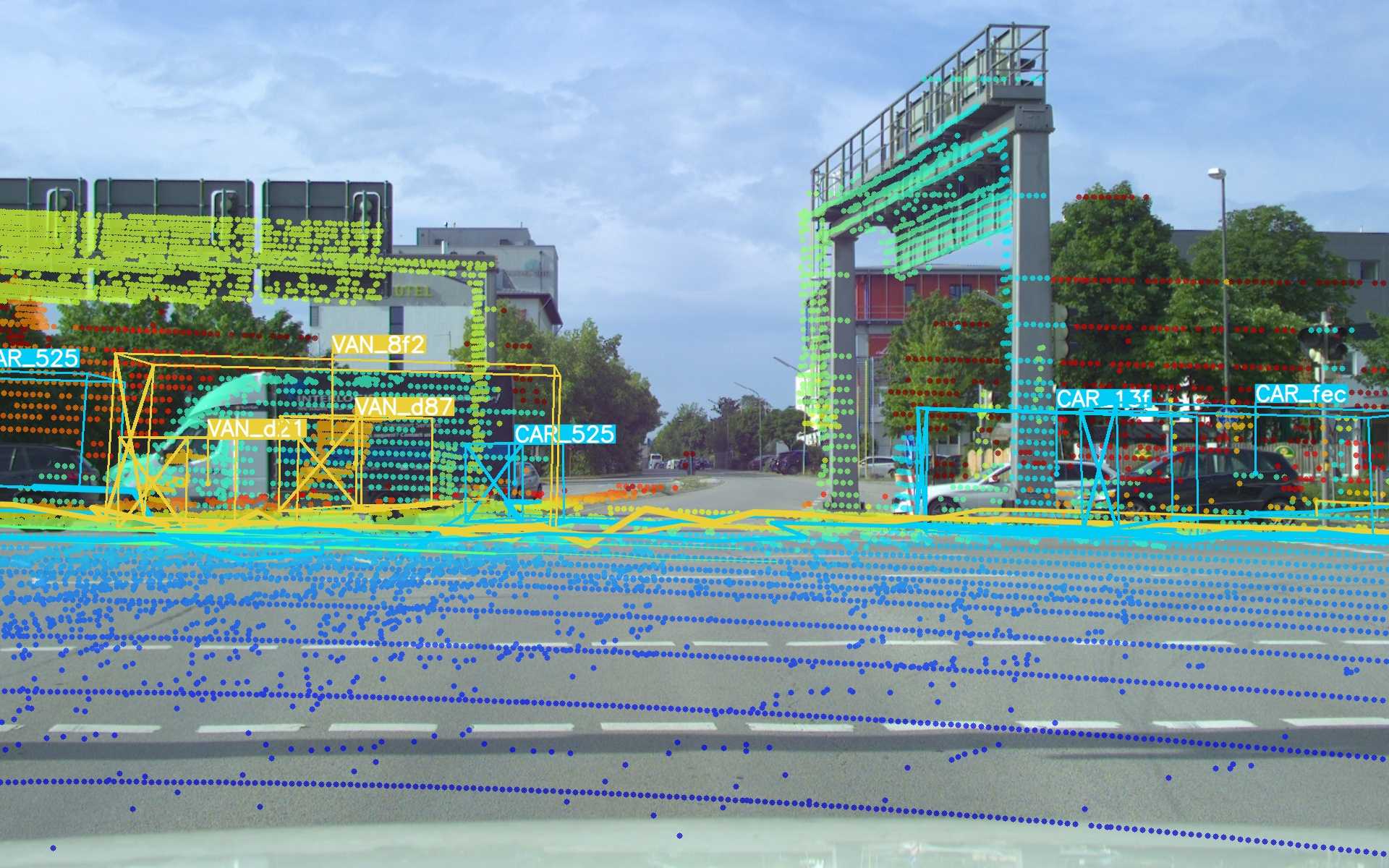}}
\endminipage
\minipage{0.25\textwidth}%
  \fbox{\includegraphics[width=.98\linewidth]{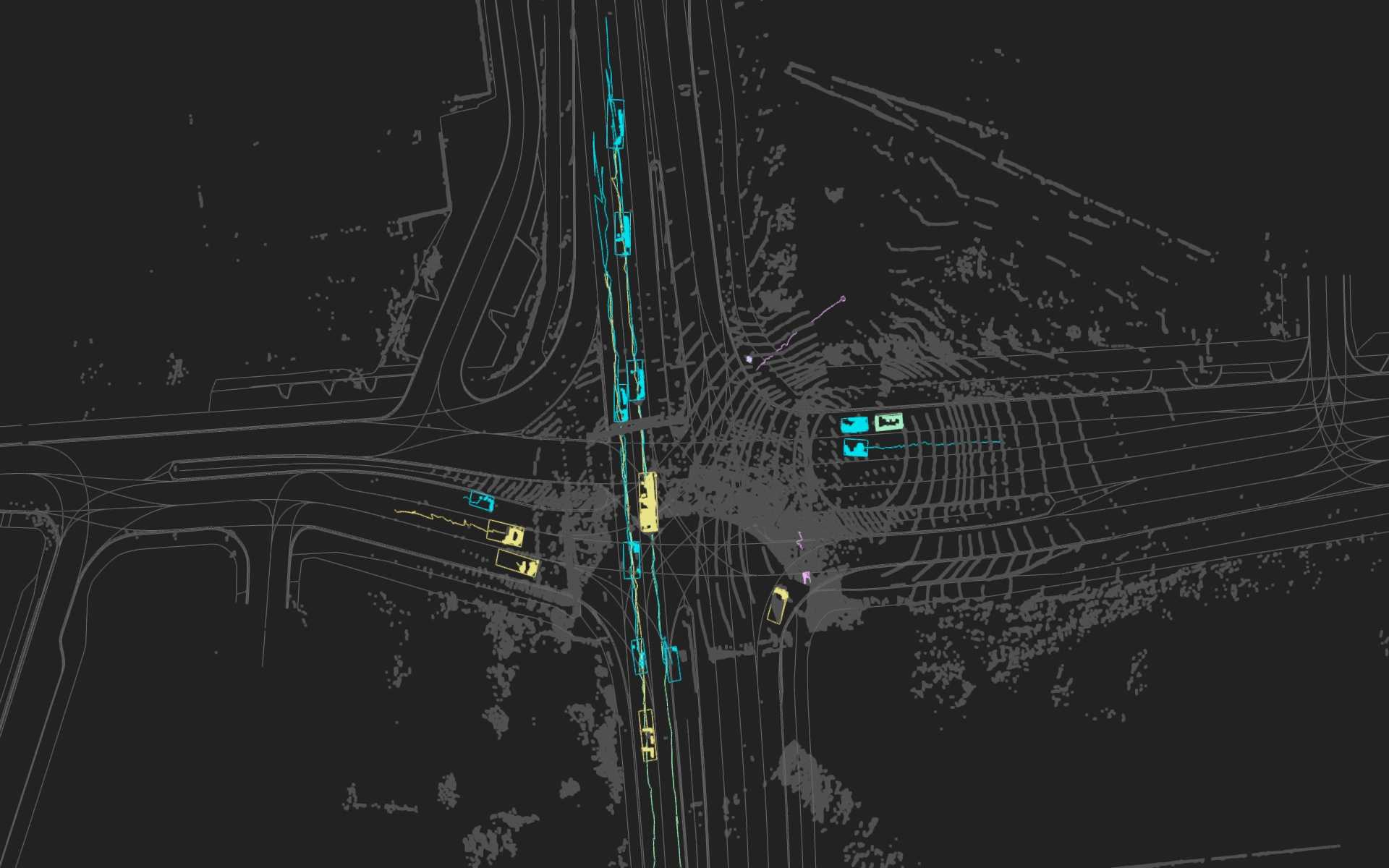}}
\endminipage
\minipage{0.25\textwidth}%
  \fbox{\includegraphics[width=.98\linewidth]{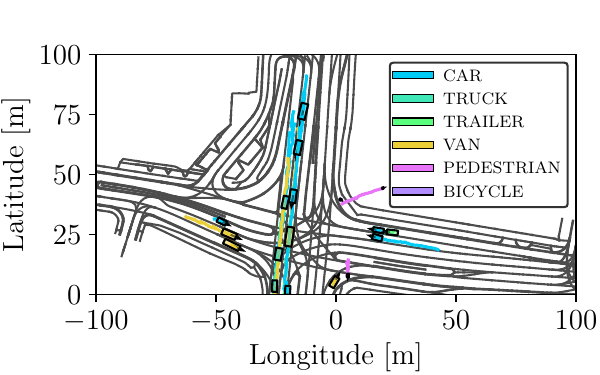}}
\endminipage
\caption{Visualization of \textbf{drive\_33} of the \textit{TUMTraf-V2X} dataset. In this scenario, a truck is occluding multiple objects that can be perceived by the roadside camera and LiDAR. Here, 2,154 3D objects were labeled.}
\label{fig:dataset_visualization_drive_33} 
\end{figure*}

\setlength{\fboxsep}{0pt}%
\setlength{\fboxrule}{1pt}%
\begin{figure*}[h!]
\centering
\minipage{0.25\textwidth}
  \fbox{\includegraphics[width=.98\linewidth]{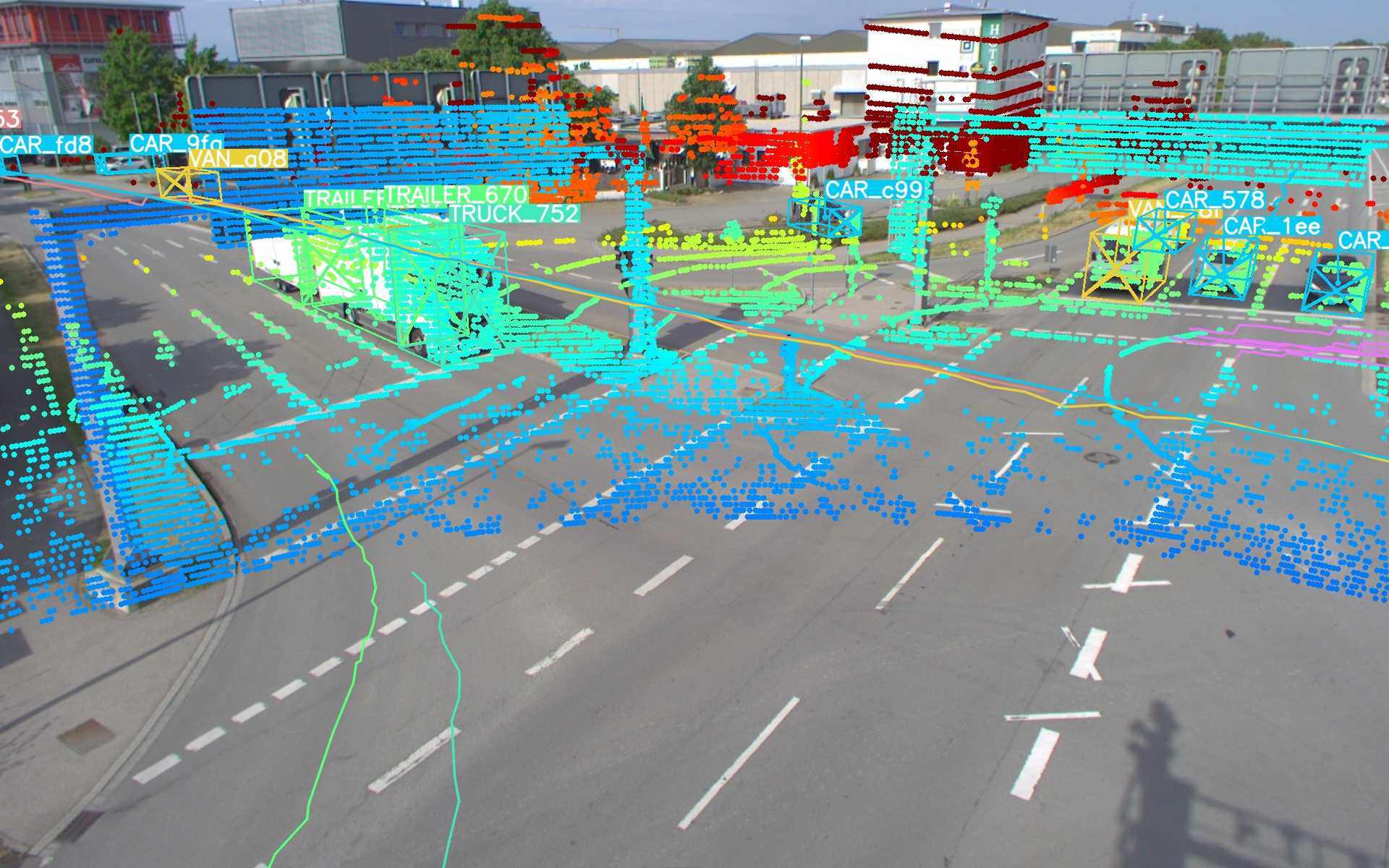}}
\endminipage
\minipage{0.25\textwidth}
  \fbox{\includegraphics[width=.98\linewidth]{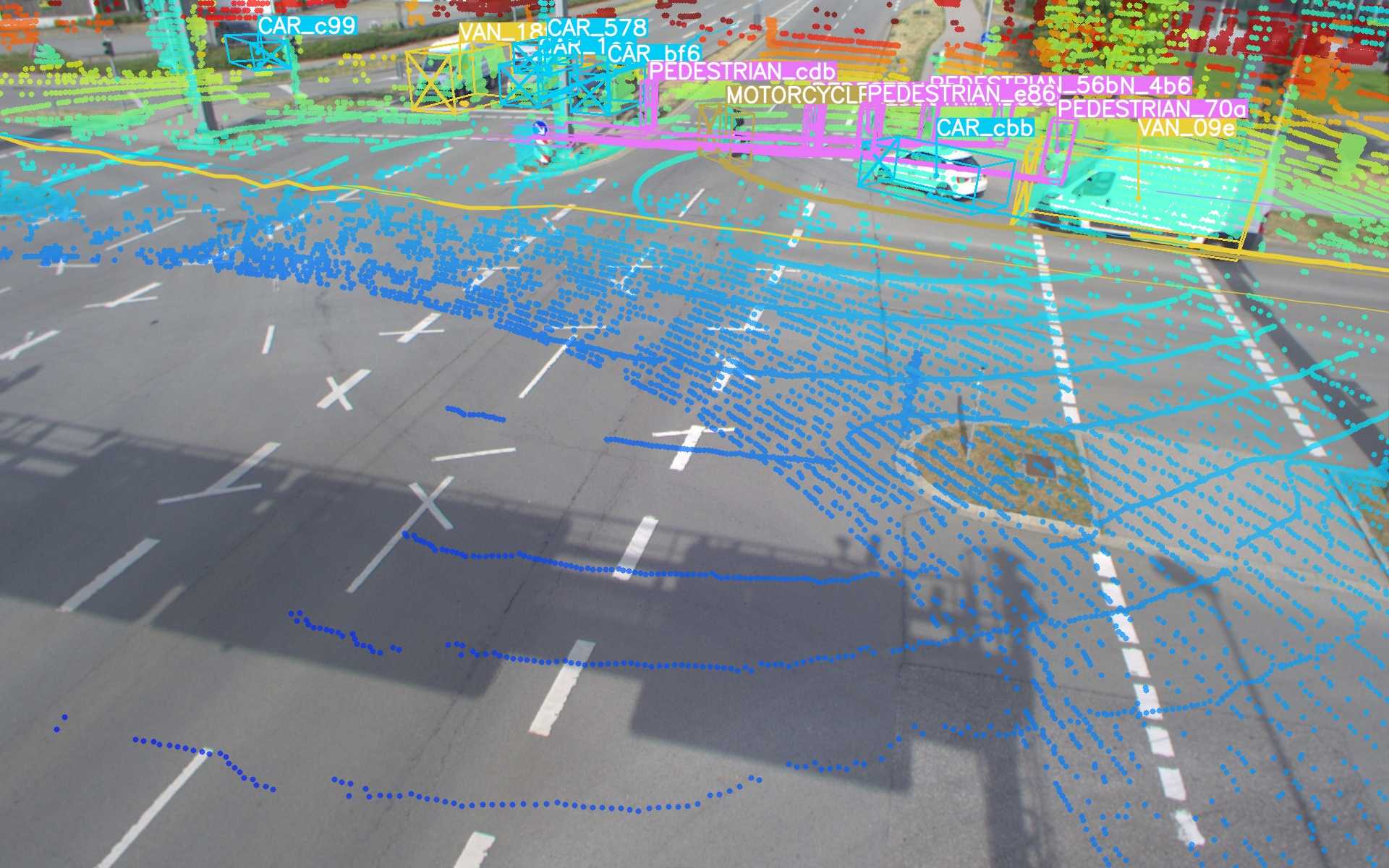}}
\endminipage
\minipage{0.25\textwidth}%
  \fbox{\includegraphics[width=.98\linewidth]{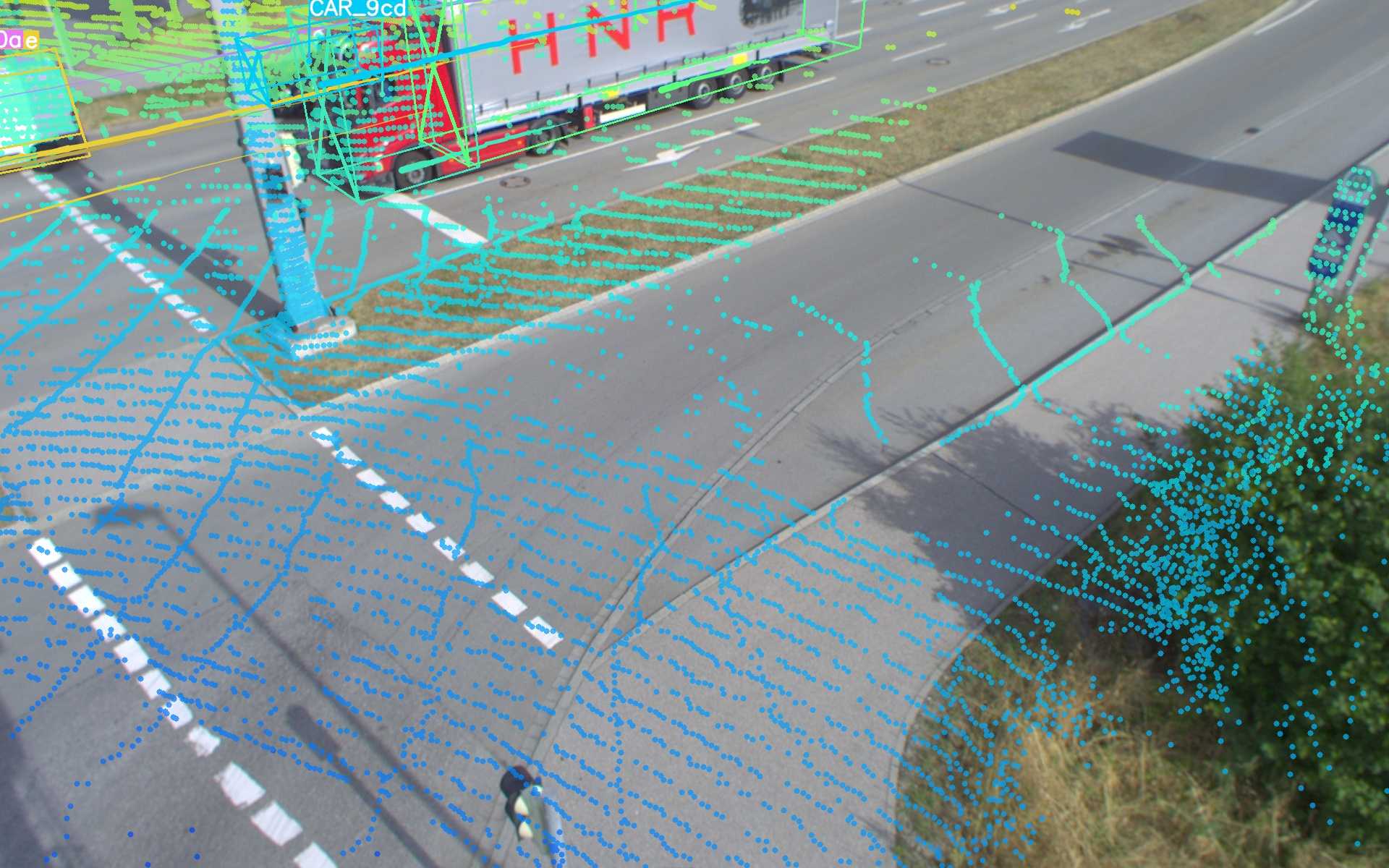}}
\endminipage
\minipage{0.25\textwidth}
  \fbox{\includegraphics[width=.98\linewidth]{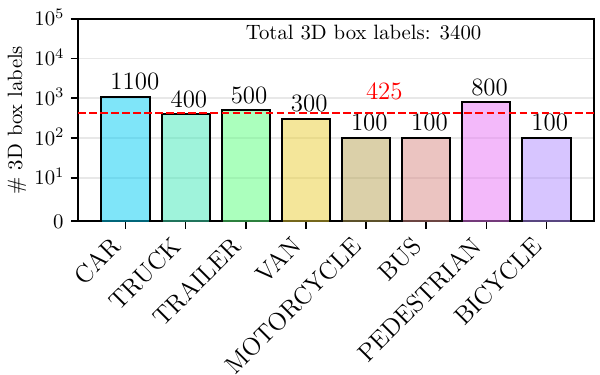}}
\endminipage\\
\vspace{-0.07cm}
\minipage{0.25\textwidth}
  \fbox{\includegraphics[width=.98\linewidth]{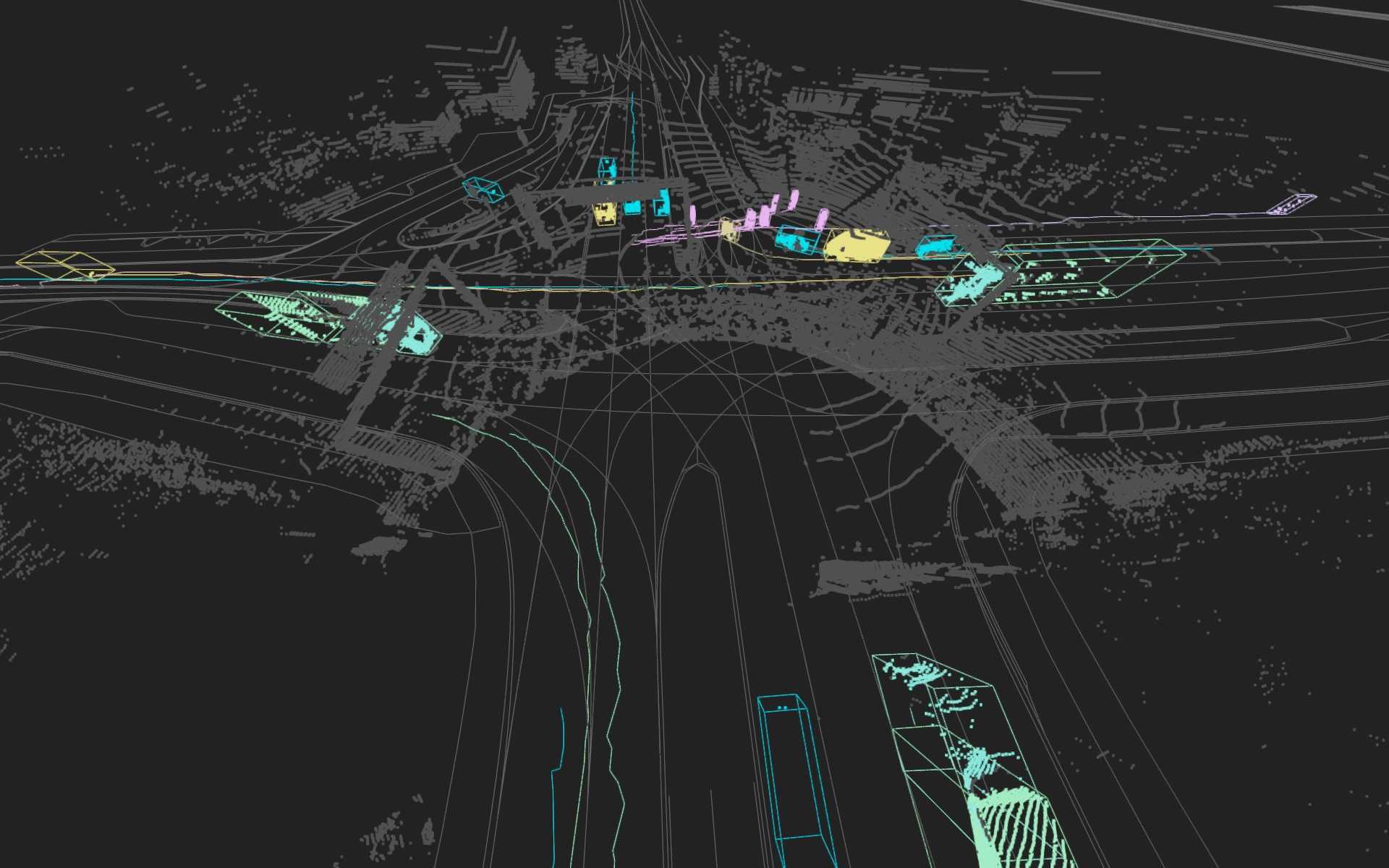}}
\endminipage
\minipage{0.25\textwidth}%
  \fbox{\includegraphics[width=.98\linewidth]{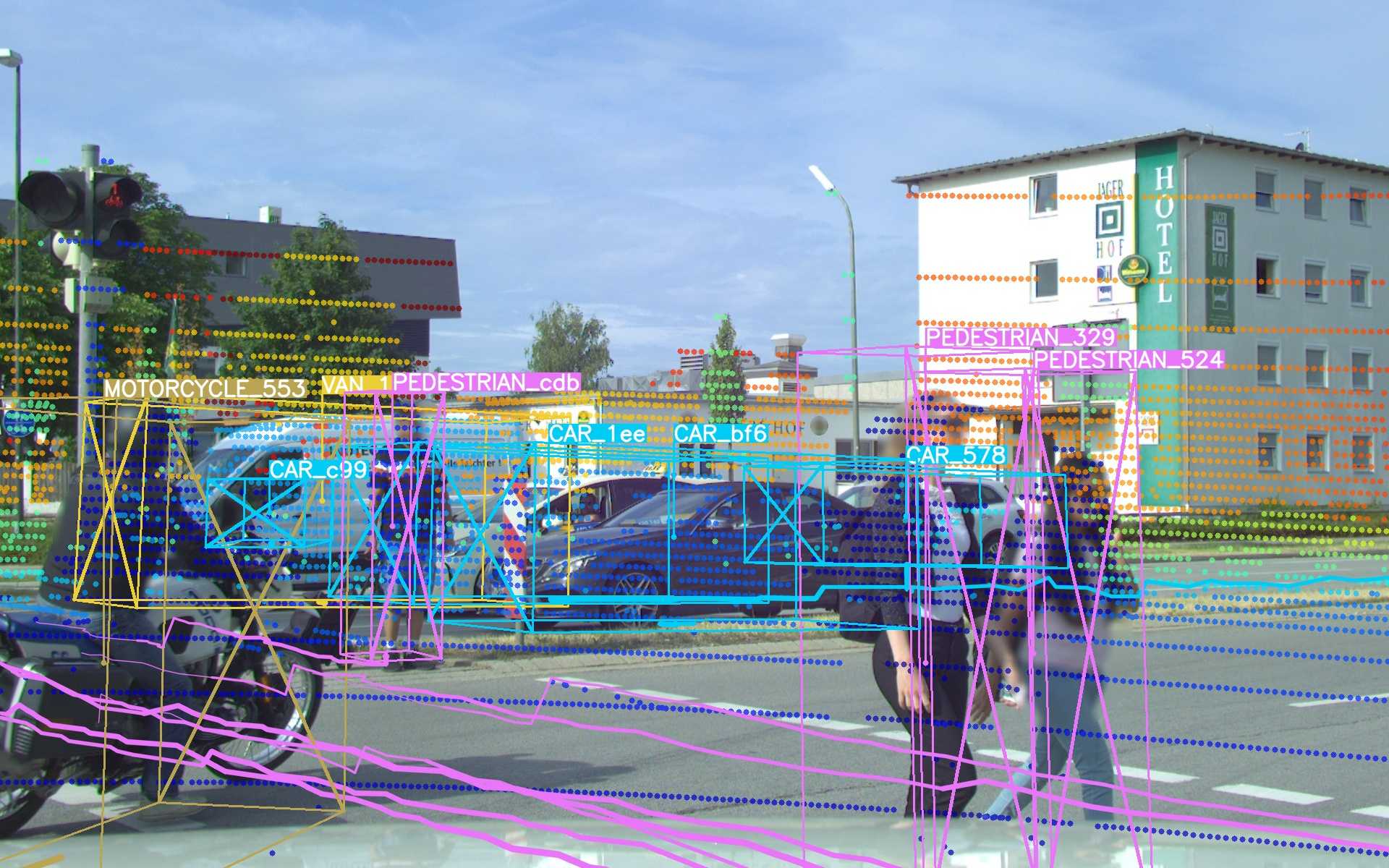}}
\endminipage
\minipage{0.25\textwidth}%
  \fbox{\includegraphics[width=.98\linewidth]{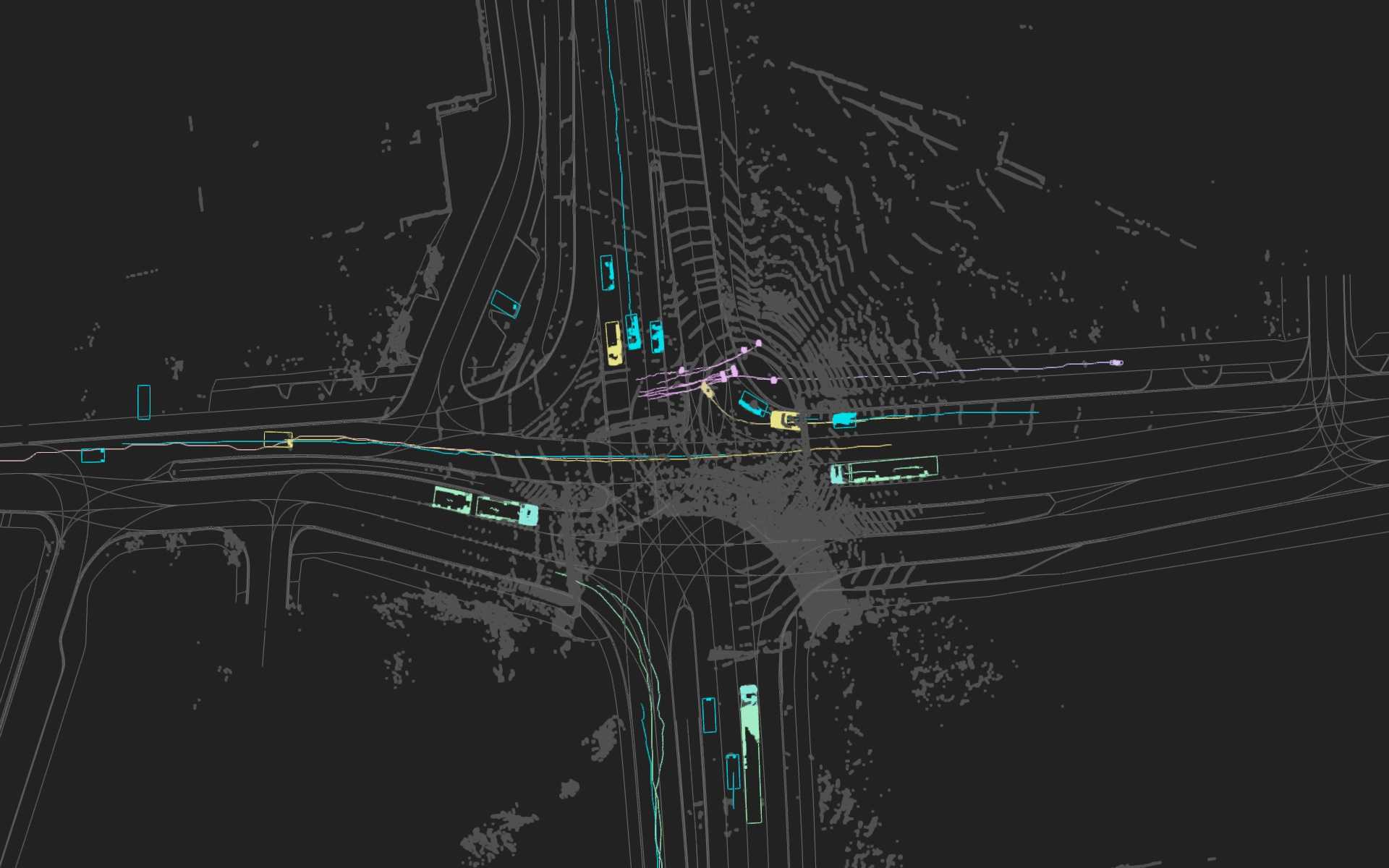}}
\endminipage
\minipage{0.25\textwidth}%
  \fbox{\includegraphics[width=.98\linewidth]{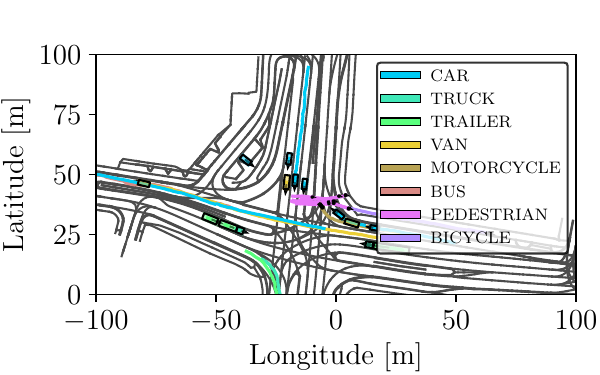}}
\endminipage
\caption{Visualization of \textbf{drive\_41} of the \textit{TUMTraf-V2X} dataset. In this example, a motorcyclist is overtaking the ego vehicle that gives way to pedestrians crossing the road. This sequence contains 3,400 labeled 3D objects.}
\label{fig:dataset_visualization_drive_41} 
\end{figure*}

\setlength{\fboxsep}{0pt}%
\setlength{\fboxrule}{1pt}%
\begin{figure*}[h!]
\centering
\minipage{0.25\textwidth}
  \fbox{\includegraphics[width=.98\linewidth]{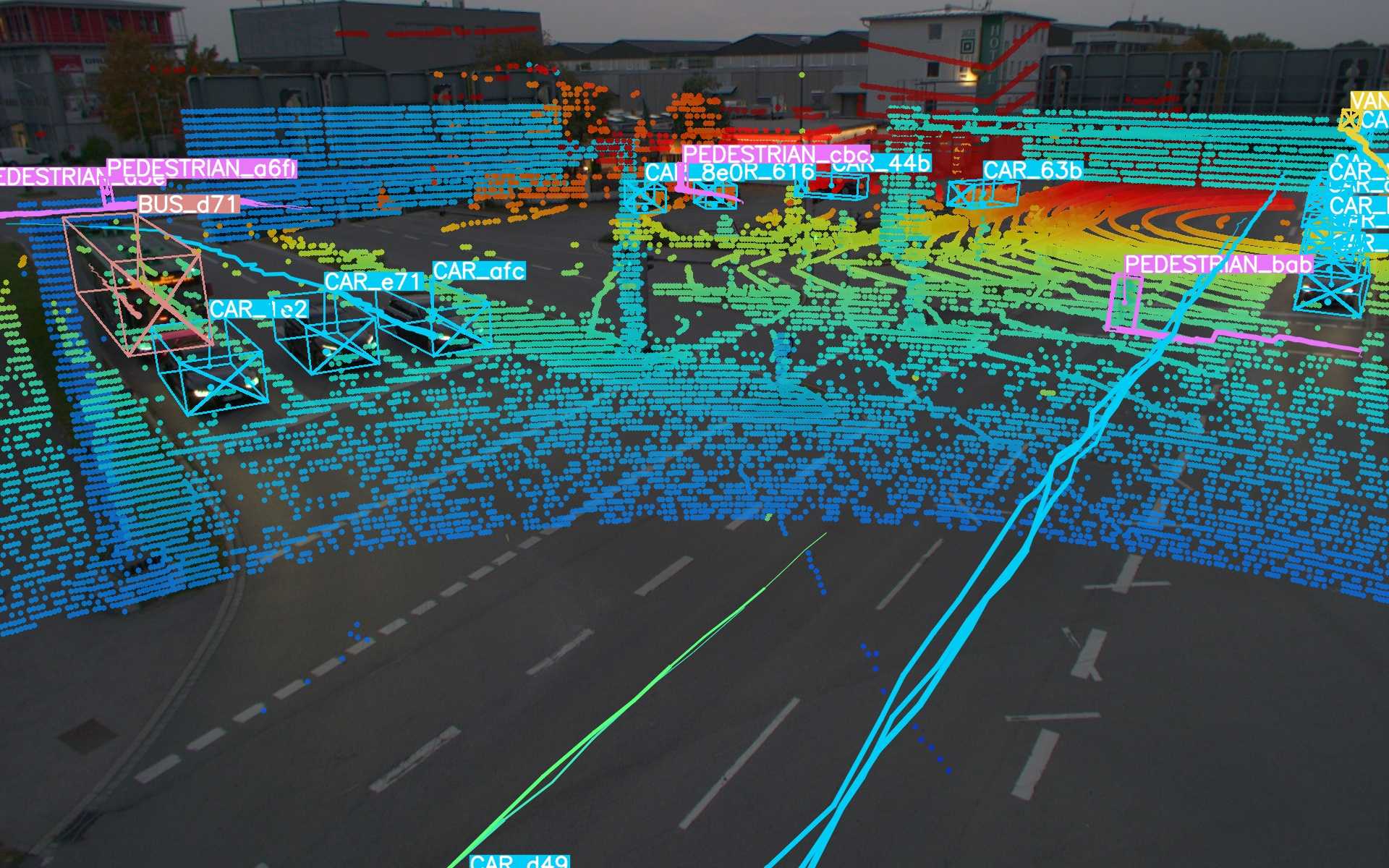}}
\endminipage
\minipage{0.25\textwidth}
  \fbox{\includegraphics[width=.98\linewidth]{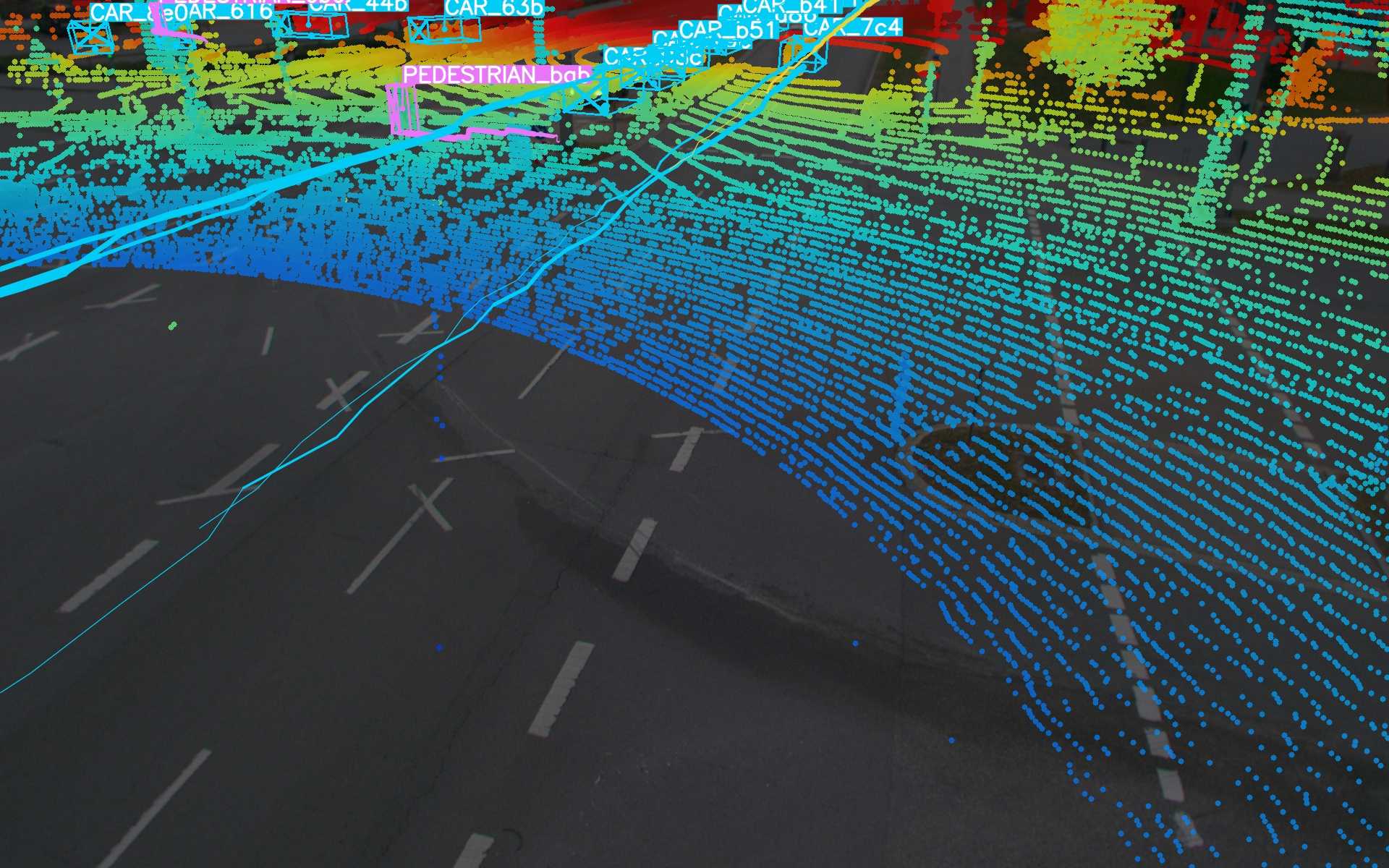}}
\endminipage
\minipage{0.25\textwidth}%
  \fbox{\includegraphics[width=.98\linewidth]{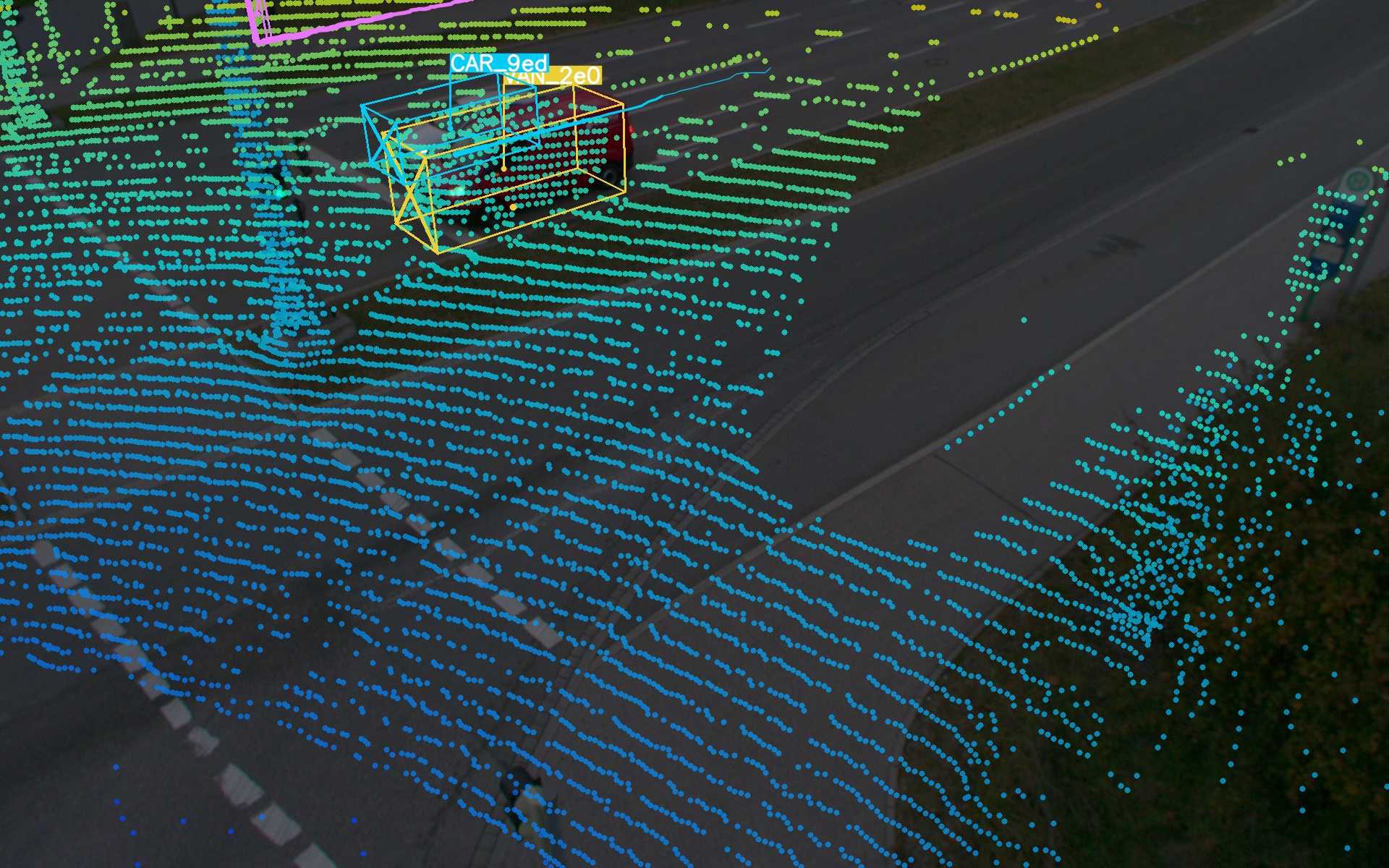}}
\endminipage
\minipage{0.25\textwidth}
  \fbox{\includegraphics[width=.98\linewidth]{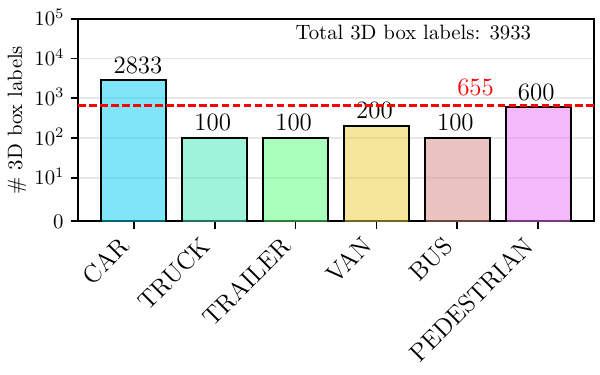}}
\endminipage\\
\vspace{-0.07cm}
\minipage{0.25\textwidth}
  \fbox{\includegraphics[width=.98\linewidth]{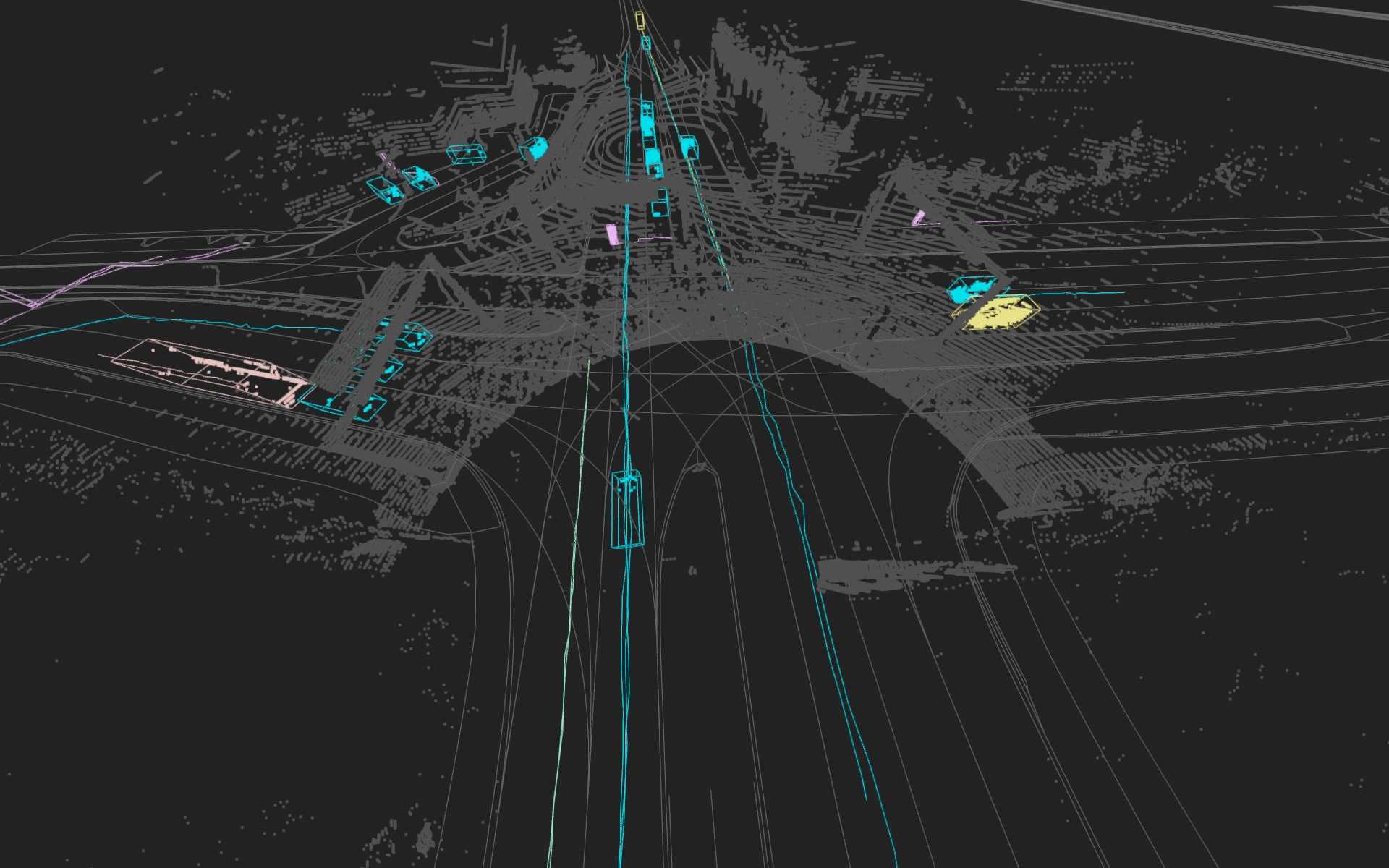}}
\endminipage
\minipage{0.25\textwidth}%
  \fbox{\includegraphics[width=.98\linewidth]{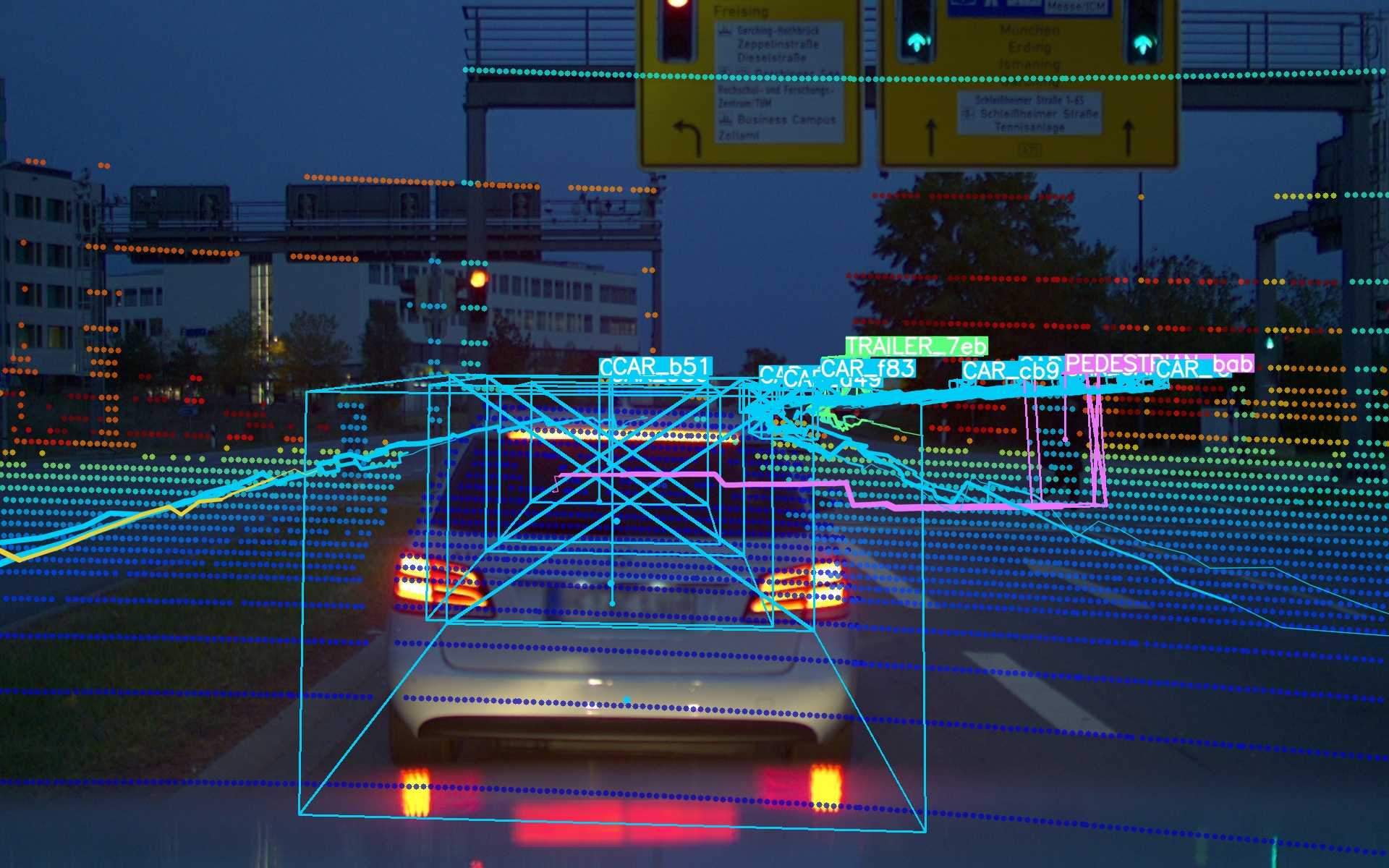}}
\endminipage
\minipage{0.25\textwidth}%
  \fbox{\includegraphics[width=.98\linewidth]{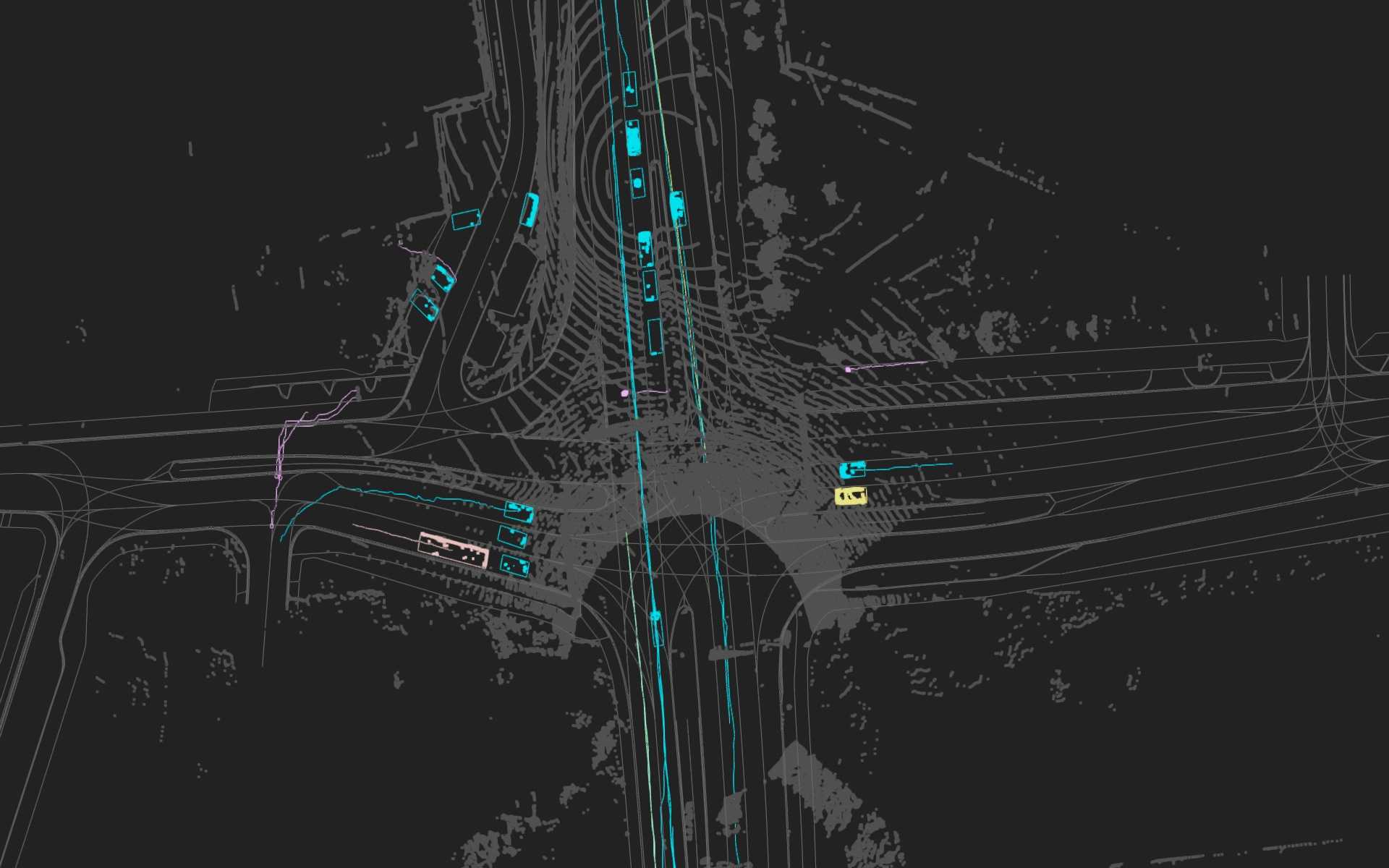}}
\endminipage
\minipage{0.25\textwidth}%
  \fbox{\includegraphics[width=.98\linewidth]{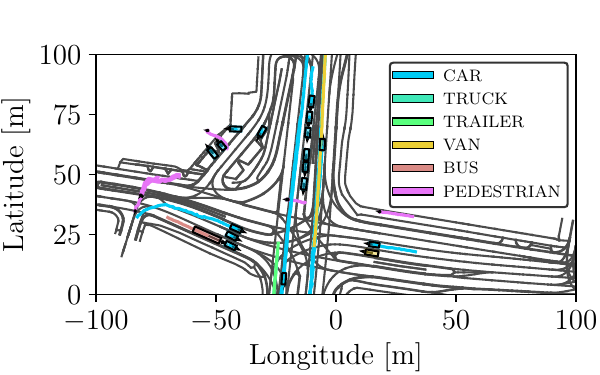}}
\endminipage
\caption{Visualization of \textbf{drive\_42} of the \textit{TUMTraf-V2X} dataset. This night scene contains a traffic violation and is the largest sequence in the dataset, with 3,933 3D objects. A pedestrian runs the red light after a fast-moving vehicle has crossed the intersection.}
\label{fig:dataset_visualization_drive_42} 
\end{figure*}
\vspace{-1\baselineskip}

\subsection{3D Object Detection}
As the most commonly utilized metric in 3D object detection tasks, \textit{mean Average Precision} (mAP) (Eq. \ref{mAP}) takes the mean value of \textit{Average Prevision} (AP) generally over the categories $\mathcal{C}$ of interest. We follow the approach of positive sample matching, introduced in \textit{nuScenes} \cite{caesar2020nuscenes}, leveraging 2D distance thresholds $\mathcal{D}$ on the ground plane between ground truth and prediction center positions, instead of using the intersection over union (IoU), to define a match (true positive). We match predictions with ground truth objects with the smallest center distance up to a certain threshold. For a given match threshold we calculate the \textit{Average Precision} (AP) by integrating the recall-precision curve for recall and precision $>0.1$. We finally average overmatch thresholds of $\mathcal{D}=\{0.5,1,2,4\}$ meters and compute the mean across all classes.

\begin{equation}
    \label{mAP}
    mAP = \frac{1}{|\mathcal{C}||\mathcal{D}|} \sum_{c \in \mathcal{C}} \sum_{d \in \mathcal{D}} AP_{c,d}
\end{equation}

\begin{table}[t]
  \caption{Evaluation results ($mAP_{BEV}$) of the \textit{CoopDet3D} and \textit{CoopCMT} model on our \textit{TUMTraf-V2X} test set in south2 FOV.}
  \label{tab:quantitativeResultsCMT}
  \centering
  \resizebox{\columnwidth}{!}{%
  \begin{tabular}{ll|rrN}
    \hline
    \multicolumn{2}{c|}{\textbf{Config.}} & \multicolumn{2}{c}{$\mathbf{mAP_{BEV}}\uparrow$} \\
    \textbf{Domain~~~~~~~~~~} & \textbf{Modality~~~~~~~~~~} & ~~~~~~~~~~CoopDet3D & ~~~~~~~~~~CoopCMT~~~~~~\\
    \hline
    Vehicle & Camera & 46.83 & \textbf{81.21} \color{ForestGreen}{(+34.38)}\\
    Vehicle & LiDAR & 85.33 &  \textbf{86.88} \color{ForestGreen}{~~(+1.55)}\\
    Infra. & Camera & 61.98 &  \textbf{79.50} \color{ForestGreen}{(+17.52)}\\
    Infra. & LiDAR & 92.86 &  \textbf{93.18} \color{ForestGreen}{~~(+0.32)}\\
    Infra. & Cam+LiDAR & 92.92 & \textbf{93.63} \color{ForestGreen}{~~(+0.71)}\\
    Coop. & LiDAR & 93.93 &  \textbf{94.27} \color{ForestGreen}{~~(+0.38)}\\
    \hline
  \end{tabular}
  }
\end{table}

\subsection{Multi-object tracking}
\textit{Multiple Object Tracking Accuracy} (MOTA) and \textit{Multiple Object Tracking Precision} (MOTP) are the most widely used metrics to evaluate tracking performance. MOTA (Eq. \ref{MOTA}) considers the main factors affecting tracking performance including \textit{False Positives} (FP), \textit{False Negatives} (FN), and \textit{ID Switches} (IDS). $GT_t$ is the number of ground truth objects at time $t$. 
\begin{equation}
    \label{MOTA}
    MOTA = 1 - \frac{\sum_t (FP_t + FN_t + IDS_t)}{\sum_t GT_t}
\end{equation}
MOTP (Eq. \ref{MOTP}) is used to measure the precision of the tracked object's position, where $d^i_t$ and $c_t$ represent the distance between the predicted object and its actual position at time $t$ and the number of matches at time $t$ respectively. 
\begin{equation}
    \label{MOTP}
    MOTP = \frac{\sum_{i,t} d^i_t}{\sum_t c_t}
\end{equation}
IDP and IDR are the ID precision and recall measuring the fraction of tracked detections that are correctly assigned to a unique ground truth ID. The IDF1 metric is the ratio of correctly identified tracked detections over the average number of ground truth objects (GT). The basic idea of IDF1 is to combine IDP and IDR into a single number. In addition, each trajectory can be classified as mostly tracked (MT), partially tracked (PT), and mostly lost (ML). A target is mostly tracked if it is successfully tracked for at least 80\% of its life span, mostly lost if it is successfully tracked for at most 20\%. All other targets are partially tracked.

\setlength{\fboxsep}{0pt}%
\setlength{\fboxrule}{1pt}%
\begin{figure*}[h!]
\centering
\minipage{0.33\textwidth}
\caption*{a) Training}
  \vspace{-0.3cm}
  \includegraphics[width=\linewidth]{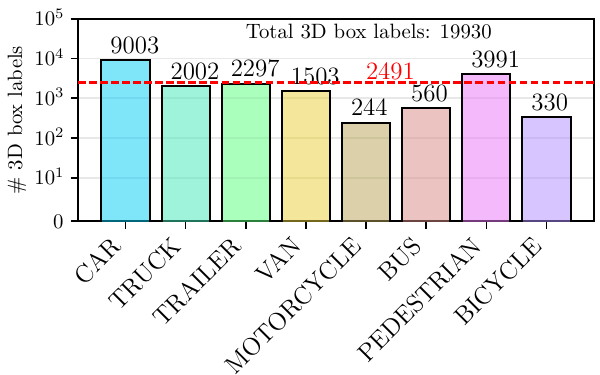}
\endminipage
\minipage{0.33\textwidth}
  \caption*{b) Validation}
  \vspace{-0.3cm}
  \includegraphics[width=\linewidth]{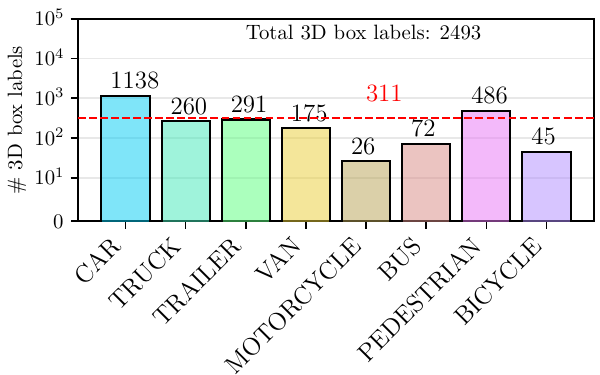}
\endminipage
\minipage{0.33\textwidth}%
  \caption*{c) Testing}
  \vspace{-0.3cm}
  \includegraphics[width=\linewidth]{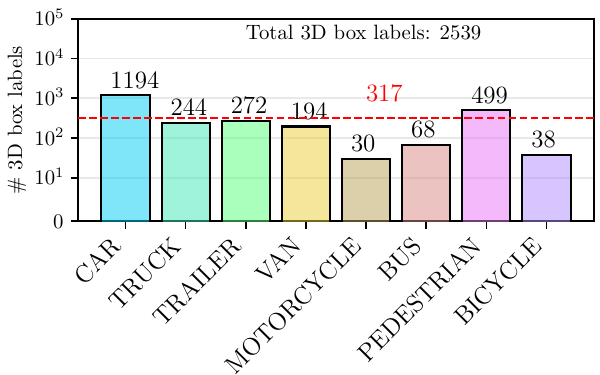}
\endminipage\\
\minipage{0.33\textwidth}
  \includegraphics[width=\linewidth]{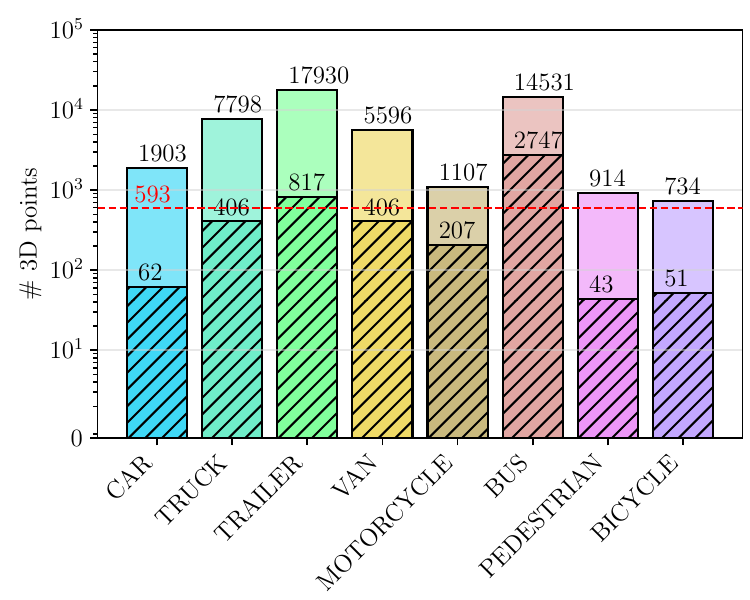}
\endminipage
\minipage{0.33\textwidth}
  \includegraphics[width=\linewidth]{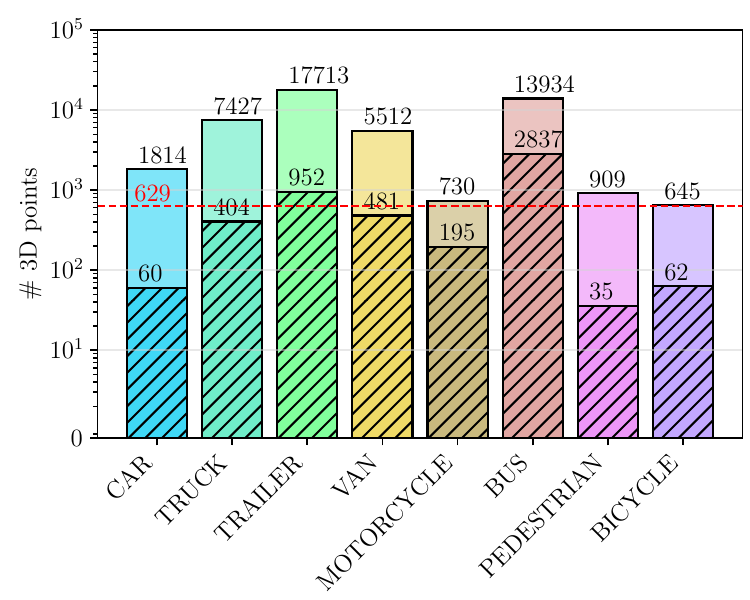}
\endminipage
\minipage{0.33\textwidth}%
  \includegraphics[width=\linewidth]{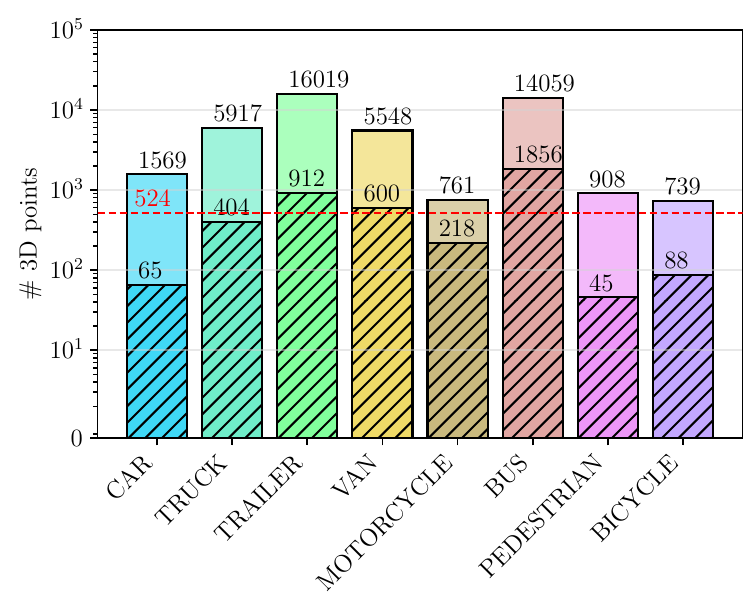}
\endminipage\\
\minipage{0.33\textwidth}
  \includegraphics[width=\linewidth]{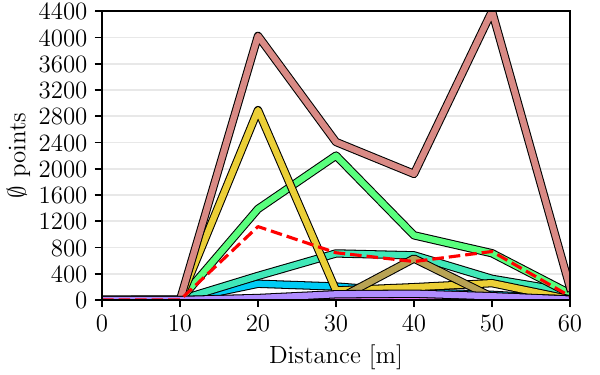}
\endminipage
\minipage{0.33\textwidth}
  \includegraphics[width=\linewidth]{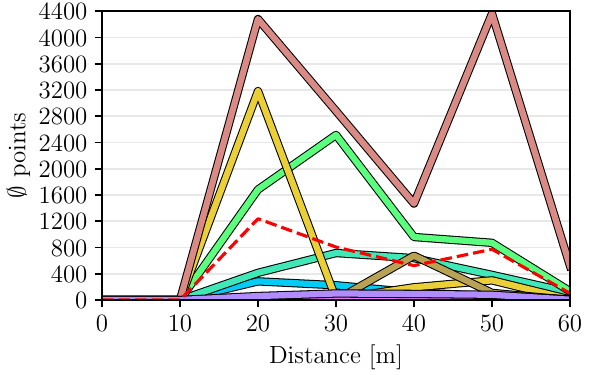}
\endminipage
\minipage{0.33\textwidth}%
  \includegraphics[width=\linewidth]{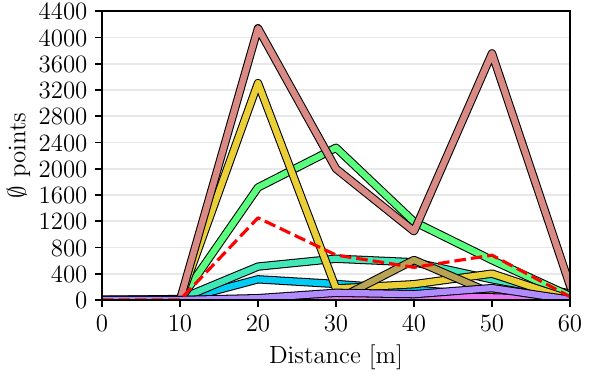}
\endminipage\\
\minipage{0.33\textwidth}
  \includegraphics[width=\linewidth]{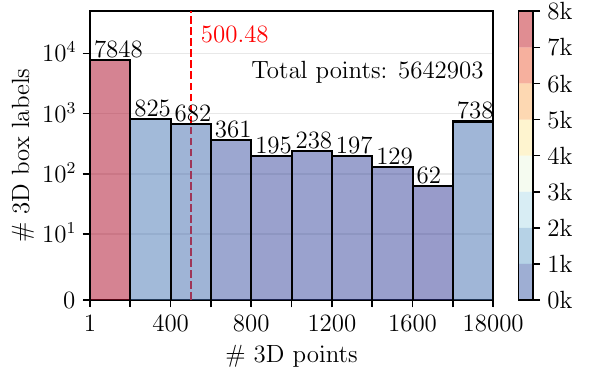}
\endminipage
\minipage{0.33\textwidth}
  \includegraphics[width=\linewidth]{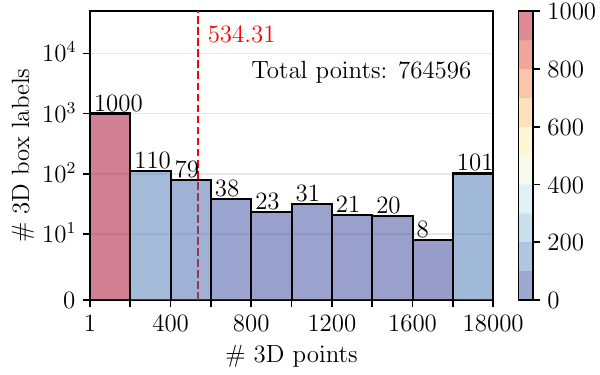}
\endminipage
\minipage{0.33\textwidth}%
  \includegraphics[width=\linewidth]{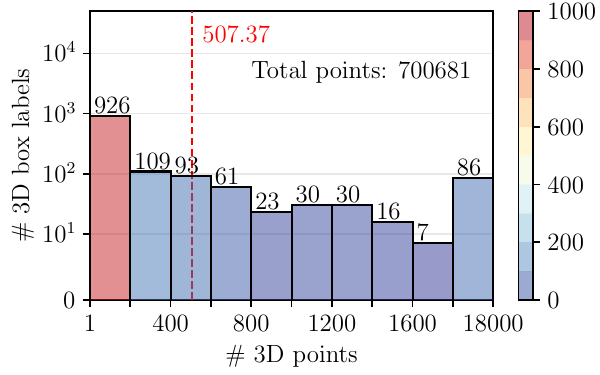}
\endminipage\\
\minipage{0.33\textwidth}
  \includegraphics[width=\linewidth]{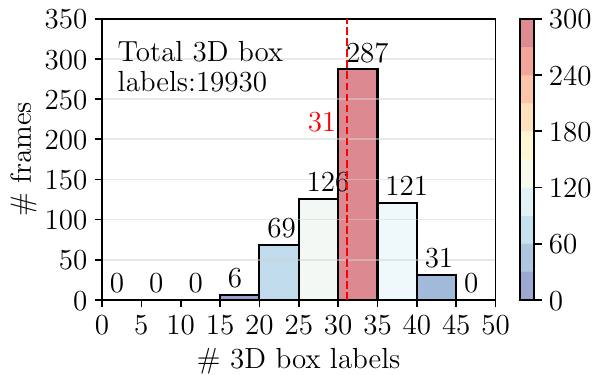}
\endminipage
\minipage{0.33\textwidth}
  \includegraphics[width=\linewidth]{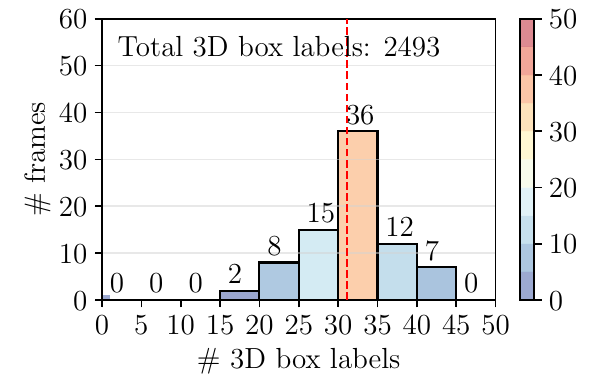}
\endminipage
\minipage{0.33\textwidth}%
  \includegraphics[width=\linewidth]{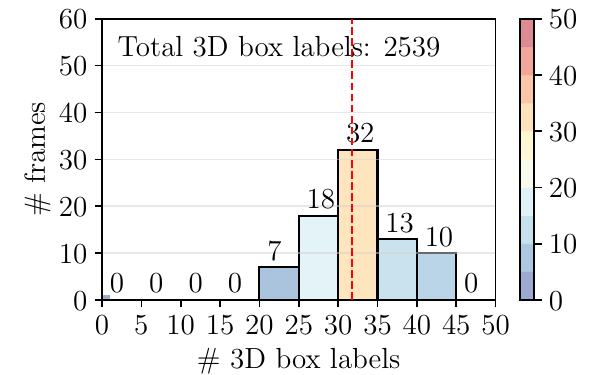}
\endminipage
\caption{Distribution of our \textit{TUMTraf-V2X} dataset (version 1.0) into a) training, b) validation, and c) test set. From top to bottom: We show the distribution of object classes within each set with the average number of 3D box labels marked in red, the distribution of 3D points for each category and each set, the labeled distance and class density for each object class and set, a histogram of 3D box densities for each set, and a histogram of frame densities for each set.}
\label{fig:distribution} 
\end{figure*}

\section{Further experiments}

We extend our experiments to consider multiple FOVs, baseline models, and different tasks made possible through the proposed \textit{TUMTraf-V2X} dataset.

\subsection{CoopDet3D}
\begin{figure*}[h!]
\centering
\minipage{0.25\textwidth}
  \fbox{\includegraphics[width=.98\linewidth]{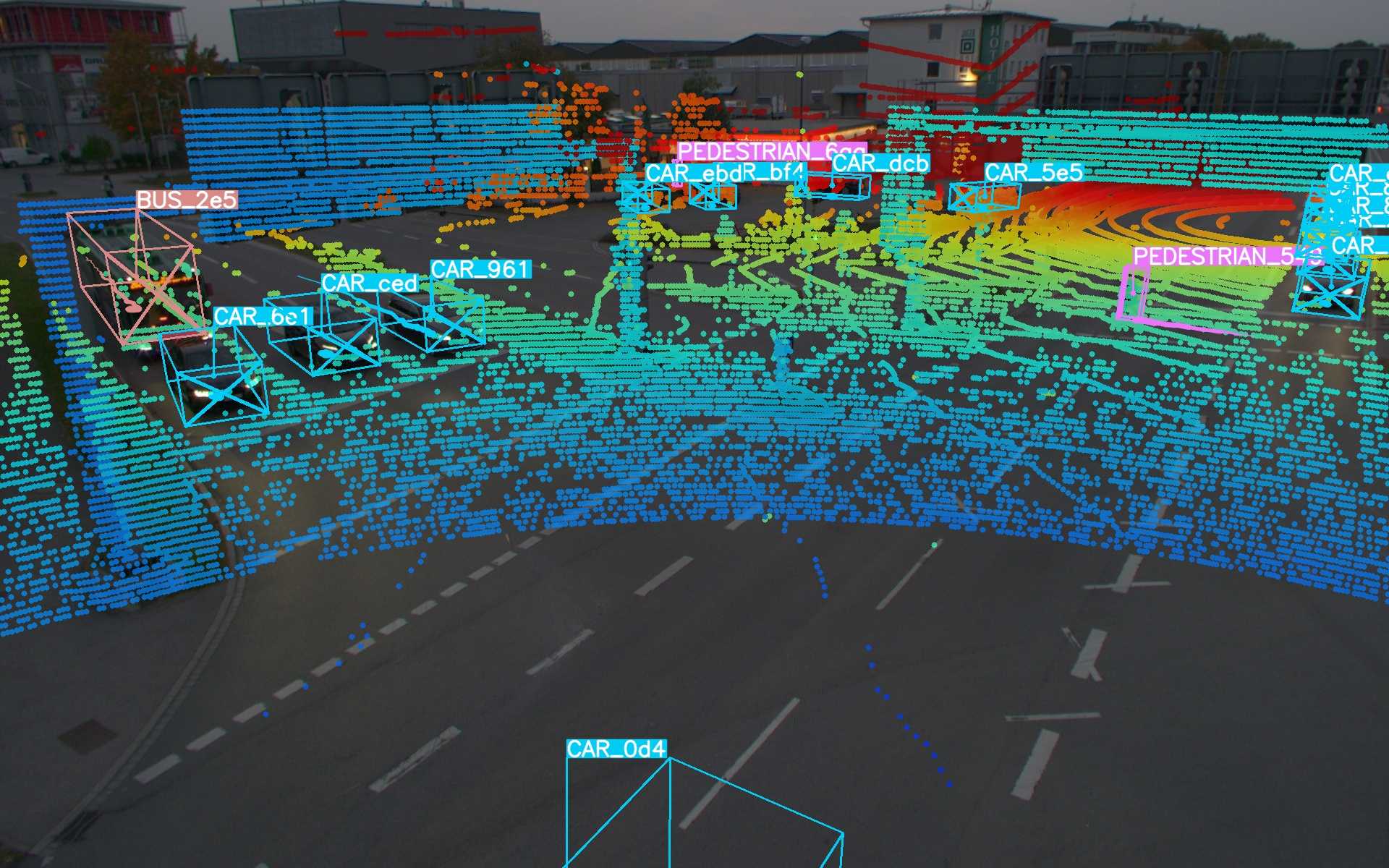}}
\endminipage
\minipage{0.25\textwidth}
  \fbox{\includegraphics[width=.98\linewidth]{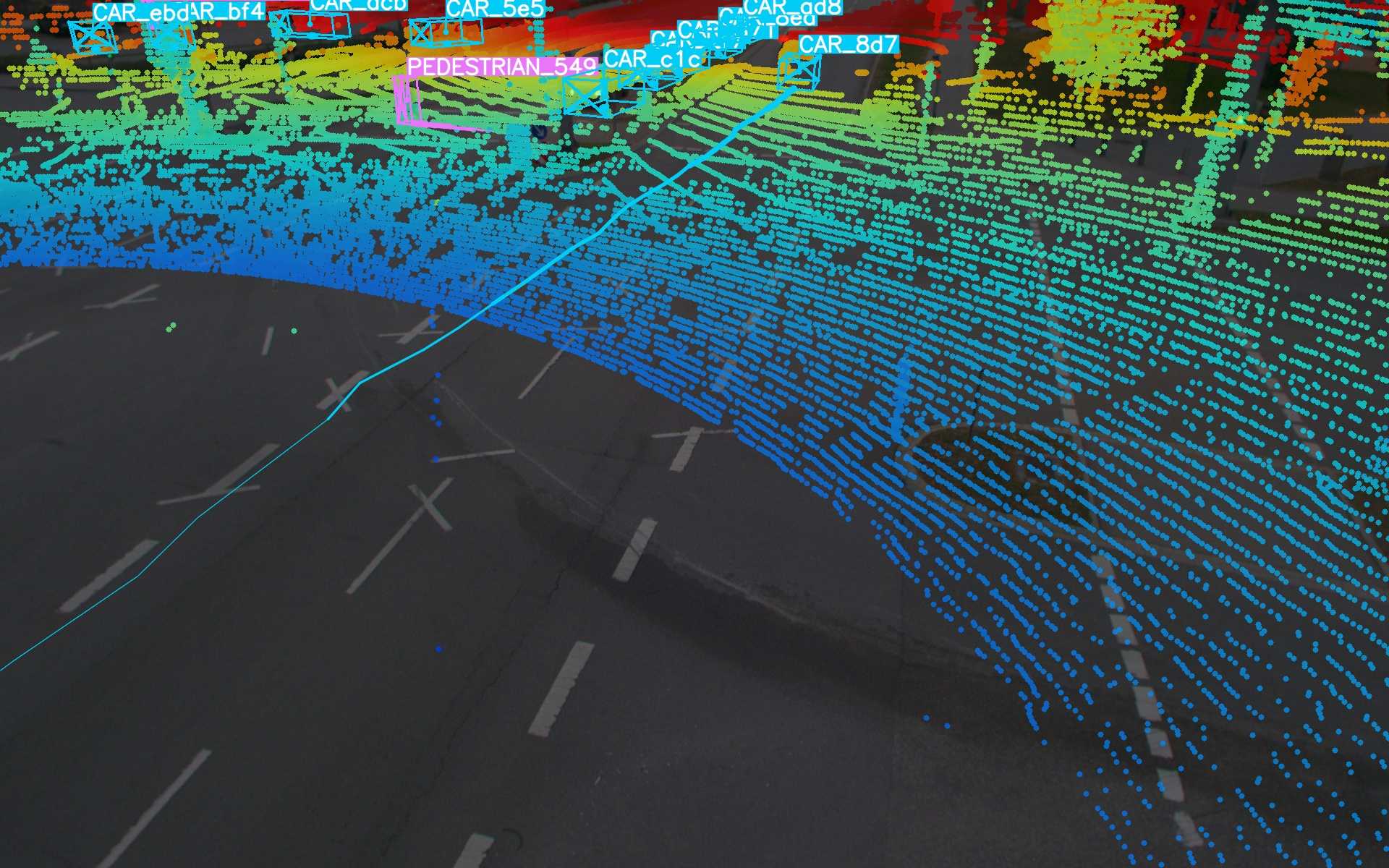}}
\endminipage
\minipage{0.25\textwidth}%
  \fbox{\includegraphics[width=.98\linewidth]{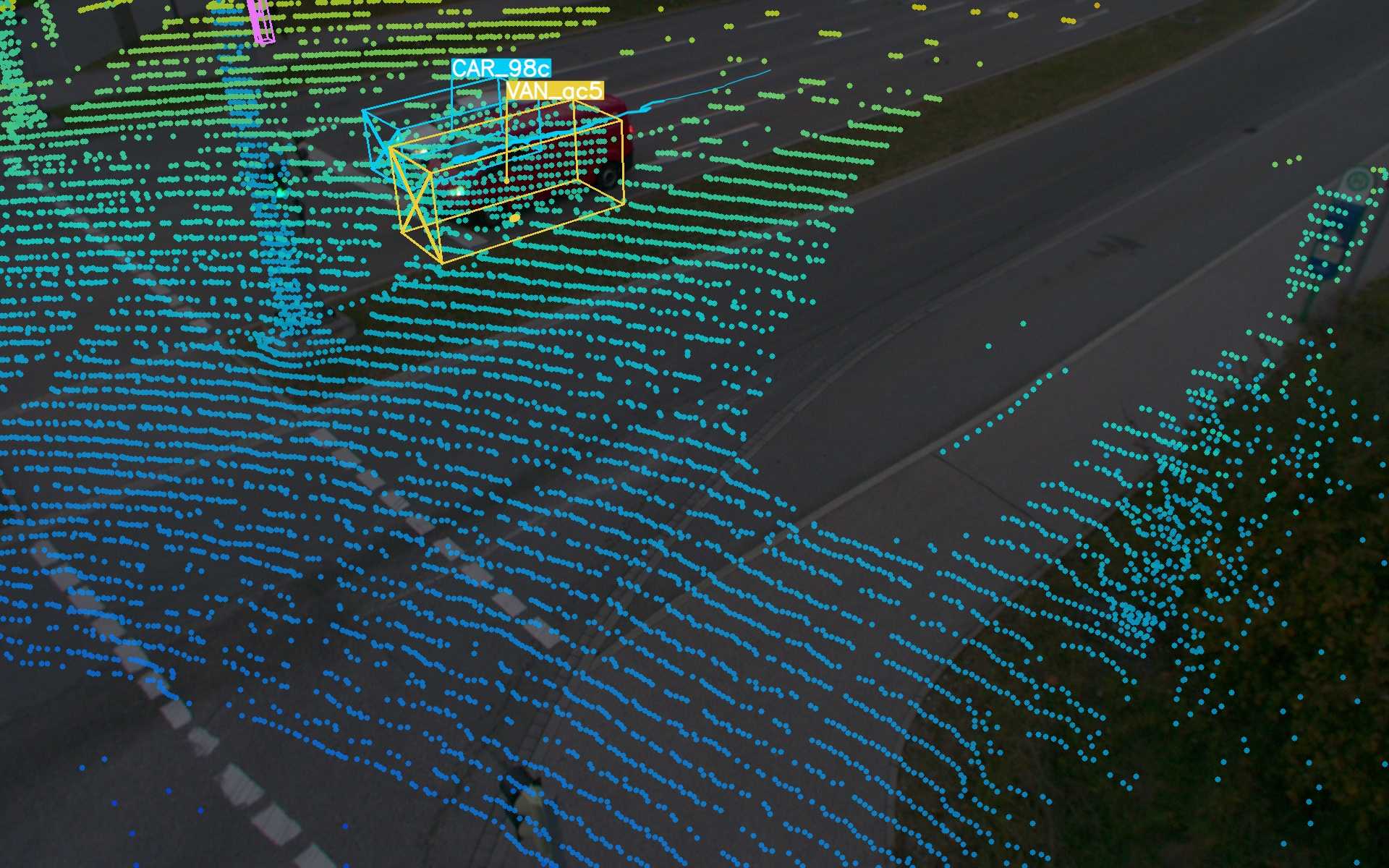}}
\endminipage
\minipage{0.25\textwidth}
  \fbox{\includegraphics[width=.98\linewidth]{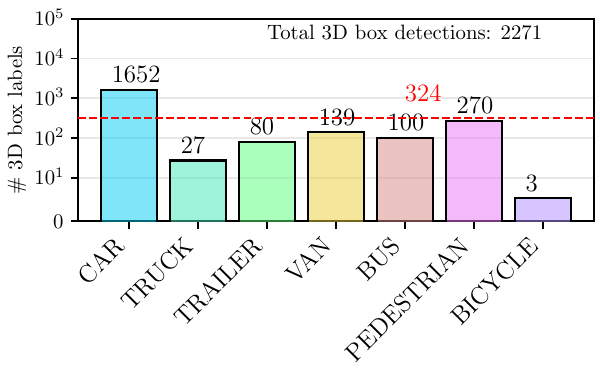}}
\endminipage\\
\vspace{-0.07cm}
\minipage{0.25\textwidth}
  \fbox{\includegraphics[width=.98\linewidth]{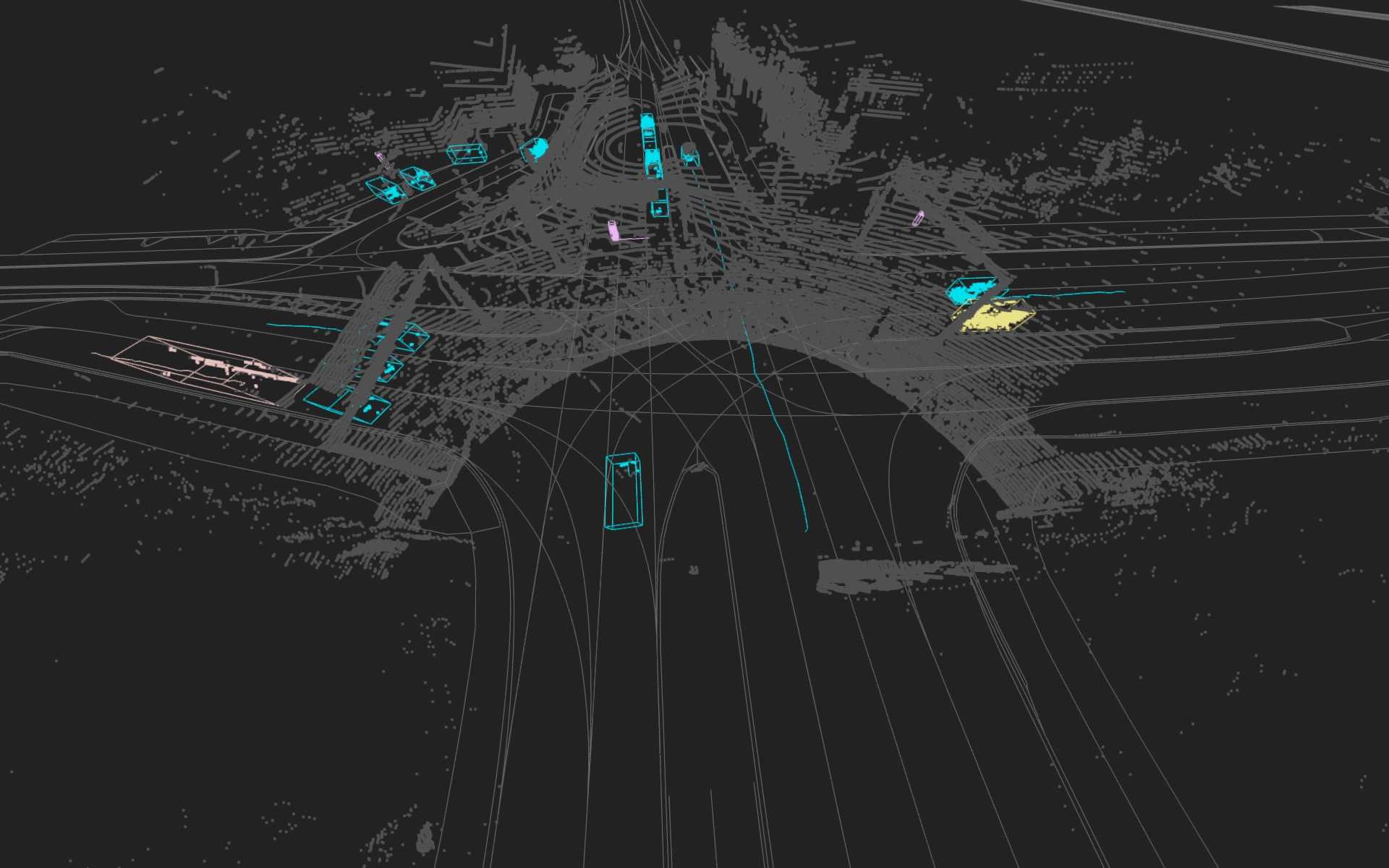}}
\endminipage
\minipage{0.25\textwidth}%
  \fbox{\includegraphics[width=.98\linewidth]{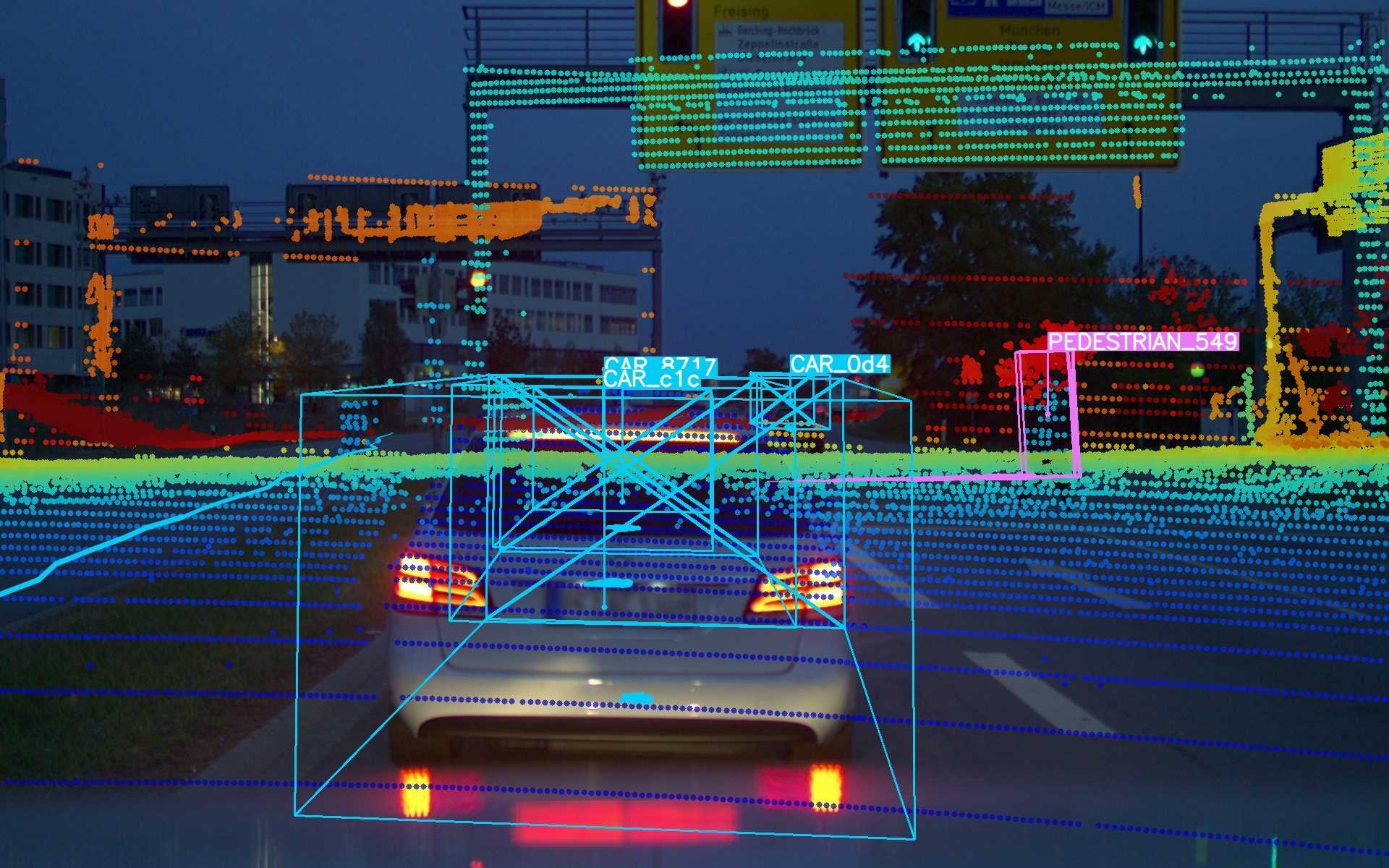}}
\endminipage
\minipage{0.25\textwidth}%
  \fbox{\includegraphics[width=.98\linewidth]{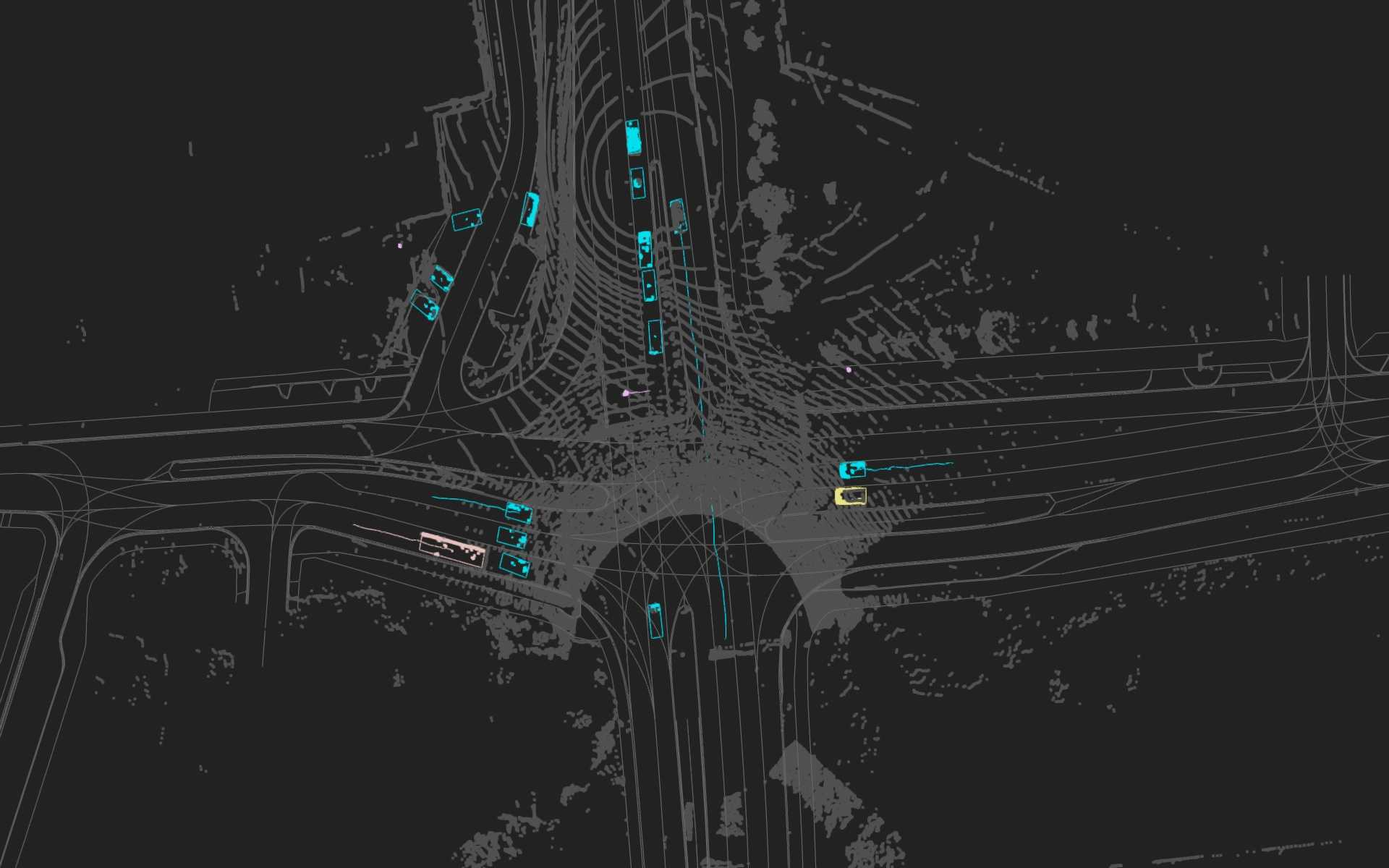}}
\endminipage
\minipage{0.25\textwidth}%
  \fbox{\includegraphics[width=.98\linewidth]{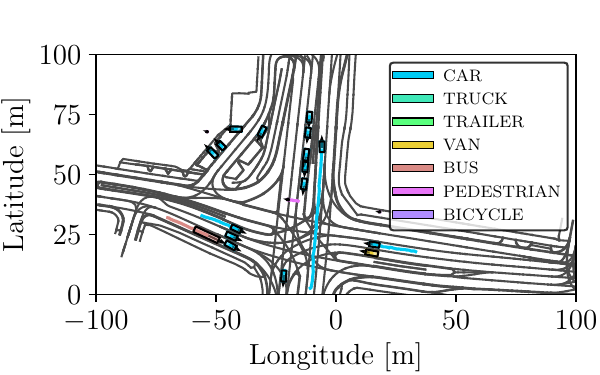}}
\endminipage
\caption{Qualitative results of \textit{CoopDet3D} on \textbf{drive\_42} of our \textit{TUMTraf-V2X} dataset of a night scene. We project the detections into point cloud scans and camera images. Moreover, we visualize object tracks in a bird's-eye view and an HD map. Finally, we show the distribution of detections in a bar chart.}
\label{fig:qualitative_results_coopdet3d} 
\end{figure*}

Previously we discussed the performance of the proposed \textit{CoopDet3D} model with \textit{PointPillars} 512\_2x and \textit{YOLOv8} backbones in South2 camera FOV. In Table \ref{tbl:quantitativeResultsS1} we show the quantitative results of the same model in South1 camera FOV. Like the South2 camera FOV, we observe that the \textit{CoopDet3D} cooperative model performs better than the vehicle-only perception model (+7.47 3D mAP). Fig. \ref{fig:qualitative_results_coopdet3d} shows qualitative results of \textit{CoopDet3D} on drive\_42.

\setlength{\fboxsep}{0pt}%
\setlength{\fboxrule}{1pt}%


\subsection{CoopCMT}
In addition to \textit{CoopDet3D}, we build another cooperative fusion model: \textit{CoopCMT} for benchmarking, based on cross-modal transformers (CMT) \cite{yan2023cross}. Similar to the proposed \textit{CoopDet3D} model, the \textit{CoopCMT} cooperative perception model uses separate vehicle and infrastructure backbones for feature extraction. Then, the extracted infrastructure and vehicle deep features are concatenated using a \textit{MaxPooling} layer (similar to \textit{PillarGrid} \cite{bai2022pillargrid}), and finally passed onto the 3D detection head. Thus, this architecture is similar to the \textit{CoopDet3D} architecture, where the \textit{BEVFusion}-based backbones and head, are replaced with the corresponding counterpart from the \textit{CMT} model. Note, that since transformer-based models require a large amount of data to be trained, the infrastructure backbone was first pre-trained on the \textit{TUMTraf Intersection} dataset \cite{zimmer2023tumtraf}, and the vehicle backbone was pre-trained on the \textit{nuScenes} dataset \cite{nuscenes2019}, to fit the domain. We compare the performance of the \textit{CoopCMT} model with \textit{CoopDet3D} in Table \ref{tab:quantitativeResultsCMT} and see that it outperforms the \textit{CoopDet3D} model in all domains and modalities.

From Table \ref{tab:quantitativeResultsCMT}, we observe a general trend in which the \textit{CoopCMT} cooperative fusion model performs better in terms of the $mAP_{BEV}$ compared to the \textit{CoopDet3D} model. However, it must be noted that the \textit{CoopCMT} model uses a transformer-based architecture, and as such, the model complexity is higher, resulting in slower inference time. For future research, the \textit{CoopCMT} model will be studied further in terms of the model complexity and FPS to ensure that this model can perform in near real-time and can be deployed on edge devices.

\subsection{3D multi-object tracking}
Next, we track the \textit{CoopDet3D} detections in a post-processing step using two different trackers: \textit{SORT} \cite{bewley2016simple} and \textit{PolyMOT} \cite{li2023poly}. The quantitative evaluation results of 15 different metrics are listed in Table \ref{tbl:tracking_results}. We use a distance threshold of 5 m for the \textit{SORT} tracker. The \textit{PolyMOT} tracker performs best in all metrics except PT and MOTP. Qualitative results are shown in Fig. \ref{fig:tracking_qualitative_results}.


\section{Statistics of all drives}
Detailed statistics of all labeled sequences are seen in \cref{fig:dataset_visualization_drive_07,fig:dataset_visualization_drive_12,fig:dataset_visualization_drive_15,fig:dataset_visualization_drive_22,fig:dataset_visualization_drive_26,fig:dataset_visualization_drive_33,fig:dataset_visualization_drive_41,fig:dataset_visualization_drive_42}. The last driving sequence (drive\_42) was recorded during nighttime and contains a traffic violation scenario in which a pedestrian is running the red light. All other sequences contain daytime traffic with heavy occlusion scenarios. We split our dataset into a training (80\%), validation (10\%), and test (10\%) set using stratified sampling to get a well-balanced split. The distribution of object classes of our training, validation, and test set is shown in Fig. \ref{fig:distribution}.

\section{Detailed dataset visualization}
We provide detailed dataset visualizations for different challenging traffic scenarios at an urban intersection, including tailgating, overtaking, U-turns, traffic violations, and occlusion scenarios. In one scene, a pedestrian runs a red light after a vehicle is crossing. We show each scenario's surround-view images, BEV projections on an HD map, point cloud visualizations, and a class distribution plot. Visualization videos for all labeled sequences are provided on our website: \url{https://tum-traffic-dataset.github.io/tumtraf-v2x}.

\section{Failure cases and limitations}
Failure cases are essential to understand the weakness of our dataset and model and to provide some guidance for future work. 
Note that, for brevity, we do not consider the network communication latency between the sensors. \\
We have tested our \textit{CoopDet3D} model in day and night scenarios in different weather conditions. Some future work will include further tests under harsh weather conditions such as heavy rain, snow, and fog.
Apart from object detection, cooperative perception poses many other challenges due to the asynchrony between the vehicle and infrastructure sensors, and the transmission delay further exacerbates this issue. While the suggested model may not fully account for these considerations, it is recommended that future research focuses on addressing these challenges through extensive live tests.

\pagebreak
\addtolength{\textheight}{-0.7cm}
{
    \small
    \bibliographystyle{ieeenat_fullname}
    \bibliography{main}
}


\end{document}